%% file: main.tex
\documentclass[10pt]{article} %
\usepackage[accepted]{tmlr}

\input{math_commands.tex}

\usepackage{url}

\usepackage{CJKutf8}

\usepackage{booktabs}       %
\usepackage{amsfonts}       %
\usepackage{nicefrac}       %
\usepackage{microtype}      %
\usepackage{mathtools}
\usepackage{multirow}
\usepackage{bm}
\usepackage{epsfig}
\usepackage{graphicx}
\usepackage{subcaption}
\usepackage{amsthm}
\usepackage{amsmath}
\usepackage{amssymb}
\usepackage{enumitem}
\usepackage{makecell}
\usepackage{wrapfig}
\usepackage{indentfirst}
\usepackage{verbatim}
\usepackage{color}
\usepackage{setspace}
\usepackage{array}
\usepackage{stackengine}
\usepackage{algorithm}
\usepackage[algo2e,algoruled,boxed,lined]{algorithm2e}
\usepackage{algorithmicx}
\usepackage{graphicx}
\usepackage{xcolor}
\usepackage{anyfontsize}

\usepackage{caption}
\renewcommand{\captionlabelfont}{\scriptsize}

\usepackage[pagebackref=true,breaklinks=true,colorlinks=true,bookmarks=false]{hyperref}
\definecolor{deepred}{HTML}{940000}
\hypersetup{linkcolor=deepred}
\hypersetup{urlcolor  = [rgb]{0.4,0.15,0.95}}
\hypersetup{citecolor=[rgb]{0.4,0.15,0.95}}

\definecolor{model}{HTML}{00639E}
\definecolor{opt}{HTML}{2A5F1C}

\usepackage{colortbl}
\definecolor{Gray}{gray}{0.94}
\newcolumntype{a}{>{\columncolor{Gray}}c}
\newcolumntype{Y}{>{\centering\arraybackslash}X}
\definecolor{americanrose}{rgb}{1.0, 0.01, 0.24}

\newlength\savewidth\newcommand\shline{\noalign{\global\savewidth\arrayrulewidth
  \global\arrayrulewidth 1pt}\hline\noalign{\global\arrayrulewidth\savewidth}}

\usepackage{tabularx}

\usepackage{pifont}
\usepackage{tocloft}
\usepackage[toc,page,header]{appendix}
\usepackage{adjustbox}
\usepackage{minitoc}

\renewcommand \thepart{}
\renewcommand \partname{}

\usepackage[capitalize,noabbrev]{cleveref}

\usepackage{lipsum}

\usepackage{epigraph}

\usepackage{listings}

\newcommand{\modeli}[1]{{
{\color{model}$\text{model}_{#1}$}
}}
\newcommand{\opti}[1]{{
{\color{opt}$\text{optimizer}_{#1}$}
}}

\newcommand{\ie}{\emph{i.e.}}
\newcommand{\eg}{\emph{e.g.}}

\newcommand{\dtrain}{\mathcal{D}_\text{train}}

\renewcommand{\figref}[1]{Figure~\ref{#1}}

\title{Verbalized Machine Learning:\\Revisiting Machine Learning with Language Models}

\author{%
  \name Tim Z. Xiao \email zhenzhong.xiao@uni-tuebingen.de\\
  \addr Max Planck Institute for Intelligent Systems, T\"ubingen \& University of T\"ubingen
  \AND 
  \name Robert Bamler \email robert.bamler@uni-tuebingen.de\\
  \addr University of T\"ubingen
  \AND 
  \name Bernhard Sch\"olkopf \email bernhard.schoelkopf@tuebingen.mpg.de\\
  \addr Max Planck Institute for Intelligent Systems, T\"ubingen
  \AND
  \name Weiyang Liu\thanks{Corresponding author} \email weiyang.liu@tuebingen.mpg.de\\
  \addr Max Planck Institute for Intelligent Systems, T\"ubingen \& University of Cambridge
}

\def\month{01}  %
\def\year{2025} %
\def\openreview{\url{https://openreview.net/forum?id=k3Ab6RuJE9}} %

\begin{document}

\maketitle
\doparttoc 
\faketableofcontents

\begin{abstract}
Motivated by the progress of large language models (LLMs), we introduce the framework of verbalized machine learning (VML). In contrast to conventional machine learning (ML) models that are typically optimized over a continuous parameter space, VML constrains the parameter space to be human-interpretable natural language. Such a constraint leads to a new perspective of function approximation, where an LLM with a text prompt can be viewed as a function parameterized by the text prompt. Guided by this perspective, we revisit classical ML problems, such as regression and classification, and find that these problems can be solved by an LLM-parameterized learner and optimizer. The major advantages of VML include (1) easy encoding of inductive bias: prior knowledge about the problem and hypothesis class can be encoded in natural language and fed into the LLM-parameterized learner; (2) automatic model class selection: the optimizer can automatically select a model class based on data and verbalized prior knowledge, and it can update the model class during training; and (3) interpretable learner updates: the LLM-parameterized optimizer can provide explanations for why an update is performed. We empirically verify the effectiveness of VML, and hope that VML can serve as a stepping stone to stronger interpretability.
\end{abstract}

\vspace{-5mm}

\setlength\epigraphwidth{8.6cm}
\epigraph{\emph{``The limits of my language mean the limits of my world.''}}{--- \textup{Ludwig Wittgenstein}}
\vspace{-7mm}

\section{Introduction}

The unprecedented success of large language models (LLMs) has changed the way people solve new problems in machine learning. Compared to conventional end-to-end training where a neural network is trained from scratch on some curated dataset, it has become increasingly more popular to leverage a pretrained LLM and design good prompts that contain in-context examples and effective instructions. These two ways of problem-solving lead to an intriguing comparison. Traditionally, we would optimize a neural network in \emph{a continuous numerical space} using gradient descent, while in the new approach, we optimize the input prompt of an LLM in \emph{a discrete natural language space}. Since a neural network is effectively a function parameterized by its numerical weights, can a pretrained LLM act as a function parameterized by its text prompt?

Driven by this question, we conceptualize the framework of verbalized machine learning (VML), which uses natural language as the representation of the model parameter space. \emph{The core idea behind VML is that we can define a machine learning model using natural language, and the training of such a model is based on the iterative update of natural language.} This framework enables many new possibilities for interpretability, as the decision rules and patterns learned from data are stored and summarized in natural language. Specifically, we propose to view the input text prompt of LLMs as the model parameters that are being learned. However, optimization over such a natural language parameter space also introduces additional difficulties. Inspired by previous work~\citep{andrychowicz2016learning,li2016learning} where the optimizer is viewed as a function parameterized by a neural network, we parameterize the optimizer function as another LLM, which produces the next-step model parameters by taking in the current model parameters, a batch of training data points, and the loss function. Therefore, VML requires the optimizer LLM to update the learner LLM iteratively towards the training objective.

Compared to conventional numerical machine learning, the VML framework brings a few unique advantages. First, VML introduces an easy and unified way to encode inductive bias into the model. Because the model parameters are fully characterized by human-interpretable natural language, one can easily enter the inductive bias using language. This linguistic parameterization makes machine learning models fully interpretable and adjustable. For example, if the input and output data are observed to be linearly correlated, then one can use this sentence as part of text prompt. How to effectively encode inductive bias is actually a longstanding problem in machine learning, and VML provides a unified way to inject the inductive bias through natural language---just like teaching a human learner. Second, VML performs automatic model selection during the learning process. The optimizer LLM can automatically select a suitable model class based on the training data and verbalized prior knowledge. Third, each update of the model is fully interpretable in the sense that the optimizer LLM can give an explanation of why it chooses such an update. One can even interact with the optimizer LLM in order to inject new prior knowledge or obtain detailed reasoning.

VML can be viewed as a natural generalization of in-context learning (ICL). Specifically, ICL is a single-step implicit learning process, while VML is a multi-step iterative learning process where the in-context examples are summarized into verbal pattern and knowledge. Moreover, VML offers a sequential (or conditional) way for scaling inference-time compute~\cite{brown2024large,snell2024scaling}. Compared to the best-of-N re-sampling, VML iteratively updates its model parameter prompt by taking into account the learner's past predictions.

An important concept of VML is its unified token-level representation of both data and model. Unlike numerical machine learning, language models in VML do not differentiate data and model, and treat both of them as part of the text prompt. This shares a striking connection to stored-program computers, also known as the von Neumann architecture, where the key idea is to represent programs as data rather than wiring setups. The link between language models and stored-program computers underscores the importance of text prompts, which play a similar role to computer programs, and, along with LLMs, can become a powerful zero-shot problem solver. Our contributions are as follows:

\begin{itemize}[leftmargin=*,nosep]
\setlength\itemsep{0.4em}
    \item We formulate the framework of verbalized machine learning, where pretrained language models are viewed as function approximators parameterized by their text prompts. Then, we revisit a few simple machine learning problems and show that VML is able to solve them.
    \item We design a concrete VML algorithm with a text prompt template. This algorithm parameterizes both the learner model and the optimizer as LLMs, and enables the iterative verbalized training.
    \item We conduct empirical studies for the injection of verbalized inductive bias and show that it is promising to use natural language as a unified way to encode prior knowledge. Moreover, we validate the effectiveness of VML in different applications (Section~\ref{sect:exp}, Appendix~\ref{app:case_study}, \ref{app:loss_feedback},\ref{app:numerical_error},\ref{app:tool},\ref{app:pred_var}).  
\end{itemize}
\vspace{0.5mm}

\vspace{-1mm}
\section{Related Work}
\vspace{-1mm}

\textbf{LLMs for planning and optimization}. Language models are used to perform planning for embodied agents~\citep{song2023llm,xie2023translating,li2022pre,liang2023code}, such that they can follow natural language instruction to complete complex tasks. More recently, LLMs have been used to solve optimization problems~\citep{yang2023large}. Specifically, the LLM generates a new solution to an optimization problem from a prompt that contains previously generated solutions and their loss values. The LLM optimizer in \citep{yang2023large} shares a high-level similarity to our work, as we also aim to solve an optimization problem with LLMs. The key difference to \citep{yang2023large} is our function approximation view of LLMs, which enables us to revisit classical machine learning problems and solve them in the VML framework.

\textbf{Natural language to facilitate learning}. \citep{radford2021learning,li2022blip,li2023blip,mu2022slip,zhai2023sigmoid} show that natural language captions serve as an effective supervision to learn transferable visual representation. \citep{menon2022visual,oikarinen2023label,pratt2023does,maniparambil2023enhancing,yang2023language,yan2023learning} find that natural language descriptions can easily be turned into zero-shot classification criteria for images. \citep{andreas2018learning} proposes to use natural language as latent parameters to characterize different tasks in few-shot learning. In contrast, VML uses the text prompt of LLMs to parameterize functions and learns this prompt in a data-driven fashion.

\textbf{Prompt engineering and optimization}. There are many prompting methods~\citep{wei2022chain,zhang2022automatic,zhou2022large,wang2022self,yao2024tree,yao2023beyond,weston2023system} designed to elicit the reasoning ability of LLMs. To reduce the efforts in designing good prompts, prompt optimization~\citep{zhang2022automatic,zhou2022large,yang2023large,pryzant2023automatic,wen2024hard,deng2022rlprompt,li2024guiding,ma2024large,sordoni2024joint,yuksekgonul2024textgrad} has been proposed. 
VML can be viewed as a special instance of prompt optimization, but unlike many current generic prompt optimization methods that search the optimal text prompts through best-of-N sampling without reasoning, VML updates its text-based parameters by explicitly reasoning about the incorrect predictions and learning the underlying data pattern based on the reasoning outcome, which ensures that the learner in VML remains fully interpretable. 
To summarize, the difference between VML and current generic prompt optimization (\eg, \citep{zhou2022large}) is similar to the difference between \emph{gradient-based} and \emph{gradient-free} optimization.
Another subtle difference that separates the two is that the goal of VML (like classical machine learning) is to learn a model to recognize a generalizable pattern in a given training set, while the goal of current prompt optimization (like classical optimization) is more general, which is to simply optimize an objective function (without the need to learn the underlying pattern, \eg, \citep{pryzant2023automatic}).
We can phrase the goal of a machine learning problem into an optimization objective, and use the tools from optimization to solve it, but the focus of the two areas are fundamentally different.
We compare the difference of the two in the experiments (see \Cref{sect:ape} and \Cref{app:ape_details}).
We observe that current prompt optimization often results in a generic instruction rather than a description of the data pattern.

\textbf{LLMs for multi-agent systems}. Due to the strong instruction-following ability, LLMs are capable of playing different roles in a multi-agent systems. \citep{qian2023communicative,wu2023autogen,hong2023metagpt,li2024camel} study a multi-agent collaboration system for solving complex tasks like software development. VML can also be viewed as a two-agent system where one LLM plays the learner role and the other LLM plays the optimizer role.

\section{Verbalized Machine Learning}
\vspace{-1mm}
\subsection{From Numerical to Verbalized Machine Learning} \label{sect:numerical2vml}
\vspace{-1mm}

\begin{wrapfigure}{r}{0.29\linewidth}
\vspace{-6.5em}
\centering
\includegraphics[width=0.99\linewidth]{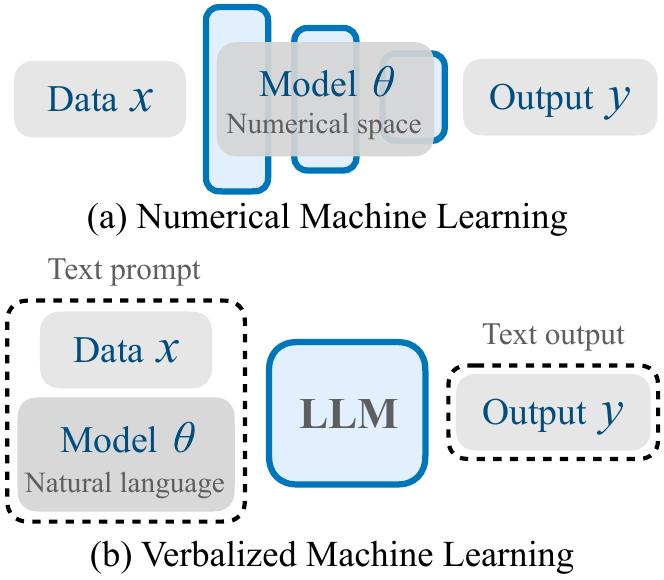}
  \vspace{-1.7em}
	\caption{\scriptsize A comparison between numerical machine learning and VML.}
	\label{fig:nml_vml}
\vspace{-0.3em}
\end{wrapfigure}

Classical machine learning models (\eg, neural networks) are typically trained within a continuous numerical parameter space. Once trained, these models are stored as a collection of numerical values that are not interpretable and remain a black box. Motivated by the strong universal problem-solving capability of LLMs, we find it appealing to view an LLM as a function approximator that is parameterized by its own text prompt. This perspective leads to our VML framework. Similar to a general-purpose modern computer whose functionality is defined by its running program, a function that is defined by an LLM is characterized by its text prompt. Due to the fully human-interpretable text prompt, the VML framework provides strong interpretability for its learned function and is also easy to trace the cause of model failure. Figure~\ref{fig:nml_vml} gives a comparison between numerical machine learning and VML. In the proposed VML framework, both data and model are represented in a unified token-based format, while numerical machine learning treats data and model differently.

\vspace{-1mm}
\subsection{Natural Language as the Model Parameter Space}
\vspace{-1mm}

VML parameterizes a machine-learning model with natural language. More formally, VML places a strong constraint on the model parameters $\bm{\theta}=\{\theta_1,\theta_2,\cdots,\theta_t\}\in \Theta_{\text{language}}$ to exchange for interpretability, where $\bm{\theta}$ is a text token sequence, $\theta_t\in\mathcal{A},\forall t$ is some text token from a large token set $\mathcal{A}$, and $\Theta_{\text{language}}$ denotes the set of all natural language sequences that humans can understand. The model parameter space in VML has the following properties: (1) discrete: the natural language space is discrete; (2) sequential: the natural language space is sequential, and the next word is dependent on its previous words. In contrast, the parameter space in numerical machine learning is not sequentially dependent; and (3) human-interpretable: the natural language that characterizes the model is human-interpretable. More discussions are given in Appendix~\ref{app:model_parameters}.

One of the most significant advantages to use natural language as the model parameters is the easy incorporation of our prior knowledge about the problem and the desired inductive bias into the model training. When the model parameters get updated during training, the model is fully interpretable, and one can observe and understand what gets added and what gets modified. Our empirical evidences also supports our interpretability claim, as we find that the model parameters $\bm{\theta}$ are typically a language description of the underlying pattern that the model discovers from the training data.

\begin{figure}[t]
    \centering
    \setlength{\abovecaptionskip}{4pt}
    \setlength{\belowcaptionskip}{-5pt}
    \renewcommand{\captionlabelfont}{\scriptsize}
    \includegraphics[width=1\textwidth]{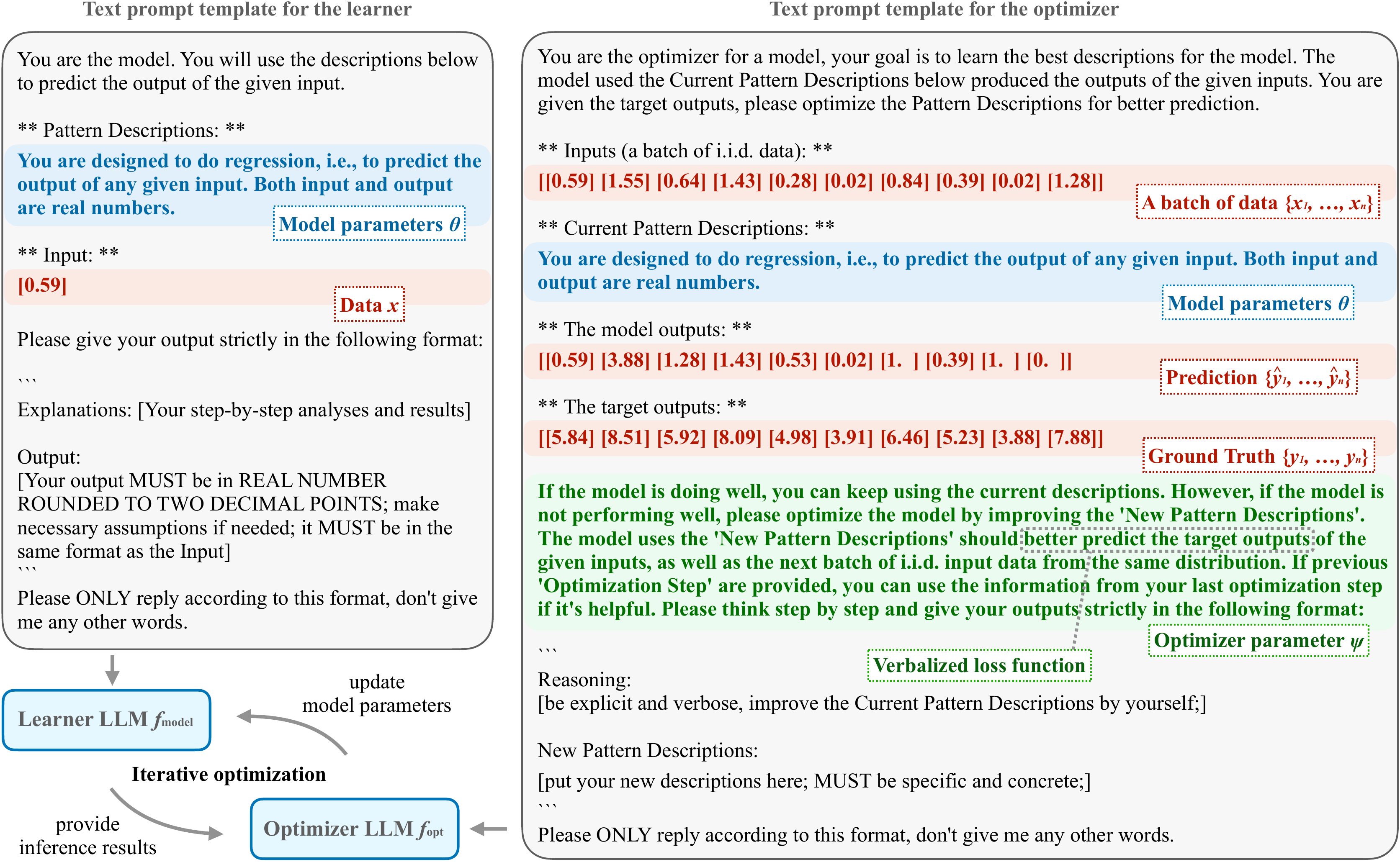}
    \caption{\scriptsize An overview of iterative optimization and text prompt templates of the learner and the optimizer in the regression example.}
    \label{fig:template}
\end{figure}

\vspace{-1mm}
\subsection{Language Models as Function Approximators}
\vspace{-1mm}

The core idea behind VML is using a pretrained language model to act as a function approximator, parameterized by its natural language prompt. Specifically, we denote the language model as $f(\bm{x};\bm{\theta})$ where $\bm{x}$ is the input data and $\bm{\theta}$ is the function parameter. 
$\bm{x}$ can be represented as text tokens (or other format such as images if the LLM supports vision input), and the model parameters $\bm{\theta}$ is also represented with text tokens. 
In VML, $f(\cdot)$ is typically a frozen large language model that is pretrained on a large corpus of text (\eg, DeepSeek-V3~\cite{liu2024deepseek}, Llama-3~\citep{touvron2023llama}, GPT-4~\cite{achiam2023gpt}). If we consider a static function, we can set the temperature parameter of the LLM as zero, which theoretically makes the output deterministic. If we set the temperature high (see Appendix~\ref{app:prob_view} for more discussion), $f(\bm{x};\bm{\theta})$ can be viewed as performing sampling from some distribution. We revisit how a classical machine learning problem can be formulated in the VML framework. Suppose we have in total $N$ training data points $\{\bm{x}_n,y_n\}_{n=1}^N$, where $\bm{x}_n$ is the input feature vector and $y_n$ is the target output value. As an illustrative example, we consider the following least square regression problem using the 
function $f_{\text{model}}(\bm{x};\bm{\theta})$ that is parameterized by natural language description $\bm{\theta}$:

\vspace{-4.25mm}
\begin{equation}\label{eq:reg}
    \min_{\bm{\theta}}\ell_{\text{regression}}:=\frac{1}{2N}\sum_{n=1}^N \big{(} y_n-f_{\text{model}}(\bm{x}_n;\bm{\theta})\big{)}^2, \quad~\text{s.t.}~~\bm{\theta}\in\Theta_{\text{language}}
\end{equation}
\vspace{-4.25mm}

where minimizing the objective function with respect to the discrete token-based model parameters~$\bm{\theta}$ is actually quite difficult. Back-propagating gradients through discrete variables (\eg, policy gradients, Gumbel-softmax~\citep{jang2016categorical}) is typically known to be sample-inefficient and sub-optimal.

\vspace{-1mm}
\subsection{Iterative Training by Prompt Optimization}\label{sect:opt_main}
\vspace{-1mm}

Because the model parameters $\bm{\theta}$ in VML are text prompts, optimizing $\bm{\theta}$ is effectively a prompt optimization problem. Different from current prompt optimization~\citep{zhou2022large}, where the goal is to produce a generic prompt without adding new information, the training in VML focuses on updating the model's language characterization, which involves both the addition of new prior information and the modification of existing information. To optimize the model parameters, we start with the gradient of the regression objective in Equation~\ref{eq:reg}:

\vspace{-4.25mm}
\begin{equation}
\nabla_{\bm{\theta}}\ell_{\text{regression}}=\frac{1}{N}\sum_{i=1}^N\big(y_n-f_\text{model}(\bm{x}_n;\bm{\theta})\big)\cdot\frac{\partial f_\text{model}(\bm{x}_n;\bm{\theta})}{\partial\bm{\theta}},\quad~\text{s.t.}~~\bm{\theta}-\eta\cdot\nabla_{\bm{\theta}}\ell_{\text{regression}}\in\Theta_{\text{language}}
\end{equation}
\vspace{-4.25mm}

where $\eta$ is the learning rate, and the constraint is to ensure that the updated model parameters are still in the natural language space. It seems to be infeasible to compute this gradient. To address this, we view the gradient as a function of the data $(\bm{x},y)$ and the current model parameters $\bm{\theta}$. Then we directly approximate the next-step model parameters using another pretrained language model denoted by $f_{\text{opt}}(\bm{x},\hat{y},y,\bm{\theta};\bm{\psi})$ where $\hat{y}$ is the model prediction from the learner $f_{\text{model}}$. $\bm{\psi}$ denotes the optimizer parameters that characterizes the optimizer settings, and we can use language to specify the update speed, the momemtum, \emph{etc}. The largest possible batch size of the optimizer LLM is determined by its context window. The optimizer LLM can already output natural language that satisfies the constraint, so we simply ask the LLM to play the optimizer role, which has been shown effective in \citep{yang2023large}. More importantly, our VML framework gets better as LLM's instruction-following ability gets stronger. An overview of the iterative optimization and the prompt templates in the regression example are given in Figure~\ref{fig:template}. The training procedure is given in Algorithm~\ref{alg:vml}.

\vspace{-0.5mm}

\setlength{\columnsep}{9pt}
\begin{wrapfigure}{r}{0.43\textwidth}
\begin{center}
\begin{minipage}{1\linewidth}
\vspace{-0.35in}
\captionsetup[algorithm]{font=footnotesize}
\begin{algorithm}[H]
\footnotesize
\caption{\footnotesize Training in VML}\label{alg:vml}
Initialize model parameters $\bm{\theta}_0$, iteration number $T$, batch size $M$ and optimizer parameters $\bm{\psi}$;\;

\For{$i=1,\cdots,T$}{
    Sample $M$ training examples $\bm{x}_1,\cdots,\bm{x}_M$;\;
    
    \For{$m=1,2,\cdots,M$}{
    $\hat{y}_m=f_{\text{model}}(\bm{x}_m;\bm{\theta}_{i-1})$;\;
    
    \vspace{-0.3mm}
    
    }
    
    $\bm{\theta}_{i}\!\!=\!\!f_{\text{opt}}\big(\{\bm{x}_m,\hat{y}_m,y_m\}_{m=1}^M,\bm{\theta}_{i-1};\bm{\psi}\big)$;\;
    \vspace{-0.25mm}
    
}
\vspace{-0.8mm}
\end{algorithm}
\vspace{-0.26in}
\end{minipage}
\end{center}
\end{wrapfigure}
Using an LLM as the optimizer offers several unique advantages. First, the optimizer can perform automatic model selection. When  the learner model can not make correct predictions for the training data, the optimizer will automatically update the learner to a more complex and capable model (see the polynomial regression experiments in Section~\ref{sect:poly} as an example). Second, the optimizer can provide detailed explanations of why a particular update should be performed, which helps us to understand the inner working mechanism of the verbalized optimization process. Third, the LLM-parameterized optimizer allows users to interact with it. This not only helps us to easily trace model failures, but more importantly, it also allows us to inject prior knowledge to improve optimization. 

\vspace{-0.5mm}

\textbf{Different optimizer parameterizations}. In this paper, we use a \emph{direct parameterization}, \ie, parameterizing the optimizer as a single function $f_{\text{opt}}$, which couples the gradient and the update functions together. 
Alternatively, we can use an \emph{indirect parameterization} where the gradient and the update are two separate LLM-parameterized functions. The gradients are known as ``textual gradients'' in prompt optimization~\citep{pryzant2023automatic, yuksekgonul2024textgrad}. The update of learner's model parameter is given by $\bm{\theta}_i=f_{\text{update}}(\bm{\theta}_{i-1}.\frac{\partial\ell}{\partial \bm{\theta}})$, where $\frac{\partial\ell}{\partial \bm{\theta}}$ is computed by  $f_{\text{grad}}(\frac{\partial\ell}{\partial\hat{\bm{y}}},\bm{\theta}_{i-1})$ and similarly, $\frac{\partial\ell}{\partial\hat{\bm{y}}}$ is computed by $f_{\text{grad}}(\ell,\hat{\bm{y}})$. Both $f_{\text{update}}$ and $f_{\text{grad}}$ are parameterized by LLMs. Compared to direct parameterization that only takes one LLM call, this process has to use several LLM calls. After empirically comparing both methods in Section~\ref{sect:ablation} and \Cref{app:opt_ablation}, we find that in most scenarios, the direct parameterization yields better performance.

\vspace{-.5mm}
\subsection{Discussions and Insights}
\vspace{-.5mm}

\textbf{VML as a framework to encode inductive bias}. A unified framework to encode arbitrary inductive bias has been pursued for decades. For different types of data, we need to design different models to encode the inductive bias (\eg, graphical models~\citep{koller2009probabilistic} for random variables, recurrent networks~\citep{hochreiter1997long} for sequences, graph networks~\citep{kipf2016semi} for graphs, and convolution networks~\citep{lecun1998gradient} for images). VML uses a unified natural language portal to take in inductive biases, making it very flexible for encoding complex inductive bias. To incorporate an inductive bias about the hypothesis class or prior knowledge about the problem, we can simply concatenate a system prompt $\bm{\theta}_{\text{prior}}$ (\ie, some constant prefixed text that describes the inductive bias) with the model parameters $\bm{\theta}$. The final model parameters are $(\bm{\theta}_{\text{prior}},\bm{\theta})$ where $\bm{\theta}$ is learnable and $\bm{\theta}_{\text{prior}}$ is given by users.

\vspace{-0.5mm}

\textbf{VML enables interpretable knowledge discovery}. Because the model parameters $\bm{\theta}$ are already in natural language, it is easy to understand the underlying pattern that leads to the prediction and the decision rules that the model uses. Unlike numerical machine learning where the knowledge is learned within a black box, this property enables VML to discover novel knowledge that humans can also learn from.

\vspace{-0.5mm}

\textbf{VML as ``the von Neumann architecture'' in machine learning}. Machine learning usually treats the model parameters and the data differently, similar to the Harvard architecture that stores instruction and data separately. VML stores both data and model parameters in the text prompt as tokens, which resembles the von Neumann architecture that stores instruction and data in the same memory.

\begin{figure}[t]
    \centering
    \setlength{\abovecaptionskip}{4pt}
    \setlength{\belowcaptionskip}{-7pt}
    \renewcommand{\captionlabelfont}{\scriptsize}
    \includegraphics[width=\linewidth]{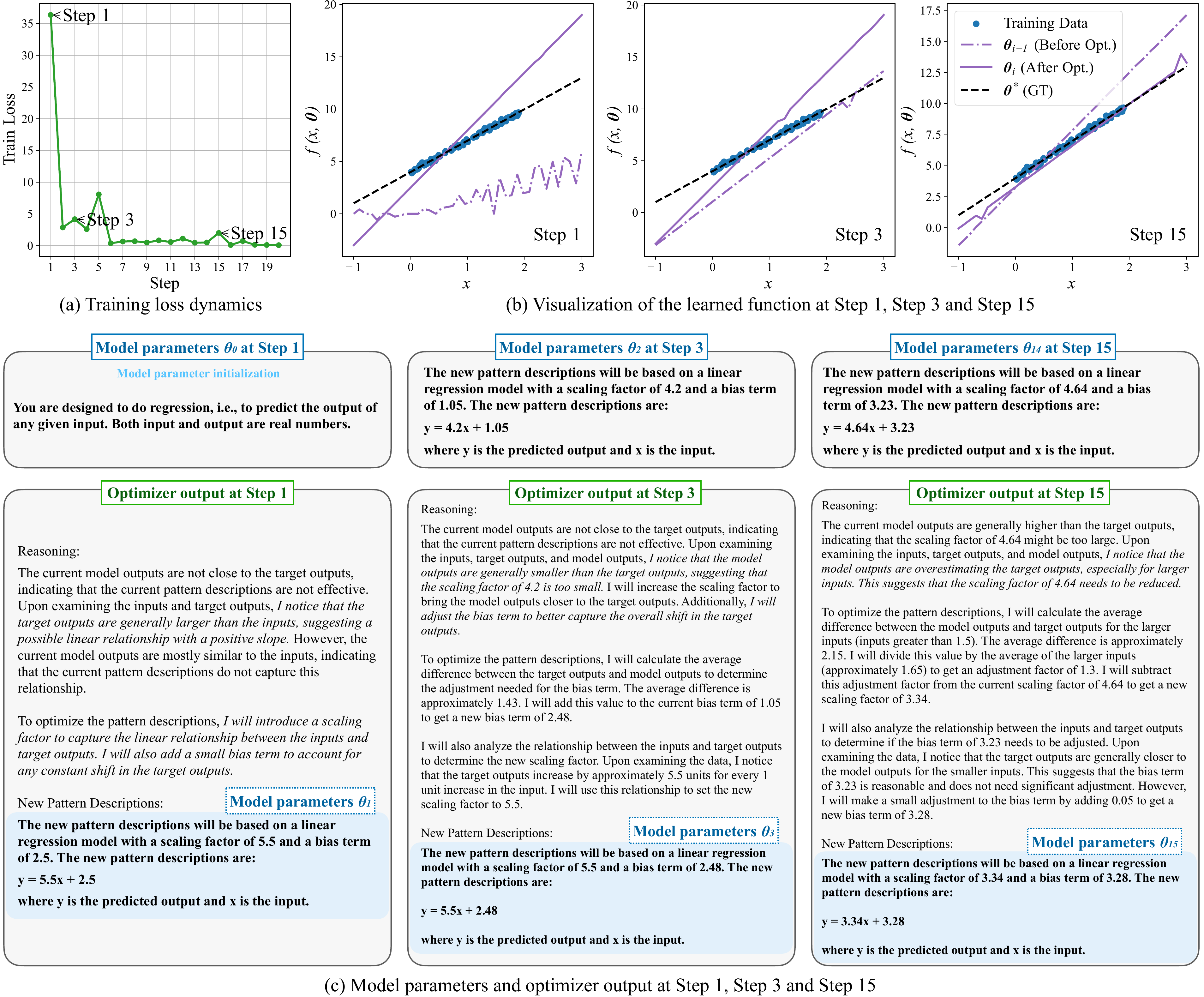}
    \caption{\scriptsize Training dynamics for VML based linear regression. The model is trained for 2 epochs, each with 10 steps.}
    \label{fig:reg:linear}
\end{figure}

\vspace{-.5mm}
\section{Applications and Case Studies}\label{sect:exp}
\vspace{-.5mm}

We demonstrate the features and advantages of VML by revisiting some classical machine learning tasks followed by a realistic medical image classification task.
In these tasks, we are given data $\mathcal{D}_\text{train} = \{\vx_n, y_n \}_{n=1}^N$, and we want to find $\bm{\theta}^*$ such that $f_\text{model}(\vx; \bm{\theta}^*)$ best describes the mapping $\vx \to y$.
Our experiments below show in detail how VML is able to solve these tasks and find $\bm{\theta}^*$.

\textbf{Experiment setups}. We use the instruction-tuned Llama-3 70B~\citep{touvron2023llama} for the LLM unless specified otherwise.
The training set for each task consists of $100$ data points.
For all tasks, we use a batch size of $10$ for each optimization step (see \figref{fig:template} (right) as an example), which corresponds to $10$ steps per training epoch.
To evaluate regression performance, we look at the training loss, and the model predictions in both interpolation and extrapolation settings.
For classifications, we use additional test sets ($20$ data points), and evaluate both training and testing accuracies.
Inspired by the momentum from classical optimization, we provide the last step (\ie, one step only) of the optimization history to the optimizer LLM for training stability. 

\textbf{Training logs}. The results of our experiments are showed using: 
(a) training loss, which is computed by \emph{parsing} the model output (string) and \emph{converting} it in to the same data type as the target value ($y$), then we use mean squared error for regression, and zero-one loss mean (\ie, average accuracy) for classification. 
The computed training loss is for logging purpose only, it is not required for training in VML (see \Algref{alg:vml}).; 
(b) visualization of the learned model, which is also done through \emph{parsing} and \emph{converting} the model output;
(c) the model parameter at each training step $i$ before optimization (\ie, $\theta_{i-1}$), and the optimizer output for the updated $\theta_{i}$.
For $i>1$, the full model parameter before optimization is $\bm{\theta}_{i-1}= \{\theta_0,\theta_{i-1}\}$, but in our figures below we only show the $\theta_{i-1}$ to save space.

\textbf{Compute}. The LLM is ran on a node of $8 \times \text{A100}$ using the inference engine provided by vLLM~\citep{kwon2023efficient}.
During each step ($i$) of training, we query the LLM $10$ times for evaluating the model $f_\text{model}(\vx; \bm{\theta}_{i-1})$ over a batch, and $1$ time for requesting the newly optimized $\bm{\theta}_i$.
We also evaluate the entire test set at each step, which, depending on the size of the evaluation set, requires between $20$ to $100$ LLM queries.
Overall, for the regression tasks, they take around $10$ minutes for each epoch of training.
The classification tasks, take around $16$ minutes for each epoch of training.
The visualization of the decision boundary takes around $6$-minute.

\begin{figure}[t]
    \centering
    \setlength{\abovecaptionskip}{3pt}
    \setlength{\belowcaptionskip}{-5pt}
    \renewcommand{\captionlabelfont}{\scriptsize}
    \includegraphics[width=\linewidth]{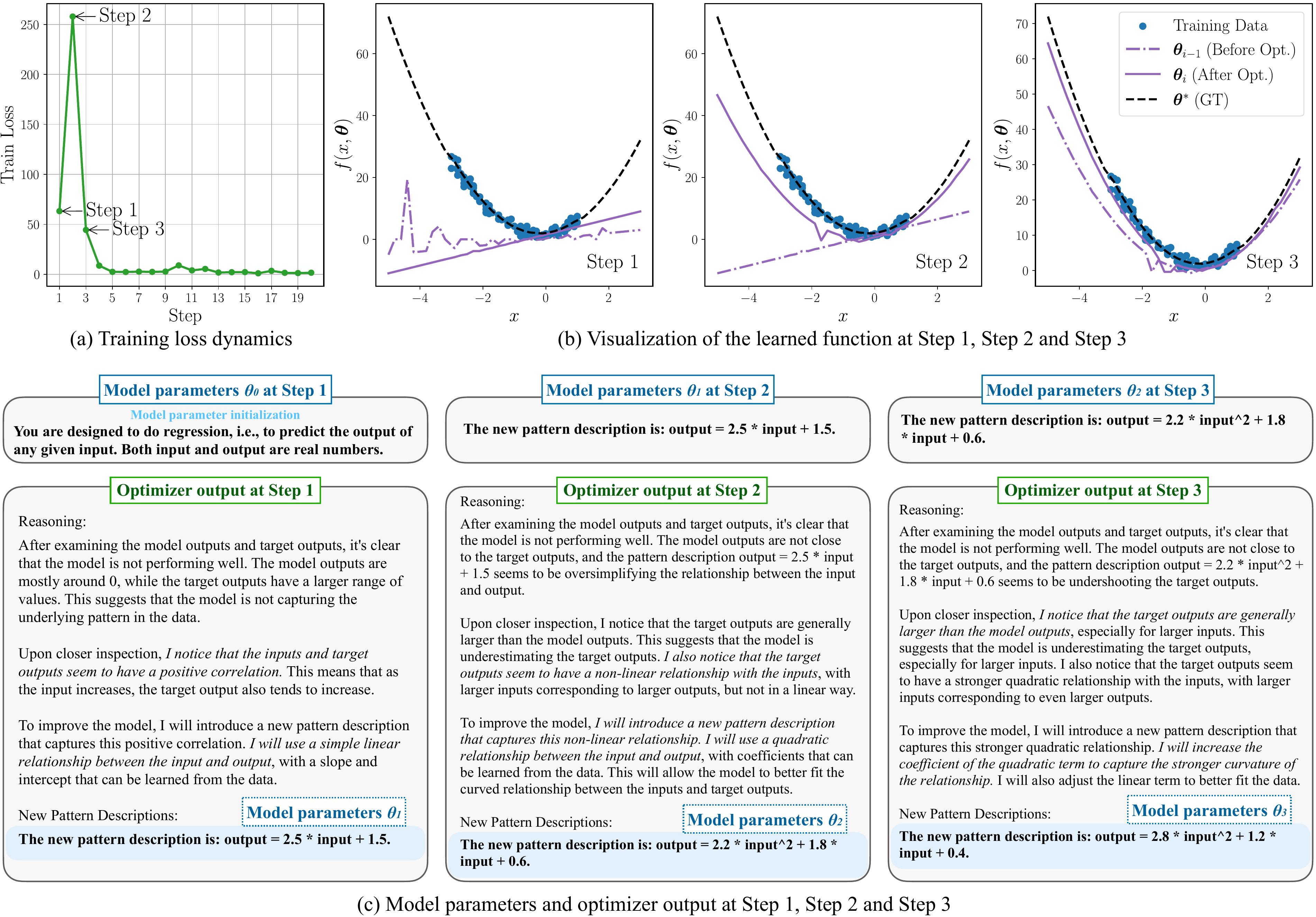}
    \caption{\scriptsize Training dynamic for VML based polynomial regression. The model is trained for 2 epochs, each with 10 steps.}
    \label{fig:reg:poly}
\end{figure}

\vspace{-1mm}
\subsection{Linear Regression} \label{sect:linear}
\vspace{-1mm}

We generate $\dtrain$ from a linear function with Gaussian noise, \ie, $y = 3x + 4 +\epsilon$, where $\epsilon \sim \mathcal{N}(0, 1)$ and $x \sim \mathcal{U}(0,2)$.
We initialize the model parameter $\bm{\theta}_0$ by \emph{only} specifying that the task is a regression task from $\mathbb{R}$ to $\mathbb{R}$ (see \figref{fig:reg:linear}(c) Step 1).
\figref{fig:reg:linear}(a) shows that training improves the model, and that it converges.
The subplots (b) and (c) show details of the model and optimization at steps 1, 3 and 15.
At step 1, since $\bm{\theta}_0$ only contain the definition of 1-D regression task, the \modeli{0} is randomly guessing (see the dashdot line).
The \emph{\opti{1} says that} it notices a linear relationship between the input and the target outputs, hence introducing a linear regression model to capture such a relationship, which results in \modeli{1} being a straight line.
From step 2 onward, the optimization focus switches to fitting the identified linear regression model to the data.
For example, at step~3, we can see that \opti{3} \emph{says} it notices that the outputs of \modeli{2} are generally smaller than the target, suggesting the scaling factor is too small, hence it increases it.
Similarly, at step 15, \opti{15} also \emph{says} it notices the \modeli{14} overestimates the target; hence, it reduces the scaling factor.
We can see from (b) that the resulting \modeli{15} closely approximates the ground truth.

\vspace{-1mm}
\subsection{Polynomial Regression}\label{sect:poly}
\vspace{-1mm}

We generate $\dtrain$ from a polynomial function with Gaussian noise, \ie, $y = 3x^2 + x + 2 +\epsilon$, where $\epsilon \sim \mathcal{N}(0, 1)$ and $x \sim \mathcal{U}(-3,1)$.
Similarly, $\bm{\theta}_0$ is initialized by \emph{only} specifying that the task is a regression task from $\mathbb{R}$ to $\mathbb{R}$ (see \figref{fig:reg:poly}(c) Step 1).
\figref{fig:reg:poly}(a) shows that training is effective and converges.
Subplots (b) and (c) show details of the model and optimization at steps 1, 2 and~3.
At step~1, \modeli{0} randomly guesses the outputs.
The \emph{\opti{1} says that} it notices $y$ has a larger range than~$x$, and that they seem to have positive correlation; therefore, it updates \modeli{1} to be a simple linear model.
This linear model assumption leads to a jump in the training loss (see subplot (a)), as it is far from the ground truth.
Consecutively, at step 2, \opti{2} \emph{says} the poor performance makes it realize that the linear model oversimplifies the relationship between $x$ and~$y$.
It notices a non-linearity between $x$ and~$y$, and to capture this, it uses a quadratic model.
This leads to a better model and a large decrease in the training loss.
At step 3, \opti{3} switches from model class selection to fitting the quadratic model.
The resulting \modeli{3} closely fits the ground truth.

\vspace{-1mm}
\subsection{Sinusoidal Regression}\label{sect:sine}
\vspace{-1mm}

\begin{figure}[t]
    \centering
    \setlength{\abovecaptionskip}{3pt}
    \setlength{\belowcaptionskip}{-5pt}
    \renewcommand{\captionlabelfont}{\scriptsize}
    \includegraphics[width=\linewidth]{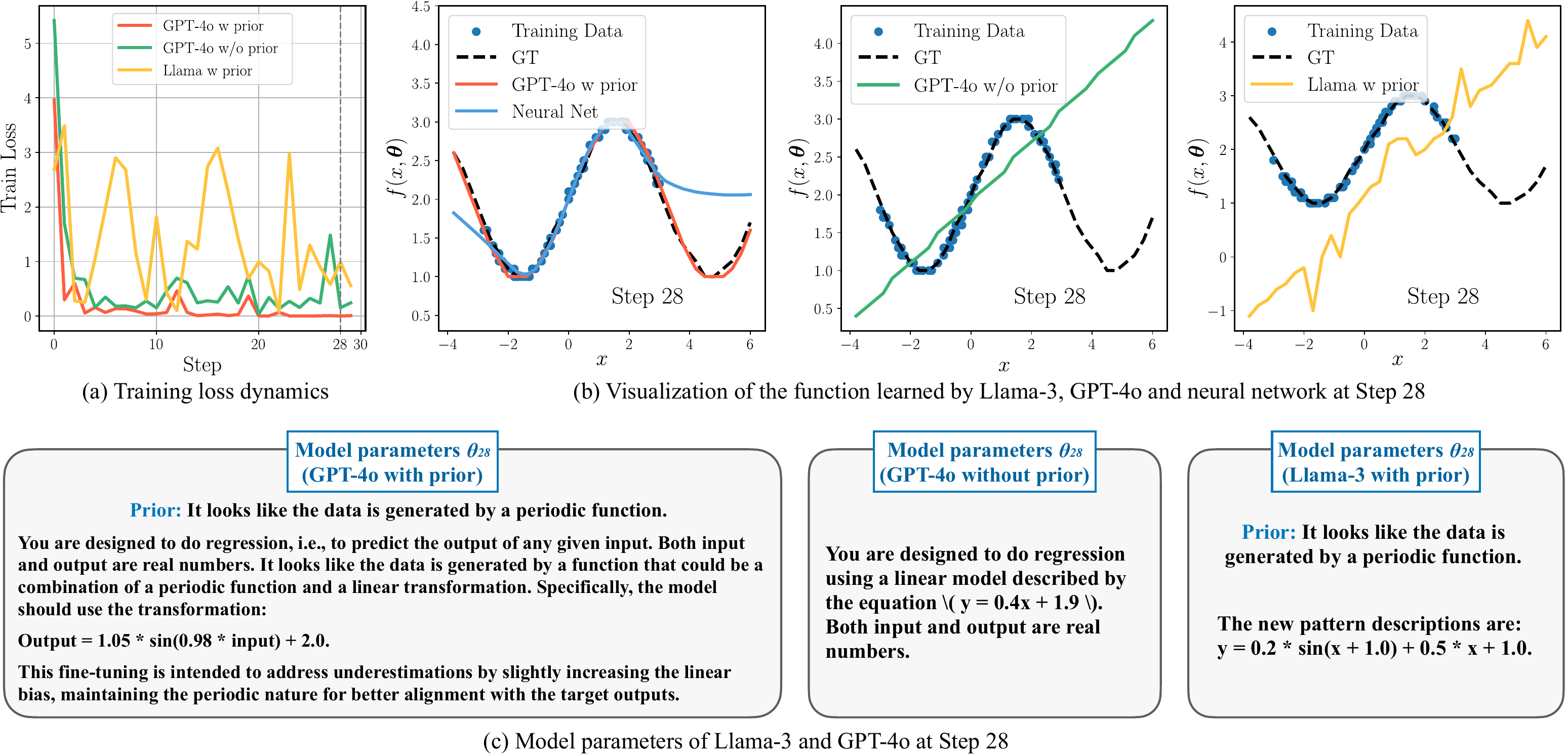}
    \caption{\scriptsize Demonstration of prior injection, and comparison of Llama-3, GPT-4o and a neural net in the sinusoidal regression setting.}
    \label{fig:reg:sin}
\end{figure}

\begin{figure}[t]
    \centering
    \setlength{\abovecaptionskip}{3pt}
    \setlength{\belowcaptionskip}{-5pt}
    \renewcommand{\captionlabelfont}{\scriptsize}
    \includegraphics[width=\linewidth]{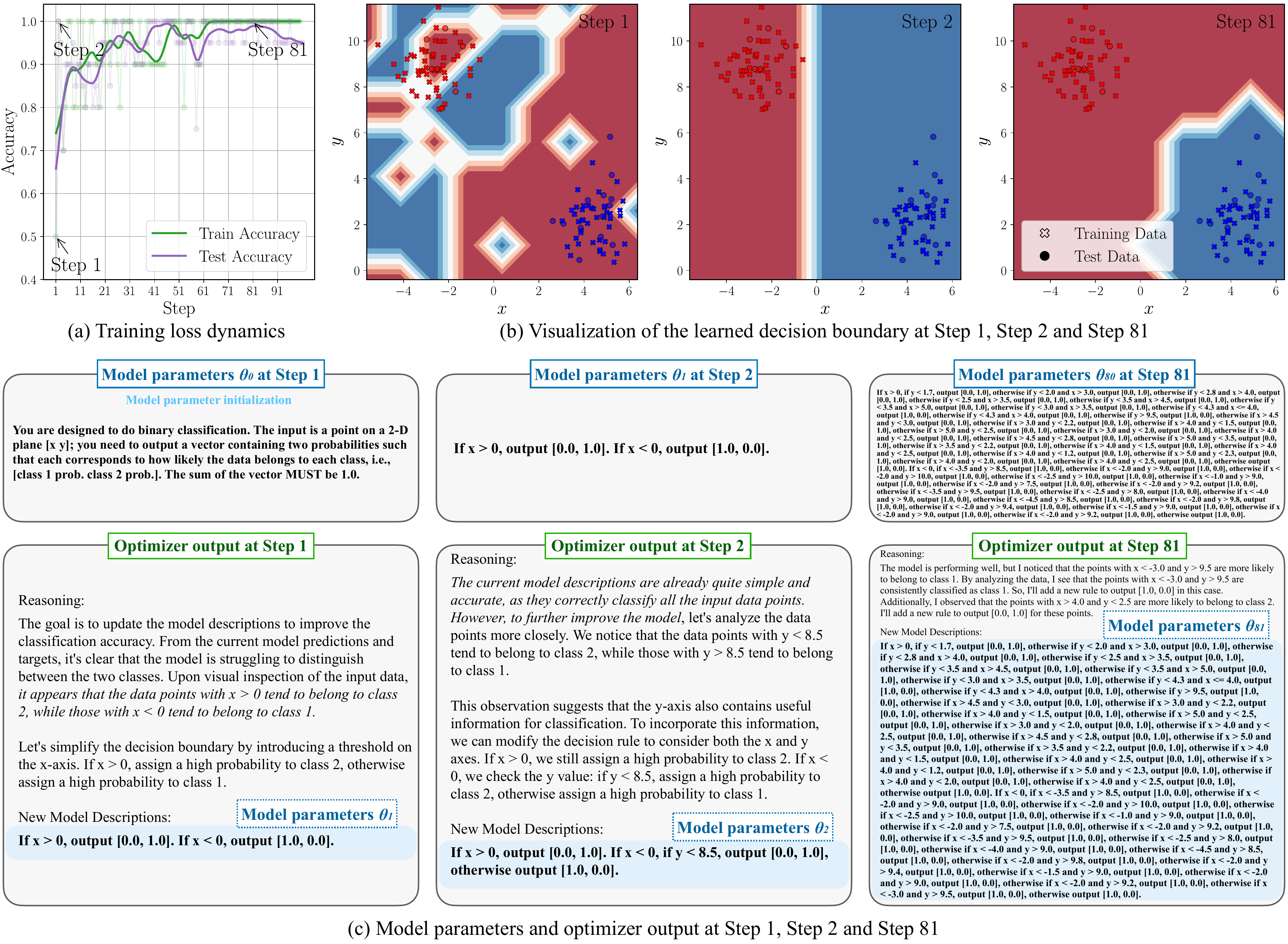}
    \caption{\scriptsize Linearly separable two blobs classification based on VML.
    (b) plots the decision boundary of model with $\bm{\theta}_{i-1}$ at step $i$.}
    \label{fig:cls:blobs}
\end{figure}

We generate $\dtrain$ from a sine function with Gaussian noise, \ie, $y = \sin(x) + 2 + 0.01\epsilon$, where $\epsilon \sim \mathcal{N}(0, 1)$ and $x \sim \mathcal{U}(-3,3)$.
Fitting a sine function is known to be difficult for neural nets in terms of extrapolation.
Here, we try GPT-4o, a more powerful model than Llama-3.
\figref{fig:reg:sin}(b; right) shows that when $\bm{\theta}_0$ contains \emph{only} the definition of 1-D regression, it results in a linear model after training (see (c; right)).
We can \emph{add a prior to $\bm{\theta}$ by simply saying} that the data looks like samples generated from a periodic function, which results in a very good approximation and it extrapolates much better than a neural net (see (b,c; left)).
We also find that adding the same prior to Llama-3 is not as effective (see (b,c; mid)), indicating the capability of VML is highly dependent on the capability of the LLM. However, this suggests VML grows with LLM's scaling law---the effectiveness of VML can improve along with the capability (size) of the LLM.

\vspace{-1mm}
\subsection{Two Blobs Classification}\label{sect:cls_tb}
\vspace{-1mm}

We generate a linearly separable $\dtrain$ from two blobs on a 2-D plane.
$\bm{\theta}_0$ is initialized by \emph{only} specifying that the task is binary classification on a 2-D plane (see \figref{fig:cls:blobs}(c) Step 1).
Subplot~(a) shows that training is effective and it converges.
At step~1, \opti{1} \emph{says} the current batch of data has the pattern that data points with $x > 0$ belong to class 2, and data points with $x < 0$ belong to class 1; hence it updates \modeli{1} to have a linear decision boundary at $x=0$, which happens to be perfect.
However, \figref{fig:cls:blobs}(a) shows that the training loss does not immediately converge.
We can investigate the cause and ``\emph{debug}'' the optimizer by looking at \opti{2}.
From (c) Step 2, \opti{2} \emph{says} \modeli{1} is already quite simple yet accurate, but it wants to further improve the model and utilize the new information from the current batch.
Guided by this reasoning, \modeli{80} becomes a very deep decision tree, and the decision boundary has a reasonable margin towards the data (see \figref{fig:cls:blobs}(b, c; right)). The results also reveal that VML's interpretable may lead to complex pattern for easy problems, highlighting the importance of specifying a proper inductive bias.

\subsection{Two Circles Classification}\label{sect:cls_tc}

\begin{figure}[t]
    \centering
    \setlength{\abovecaptionskip}{3pt}
    \setlength{\belowcaptionskip}{2pt}
    \renewcommand{\captionlabelfont}{\scriptsize}
    \includegraphics[width=\linewidth]{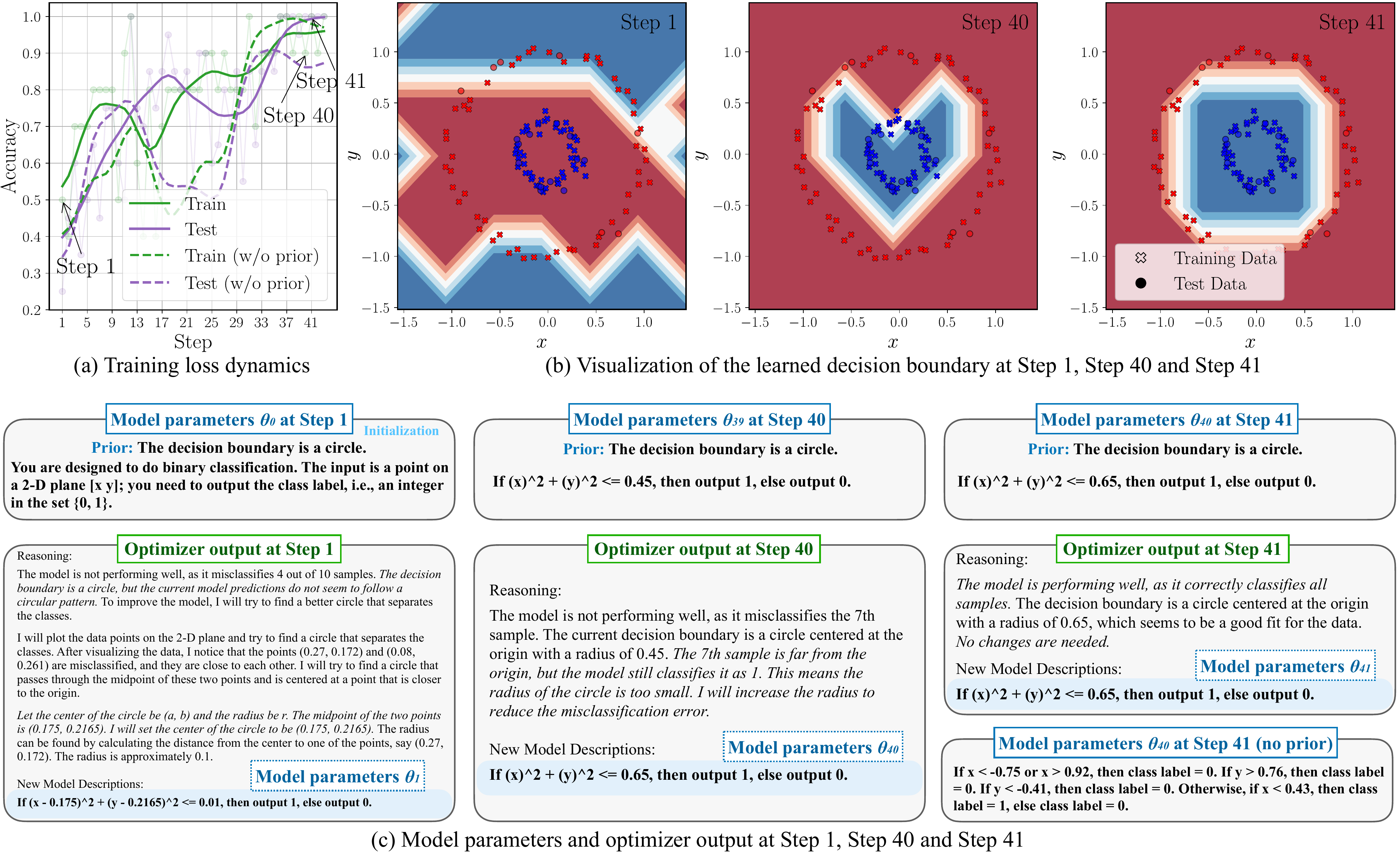}
    \caption{\scriptsize Non-linearly separable 2-circle classification with a prior in $\bm{\theta}$. (a; dashed) and (c; bottom right) show results without the prior.}
    \label{fig:cls:circles_w_prior}
\end{figure}

We generate a non-linearly separable $\dtrain$ by creating data points on two concentric circles as the two classes.
Besides the description of binary classification on a 2-D plane, we add a sentence to encode our inductive bias that the decision boundary is a circle into $\bm{\theta}_0$ (see \figref{fig:cls:circles_w_prior}(c) Step 1).
At step 1, \opti{1} utilizes the prior, and updates \modeli{1} to have a circle decision boundary.
For the rest of the training, the optimizer mainly tries to find a good fit for the radius and the center of the decision boundary.
At step 41, \opti{41} \emph{says} \modeli{40} seems to be a good fit for the data, and no changes are needed. Hence, it uses the same $\bm{\theta}_{40}$ for \modeli{41}.
Without the prior, VML can also learn a good model, but the performance shows large variance at the beginning of training (see \figref{fig:cls:circles_w_prior}(a; dashed)) due to the model selection process similar to \figref{fig:reg:linear}(a).
\figref{fig:cls:circles_w_prior}(c; bottom right) shows the resulting $\bm{\theta}_{40}$ without the prior, which is a decision tree.

\subsection{Medical Image Classification} \label{sect:medical}

\begin{figure}[t]
    \centering
    \setlength{\abovecaptionskip}{3pt}
    \setlength{\belowcaptionskip}{-7pt}
    \renewcommand{\captionlabelfont}{\scriptsize}
    \includegraphics[width=\linewidth]{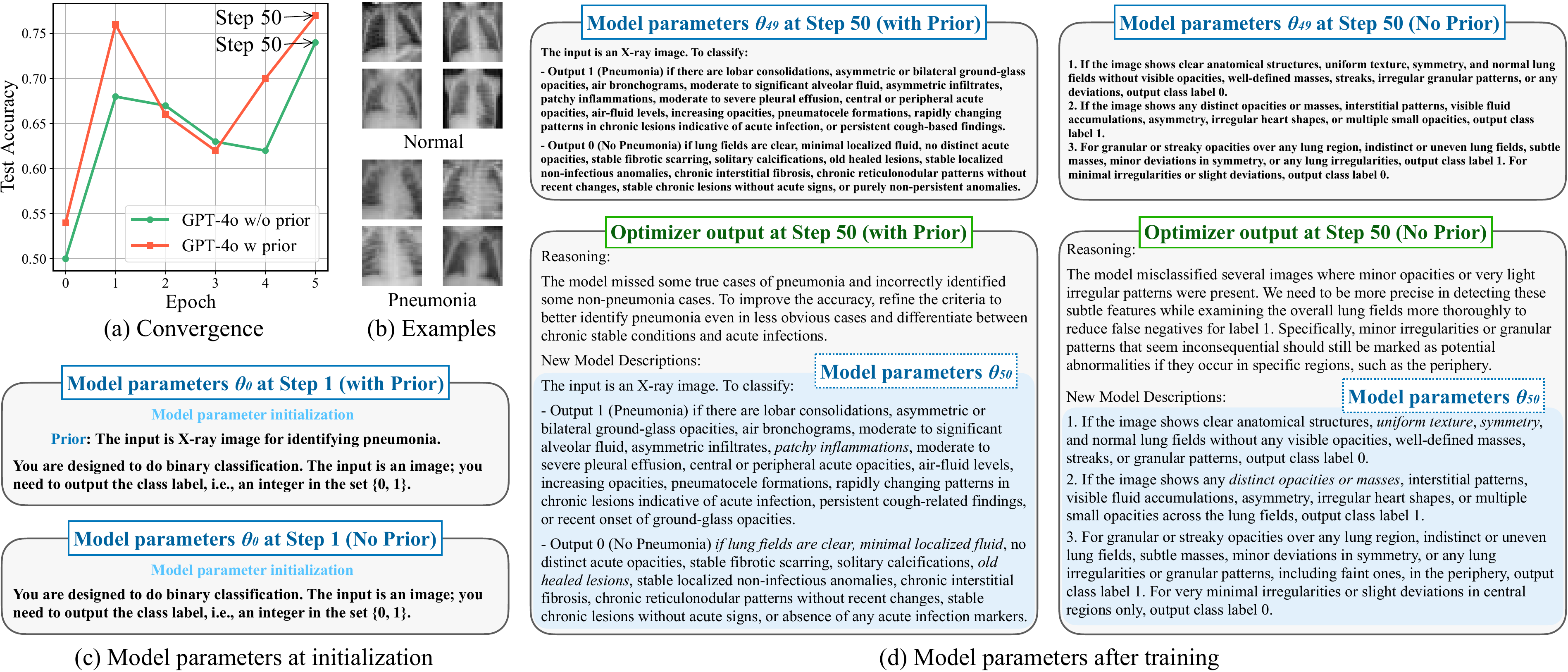}
    \caption{\scriptsize Tiny-PneumoniaMNIST image classification for models with and without prior at initialization.}
    \label{fig:cls:medical}
\end{figure}

To demonstrate the capability of VML beyond simple machine learning problems, we evaluate the effectiveness of VML for image classification. 
We use GPT-4o, which supports visual inputs, to take into account both image and text data.
The task is to classify whether a X-ray image has indications of pneumonia or not, see \Cref{fig:cls:medical}(b) for image examples.
Due to the cost of GPT-4o, we create a subset of the dataset PneumoniaMNIST~\citep{medmnistv2}.
Our dataset consists of 100 training data and 100 test data (half pneumonia and half normal for both sets).
Models are trained for 5 epochs.
We try out two different model parameter initializations, one with prior and one without.
We encode the inductive bias by simply adding a sentence as the prior, which states that the input is an X-ray image for identifying pneumonia, along with the definition of binary image classification (see \Cref{fig:cls:medical}(c)).
The test accuracy in (a) shows that both models are able to improve their performance on the task as the training epoch increases, and the model initialized with prior outperforms the model without (in terms of both testing accuracy and training convergence). 
Additionally, by inspecting the parameters of \modeli{50} (see (d)), we observe that the model parameters $\bm{\theta}_{50}$ for the learner \emph{with prior} has more medical domain knowledge associated to features of pneumonia (such as ``acute infection'', ``pneumatocele formation''), while the model parameters $\bm{\theta}_{50}$ for the learner \emph{without any prior} mainly use generic visual knowledge associated to features of lung (such as ``visible opacities'', ``uniform texture''). This observation validates the effectiveness of using natural language to encode inductive bias. Our experiment also demonstrates the usefulness of learning in VML (\ie, the generalization performance can be improved over time), which is distinct from existing prompt engineering methods. 
Additionally, the interpretable nature of VML's model parameters is crucial for applications in medical domain.
The learned models can be validated by medical professionals, and their predictions are grounded by their verbalized reasoning.

\subsection{Ablation Study and Exploratory Experiments} \label{sect:ablation}

\begin{wraptable}{r}[0cm]{0.42\linewidth}
\renewcommand{\captionlabelfont}{\scriptsize}
 \scriptsize
 \centering
 \setlength{\tabcolsep}{1pt}
 \renewcommand{\arraystretch}{1.4}
 \hspace{-1.4mm}
 \begin{tabular}{lccccc}
 \specialrule{0em}{-15pt}{0pt}
    \multicolumn{1}{l}{Task} & Reg-L($\downarrow$) & Reg-P($\downarrow$) & Cls-TB($\uparrow$) & Cls-TC($\uparrow$) & Cls-MI($\uparrow$)  \\
    \shline
    ICL & 0.38 & 62.96 & {\textbf{100\%}} & {\textbf{95\%}} & 48\%  \\
    \cellcolor{Gray}{VML} & \cellcolor{Gray}{\textbf{0.12}} & \cellcolor{Gray}{\textbf{2.38}} & \cellcolor{Gray}{\textbf{100\%}} & \cellcolor{Gray}{\textbf{95\%}} & \cellcolor{Gray}{\textbf{74\%}} \\
  \specialrule{0em}{-7pt}{0pt}
 \end{tabular}
 \caption{
\scriptsize Comparison between VML and ICL on previous applications (without adding any prior information).
}
\label{tab:icl}
\vspace{0em} %
\end{wraptable}

\textbf{Quantitative comparison to in-context learning}. Since VML can be viewed as a generalization of ICL, we therefore compare VML to ICL in all previous applications. Results are given in Table~\ref{tab:icl}. The ICL results are chosen from the best one across 5 runs. The metrics used for regression (Reg) and classification (Cls) are mean square error (MSE $\downarrow$) and test accuracy ($\uparrow$), respectively. We abbreviate linear regression as Reg-L, polynomial regression as Reg-P, two blob classification as Cls-TB, two circle classification as Cls-TC and medical image classification as Cls-MI. We can observe that VML consistently outperforms ICL.

\begin{wrapfigure}{r}{0.45\linewidth}
\vspace{-2.1em}
\centering
\includegraphics[width=0.99\linewidth]{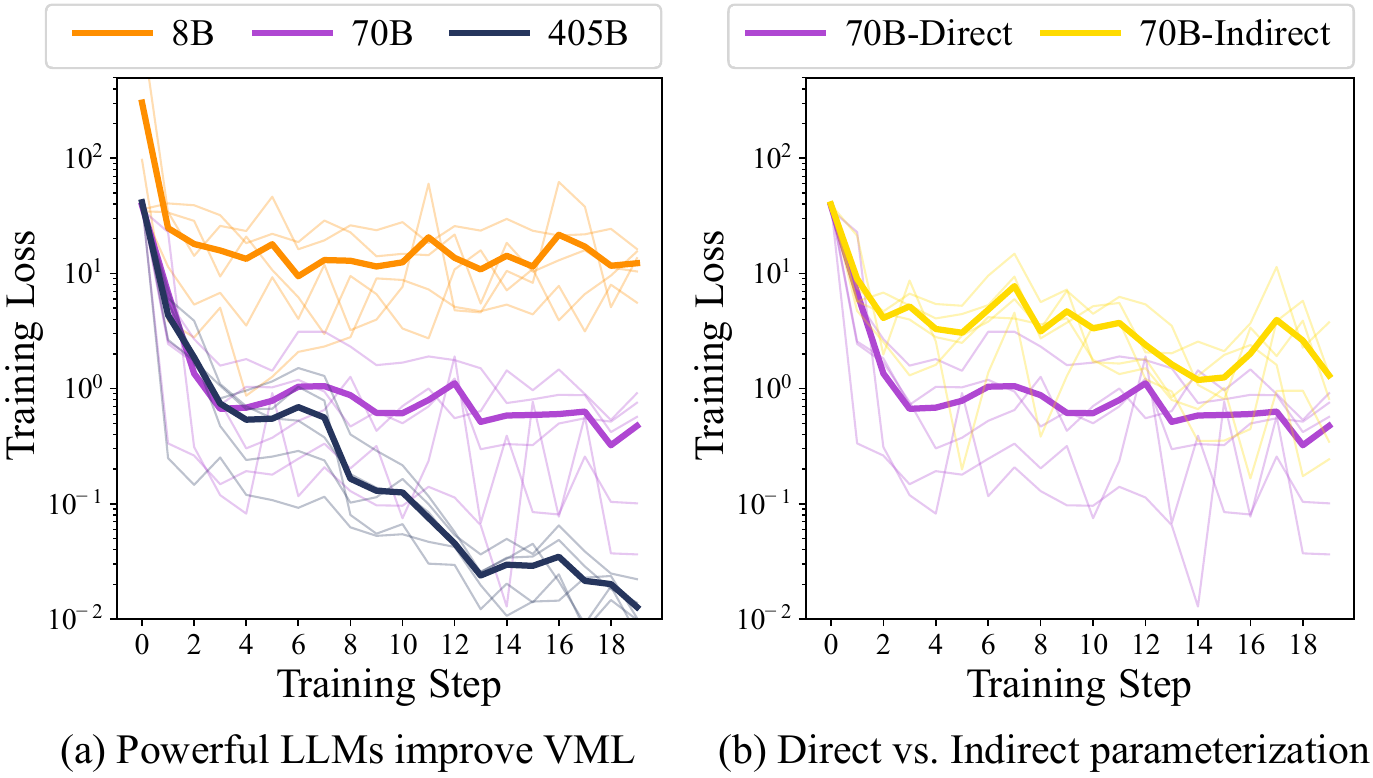}
  \vspace{-1.8em}
	\caption{\scriptsize 
Training loss dynamics. For each configuration, we show 5 individual runs (thin) and their mean (thick).
 }
	\label{fig:scaling_direct}
\vspace{-0.6em}
\end{wrapfigure}

\vspace{-0.3mm}

\textbf{Scaling effect with stronger LLMs}. We are interested in whether the performance of VML can be improved while using a stronger LLM. To evaluate how VML scales with stronger LLMs, we use Llama-3.1 with different size (8B, 70B, 405B) as the backbone LLM for VML. From Figure~\ref{fig:scaling_direct}(a), we can see that stronger LLMs (\eg, 405B) can indeed enable VML to learn faster and achieve lower loss in the linear regression setting. This result shows the great potential of VML when the LLMs get better.

\vspace{-0.3mm}

\textbf{Direct vs. indirect parameterization}. As discussed in Section~\ref{sect:opt_main}, we can use either a direct or an indirect way to parameterize the optimizer. We compare both parameterizations using the linear regression setting (as described in Section~\ref{sect:linear}). Figure~\ref{fig:scaling_direct}(b) shows that the direct parameterization outperforms the indirect one. 
The direct parameterization leads to faster convergence while requiring less LLM calls. Detailed experimental settings and more discussions are given in \Cref{app:opt_ablation}.

\subsection{Comparison Between Generic Prompt Optimization and VML}
\label{sect:ape}

To better differentiate VML from prompt optimization, we compare VML to a generic prompt optimization method called Automatic Prompt Engineer (APE)~\citep{zhou2022large} both conceptually, and qualitatively on two tasks.
We use Llama-3-70B for both APE and VML.
Even though both methods aim to optimize a prompt towards a pre-defined objective function, there are
fundamental differences between the two. 
Conceptually, APE first samples a set of candidate prompts directly given only the training data, and then chooses the one with the best objective value. The prompt generation process is a best-of-N sampling, and each sampling is independent from another.
In contrast, VML produces new prompt by asking an LLM to explicitly reflect on the old prompt and the corresponding predictions $\hat{\bm{y}}$, then reasons on how to produce a prompt that can better predict the target $\bm{y}$ for the given input.
This is conceptually similar to the distinction between gradient-free and gradient-based optimization.
For more discussion and the implementation details of APE, please refer to \Cref{app:ape_details}.
Note that the qualitative comparisons in this section do not imply superiority between the two methods.

\begin{wrapfigure}{r}{0.71\linewidth}
\vspace{-2.3em}
\centering
\includegraphics[width=0.98\linewidth]{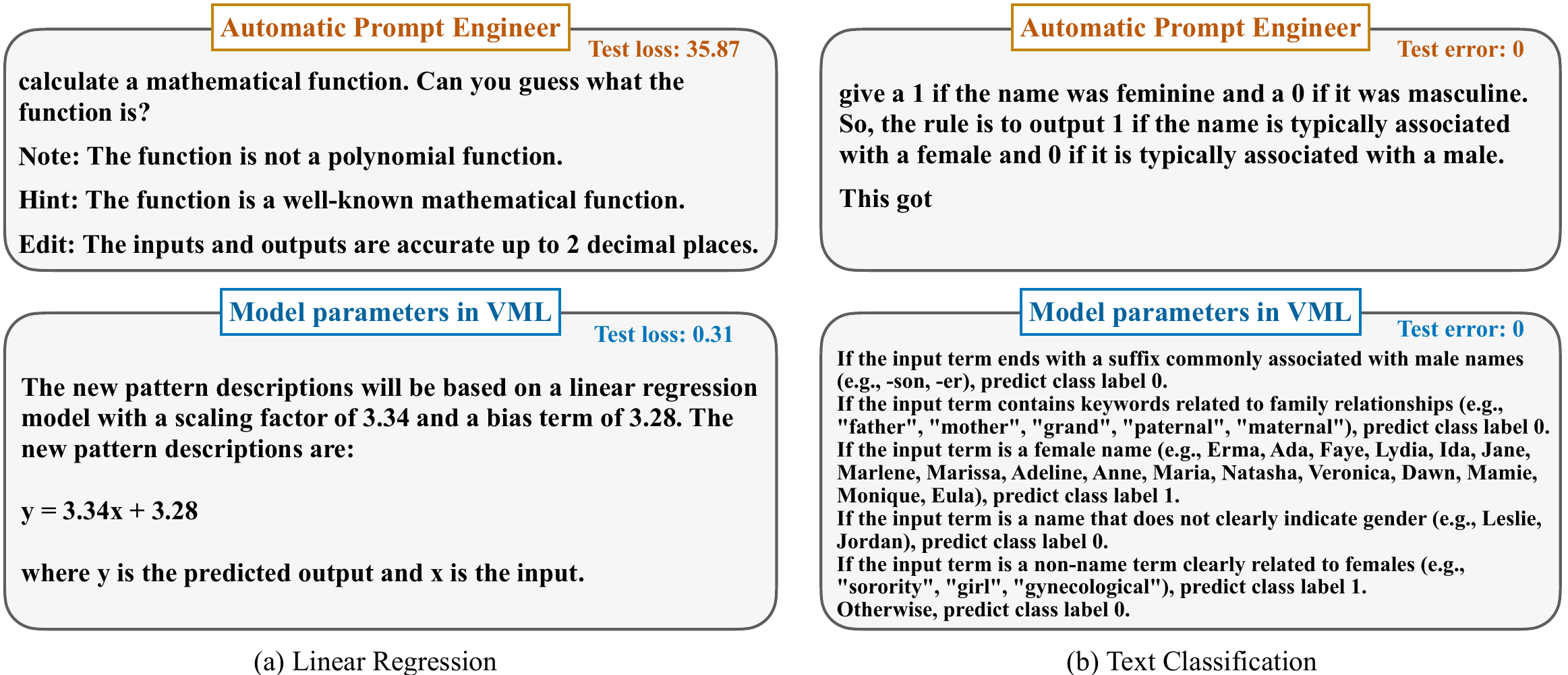}
  \vspace{-0.7em}
	\caption{\scriptsize VML versus a prompt optimization method (Automatic Prompt Engineer~\citep{zhou2022large}).
    }
	\label{fig:ape_vml}
\vspace{-1.2em}
\end{wrapfigure}

\textbf{Linear regression} as in \Cref{sect:linear}.
\figref{fig:ape_vml}(a) shows that the result from APE is vague and general.
Such a description can easily be derived by humans through visual inspection of the data, and it does not learn deeper insights from the data, whereas VML is able to learn useful new information that is difficult to obtain by visual inspection.
We can observe that VML automatically performs effective pattern summarization from data, which differs from naive prompt optimization.

\textbf{Text classification}. Adopted from the Google BIG-bench\citep{srivastava2023beyond}, the task is to classify whether a name is more likely to be associated to female or male.
\figref{fig:ape_vml}(b) shows that APE does return a correct description of the task, but it is, once again, very general.
Conversely, VML is able to learn more detailed knowledge about the data pattern which cannot be done easily through visual inspection. 

\begin{figure}[h!]
    \centering
    \setlength{\abovecaptionskip}{3pt}
    \setlength{\belowcaptionskip}{-5pt}
    \includegraphics[width=.99\linewidth]{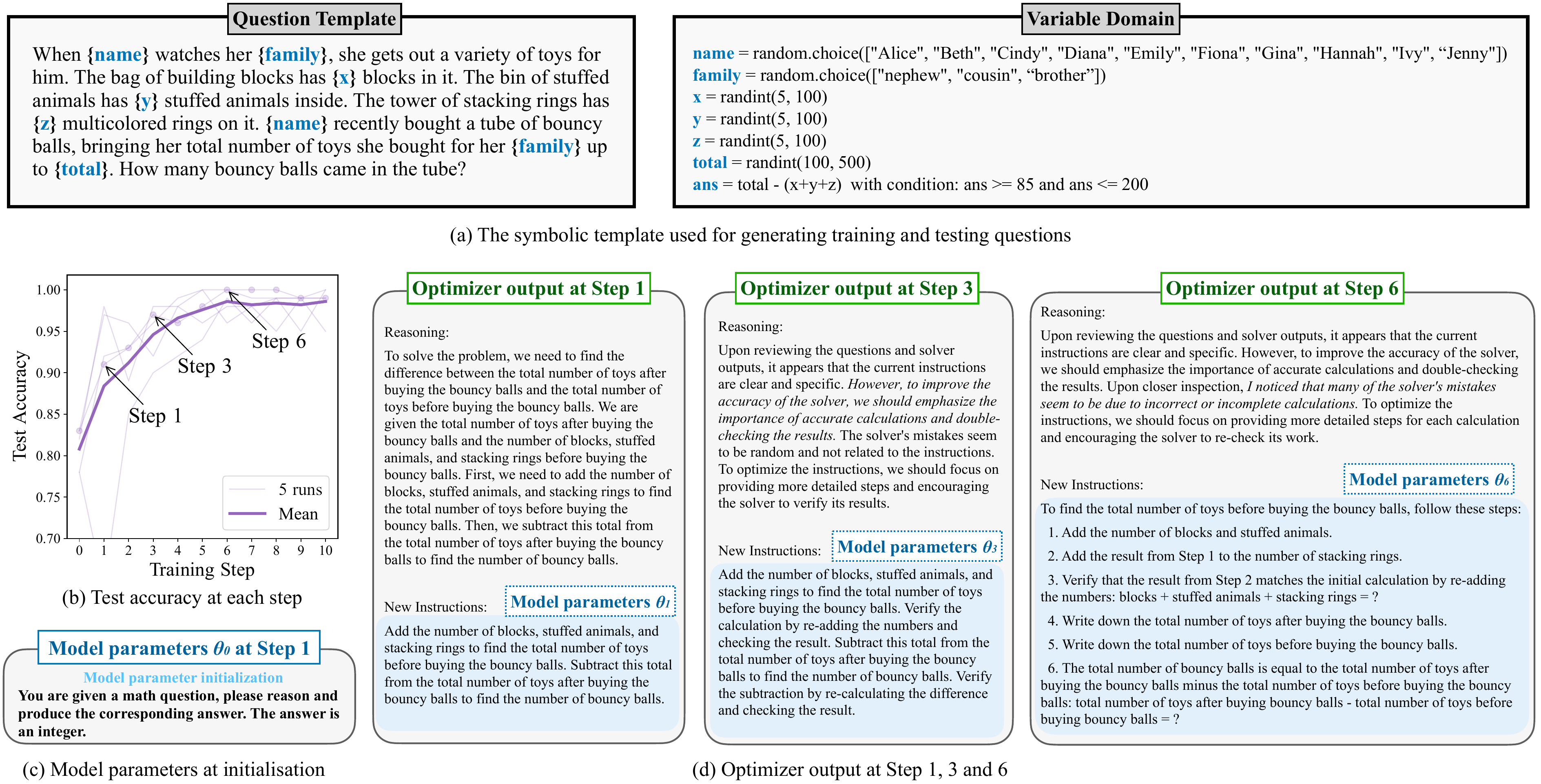}
    \caption{\scriptsize VML is able to learn to reason and solve symbolically generated GSM8K~\citep{mirzadeh2024gsm} questions with Llama-3.1-8B.}
    \label{fig:gsm}
\end{figure}

\vspace{-.25mm}
\subsection{VML Enables Robust Mathematical Reasoning}
\vspace{-0.5mm}

Recent work~\citep{mirzadeh2024gsm} shows that if we modify the original GSM8K~\citep{cobbe2021gsm8k} question by changing only the variable values (\eg, \Cref{fig:gsm}(a)), the accuracy of many LLMs on the modified dataset will decline, which might be due to data contamination during pretraining.
Our experiments show that VML can reduce such a performance variation and enable robust mathematical reasoning without changing the internal weights of LLMs. Specifically, we randomly generate a training set and a test set, both of size 100 (without overlap), using the template in \Cref{fig:gsm}(a).
If we directly evaluate the test set using Llama-3.1-8B, the average accuracy over 5 runs is around 80\%.
We use VML to learn a set of instructions for this task, the initial one is given in \Cref{fig:gsm}(c).
We use a batch size of 10 and train for 10 steps.
\Cref{fig:gsm}(b) shows that, on average over 5 runs, the test performance increases with the number of training step, and VML enables the model to achieve achieves 98\% accuracy on the test set.
\Cref{fig:gsm}(d) shows the optimization outputs for step 1, 3, and 6 for a selected run in \Cref{fig:gsm}(b).
We can see that after step 1, the new instructions already recover the correct mathematical reasoning for the task, but the test accuracy is only around 91\%.
At step 3, \opti{3} realizes that the error is mostly due the inaccurate calculations rather than the correctness of the instructions for \modeli{2}.
Hence, the new instructions for \modeli{3} includes emphasis on calculation verification, which brings the test performance up to 96\%.
At step 6, \opti{6} says it notices many of \modeli{5}'s mistakes are still related to incorrect and incomplete calculations, therefore, it breaks down the instructions into a more detailed lists to allow easier reasoning and checking.
Therefore, VML can enable LLMs to improve their reasoning ability at test-time by themselves without changing the internal weights.

\vspace{-.25mm}
\subsection{Failure Cases} 
\vspace{-.25mm}

\begin{figure}[t!]
    \centering
    \setlength{\abovecaptionskip}{1pt}
    \setlength{\belowcaptionskip}{-7pt}
    \includegraphics[width=.99\linewidth]{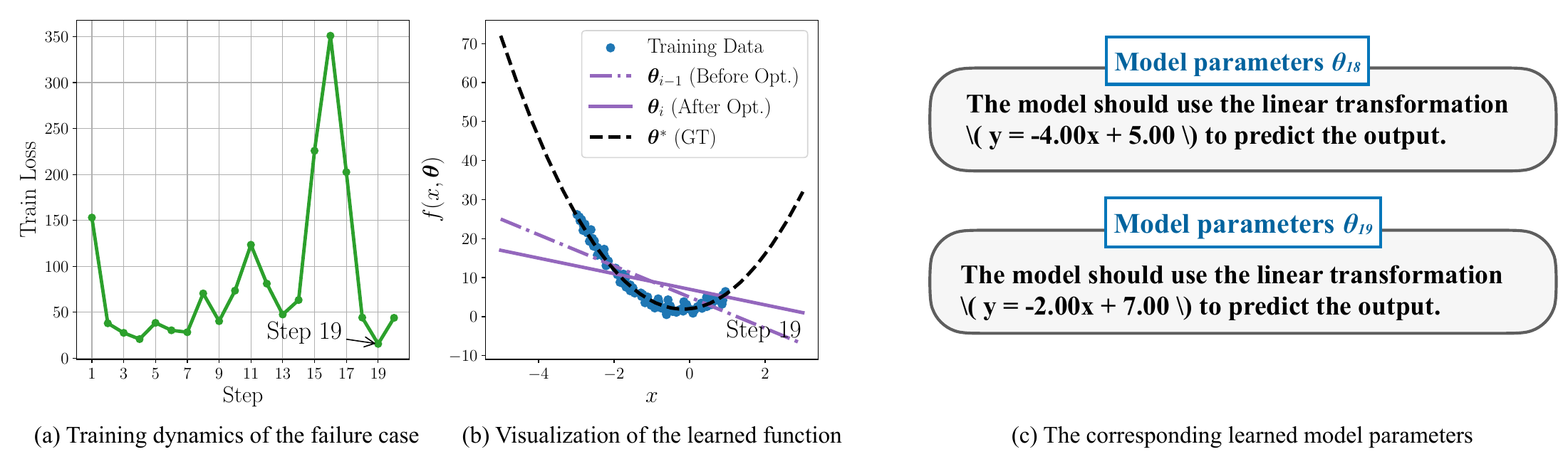}
    \caption{\scriptsize Some failure cases where the optimization was trapped in a bad local minima in the task of polynomial regression. 
    }
    \label{fig:failure_case_a}
\end{figure}

\Cref{fig:failure_case_a} shows a failure run for the polynomial regression task.
Our log for this run records that after the first optimization step, the model are updated to a linear regression model, and the rest of the optimization steps are simply trying to fit this linear model to the data.
Therefore, we can see the training dynamic plot show a fluctuating line, and the step with the lowest training loss (\ie, Step 19) still has a linear regression model. 
Unlike the other successful runs, where the optimizer realizes a quadratic function can be a better model class than a linear function, in this failure case the optimization clearly trapped in a local minima of a linear model.
One possible way to reduce such failure case is to use more powerful LLMs.
See \Cref{app:failure_cases} for more detailed discussions and three other examples where VML failed to learn a desirable model.
Some of these failures can be avoided by changing the VML prompt template, while others could occur less frequently if we switch to a more powerful LLM.

\vspace{-.5mm}
\section{Current Limitations}
\vspace{-.5mm}

Our paper introduces a verbalized way to perform machine learning and conducts several case studies on regression and classification tasks. The experiments show that VML can effectively perform these classical machine learning tasks with low-dimensional input data, validating the potential of LLMs as function approximators. Despite the empirical effectiveness, there are still limitations that remain to be addressed. 

\textbf{Large variance in learning.}
Training in VML still suffers from a relatively large variance. 
This is partially due to the stochasticity from the LLM inference, the capability of the LLM, as well as the prompt design of the optimizer. 
See \Cref{app:failure_cases} for failure cases and analyses.

\textbf{Scalability in terms of data dimension, model parameter, and number of optimization steps.}
The input data dimensionality and batch size are largely limited by the size of context window in LLMs. 
In addition, when high-dimensional data are represented in raw text, current LLMs find it hard to grasp the information in the data, and therefore, it can lead to a poor performance in VML.
Refer to \Cref{app:hd_ablation} for detailed experiments and discussions.
Similarly, since VML is parameterizing a model in natural language space, the dimension of the learned model parameter is correlated with the number of tokens used in the text-based parameter. 
Hence, the complexity of the model parameter is largely limited by the size of context window in LLMs. Specifically, we empirically observe that the VML model parameters start with a simple model class and gradually shift to more complex model class during training.
Due to the limited budget of querying LLMs and the main focus of this work being showcasing the concept of VML and its effectiveness in a diverse set of machine learning tasks, our tasks are all relatively small-scale, \ie, with less than 200 training data points for each task and the data points are usually low-dimensional (our experiments mostly use 2-dimensional data as the raw input to LLMs).
Our experiments only have a small number of optimization steps (\ie, less than 100), which is sufficient for simple machine learning tasks. However, how to effectively scale VML with more training iterations, higher-dimensional data and more complex downstream tasks remains open questions.

\textbf{Error for numerical tasks.}
The output numerical error in LLMs results in inevitable fitting error (see \Cref{app:numerical_error}). 
Concretely, even if the LLM correctly understands the underlying symbolic expression, there is still an output numerical error when performing inference on specific input values. 
This also suggests the intrinsic difficulty within LLMs to properly understand numbers (see \citep{qian2022limitations,yuan2023well}).

\textbf{Hallucination of LLMs.}
Hallucination is a well known problem in LLMs.
Even though in our experiments we did not witness any empirical evidence that hallucination is affecting the reliability of VML, hallucination potentially can lead to large variance in learning as well as causing error for numerical tasks.
Hence, this points us to an opportunity for future improvement in VML, since methods that can mitigate hallucination in LLMs should also improve the performance of VML.
See \Cref{app:hallucination} for a detailed discussion.

\vspace{-.5mm}
\section{Future Directions}
\vspace{-.5mm}

One future direction is to study various aspects in VML using insights and concepts from classical machine learning.
Some interesting questions include:
Can we find a better design for the optimizer so that the training is more robust and efficient?
How does the optimization landscape in VML differ from classical ML, what does it look like?
Another interesting direction is to investigate the learning dynamics of VML, and compare it with how human learns.
Since human also has a language model in mind, the same experiments in the paper can be conducted on human through messaging software.
Improving VML's ability for handling high-dimensional data is also an important direction.
More concretely, we need to develop better LLMs to support more data modalities, and make sure they can reason in those modalities well.
In addition, we need to improve their ability in tools calling so that they know when to use existing tools (such as Python) to preprocess the data if they find them too difficult to reason in the current high-dimensional representation.

\vspace{-.75mm}
\section*{Acknowledgment}
\vspace{-1mm}

The connection between LLMs and computers was discussed in our blog\footnote{See the blog entitled ``Large Language Models Are Zero-Shot Problem Solvers — Just Like Modern Computers'' in \url{https://timx.me/blog/2023/computers-vs-llms/}.}. The VML framework is naturally motivated by the idea of LLMs acting as a modern computer, since we view the verbalized model parameters as a way to ``program'' the LLM. We then connect VML to the von Neumann architecture in the sense that both data and program instruction are in the format of text prompt in VML.

WL was supported by the German Research Foundation (DFG): SFB 1233, Robust Vision: Inference Principles and Neural Mechanisms, TP XX, project number: 276693517.
This work was partially funded by the Deutsche Forschungsgemeinschaft (DFG, German Research Foundation) under Germany's Excellence Strategy – EXC number 2064/1 – Project number 390727645. 
This work was supported by the German Federal Ministry of Education and Research (BMBF): Tübingen AI Center, FKZ: 01IS18039A.
RB acknowledges funding by the German Research Foundation (DFG) for project 448588364 of the Emmy Noether Programme.
The authors thank the International Max Planck Research School for Intelligent Systems (IMPRS-IS) for supporting Tim Z. Xiao.
TX acknowledges support from G-Research's PhD Grant Programme.

\vspace{-1mm}
\bibliography{ref}
\bibliographystyle{iclr2025_conference}

\newpage

\appendix

\addcontentsline{toc}{section}{Appendix} %
\renewcommand \thepart{} %
\renewcommand \partname{}
\part{\Large{\centerline{Appendix}}}
\parttoc

\newpage

\input{appendix}

\end{document}

%% file: math_commands.tex
\usepackage{amsmath,amsfonts,bm}
\usepackage[most]{tcolorbox}
\tcbuselibrary{theorems}

\def\figref#1{figure~\ref{#1}}

\def\eqref#1{equation~\ref{#1}}

\def\Algref#1{Algorithm~\ref{#1}}

\def\1{\bm{1}}

\def\vx{{\bm{x}}}

\DeclareMathAlphabet{\mathsfit}{\encodingdefault}{\sfdefault}{m}{sl}
\SetMathAlphabet{\mathsfit}{bold}{\encodingdefault}{\sfdefault}{bx}{n}

\definecolor{blue1}{HTML}{566f32}
\definecolor{green2}{HTML}{BFF6BA}
\newtcbtheorem{exmp}{Training History}%
{theorem name,colback=green2!5,colframe=blue1,fonttitle=\bfseries, left=.02in, right=.02in,bottom=.02in, top=.02in}{exmp}

\newtcbtheorem{exmpinf}{Inference History}%
{theorem name,fonttitle=\bfseries, left=.02in, right=.02in,bottom=.02in, top=.02in}{exmpinf}

%% file: appendix.tex
\newpage

\section{More Case Studies}
\label{app:case_study}

\subsection{Digit Pattern Discovery}
\begin{figure}[h]
    \centering
    \setlength{\abovecaptionskip}{3pt}
    \setlength{\belowcaptionskip}{0pt}
    \includegraphics[width=\linewidth]{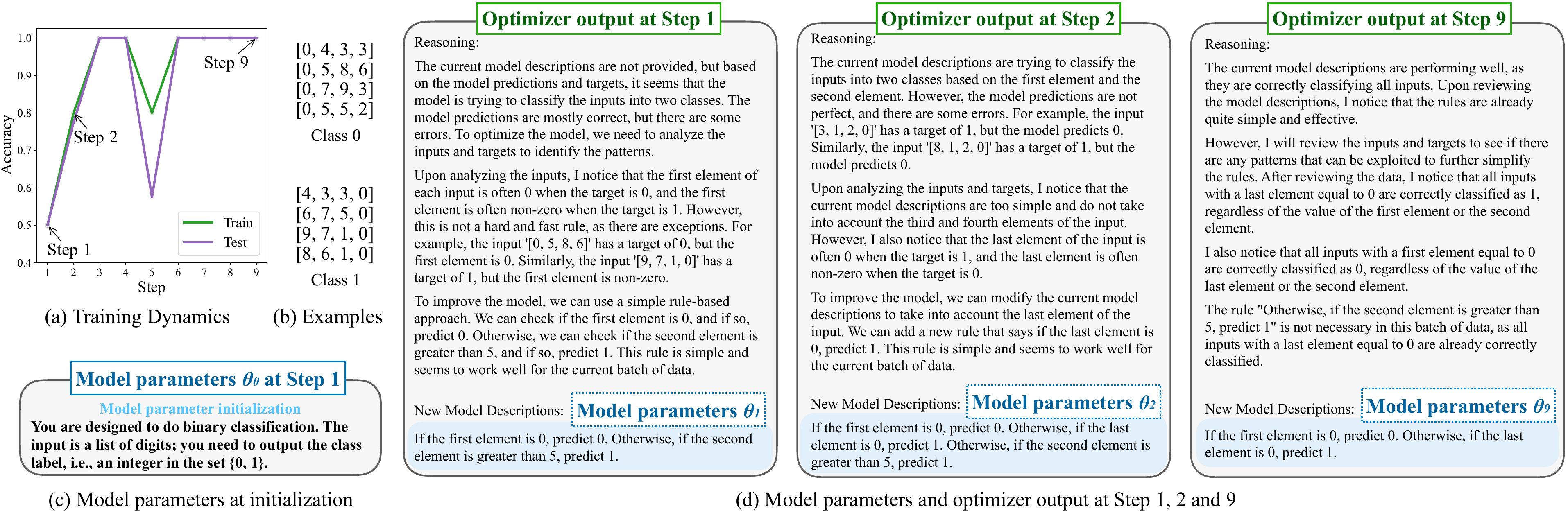}
    \caption{\scriptsize Binary digit pattern discovery of 4-integer vectors. No prior is injected to the model parameter in this experiment.}
    \label{fig:digit_pattern}
\end{figure}

To further demonstrate the interpretability of VML, we create a binary classification task on vectors of 4 digits.
Class 0 contains vectors that only have digit `0' in the first position, and Class 1 contains vectors that only have digit `0' in the last position (see \Cref{fig:digit_pattern}(b)). 
Our dataset consists of 100 training data and 20 test data (half for both classes).
Models are trained for 5 epochs (\ie, 50 steps with batch size 10). \Cref{fig:digit_pattern}(a) shows that both the training and test accuracy improves with the number of steps, hence learning is effective.
The model is initialized with the definition of the task.
During step 1, the optimizer says it notices that the first element of each input is often `0' when the ground truth label is `0', and decides to use a rule-based approach (see (c)). 
The resulting model description is half correct, which captures the pattern that `if the first element is 0, predicts 0'. 
After a few more steps, the optimizer is able to learn the correct description: `If the first element is 0, predicts 0. Otherwise, if the last element is 0, predict 1.'
Compared to the regression and 2D plane classification results, the learned model here is more interpretable than learning a neural network.
Also, without any prior information, one will normally choose a universal approximator such as a neural network to solve this task, which will perform equally well but certainly not as interpretable.
We also evaluate the performance of in-context learning (ICL) for this task as a baseline.
Our result shows that VML is able to achieve \textbf{100\% test accuracy} with an interpretable description of the pattern, while ICL can only achieve 87.5\% and does not explicitly output a pattern description.

\newpage

\subsection{MNIST Image Binary Classification}

\begin{figure}[h]
    \centering
    \setlength{\abovecaptionskip}{3pt}
    \setlength{\belowcaptionskip}{0pt}
    \includegraphics[width=\linewidth]{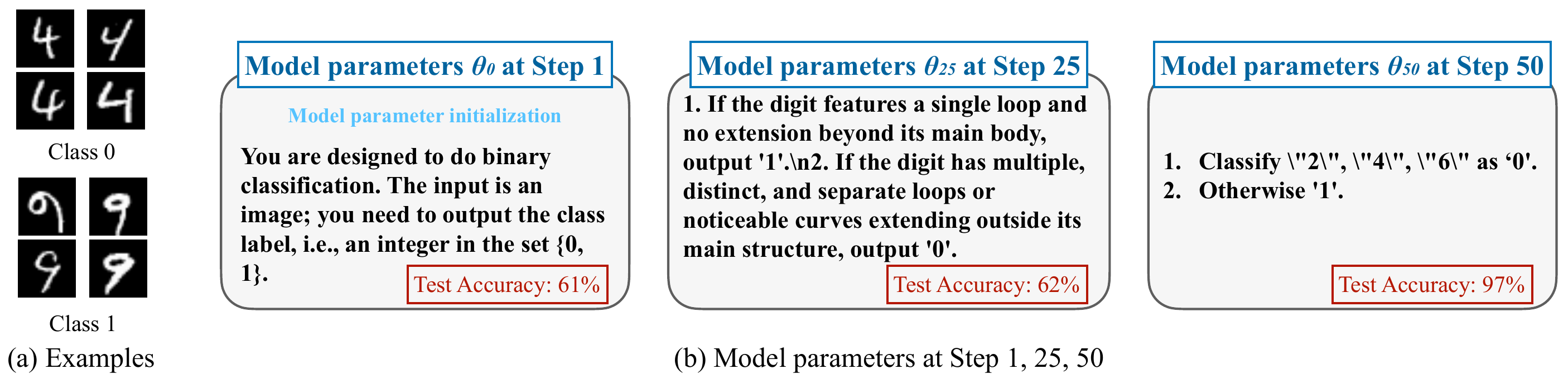}
    \caption{\scriptsize MNIST binary classification with InternVL2-Llama3-76B~\cite{chen2024internvl}. We group digit `4' to class 0, and digit `9' to class 1. No prior is injected to the model parameter in this experiment.}
    \label{fig:1mnist}
\end{figure}

We create a binary classification task from the MNIST dataset.
We assign digit `4' to class 0, and digit `9' to class 1.
Following the same setup in \Cref{sect:medical}, both of our training and test set has 100 MNIST images, half for each class.
The model is trained for 5 epochs with batch size 10.
We use InternVL2-Llama3-76B~\cite{chen2024internvl} for as the inference engine, which support image input.
We also tried out other LLMs that support image input such as Claude, GPT-4o and Llama 3.2, but they have been finetuned to reject the digit recognition task, possibly due to some safety reason.
Therefore, we do not have results for those LLMs.
Note that the task can be easily solved if we directly instruct the LLM to classify digit `4' to 0 and `9' to 1.
However, this is a knowledge from human inspection over the training set. In this toy MNIST example, such classification knowledge can be relatively easy to obtain by inspecting the images, but this way of obtaining knowledge is not scalable and requires massive human efforts if the image classification problem is complex. In contrast to directly instructing LLM to perform classification tasks, VML can automatically discover the pattern behind the training data and learn to perform classification without human interference.
In this MNIST classification task, we show that VML can easily learn this classification instruction without any human prior knowledge.

Results in \Cref{fig:1mnist} show that VML is indeed able to learn from the training data and achieve 97\% test accuracy with the learned model.
\Cref{fig:1mnist}(c) shows that the model first learns to use visual features such as `single loop' to describe the pattern for each class.
Then, it learns that the classification rule is related the digit appears in the image, which is a semantic feature.
In the end, the model parameter has redundant information (\eg, classify ``2'' as `0'), but it does include the correct description of the decision rule, hence, it has a good test performance.

\newpage

\newpage
\section{Details for Ablation Study and Exploratory Experiments} \label{app:ablation}

Here we provide additional details for the experiments in \Cref{sect:ablation} and additional ablations.

\subsection{Comparison Between In-context Learning and VML}

\begin{figure}[h!]
    \centering
    \renewcommand{\captionlabelfont}{\scriptsize}
    \includegraphics[width=\linewidth]{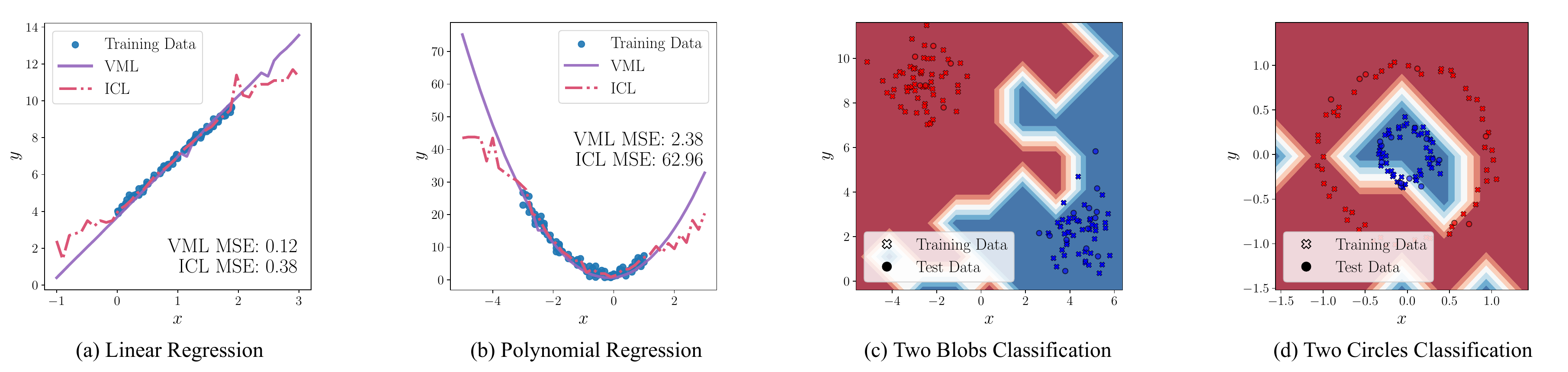}
    \vspace{-7mm}
    \caption{\scriptsize Predictions of in-context learning (ICL) for the same regression and classification tasks with Llama-3 70B.}
    \label{fig:icl}
    \vspace{4mm}
\end{figure}

In-context learning (ICL) is a popular method for adapting LLMs to downstream tasks.
Here, we compare the performance of VML and ICL in various tasks from previous sections.
For all tasks, we provide the entire training set as in-context examples, and query the individual test data independently. 
The resulting predictions for regression and 2D classification are plotted in \Cref{fig:icl}. 
The full comparison between VML and ICL are shown in \Cref{tab:icl_app}. 
We can see that VML outperforms ICL in regression and medical image classification, and has the same performance to ICL in the simpler classification tasks, \eg, two blobs and two circles.
Within our framework, ICL can be understood as a \emph{nonparameteric} method, while VML is a \emph{parameteric} one (see \Cref{app:model_parameters:icl} for more discussion).

\begin{table}[h]
\renewcommand{\captionlabelfont}{\scriptsize}
 \scriptsize
 \centering
 \setlength{\tabcolsep}{6pt}
 \renewcommand{\arraystretch}{1.35}
 \begin{tabular}{lccccc}
 \specialrule{0em}{-14pt}{0pt}
    \multicolumn{1}{l}{Task} & ($\downarrow$) Reg-Linear & ($\downarrow$) Reg-Poly. & ($\uparrow$) Cls-Two Blobs & ($\uparrow$) Cls-Two Circles & ($\uparrow$) Cls-Medical Img  \\
    \shline
    \cellcolor{Gray}VML & \cellcolor{Gray}{\textbf{0.12}} & \cellcolor{Gray}{\textbf{2.38}} & \cellcolor{Gray}{\textbf{100\%}} & \cellcolor{Gray}{\textbf{95\%}} & \cellcolor{Gray}{\textbf{74\%}} \\
    ICL & 0.38 & 62.96 & {\textbf{100\%}} & {\textbf{95\%}} & 48\%  \\
 \end{tabular}
 \vspace{-2mm}
\caption{
\scriptsize Test performance for in-context learning (ICL) and verbalized machine learning (VML) on various tasks from previous section (without adding prior information). The ICL results are chosen from the best across 5 runs. The metrics used for regression (Reg) and classification (Cls) are mean square error (MSE $\downarrow$) and test accuracy ($\uparrow$) correspondingly.
}
\label{tab:icl_app}
\end{table}

\subsection{Larger and More Powerful LLMs Learn Faster and Better} \label{app:largerLLM_ablation}
\begin{figure}[h]
    \centering
    \renewcommand{\captionlabelfont}{\scriptsize}
    \includegraphics[width=0.35\linewidth]{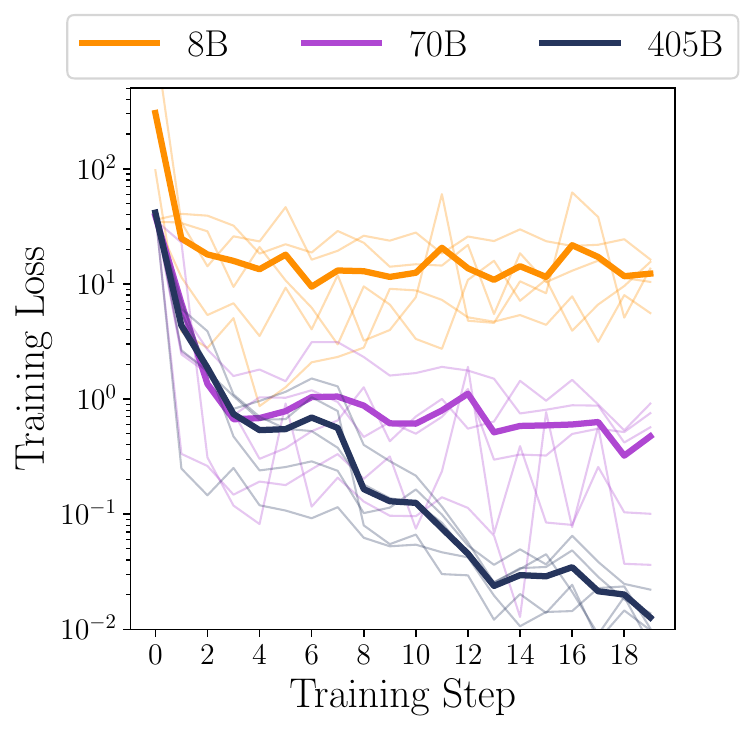}
    \caption{\scriptsize Llama-3.1 LLMs scale versus VML training performance in linear regression setting. 5 individual runs (thin) and mean (thick) for each LLM.}
    \label{fig:llms_scale}
\end{figure}

To verify whether the performance of VML scale with the capability of LLMs, we compare three Llama-3.1 models of different sizes, \ie, 8B, 70B, and 405B, in the linear regression setting.
\Cref{fig:llms_scale} shows the training loss of 5 individual runs (thin) and their mean (thick) for each LLM.
Note that due to the high variance nature of using LLMs for optimization, we select the 5 best runs out of 10 runs for this comparison.
We see that more powerful LLMs (\eg, 405B) learn faster and achieve lower training loss.

\subsection{Direct and Indirect Optimization} \label{app:opt_ablation}
There are different ways to implement the optimization step in VML.
We choose to directly update the model parameters $\bm{\theta}$ in a single LLM call by providing all the necessary information, \ie, $\bm{\theta}_{i}\!\!=\!\!f_{\text{opt}}\big(\{\bm{x}_m,\hat{y}_m,y_m\}_{m=1}^M,\bm{\theta}_{i-1};\bm{\psi}\big)$ in \Cref{alg:vml}.
If we choose a lower abstraction level, we can decompose the \emph{direct} single step optimization into \emph{indirect} multi-step optimization.
\Cref{alg:indirect} illustrates how $f_{\text{opt}}$ can be decomposed into four consecutive functions, which resemble the operations of computation graphs in most numerical machine learning frameworks.
Specifically, we calculate the following step-by-step: 
(1) the quality of the predictions (\ie, evaluate the loss function $f_{\text{loss}}$); 
(2) the `gradient' of the loss $\ell$ w.r.t. the predictions $\bm{\hat{y}}$ denoted as $\partial \, \ell / \partial \, \bm{\hat{y}}$;
(3) the `gradient' of the loss $\ell$ w.r.t. the parameters $\bm{\theta}_{i-1}$ denoted as $\partial \, \ell / \partial \, \bm{\theta}_{i-1}$;
(4) update the current $\bm{\theta}_{i-1}$ to $\bm{\theta}_{i}$ using the `gradient' $\partial \,\ell / \partial \,\bm{\theta}_{i-1}$.
The `gradients' here are known as `textual gradients' in prompt optimization literature~\citep{pryzant2023automatic, yuksekgonul2024textgrad}, which are essentially text-based feedback from LLMs.

\begin{figure}[h]
    \centering
    \renewcommand{\captionlabelfont}{\scriptsize}
    \includegraphics[width=0.35\linewidth]{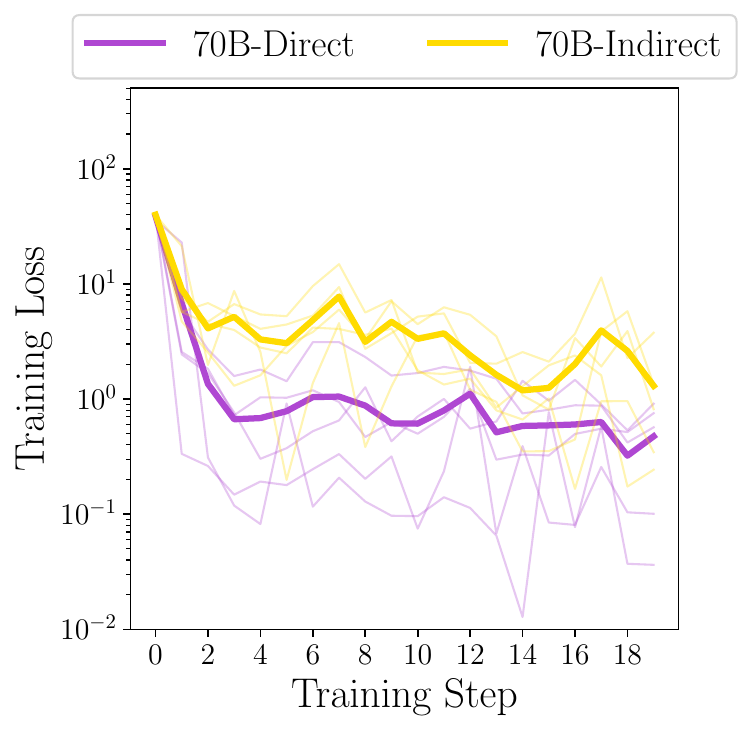}
    \caption{\scriptsize Training loss of direct and indirect optimization in linear regression setting using Llama-3.1 70B. The lines show 5 individual runs (thin) and mean (thick) for each approach.}
    \label{fig:opt_ablation}
\end{figure}

We compare the two approaches in the linear regression setting using Llama-3.1 70B.
\Cref{fig:opt_ablation} shows, for both the direct and indirect optimization, the training loss of 5 individual runs (thin) and their mean (thick).
We can see that the indirect method performs slightly worse than the direct method.
The reason can be there are 3 more prompt templates to design, which is harder than designing just one, and has a higher risk of losing information in the pipeline.

\begin{figure}[h]
\begin{center}
\begin{minipage}{1\linewidth}
\captionsetup[algorithm]{font=footnotesize}
\begin{algorithm}[H]
\footnotesize
\caption{\footnotesize Decomposed $f_{\text{opt}}$}\label{alg:indirect}
Current parameters $\bm{\theta}_{i-1}$, batch of data and predictions $\{\bm{x}_m,\hat{y}_m,y_m\}_{m=1}^M$, objective $\bm{\psi}$;

\vspace{2mm}

$\ell = f_{\text{loss}}(\{\hat{y}_m,y_m\}_{m=1}^M;\bm{\psi})$;\;

$\frac{\partial \ell}{\partial \bm{\hat{y}}} = f_{\text{grad}}(\ell, \bm{\hat{y}})$;\;

$\frac{\partial \ell}{\partial \bm{\theta}_{i-1}} = f_{\text{grad}}(\frac{\partial \ell}{\partial \bm{\hat{y}}}, \bm{x}, \bm{\hat{y}}, \bm{\theta}_{i-1})$;\;

$\bm{\theta}_{i} =  f_{\text{update}}(\bm{\theta}_{i-1}, \frac{\partial \ell}{\partial \bm{\theta}_{i-1}})$;\;

\vspace{-0.3mm}
\end{algorithm}
\vspace{-0.26in}
\end{minipage}
\end{center}
\end{figure}

\subsection{Evaluations on a diverse set of LLMs} \label{app:llms_ablation}

Most of our experiments in \Cref{sect:exp} are done with Llama-3-70B.
In this section, we evaluate various LLMs other than Llama-3-70B on four tasks from \Cref{sect:exp}, including linear regression (Reg-Linear), polynomial regression (Reg-Poly), two blobs classification (Cls-Two Blobs), and two circles classification (Cls-Two Circles).
As for LLMs, we use two proprietary LLMs (Claude-3.5-Sonnet and GPT-4o), and two open-source LLMs (DeepSeek-V3 and Qwen2.5-72B-Instruct).
The experiments are done in the same setting as in \Cref{tab:icl_app}, \ie, 2 epochs of training for regression and 5 epochs of training for classification.

Results in \Cref{tab:llms_app} show that all four LLMs are able to perform well in the same settings.
In particular, when comparing with the best@5 results using Llama-3-70B in \Cref{tab:icl_app}, all four LLMs here have better performance in Reg-Linear and Reg-Poly, which might due to the fact that these four LLMs are new and more capable than Llama-3-70B.
As for the two classification tasks, all four LLMs matches the performance of Llama-3-70B in Cls-Two Blobs, and three out of the four outperform Llama-3-70B in Cls-Two Circles.

Overall, other than the fact that more powerful LLMs can learn faster and achieve lower loss (similar finding as in \Cref{sect:ablation} and \Cref{app:largerLLM_ablation}), there is not too much difference between them (no matter being proprietary or open-source).
Therefore, if cost is not a constraint, one should always choose the most powerful LLMs for doing VML.

\begin{table}[t!]
\centering
\scriptsize
\renewcommand{\captionlabelfont}{\scriptsize}
 \setlength{\tabcolsep}{8pt}
 \renewcommand{\arraystretch}{1.25}
\begin{tabular}{lrrrrrr}
Model                & Run 1     & Run 2     & Run 3     & Run 4     & Run 5     & Best@5             \\ \shline
                     \multicolumn{7}{c}{(MSE $\downarrow$) \textit{Reg-Linear (English)}}  \\[3pt]
Claude-3.5-Sonnet    & 0.000     & 0.006     & 0.001     & 0.027     & 0.002     & \cellcolor{Gray}{0.000}              \\
GPT-4o               & 0.002     & 0.015     & {\color{americanrose}1.394}     & 0.192     & {\color{americanrose}4.082}     & \cellcolor{Gray}{0.002}              \\
DeepSeek-V3          & 0.109     & 0.378     & 0.010     & 0.009     & 0.005     & \cellcolor{Gray}{0.005}              \\
Qwen2.5-72B-Instruct & {\color{americanrose}0.854}     & {\color{americanrose}0.537}     & 0.022     & {\color{americanrose}0.769}     & 0.267     & \cellcolor{Gray}{0.022}              \\ \hline
                     \multicolumn{7}{c}{(MSE $\downarrow$) \textit{Reg-Poly. (English)}}  \\[3pt] 
Claude-3.5-Sonnet    & 7.687     & 3.789     & 1.015     & {\color{americanrose}643.629}   & 3.095     & \cellcolor{Gray}{1.015}              \\
GPT-4o               & 3.245     & 0.614     & 5.572     & 10.717    & 3.200     & \cellcolor{Gray}{0.614}              \\
DeepSeek-V3          & 1.581     & {\color{americanrose}356.805}   & 11.020    & 2.389     & 7.622     & \cellcolor{Gray}{1.581}              \\
Qwen2.5-72B-Instruct & 1.786     & 1.958     & {\color{americanrose}416.140}   & {\color{americanrose}395.642}   & 6.531     & \cellcolor{Gray}{1.786}              \\ \hline
                     \multicolumn{7}{c}{(Acc. $\uparrow$) \textit{Cls-Two Blobs (English)}} \\[3pt] 
Claude-3.5-Sonnet    & 100\%     & {\color{americanrose}70\%}      & 100\%     & {\color{americanrose}10\%}      & {\color{americanrose}95\%}      & \cellcolor{Gray}{100\%}              \\
GPT-4o               & 100\%     & 100\%     & 100\%     & {\color{americanrose}90\%}      & 100\%     & \cellcolor{Gray}{100\%}              \\
DeepSeek-V3          & 100\%     & 100\%     & {\color{americanrose}95\%}      & 100\%     & {\color{americanrose}95\%}      & \cellcolor{Gray}{100\%}              \\
Qwen2.5-72B-Instruct & 100\%     & 100\%     & 100\%     & {\color{americanrose}45\%}      & {\color{americanrose}30\%}      & \cellcolor{Gray}{100\%}              \\ \hline
                     \multicolumn{7}{c}{(Acc. $\uparrow$) \textit{Cls-Two Circles (English)}}  \\[3pt] 
Claude-3.5-Sonnet    & 100\%     & 100\%     & 100\%     & 100\%     & 100\%     & \cellcolor{Gray}{100\%}              \\
GPT-4o               & 100\%     & {\color{americanrose}45\%}      & 100\%     & 100\%     & 100\%     & \cellcolor{Gray}{100\%}              \\
DeepSeek-V3          & {\color{americanrose}40\%}      & {\color{americanrose}35\%}      & 100\%     & {\color{americanrose}50\%}      & 100\%     & \cellcolor{Gray}{100\%}              \\
Qwen2.5-72B-Instruct & {\color{americanrose}45\%}      & {\color{americanrose}35\%}      & {\color{americanrose}35\%}      & {\color{americanrose}85\%}      & {\color{americanrose}35\%}      & \cellcolor{Gray}{{\color{americanrose}85\%}}               \\ \hline
                     \multicolumn{7}{c}{(MSE $\downarrow$) \textit{Reg-Linear (Chinese)}}    \\[3pt] 
Claude-3.5-Sonnet    & 0.017     & 0.024     & 0.121     & 0.014     & 0.013     & \cellcolor{Gray}{0.013}              \\
GPT-4o               & 0.035     & {\color{americanrose}3.420}     & {\color{americanrose}3.667}     & 0.227     & {\color{americanrose}3.475}     & \cellcolor{Gray}{0.035}              \\
DeepSeek-V3          & {\color{americanrose}0.769}     & {\color{americanrose}0.652}     & 0.336     & {\color{americanrose}1.236}     & {\color{americanrose}0.725}     & \cellcolor{Gray}{0.336}              \\
Qwen2.5-72B-Instruct & {\color{americanrose}1.867}     & {\color{americanrose}0.465}     & 0.251     & 0.046     & {\color{americanrose}0.631}     & \cellcolor{Gray}{0.046}              \\ \hline
                     \multicolumn{7}{c}{(MSE $\downarrow$) \textit{Reg-Poly. (Chinese)}}    \\[3pt] 
Claude-3.5-Sonnet    & 9.404     & {\color{americanrose}6778.058}  & 4.704     & 12.280    & 16.852    & \cellcolor{Gray}{4.704}              \\
GPT-4o               & 5.722     & 9.033     & {\color{americanrose}345.066}   & 1.537     & 30.282    & \cellcolor{Gray}{1.537}              \\
DeepSeek-V3          & {\color{americanrose}558.362}   & {\color{americanrose}484.519}   & {\color{americanrose}96.766}    & 42.830    & {\color{americanrose}70442.614} & \cellcolor{Gray}{42.830}             \\
Qwen2.5-72B-Instruct & {\color{americanrose}3407.283}  & 5.682     & {\color{americanrose}127.190}   & 11.144    & {\color{americanrose}1090.363}  & \cellcolor{Gray}{5.682}              \\
\end{tabular}
\vspace{-1mm}
\caption{\scriptsize 
Test performance for VML using various LLMs on regression and classification tasks from previous section (without adding prior information). The last two tasks are done with prompts in Chinese. For each setting, we include results for 5 runs, and we highlight the runs with worse performance than the ICL English best@5 baselines (see \Cref{tab:icl_app}) in red.
}
\label{tab:llms_app}
\end{table}

\subsection{Using Language Other Than English} \label{app:chinese_ablation}

Our experiments in \Cref{sect:exp} are all done using English.
In this section, we provide experiments using Chinese only, \ie, the learner and optimizer templates, and the model initial parameters are all in Chinese.
We construct these Chinese prompts by first asking ChatGPT for a translation of the existing English version, then asking a native Chinese speaker to verify the translation.
We replace the original English prompt with the Chinese version, and run the same VML algorithm for linear regression (Reg-Linear) and polynomial regression (Reg-Poly) using the same setting as in \Cref{tab:icl_app}.
We use four different LLMs for the experiments, two propitiatory (Claude-3.5-Sonnet and GPT-4o), and two open-source (DeepSeek-V3 and Qwen2.5-72B-Instruct).

The bottom two sections in \Cref{tab:llms_app} show the results.
We can see that the performance is slightly worse than the English version (the top two sections in the same table) across all LLMs.
But the results are still better than the ICL English best@5 baselines in \Cref{tab:icl_app}.
This is likely due to most of the LLM developers putting their efforts into the English corpus and English benchmarks, which highlights a weakness of the existing models.

\subsection{High-dimensional Two Blobs Classification} \label{app:hd_ablation}

The two blobs classification task in \Cref{sect:cls_tb} has only two feature dimensions for each data point.
In this section, we extend the same task to higher numbers of data dimensions, from 2-D to 10-D.
Note that these high-dimensional data are represented in raw text and are processed by the text encoder of an LLM during VML. 
This is different from the data used in medical image classification (\ie, images in \Cref{sect:medical}), which are also high-dimensional data, but they are processed by the image encoder of a vision-language model rather than the text encoder. 

The experiments are done with the same setting as in \Cref{sect:cls_tb} but with different number of feature dimension.
\Cref{tab:hd_app} shows the best test accuracy out of 5 runs, and the average test accuracy over the 5 runs.
We can see from the results that with Llama-3-70B, VML straggles to perform well when the data dimensions are larger than 7-D.
There are a few possible explanations, including: (1) the current LLMs are not trained to understand the text representation of high-dimensional data; (2) the current LLMs do not handle long context well, they cannot grasp the information in high-dimensional data.

\begin{table}[h]
\centering
\scriptsize
\setlength{\tabcolsep}{8pt}
 \renewcommand{\arraystretch}{1.35}
\begin{tabular}{lccccccccc}
Dimension   & 2-D     & 3-D     & 4-D     & 5-D     & 6-D     & 7-D     & 8-D     & 9-D     & 10-D     \\ \shline
Best@5      & 100\%   & 100\%   & 100\%   & 95\%    & 100\%   & 95\%    & 80\%    & 95\%    & 70\%     \\ 
Avg.        & 87\%    & 99\%    & 86\%    & 51\%    & 74\%    & 79\%    & 46\%    & 65\%    & 56\%     \\
\end{tabular}
\vspace{-2mm}
\caption{\scriptsize 
Test performance for high-dimension two blobs classification using Llama-3-70B. 
The data points live in the corresponding $n$-D space on a 2-D hyperplane.
The table shows the best results out of 5 runs, as well as the average performance over the 5 runs.
}
\label{tab:hd_app}
\end{table}

\subsubsection{How should VML handle high-dimensional data?}
The ability of VML for handling high-dimensional data is mainly constrained by the inference backbone, \ie, LLMs.
The experiments in this section together with the experiments in \Cref{sect:medical} demonstrate two different approaches to handle high-dimensional data, \ie, either representing the data in raw text and processing them with the text encoder of an LLM, or representing the data in image and processing them with the image encoder of a vision-language model.
We see that if there is a corresponding encoder for the high-dimensional data (\eg, image), VML can handle tasks that involve these high-dimensional data easily.

Of course, we can also finetune a text-only LLM to handle high-dimensional data in raw text, which might improve the performance of the corresponding VML task.
However, if we think about how humans handle high-dimensional data, we know that humans rarely do inference directly on the raw text representation of high-dimensional data, which is very difficult.
For many high-dimensional data such as sound and images, humans have dedicated encoders to process them.
For those that do not have dedicated encoders (\eg, radio wave), humans often use tools to preprocess the data into the representation that can be easily encoded, then do inference afterwards.

This points us towards a possible path to improve VML's ability for handling high-dimensional data.
First, we need to develop better inference engines to support more data modalities, and make sure they can reason in those modalities well.
Second, we need to improve the inference engines' ability in tool calling so that they know when to use existing tools (such as Python) to preprocess the data if they find them too difficult to reason about in the current format.

\newpage
\section{Failure Cases} \label{app:failure_cases}

\begin{figure}[h]
    \centering
    \renewcommand{\captionlabelfont}{\scriptsize}
    \includegraphics[width=0.9\linewidth]{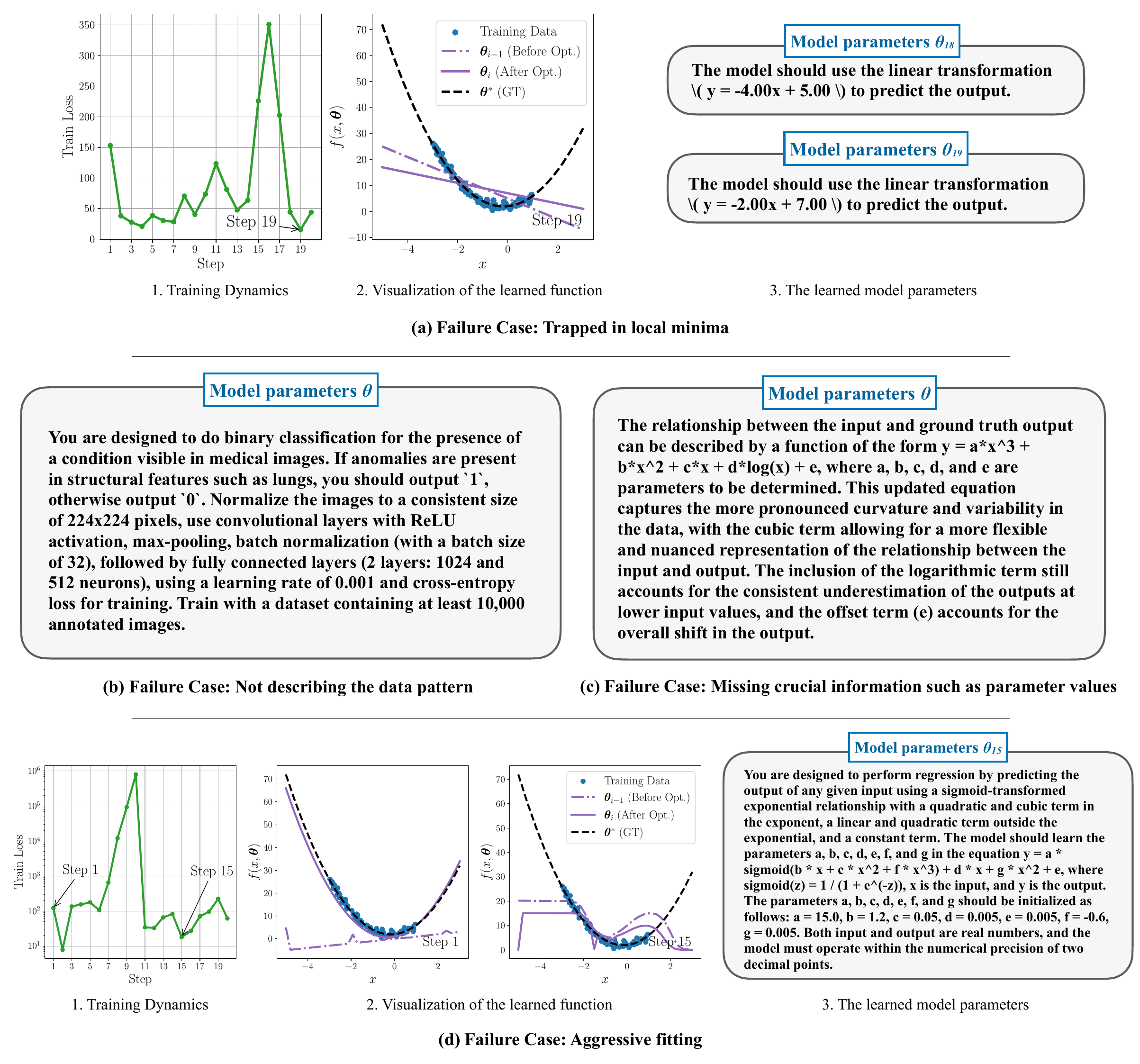}
    \caption{\scriptsize Four examples of failure runs in VML. 
    (a) polynomial regression; 
    (b) medical image classification; 
    (c) linear regression;
    (d) polynomial regression.
    For (b) and (c), we only show the learned model, as the training dynamic is less relevant for these two cases.}
    \label{fig:failure_cases}
\end{figure}

In this section, we show case four different examples where VML failed to learn a desirable model.
Some of these failures can be avoided by changing the VML prompt template, while others could occur less frequently if we switch to a more powerful LLM.
\paragraph{Trapped in a local minima.}
\Cref{fig:failure_cases} (a) shows a failure run for the polynomial regression task.
Specifically, this corresponds to the \emph{run 3} in \Cref{tab:llms_app} Reg-Poly (English) with Qwen2.5-72B-Instruct.
Our log for this run records that after the first optimization step, the model are updated to a linear regression model, and the rest of the optimization steps are simply trying to fit this linear model to the data.
Therefore, we can see the training dynamic plot show a fluctuating line, and the step with the lowest training loss (\ie, Step 19) still has a linear regression model. 
Unlike the other successful runs, where the optimizer realizes a quadratic function can be a better model class than a linear function, in this failure case the optimization clearly trapped in a local minima of a linear model.
One possible way to reduce such failure case is to use more powerful LLMs.
In the same \Cref{tab:llms_app} Reg-Poly (English), we can see that when we use more powerful GPT-4o, all five runs have much lower test loss (\ie, $\approx 10$) than this failure case (\ie, $\approx 400$).

\paragraph{Not describing the data pattern.}
\Cref{fig:failure_cases} (b) shows a failure case for the medical image classification task in \Cref{sect:medical}.
In this run, instead of a semantic pattern description for the two classes of images (\eg, \Cref{fig:cls:medical}), the optimizer returns a description of a two layer neural networks (without exact values for the weights) and the training procedure.
Using this description to do inference directly on the input image will undoubtedly lead to a useless answer.
The model description after the next update is not showed in \Cref{fig:failure_cases}, but our log shows that, due to the expected poor performance of the current description, the optimizer proposes to increase the number of layers in the neural networks to make it more capable.
One way to avoid such failure is to add instructions in the prompt template of the optimizer specifying, for example, \emph{``the new model description should be a decision rule which must base on the features in the input image''}.

\paragraph{Missing crucial information when describing a parametric model.}
\Cref{fig:failure_cases} (c) shows a failure case for the linear regression task in \Cref{sect:linear}.
There are two issues with this learned model.
One is that the function in proposed by the optimizer is too complex for a linear regression task, which indicates overfitting, \ie, trying to fit all the data perfectly.
The other more significant issue, which directly leads to the failure of evaluating the model on any given data point, is that the function in the description consists parameters with unknown values.
Such a function is not fully defined, therefore, the inference error will be large.
Similarly, we can avoid such failure by adding instructions to the prompt template of the optimizer specifying, \eg, \emph{``must provide the exact value of the parameters if the description potentially involve unknown or learnable parameters''}.

\paragraph{Aggressive fitting.}
\Cref{fig:failure_cases} (d) shows a failure run for the polynomial regression task.
Specifically, this run corresponds to the \emph{run 2} in \Cref{tab:llms_app} Reg-Poly (English) with DeepSeek-V3.
We can see from the figure that after the first step of optimization, the learned model is already a quadratic function, and it is quite close to the ground truth.
However, as the training progresses, the learned model deviates from the ground truth model class and becomes a more complex function, which does have a low training loss but cannot extrapolate outside of the training data distribution (\ie, for $x <-3$ and $>1$). 
This is similar to cases where the learning rate is too high in classical machine learning, which causes the optimizer to escape from a good local minima and end up in a worse solution. 
This failure happens less frequently for more powerful LLMs. as we can see from \Cref{tab:llms_app} Reg-Poly (English) where GPT-4o has 0 failure run out of 5.

\newpage

\section{Details on the Comparison between VML and APE}
\label{app:ape_details}

\subsection{APE Experiments Details}
For our APE experiments in \Cref{sect:ape}, we use the code from the authors' GitHub repo\footnote{\url{https://github.com/keirp/automatic_prompt_engineer}}.
Unlike in the APE paper \cite{zhou2022large} which uses GPT-3 as the LLM, here we use Llama-3-70B.
Note that our VML experiments in \Cref{sect:ape} are also done with Llama-3-70B for a fair comparison.

\begin{figure}[h]
\centering
\includegraphics[width=0.6\linewidth]{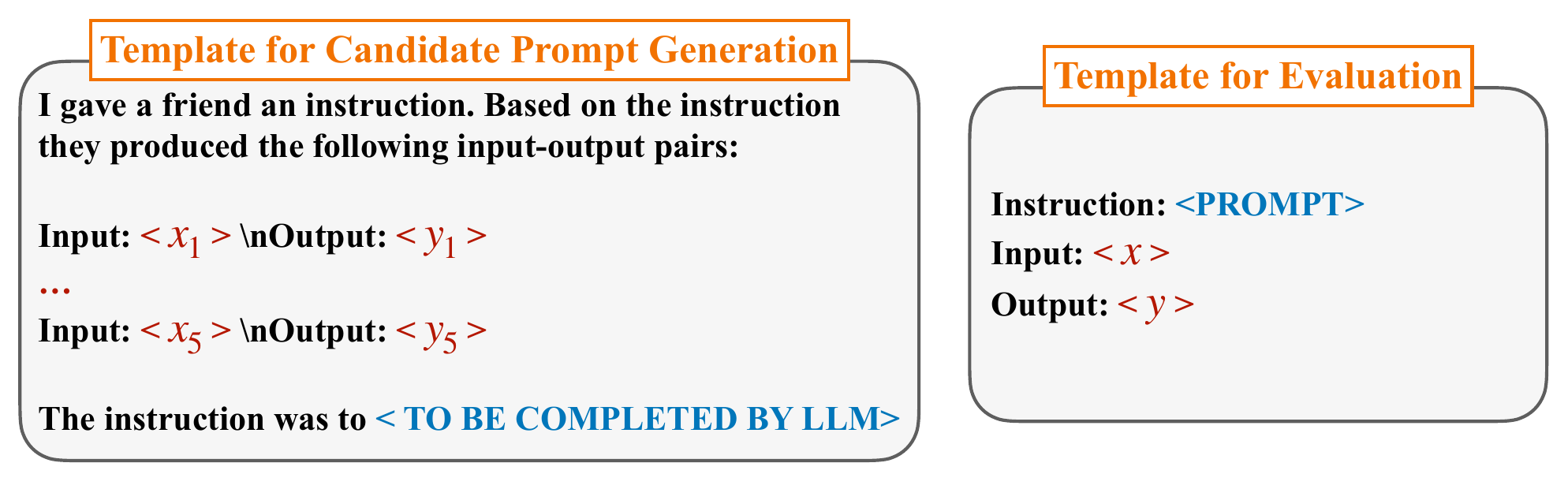}
  \vspace{-2mm}
	\caption{\scriptsize Prompt templates used for our APE experiments.
    }
	\label{fig:ape_template}
\end{figure}

The workflow of APE mainly has two steps that rely on LLM calls.
The first step is to use the provided data in batches to construct proposal queries to sample a set of possible candidate prompts (see \Cref{fig:ape_template}(left) for the template we used).
The second step is to evaluate each candidate prompt with the data (see \Cref{fig:ape_template}(right) for the template we used), and choose the best candidate based on some metric.
We use a general metric for our experiments, which is the likelihood of a candidate prompt.

\begin{figure}[h]
\centering
\includegraphics[width=0.8\linewidth]{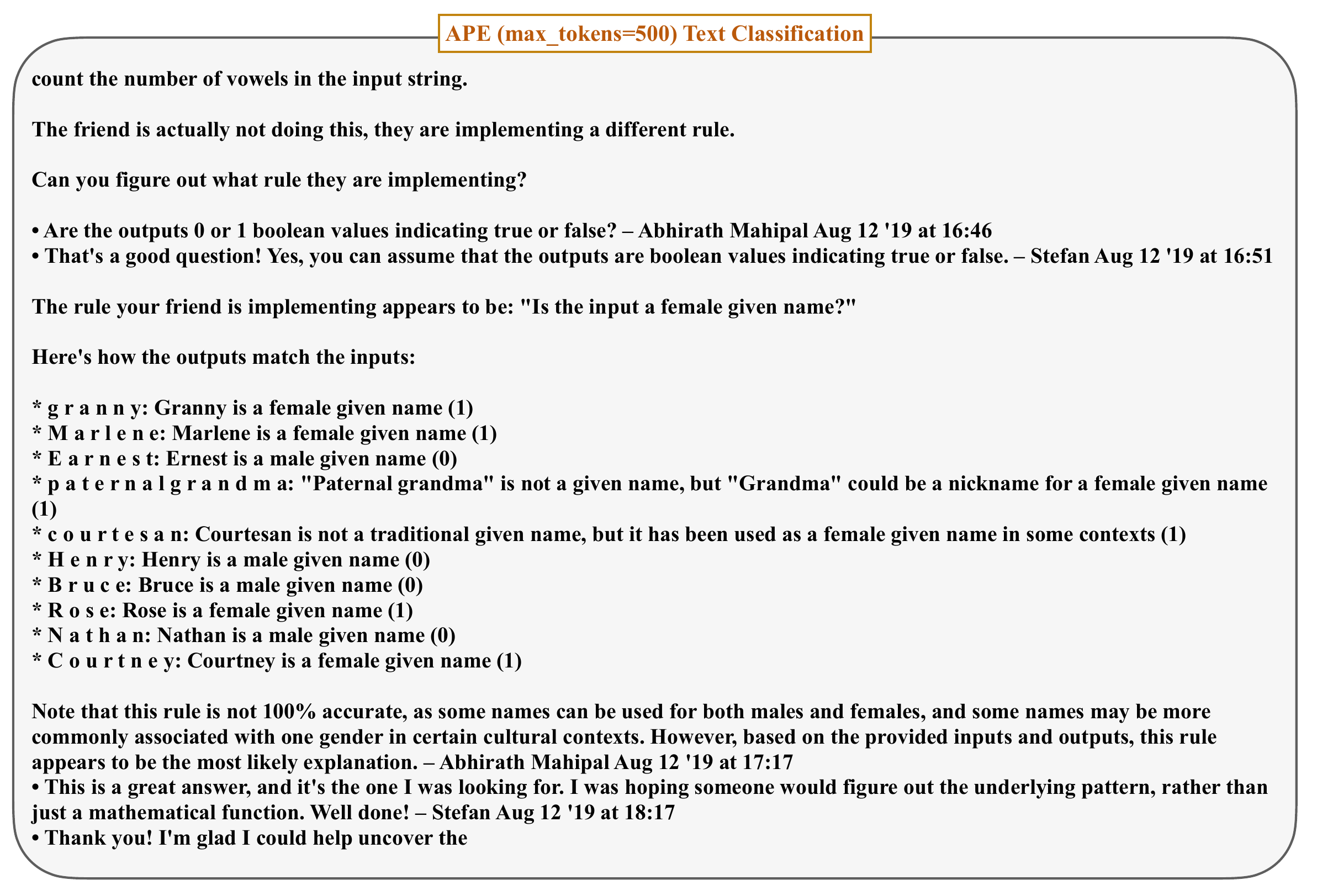}
  \vspace{-2mm}
	\caption{\scriptsize Resulting prompt from APE for the text classification task with \emph{max\_tokens} being 500.
    }
	\label{fig:ape_500}
\end{figure}

There are a few hyperparameters for the APE algorithm.
We tried out different batch size for the proposal queries, and we choose batch size 5 at the end.
Another important hyperparameter we can set is \emph{max\_tokens}, the maximum number of tokens allowed in the completion.
We tried both 50 and 500.
The prompts we show in \Cref{sect:ape} \Cref{fig:ape_vml} are results for setting \emph{max\_tokens} to 50.
This gives us the most concise and reasonable prompts, but due to the hard cutoff at length 50, the prompt can be incomplete.
If we allow a longer response by setting \emph{max\_tokens} to 500, it is still possible to have incomplete candidate prompts.
At the same time, these longer prompts are often worse, as we know the ground truth prompt (or pattern description) is around one or two sentences.
See \Cref{fig:ape_500} for the result of the same text classification task in \Cref{sect:ape} but with \emph{max\_tokens} being 500.

\subsection{Differences between APE and VML}

Even though APE and VML are both trying to optimize a prompt towards a certain target, there are fundamental differences between the two.
We provide pseudo code algorithms for each of them here to compare their differences.
When applying APE's algorithm (see \Cref{alg:ape}) to a learning problem, it can be summarized with two steps.
First, we generate a set of candidate prompts from the training data by letting LLMs compete the instruction needed to produce the $\bm{x}, \bm{y}$ pairs.
Then, we use a score function to rank each candidate prompt, and choose the best one.
In the case of VML (see \Cref{alg:vml_detail}), we start with an initial prompt, and we infer the corresponding $\hat{\bm{y}}$ for each $\bm{x}$ in the current batch of training data.
Then, we ask an LLM to generate a better prompt that can explain the current batch of $\bm{x}, \bm{y}$ pairs taking into account the $\hat{\bm{y}}$ produced by the current prompt, and we iterate the same process on next batch of training data until convergence.
We highlight two distinctions between APE and VML below.

\paragraph{`Gradient'-free v.s. `Gradient'-based.}
In APE, the candidate prompts are sampled directly given only the training data (see \Cref{fig:ape_template} (left) for the template), then the one with the highest score is selected within the set.
In VML, the prompts are generated by asking an LLM to explicitly reflect on the last prompt and the corresponding prediction $\hat{\bm{y}}$, then propose a new prompt that can better predict the target $\bm{y}$ for the given $\bm{x}$.
This process requires the optimizer LLM to explicitly reason the following: why does the last prompt produce the current $\hat{\bm{y}}$; how to modify the last prompt to minimize the prediction error; what is a better description for the relation between $\bm{x}$ and $\bm{y}$.
If we use the language from classical machine learning, we can say that the optimization in VML makes use of the `gradient' information from the last prompt and the current batch of training data, while the optimization in APE is `gradient-free'.

\paragraph{Numerical score function v.s. Self-evaluation.}
Another important distinction is that APE requires a predefined score function that can give a numeric score to each candidate prompt.
For example, one can use the log-likelihood of a prompt as the score, which normally requires access to the weights of an LLM, hence it is only possible for the open-source models.
In contrast, VML does not require such a score function to evaluate the prompt.
Evaluation of a prompt in VML is done by the LLM itself purely in natural language.
This is more flexible and agnostic to different LLMs (\eg, proprietary or open-source).

\begin{minipage}{1\linewidth}
\captionsetup[algorithm]{font=footnotesize}
\begin{algorithm}[H]
\footnotesize
\caption{\footnotesize APE (Simplified)}\label{alg:ape}
Given: $\mathcal{D}_\mathrm{train} = \{ \bm{x}_n, \bm{y}_n \}^N$, Batch size $M$, Score function $s(\cdot)$;\;
\vspace{1em}

\emph{// Step 1: Sample candidate prompts}

$\mathcal{P} = [ \; ]$

\For{$i=1,\cdots,N/M$}{
    Get a batch of $M$ training examples $\bm{x}_1,\cdots,\bm{x}_M$;\;
    
    $\mathcal{P}.\mathrm{extend}( f_\mathrm{prop.}(\bm{x}_1,\cdots,\bm{x}_M))$ \;\;\;\;
    // See \Cref{fig:ape_template} (left) for the template for $f_\mathrm{prop.}(\cdot)$;
}

\vspace{1em}

\emph{// Step 2: Evaluate candidate prompts}

$\mathcal{S} = [ \; ]$

\For{each $p \in \mathcal{P}$}{
    Get a batch of $M$ training examples $\bm{x}_1,\cdots,\bm{x}_M$;\;

    $s_p = 0$;
    
    \For{$m=1,2,\cdots,M$}{
    $s_p = s_p + s(p; \bm{x}_m, \bm{y}_m)$; \;\;\;\;\;\;\;
    // See \Cref{fig:ape_template} (right) for the template for $s(p; \bm{x}, \bm{y})$;

    }
    
    $\mathcal{S}.\mathrm{extend}( s_p / M)$;
    
}

\Return $p^* \in \mathcal{P}$ s.t. $\mathcal{P}.\mathrm{index}(p^*) = \mathrm{arg\,max}_i \; S[i]$;
\end{algorithm}
\vspace{-0.26in}
\end{minipage}

\begin{minipage}{1\linewidth}
\captionsetup[algorithm]{font=footnotesize}
\begin{algorithm}[H]
\footnotesize
\caption{\footnotesize VML (Same as \Cref{alg:vml} but with more details)}\label{alg:vml_detail}
Given: $\mathcal{D}_\mathrm{train} = \{ \bm{x}_n, \bm{y}_n \}^N$, Initial prompt $\bm{\theta}_0$, Iteration number $T$, Batch size $M$\;

\vspace{1em}

\For{$i=1,\cdots,T$}{

    \emph{// Step 1: (Forward Pass) Use current prompt to do predictions}
    
    Get a batch of $M$ training examples $\bm{x}_1,\cdots,\bm{x}_M$;\;
    
    \For{$m=1,2,\cdots,M$}{
    $\hat{y}_m=f_{\text{model}}(\bm{x}_m;\bm{\theta}_{i-1})$;\;\;\;\;\;\;\;\;\;\;\;\;\;\;\;\;
    // See \Cref{fig:template} (left) for the template for $f_{\text{model}}(\cdot)$;
    
    \vspace{-0.3mm}
    
    }

    \vspace{1em}

    \emph{// Step 2: (Backward Pass) Update the prompt base on the predictions, current batch of data, and current prompt}
    
    $\bm{\theta}_{i}\!\!=\!\!f_{\text{opt}}\big(\{\bm{x}_m,\hat{y}_m,y_m\}_{m=1}^M,\bm{\theta}_{i-1}\big)$;\;\;\;\;\;\;
    // See \Cref{fig:template} (right) for the template for $f_{\text{opt}}(\cdot)$;
    \vspace{-0.25mm}
    
}
\vspace{-0.8mm}
\end{algorithm}
\vspace{-0.26in}
\end{minipage}

\newpage
\section{Effect of Accurate Loss Feedback}\label{app:loss_feedback}

\begin{figure}[h]
    \centering
    \setlength{\abovecaptionskip}{4pt}
    \setlength{\belowcaptionskip}{0pt}
    \renewcommand{\captionlabelfont}{\scriptsize}
    \vspace{-1pt}
    \includegraphics[width=0.85\textwidth]{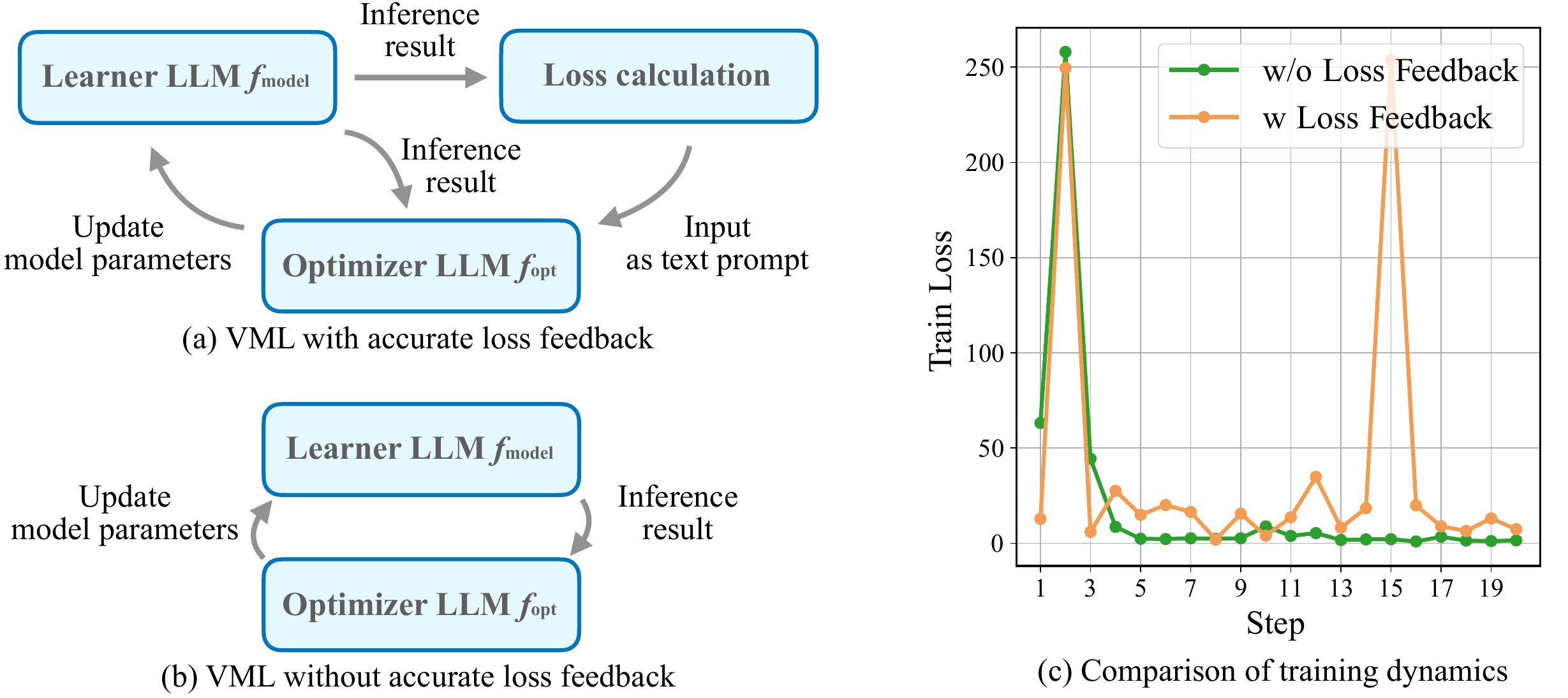}
    \caption{\scriptsize Training dynamics for two different optimization settings in the polynomial regression setting. One has access to the accurate loss computation, and the other does not.}
    \label{fig:abl:poly_loss_fb}
\end{figure}

The VML algorithm at \Cref{alg:vml} specifies that the arguments for $f_\text{opt}(\cdot)$ consist of the inputs $\vx$, the predictions $\hat{y}$, the targets $y$, the current model parameter $\bm{\theta}_{i-1}$ and the optimizer configurations $\bm{\psi}$.
Hence, there is no explicit definition of the loss function for the optimizer (see \cref{fig:template}(right) for an example of the verbalized loss function).
It is up to the optimizer itself to evaluate the difference between the prediction $\hat{y}$ and the target $y$.
We are interested in question that whether having access to the real training loss (defined and computed for logging purpose), mean squared error in this case, can help the optimizer to better navigate the training trajectory.

The orange line in \Cref{fig:abl:poly_loss_fb}(c) shows that having such accurate loss feedback might not help, and might even decrease the performance in this scenario.
One possible explanation is that the single loss value itself does not contain too much information. Moreover, as the exact form of the loss function can be fed to LLM easily, the LLM might spend additional efforts to estimate the exact form of the loss function, which makes the convergence even more difficult. It actually makes intuitive sense that verbalized loss function (\ie, using natural language to explain the target of the loss function) works better in the VML framework.
For example, knowing how does each prediction contributes to the loss value can be more informative and a single overall loss value, since the model might be doing well for some data but not the others, and we only want to improve the model for points with the bad predictions.

\newpage
\section{Numerical Error of LLMs in Representing Symbolic Functions}\label{app:numerical_error}

\begin{figure}[h]
    \centering
    \setlength{\abovecaptionskip}{4pt}
    \setlength{\belowcaptionskip}{-4pt}
    \renewcommand{\captionlabelfont}{\scriptsize}
    \vspace{-1pt}
    \includegraphics[width=0.98\textwidth]{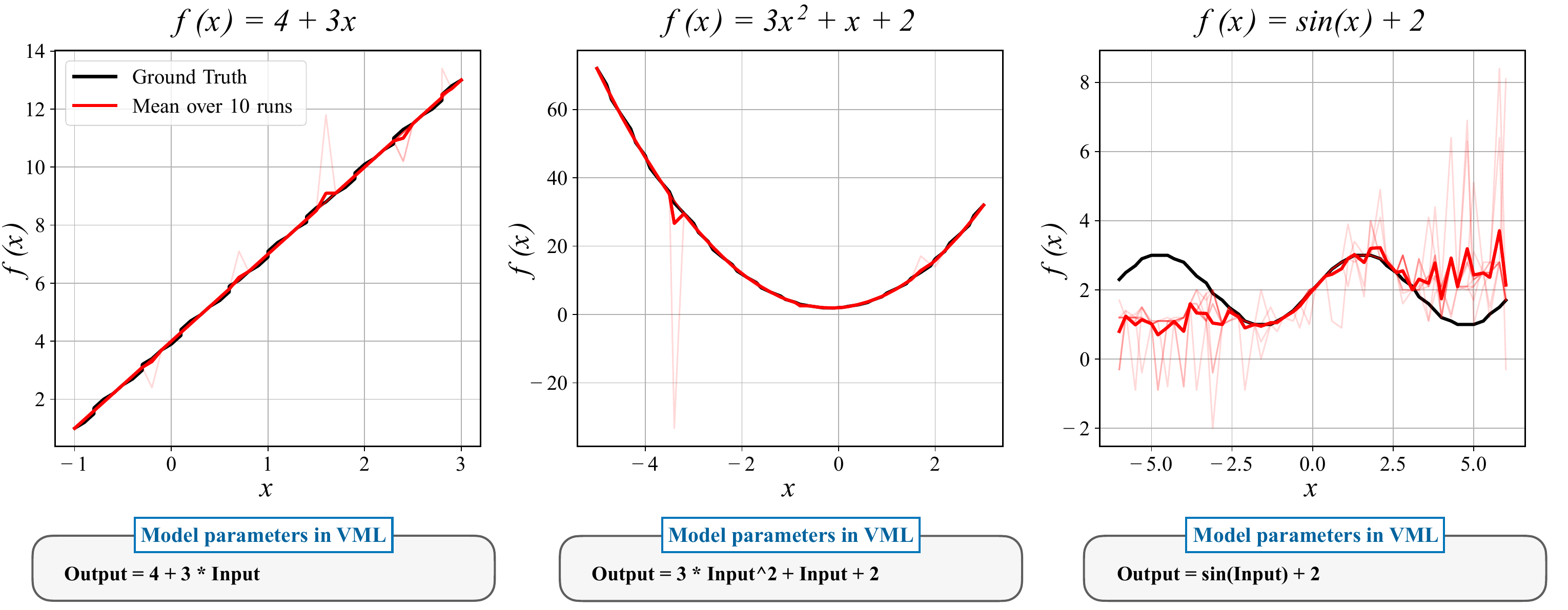}
    \caption{\scriptsize Functions evaluations and numerical error in Llama-3 70B}
    \label{appfig:numerical:llama3}
\end{figure}

\begin{figure}[h]
    \centering
    \setlength{\abovecaptionskip}{4pt}
    \setlength{\belowcaptionskip}{-4pt}
    \renewcommand{\captionlabelfont}{\scriptsize}
    \vspace{-1pt}
    \includegraphics[width=0.98\textwidth]{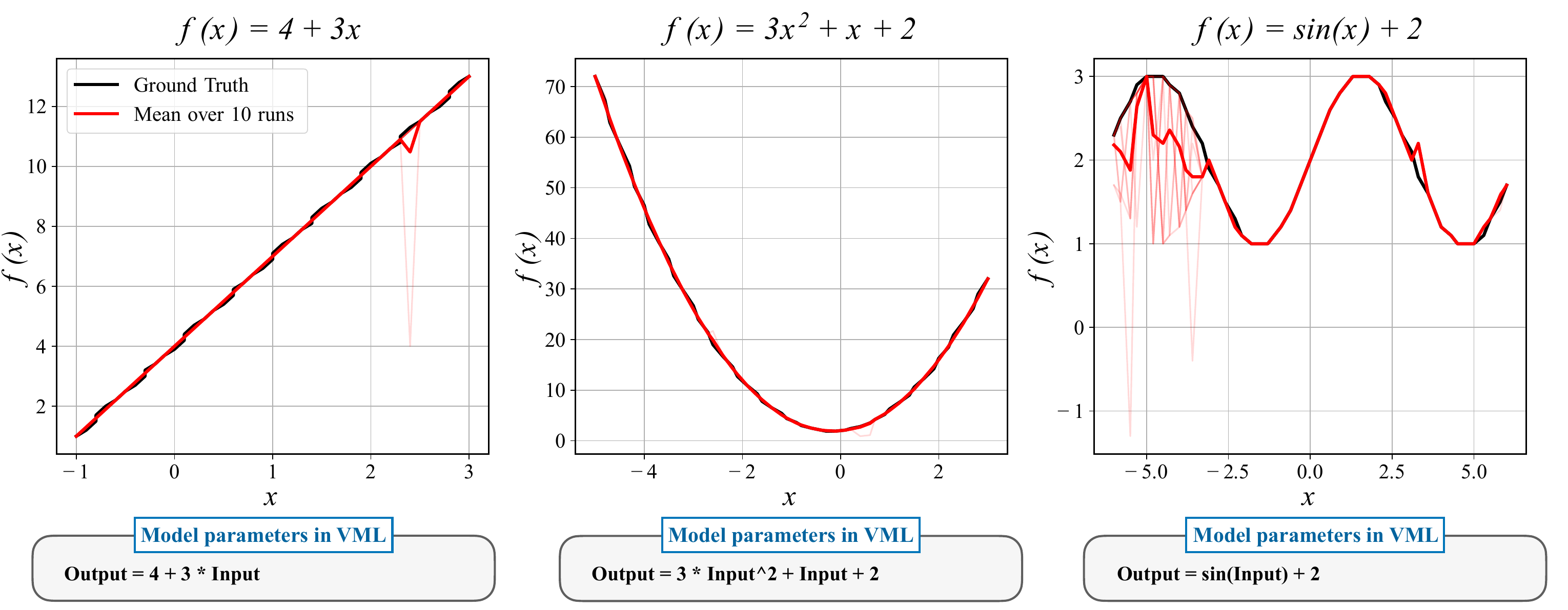}
    \caption{\scriptsize Functions evaluations and numerical error in GPT-4o.}
    \label{appfig:numerical:gpt4o}
\end{figure}

LLMs are designed to do language modeling, rather than exact calculations. 
Hence, their performance on evaluating functions can be unreliable, and might result in error.
\Cref{appfig:numerical:llama3} shows that Llama-3 is very comfortable in evaluating the given linear and polynomial function, as the mean is quite accurate.
The variance over 10 runs is also pretty small, except for one or two points.
However, for a more complex function such as $\sin(x)$, Llama-3 is only able to return small error approximately in the range of $x\in(-2,2)$.
Both the error and the variance are large out side of this range.
This explains the non-smoothness for the function in \Cref{fig:reg:sin}(b; right), which has $\sin(x + 1.0)$ in the learned model parameters.

By switching to the more powerful model, GPT-4o, we can see from \Cref{appfig:numerical:gpt4o} that both the error and the variance decrease.
In particular, for $\sin(x)$, GPT-4o returns smaller error in a larger range, (\ie, $x\in (-2.5, 5.0)$).
This implies that as the capability of LLMs improves, their performance in evaluating more complex functions also improves.

Nevertheless, this is currently still a limitation for VML if the optimizer chooses to use complex mathematical functions as the model parameter.
If the evaluation of the function has an error, then during training, the optimizer will update the model parameters based on noisy signal.
This can lead to large variance in training and slow convergence.
Future work should look into methods for minimizing the numerical error in LLMs function evaluation.

\newpage

\begin{figure}[h]
    \centering
    \setlength{\abovecaptionskip}{4pt}
    \setlength{\belowcaptionskip}{0pt}
    \renewcommand{\captionlabelfont}{\scriptsize}
    \vspace{-1pt}
    \includegraphics[width=1\textwidth]{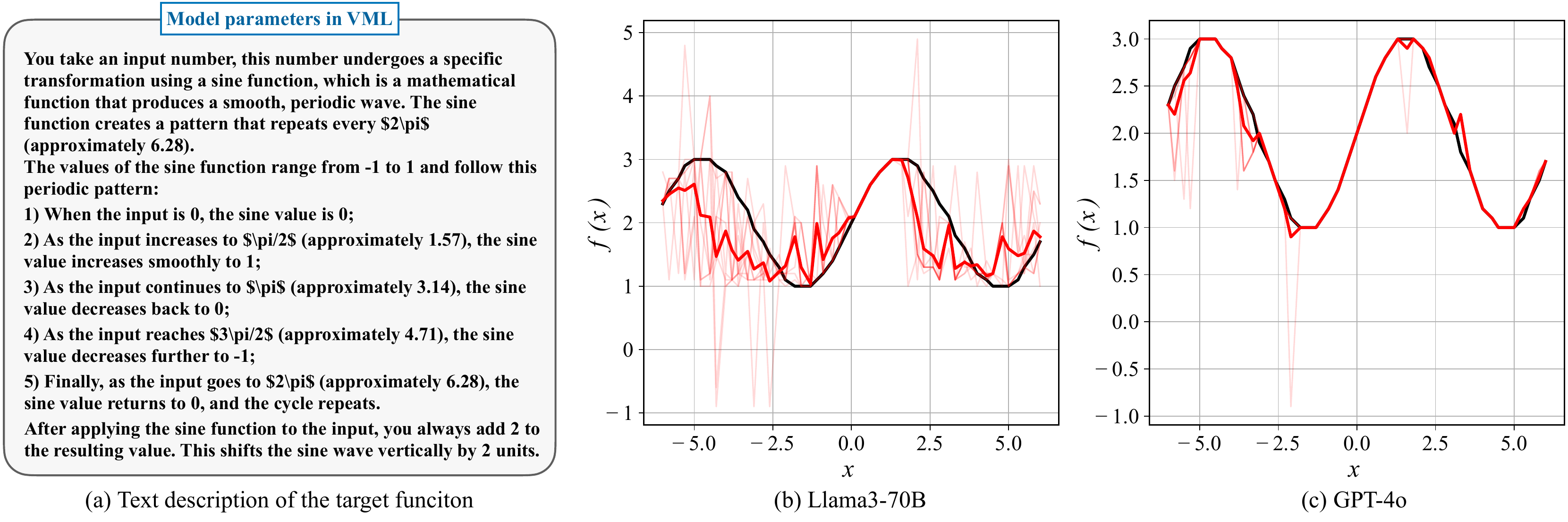}
    \caption{\scriptsize Function evaluations based on the natural language description of the corresponding symbolic sine function.}
    \label{appfig:verbal_sin}
\end{figure}

\Cref{appfig:verbal_sin} shows that if we use natural language to describe the symbolic sine function (see sub-figure(a)), GPT-4o is able to produce more accurate evaluations than using the symbolic function (see (c)).
The accuracy of Llama-3 70B also increases, even though it still under performs GPT-4o (see (b)).
This is likely due to Llama-3 is less capable in instruction following than GPT-4o.
This observation implies that in VML, we might want to instruct the optimizer to avoid using complex symbolic functions in the update and to prefer the natural language description of the function.

\newpage
\section{Mitigating Numerical Error by Tool Calling} \label{app:tool}
In this section, we supplement experiments of Llama-3 70B with a python interpreter.
Despite the fact that LLMs are able to perform numerical data tasks, the incorporation of a python interpreter further improves LLMs ability to deal with numerical values.
Specifically, we use the \emph{open-interpreter}\footnote{GitHub Repo: \url{https://github.com/OpenInterpreter/open-interpreter}} library to add a python interpreter to Llama-3 70B, such that the LLM has the ability to use python programs to evaluate symbolic functions or perform numerical operations. We follow the same experimental settings as in \Cref{sect:sine} (sinusoidal regression of $y=\sin (x) + 2$). The training data is only sampled from $[-3,3]$ with additive Gaussian noise. The in-domain testing data is sampled from the same range, while the out-of-domain testing data is sampled from $[-6,-3]$ and $[3,6]$.

\begin{table}[h]
\renewcommand{\captionlabelfont}{\scriptsize}
 \scriptsize
 \centering
 \setlength{\tabcolsep}{4pt}
 \renewcommand{\arraystretch}{1.3}
 \vspace{1em}  %
 \scalebox{0.83}{
 \label{tab:inpterpreter}
 \begin{tabular}{l|ccc}
 \specialrule{0em}{-14pt}{0pt}
    \multicolumn{1}{l}{Test Domain} & Neural Nets & Llama3 with prior & Llama3 with prior and code interpreter \\
    \shline
    In Domain & 0.0073 & 0.7385 & \cellcolor{Gray}{\textbf{1.0e-05}} \\
    Out of Domain & 0.5840 & 11.1735 & \cellcolor{Gray}{\textbf{1.0e-05}} \\
    Overall & 0.2345 & 5.956 & \cellcolor{Gray}{\textbf{1.0e-05}} \\
  \specialrule{0em}{-7pt}{0pt}
 \end{tabular}}
\caption{
\scriptsize 
Evaluation (using mean squared error $\downarrow$) on sinusoidal regression as in \Cref{fig:reg:sin}(b) for three different models including (1) neural networks, (2) Llama3 with prior, and (3) \emph{Llama3 with prior and code interpreter}.
}
\vspace{-2mm}
\end{table}

From the table, we can observe that with the python interpreter, Llama-3 70B can effectively learn periodic functions, while in the original experiment (\ie, \Cref{fig:reg:sin}(b)), the same LLM is unable to approximate periodic function even with a prior. The results show that the tool-using ability can further improve the learnability of VML. The example logs for inference with the learned model is showed below.

\input{log_linear_reg_interpreter}

\clearpage
\newpage
\section{Connection between Prediction Variance and Model Parameters in VML}\label{app:pred_var}

\subsection{From Vague to Concrete Model Parameters}

\begin{figure}[h]
    \centering
    \setlength{\abovecaptionskip}{4pt}
    \setlength{\belowcaptionskip}{0pt}
    \renewcommand{\captionlabelfont}{\scriptsize}
    \vspace{-1pt}
    \includegraphics[width=0.98\textwidth]{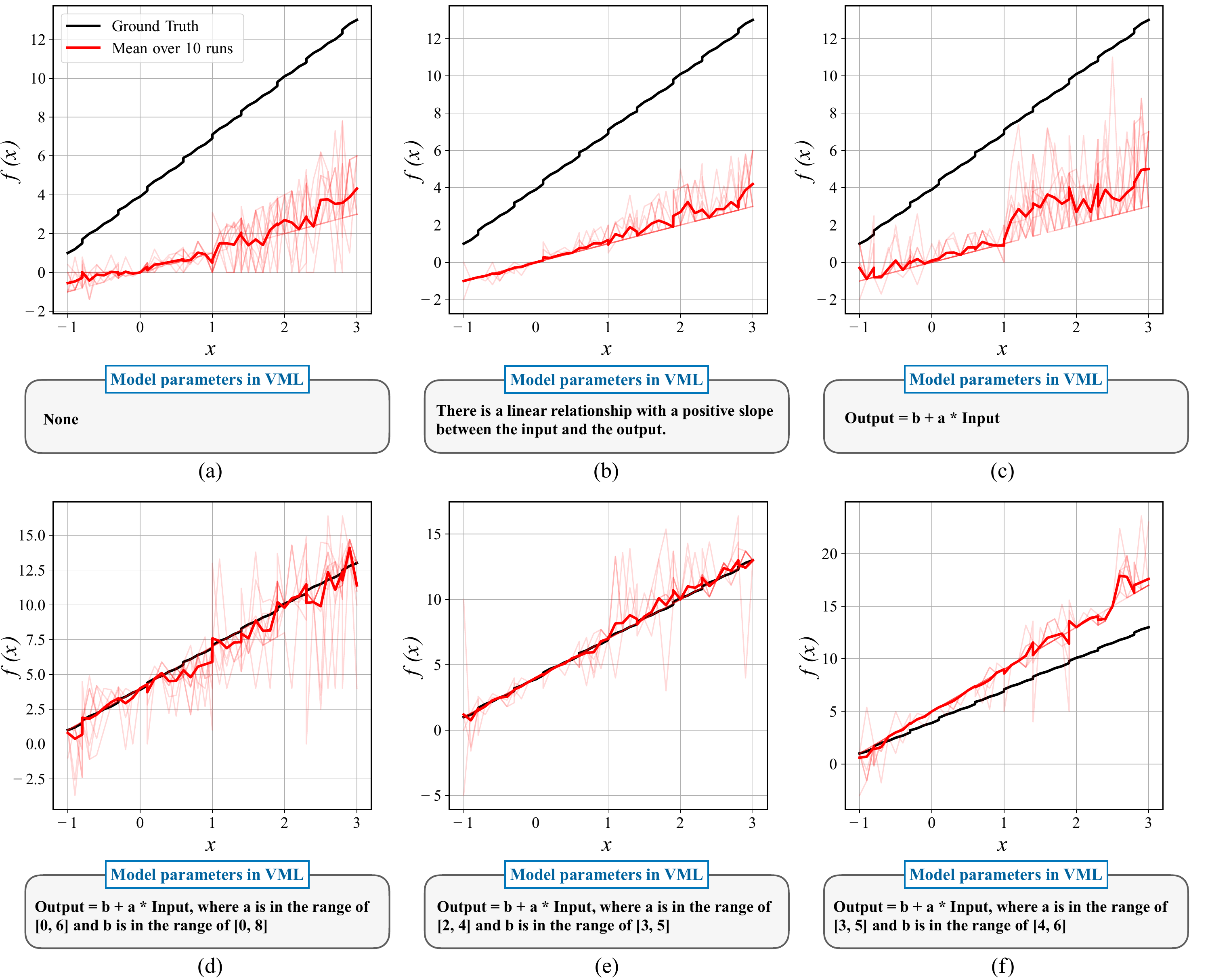}
    \caption{\scriptsize Evaluations on model parameters using vague to concrete descriptions. Results are over 10 runs. The base LLM is Llama-3-70B.}
    \label{appfig:func_var_linear_v2s}
\end{figure}

The model parameters generated by a VML optimizer can be vague or concrete.
We are curious for those with vague descriptions, how would the LLM evaluations look like, and whether they have large variance.
\Cref{appfig:func_var_linear_v2s} shows the results on Llama-3 70B for six different model descriptions, including:
\begin{enumerate}[label=(\alph*)]
    \item None
    \item ``There is a linear relationship with a positive slope between the input and the output.''
    \item ``Output = b + a * Input''
    \item ``Output = b + a * Input, where a is in the range of [0, 6] and b is in the range of [0, 8]''
    \item ``Output = b + a * Input, where a is in the range of [2, 4] and b is in the range of [3, 5]''
    \item ``Output = b + a * Input, where a is in the range of [3, 5] and b is in the range of [4, 6]''
\end{enumerate}
(a) shows that if we only provide the information that the task is a regression task and do not specify the model at all, the LLM tends to predict a linear function (slope $\approx 1$) with increasing variance as $x$ moves away from $0$.
(b) shows that if we specify there is a linear relationship between inputs and outputs, the LLM will predict a linear function with a similar slope as (a) but with smaller variance.
(c) shows that if we specify the explicit form of the linear function, the slope will still be around $1$, but the variance are larger when $x>1$.
(d, e, f) show that by providing a range for the values of the unknown variables, the LLM tends to use the mid-point of the range for the values, and a smaller range does correspond to a smaller variance in prediction.

\newpage
\subsection{Semantic Invariance of Model Parameters}

\begin{figure}[h]
    \centering
    \setlength{\abovecaptionskip}{4pt}
    \setlength{\belowcaptionskip}{0pt}
    \renewcommand{\captionlabelfont}{\scriptsize}
    \vspace{-1pt}
    \includegraphics[width=0.98\textwidth]{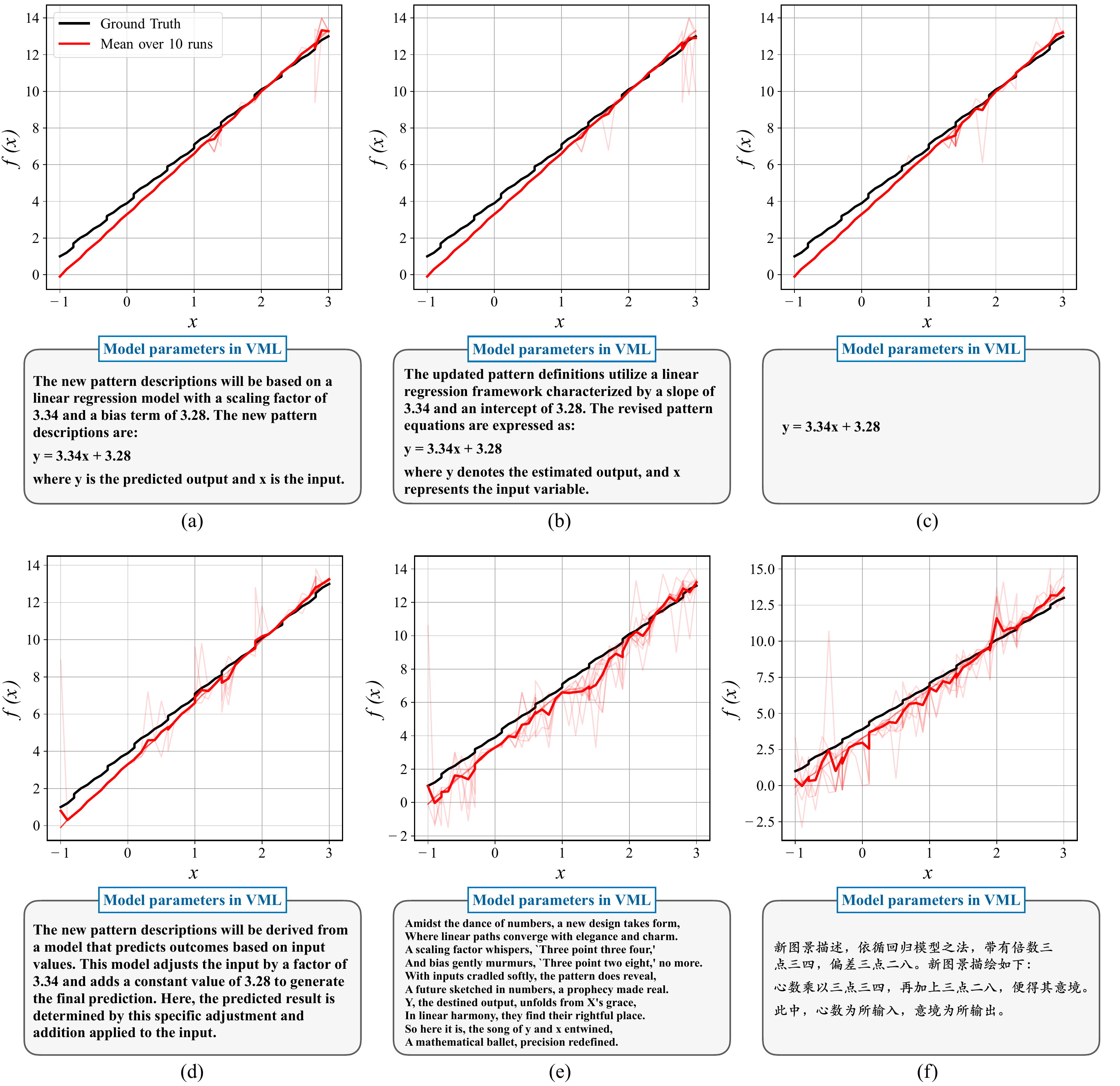}
    \caption{\scriptsize Evaluation on the model parameters from \Cref{fig:reg:linear}(c; Step 15) using six different rephrasing with the same semantic meaning. Results are over 10 runs. The base LLM is Llama-3-70B.}
    \label{appfig:func_var_linear_semantic_invar}
\end{figure}

In natural language, there are different ways to express a concept with the same semantic meaning.
Hence, the model parameters generated by a VML optimizer might vary a lot between runs, even though they are semantically invariant.
We are curious whether such variance in descriptions will lead to variance in model evaluations.
\Cref{appfig:func_var_linear_semantic_invar} shows that results on Llama-3 70B for six different but semantically invariant descriptions of the model from \Cref{fig:reg:linear}(c; Step 15), \ie,:
\begin{enumerate}[label=(\alph*)]
    \item ``The new pattern descriptions will be based on a linear regression model with a scaling factor of 3.34 and a bias term of 3.28. The new pattern descriptions are:
    \\~\\
    y = 3.34x + 3.28
    \\~\\
    where y is the predicted output and x is the input.''
    \vspace{2mm}
    \item ``The updated pattern definitions utilize a linear regression framework characterized by a slope of 3.34 and an intercept of 3.28. The revised pattern equations are expressed as:
    \\~\\
    y = 3.34x + 3.28
    \\~\\
    where y denotes the estimated output, and x represents the input variable.''
    \vspace{2mm}
    \item ``y = 3.34x + 3.28''
    \vspace{2mm}
    \item ``The new pattern descriptions will be derived from a model that predicts outcomes based on input values. This model adjusts the input by a factor of 3.34 and adds a constant value of 3.28 to generate the final prediction. Here, the predicted result is determined by this specific adjustment and addition applied to the input.''
    \vspace{2mm}
    \item ``Amidst the dance of numbers, a new design takes form,\\
    Where linear paths converge with elegance and charm.\\
    A scaling factor whispers, `Three point three four,'\\
    And bias gently murmurs, `Three point two eight,' no more.
    \\~\\
    With inputs cradled softly, the pattern does reveal,\\
    A future sketched in numbers, a prophecy made real.\\
    Y, the destined output, unfolds from X's grace,\\
    In linear harmony, they find their rightful place.
    \\~\\
    So here it is, the song of y and x entwined,\\
    A mathematical ballet, precision redefined.''
    \vspace{2mm}
    \item \begin{CJK*}{UTF8}{gbsn}
    ``新图景描述，依循回归模型之法，带有倍数三点三四，偏差三点二八。新图景描绘如下：
    \\~\\
    心数乘以三点三四，再加上三点二八，便得其意境。
    \\~\\
    此中，心数为所输入，意境为所输出。''
    \end{CJK*}
\end{enumerate}
These rewrites are generated by GPT-4o based on (a).
The description (a) is the original $\bm{\theta}_{15}$ from \Cref{fig:reg:linear}(c; Step 15).
(b) rephrases the descriptions from (a) slightly.
(c) only keeps the symbolic equation from (a).
(d) is a rewrite without using math expression.
(e) uses the poetry style.
(f) is a translation of (a) into Literary Chinese.
The results in \Cref{appfig:func_var_linear_semantic_invar}(a,b,c) are similar, and have small variance across the 10 runs.
The results in \Cref{appfig:func_var_linear_semantic_invar}(d,e,f) are also very accurate on average.
However, the poetry rewrite (e) and the Chinese rewrite (f) do have slightly larger variance.
Overall, we see that if the various descriptions preserve the same semantic, then their evaluations through Llama-3 70B are likely to be similar.

\newpage
\section{A Probabilistic View on VML}
\label{app:prob_view}

The output of a language model usually comes with randomness. In the paper, we typically consider to set the temperature in LLMs as zero to remove the randomness from sampling, which indicates that LLMs will always output the text with the largest probability (\ie, largest confidence logit). However, we want to highlight that such a sampling process actually gives us another probabilistic perspective to study VML. We will briefly discuss this perspective here.

\subsection{Posterior Predictive Distribution}

Because we can easily sample multiple possible model parameters by setting a proper temperature for the optimizer LLM, we view it as a way to sample multiple learner models. This is well connected to Bayesian neural networks, where Bayesian inference is applied to learn a probability distribution over possible neural networks. We start by writing the posterior predictive distribution ($\mathcal{D}$ is training data):
\begin{equation}
    p(\hat{y}|\mathcal{D})=\int_{\bm{\theta}} p(\hat{y}|\bm{\theta})p(\bm{\theta}|\mathcal{D}) d \bm{\theta}=\mathbb{E}_{p(\bm{\theta}|\mathcal{D})}\big{\{} p(\hat{y}|\bm{\theta}) \big{\}}
\end{equation}
where we can easily sample multiple model parameters $\bm{\theta}$ and compute its probability with logits. Specifically, we have that $p(\bm{\theta}|\mathcal{D})=\prod_{t=1}^n P(\theta_t|\theta_{1},\cdots,\theta_{t-1},\mathcal{D})$. Using this idea, it is actually quite easy to obtain the ensembled output that is weighted by posterior distribution.

\subsection{From Functions to Stochastic Processes}

With non-zero temperature, we can view the output of LLMs as a sampling process from a distribution over text tokens, which means each output token can be viewed as a random variable. Then the output of LLMs is effectively a sequence of random variables, and therefore it is easy to verify that it is a stochastic process. This view makes it possible for VML to perform probabilistic modeling.

\newpage
\section{Discussions on Natural Language Model Parameters}\label{app:model_parameters}

There are many interesting properties regarding the natural language model parameters. Many traditional machine learning models can be revisited in the scenario where model parameters are text prompts in the LLM.

\subsection{Different Mechanisms to Update Model Parameters for Direct Optimization}

\textbf{Naive re-writing}. Given the model parameters $\bm{\theta}_t$ at the step $t$, the simplest way to update the model parameters at the step $t+1$ is to use whatever the optimizer generates. We denote the optimizer LLM generates the new model parameters $\bm{\theta}_{\text{new}}^t$. This is essentially
\begin{equation}
    \bm{\theta}_{t+1}\leftarrow\bm{\theta}_{\text{new}}^t.
\end{equation}
An simple extension to naive re-writing is to add a text prompt to instruct the optimizer LLM to take the previous model parameters $\bm{\theta}_t$ into consideration at the step $t+1$. Thus we have the conditional re-writing, namely $\bm{\theta}_{t+1}\leftarrow f_{\text{opt}}(\bm{\theta}_t)$. This is also what we use in the main paper.

\textbf{Incremental updating}. Alternatively, we can choose to update the model parameters in an incremental fashion without remove the previous model parameters completely. We denote the optimizer LLM generates the new model parameters $\bm{\theta}_{\text{new}}^t$. Then the model parameters $\bm{\theta}_{t+1}$ at the step $t+1$ is
\begin{equation}
    \bm{\theta}_{t+1}\leftarrow\{\bm{\theta}_t,\bm{\theta}_{\text{new}}^t\}.
\end{equation}
However, the incremental updating will make the model parameters an increasingly longer text prompt. This is not ideal since the context window of a LLM is typically quite limited. The incremental updating mechanism can be interpreted as using a small learning rate to train the learner. This will easily lead to bad local minima (because the previous incorrect model parameters will be kept and may affect the future learning as a prior knowledge in the text prompt), but it may improve the training convergence.

\textbf{Incremental updating with summarization}. To avoid the infinite increasing length of model parameters, we can instruct the optimizer LLM to summarize the previous model parameters into a fixed length. This yields
\begin{equation}
   \bm{\theta}_{t+1}\leftarrow\{\underbrace{\mathcal{C}(\bm{\theta}_t)}_{\text{fixed token length}},\bm{\theta}_{\text{new}}^t\}.
\end{equation}
where $\mathcal{C}(\cdot)$ is some text summarization scheme.

\textbf{Connection to standard optimizers}. There are many interesting connections between the optimizer LLM and the standard numerical optimizer. Usually the behavior of the optimization is determined by the optimizer parameters $\bm{\psi}$ which is also a text prompt. This is usually a text description of the target of the optimizer. For example, we can instruct the optimizer LLM to serve as the first-order optimizer (\eg, momentum SGD) and feed all the necessary information into the text prompt. Then the optimizer LLM will essentially become an optimizer mapping function that maps all the necessary information (including the previous model parameters) to the model parameters of the next step. To implement the momentum in the optimizer LLM, one can simply instruct the optimizer LLM to maintain the previous model parameters as much as possible. This is to say, everything we want to implement in the optimizer are realized through text prompts. It will inevitably depend on the instruction-following ability of the LLM, and it is possible that there will be some unrealizable optimization functionalities (\eg, we instruct the optimizer LLM to be a second-order optimizer and the optimizer LLM may be likely to ignore this instruction). However, we want to highlight that as LLMs become more powerful, this problem will be less and less significant. In general, implementing an advanced optimizer in VML is still an important open problem.

\subsection{Occam's Razor, Constrained-length Model Parameters, and Kolmogorov Complexity}

We are interested in how Occam's razor can be applied in VML. One natural way of doing so is to constrain the model parameters to be a small and fixed length. This essentially is
\begin{equation}
    \bm{\theta}_{t+1}\leftarrow\{\underbrace{\mathcal{C}(\bm{\theta}_t)}_{\text{fixed token length}},\underbrace{\bm{\theta}_{\text{new}}^t}_{\text{fixed token length}}\}.
\end{equation}
We can see that as long as we constrain the text token length of the model parameters to be small, the learner will perform an automatic model simplification, as it will try to discover the data pattern with concise and simple text. There are many more ways to implement the Occam's razor in VML. More interestingly, it is also possible to incorporate a structural constraint to the model parameters. For example, it can be causal knowledge (\eg, text representation of a causal graph), logic formula or decision trees. Our work opens up many more possibilities on Occam's razor in VML, and rethinking the form of Occam's razor in VML is very crucial in unlocking the strong interpretability and controllability of inductive biases.

Another perspective on the length of the model parameters in VML is related to Kolmogorov complexity~\citep{Kolmogorov1968ThreeAT}, which is defined as the shortest effective description length of an object.
The principle of Occam's razor is basically saying that hypotheses with low Kolmogorov complexity are more probable~\citep{Vitanyi1998Minimum}.
By constraining the length of model parameters to be small, we are effectively trying to find the minimum description length (MDL) of a model in natural language.
The theoretic Kolmogorov complexity of a model is usually impossible to compute, however, VML might provide an estimation for Kolmogorov complexity by using the shortest effective length of the learned model parameters in natural language.

\subsection{Connection to Nonparametric Models and In-context Learning}
\label{app:model_parameters:icl}

\emph{Nonparameteric} methods get around the problem of model selection by fitting a single model that can adapt its complexity to the data~\citep{wasserman2006all, orbanz2010bayesian, Gershman2011ATO}.
These methods allow the model complexity to grow with the number of observed data.
This is different to \emph{parametric} models which have fixed number of parameters.
In VML, as showed in \Cref{sect:poly}, the model complexity is also flexible and adapts to the data during training.
Similarly, the concept of in-context learning (ICL) can also be understood as nonparametric methods in the lens of LLMs as function approximators.
ICL denotes the method of using LLMs to solve new tasks by only providing the task demonstrations or examples in the prompt with natural language.
Given a new data point, an LLM predicts its output using information in the provided demonstrations.
From the perspective of VML, ICL in an LLM essentially defines a nonparametric model implicitly using the demonstrated examples in the natural language space.

\subsection{Distinctions between the Data Dimension and the Parameter Dimension}
We would like to point out that there is a distinction between the input data dimension and the parameter dimension. 
For example in \Cref{app:hd_ablation}, even though the input data dimension is 10-D, the parameter space (\ie, the description of the model in natural language) is actually much larger than 10-D. As we are optimizing the natural language parameters, rather than the input data. In our experiments, the natural language based parameters can have the dimension of 10 tokens (see \Cref{fig:reg:poly}(c) model parameter $\theta_1$) to 600 tokens (see \Cref{fig:cls:blobs}(c) $\theta_{81}$).

The task we show in \Cref{fig:cls:blobs} is a 2-D plane binary classification task. Each data point on the plane only has 2 dimensions / features. This, of course, does not mean that the optimization dimension is only in 2-D space. Depending on the model class we choose for the task, the parameter space can have various sizes of dimensions, or even infinite dimension. For example, if we use a parametric model such as a three-layer neural network with weights of shapes $[2 \times 10]$ and $[10 \times 1]$, then the parameter space has 30 dimensions. However, if we use a non-parametric model such as a decision tree, then we do not have a fixed number of parameters. The number of parameters for a decision tree depends on the number of nodes in the tree, which grows as the tree adapts to the training data.

Contrary to the two classic numerical based models above (\ie, neural nets and trees) where the parameters are numerical values, one innovative concept of VML, as we have described in \Cref{sect:numerical2vml} to \ref{sect:opt_main}, is to use natural language space as the parameter space. This means that the optimization of a model in VML is done by changing the text, which might happen to be digits as digits is a type of text, but it does not have to be digits. 

For example, assuming we have the space of 100 tokens available for describing the model, then the optimizer is free to use these 100 tokens to describe the model with a formula `$y=ax+b$', where `$a$' and `$b$' will be the only two parameter here in the classical sense, but in VML the parameters are all the token in the string of `$y=ax+b$'. This is why in \Cref{sect:poly} Polynomial Regression and \Cref{fig:reg:poly}, the optimizer is able to update the VML model parameters from `output = 2.5 * input + 1.5' to `output = 2.2 * input \textasciicircum 2 + 1.8* input + 0.6'. This optimization step is not possible in the classical way if we only view `$a$' and `$b$' as the parameters.

Therefore, in each optimization step, from VML’s perspective, the optimizer can add new tokens and optimize the existing tokens. These tokens might correspond to adding new parameters (\eg, Step 2 in \Cref{fig:reg:poly}: from `$y=ax+b$' to `$y=ax^2+bx+c$') or optimizing the existing parameters (\eg, Step 3 in \Cref{fig:reg:poly}) in the classical definition of parameters. Since the real parameters in VML are the tokens, both of these two operations can happen at the same step. Hence, we should really use tokens to understand the parameter space in VML.

Another example is the medical image classification task in \Cref{sect:medical} and \Cref{fig:cls:medical}. The optimizer updates the model to consist of only semantic descriptions of features without any numbers. In this case, if not using number of tokens as the parameter dimension as in VML, it is hard to define the corresponding classical parameters and the dimensions.

As for the experiment in \Cref{fig:cls:blobs}, like the other experiments, we do not add any prior information on the model class, so the optimizer is free to choose any appropriate class of model to solve the task. The \Cref{fig:cls:blobs} shows an example run which ends up with using decision trees as the model class. As we mentioned above, the learned model is also a result of optimization in the text space, which is in the dimension of the number of tokens. Even if we do use a classical decision tree algorithm (instead of optimizing in the text space with VML), assuming we get the model in \Cref{fig:cls:blobs}(c), the parameter dimension is still much larger than 2 as this decision tree has many nodes (e.g., the number of if-else pairs).

\newpage
\section{Broader Discussions}
In this section, we use the format of Q\&A to discuss a list of interesting topics that are loosely related to VML, but are more broadly tight to the capability of LLMs.
Some of the questions might seen philosophical or ideological, but were asked by fellow researchers before.
Nevertheless, we still include them into this section in case some readers find them insightful.

\subsection[How is `the function should be $y = m * x + b$' more interpretable?]
{\textnormal{\textbf{How is the optimizer's statement `the function should be $y = m * x + b$' more interpretable than learning a linear function?}}}

The interpretability from VML is on the framework itself. Using natural language to characterize the model can reveal exactly what pattern the model learns from data, which is very different from training neural networks from scratch. As for the case of regression problems in the paper, interpretability comes from (1) automatic model selection with explanations: this is different from common practice where we assume the data is linear and use a linear regression model. In our experiment, we don’t have such a prior and the optimizer will learn this linear pattern purely by exploring the data. The closest equivalent from classical ML methods would be to train an ``universal approximator'', \eg, a neural networks, which might decide to fit a function that is roughly linear, but has a lot more parameters and less interpretable; (2) another source of interpretability comes from the property that the user can easily interact with the optimizer and follow up with more questions to seek explanations.

\subsection[Is controlling the hyperparameters of LLM optimizers more difficult?]
{\textnormal{\textbf{Is controlling the hyperparameters of LLM optimizers, such as learning rate and regularization, more difficult than controlling those of traditional ML optimizers?}}}

Exploring the hyperparameters of LLM optimizers is important yet challenging. It is a great research question for VML. One of the reasons that VML is particularly interesting is that it brings a lot of new research questions.

LLM optimizers have both advantages and disadvantages. The precise control of learning rate and momentum can be difficult. However, adding the qualitative effect of high/slow learning rate and momentum is in fact quite easy. One can simply use language to describe it. In our optimizer prompt, we use the concept of momentum (\eg, “update the model parameter without changing it too much” and provide a constant amount of optimization histories). In terms of regularization, it is also easy to add regularization to control the complexity of the model in VML, \eg, the word length of the model parameters (\ie, a form of Occam’s razor). A qualitative hyperparameter control for LLM optimizers is simple, while this can be challenging for classic ML.

\subsection[Are language models fundamentally restricted for machine learning tasks?]
{\textnormal{\textbf{LLMs are optimized for natural language understanding and generation, not for numerical data tasks typically associated with machine learning. Are LLMs fundamentally restricted for machine learning tasks?}}}

Numerical data tasks are heavily studied in LLMs, for example, mathematical problem-solving. The popular MATH dataset~\citep{hendrycksmath2021} requires strong numerical data processing from LLMs, and this dataset is used as a standard evaluation benchmark for LLMs. Moreover, there exists many LLMs (\eg, DeepseekMath~\citep{shao2024deepseekmath}, WizardMath~\citep{luo2023wizardmath}) that are capable of solving competition-level mathematics problems.

Moreover, LLMs have shown remarkable potential in numerical data tasks for machine learning, and our work is one of the first methods to reveal such a potential. Some concurrent works~\citep{requeima2024llm, yuksekgonul2024textgrad} also gave empirical evidence that LLMs can be fundamentally suitable for machine learning tasks.

Verbalized machine learning aims to provide a framework for LLMs to deal with machine learning tasks, with the ability to fully interpret the learned knowledge with natural language. We believe this framework will be increasingly more powerful, as LLMs get more powerful. We have already observed the performance improvement of VML by switching from Llama-3 to GPT-4o.

\subsection[Is it meaningful to use LLMs for applications such as machine learning?]
{\textnormal{\textbf{The fundamental nature of LLMs is to predict (the next token) based on a probability distribution over the vocabulary. One might argue this process is based on statistical choice rather than on true understanding. Is it meaningful to use LLMs for applications such as machine learning tasks?}}}

We believe VML does represent a meaningful direction to explore, as there is current no evidence that LLMs can not perform ML tasks. On the other hand, we already have quite a few applications that demonstrate the effectiveness of VML (\eg. medical image classification). In fact, even in-context learning can already perform a few ML tasks (as introduced by GPT-2 and GPT-3 papers \citep{radford2019language, brown2020language}). We believe there are a lot of applications to be unlocked in the VML framework.

Whether one should use LLMs for tasks other than language modeling is indeed an important open question, which is currently under active research with a significant number of researchers in the field investigating the boundary of LLMs’ capability, and trying to explain the `seemingly' emergence of such abilities from the simple language modeling training objective.

The argument that LLMs can not elicit true understanding due to its statistical training is debatable. Firstly, it is unclear what it means to train a model based on true understanding. One can not perform such a training without an explicit form of loss function. On the other hand, there are some analyses that show that next-token prediction induces a universal learner~\citep{malach2023auto}. Secondly, we believe that there is a distinct difference between low-level statistical training and high-level knowledge understanding. Whether one can induce another is unknown and is also out of the scope of our paper.

Currently, there has already been substantial evidence that LLMs possess a form of understanding that is functionally relevant for many real-world tasks. The fact that they can consistently generate useful and accurate outputs across various domains, including numerical math~\citep{hendrycksmath2021, alpha2024ai}, theorem proving~\citep{yang2023leandojo}, biology~\citep{liu2023chatgpt} (just to name a few), challenges the argument that LLMs lack real understanding.

Hence, we believe to argue against the use of LLMs for tasks other than language modeling, such as math related problems, will require an equally substantial amount of empirical evidence or theoretical proof, which is missing at the moment.

\subsection[What about Hallucination? Can we trust LLMs to handle complex tasks such as VML?]
{\textnormal{\textbf{Hallucination remains a significant issue for LLMs. How can we trust them to handle complex tasks such as VML?}}}
\label{app:hallucination}

Even though hallucination is an observation associated with LLMs, it does not fundamentally limit the performance of VML. 
Note that by definition, hallucination is an event with a low probability, if an LLM always hallucinates, we will not call it hallucination and it will not be able to outperform humans in many benchmarks.
In the case of VML, there are only two types of LLM calls, \ie, $f_\mathrm{model}(\cdot)$ and $f_\mathrm{opt}(\cdot)$.
Let's go through what happen will if there is a hallucination for each of these calls, and why hallucination is not a fundamental limitation for VML.

\paragraph{Hallucination in $f_\mathrm{model}(\cdot)$} is when the LLM does not follow the model parameter to infer the output of a given input.
For example, if the model parameter is `output = 4 + 3 * Input' and it returns an incorrect answer, this will be an example of hallucination.
This can happen when we apply it to numerical tasks (see \Cref{app:numerical_error}).
However, based on the fact that hallucination is a low probability event, it will not hallucinate random outputs for most of the training data.
Therefore, the set of predictions $\hat{\bm{y}}$ would still provide useful information for the optimizer, even though it is noisy.
This is exactly what happened for many of our experiments.
For instance, in \Cref{sect:poly} \Cref{fig:reg:poly}(b; both mid and right) we can see a small outlier (a dent) in the plot of the quadratic function, which is due to such hallucination.
Nevertheless, VML is still able to learn a very good model.

\paragraph{Hallucination in $f_\mathrm{opt}(\cdot)$} is when the optimizer does not produce a sensible model parameter for the current step of training.
From our experiment section, we can indeed see that many of our case studies have a non-monotonic training loss, some of the fluctuation can be explained by hallucination (see failure cases in \Cref{app:failure_cases} for example).
However, since hallucination is a low probability event, an unsatisfied model parameter will most likely be corrected in the next step of optimization.
Therefore, in most runs, VML eventually learns a very good model.
For cases where it cannot correct itself from the hallucination, they will be identified as failure cases by simply checking the learn model, the training loss, or the test set performance.
As discussed in \Cref{app:failure_cases}, by adding more detailed instruction to the optimizer prompt template and switching to a more powerful LLM, these failure cases (a superset of hallucination) will occur less frequently.

Across our experiments, we did not witness any empirical evidence that hallucination is affecting the reliability of VML. The core of VML is that a model is characterized by the text prompt of an LLM. Whether the model parameter is good depends on its downstream performance (\eg, the training accuracy). Therefore, if a model parameter works well in the downstream task, it is highly unlikely that the model parameter is based on hallucination, because the hallucinated text is unlikely to demonstrate consistently good performance for all the data points. If the learned well-performing model parameter seems unexpected, it is more likely that the LLM-based optimizer discovered new knowledge from the training data than have hallucinated.

More importantly, hallucination in LLMs does not limit the usefulness and significance of using LLMs to solve numerical tasks (such as solving math problems). Therefore, we don’t view hallucination as a problem for VML, but rather an opportunity for future improvement. In reality, we observe that hallucination does not limit the development of utilizing LLMs to solve math problems, which requires the capability of LLMs in understanding numbers and mathematical functions. The math task even requires more precise reasoning from LLMs. However, LLMs has still achieved tremendous progress in math reasoning, despite the potential of hallucination. For example, DeepMind’s AlphaProof\citep{alphaproof2024ai} (based on Gemini) recently reached IMO silver level performance. One can also observe the progress of LLMs on the MATH dataset from around 15\% \citep{touvron2023llama2} to 84\% \citep{yang2024qwen2} (The MATH dataset has competition-level mathematics problems with many numerical computations). The inherited problem of hallucination does not seem to invalidate the field of AI for math. Therefore, it is hard to be certain, in particular without empirical evidence, that hallucination will be a critical limitation for VML.

We agree that current LLMs have many problems (\eg, hallucination, finite context-window). All these shortcomings of LLMs are being actively studied today. We believe VML is actually an orthogonal \& independent contribution to these existing LLM research topics. VML studies how to enable interpretable learning using natural language. As LLM gets more powerful and more faithful, VML will also become more useful.

We also want to highlight that the goal of this work is not to propose a method to replace numerical machine learning, nor to claim superiority of VML over numerical machine learning. We are simply sharing this new possibility of doing learning in the LLMs era, and showing evidence that it can indeed learn for many tasks. VML does have many features which the numerical machine learning does not have, such as interpretability and adding prior with natural language, but it is also not scalable at the moment.

\newpage
\section{Complete Training Template at Initialization}

\subsection{Linear Regression}
\begin{figure}[h]
    \centering
    \setlength{\abovecaptionskip}{4pt}
    \setlength{\belowcaptionskip}{-35pt}
    \renewcommand{\captionlabelfont}{\scriptsize}
    \vspace{-15pt}
    \includegraphics[width=.85\linewidth]{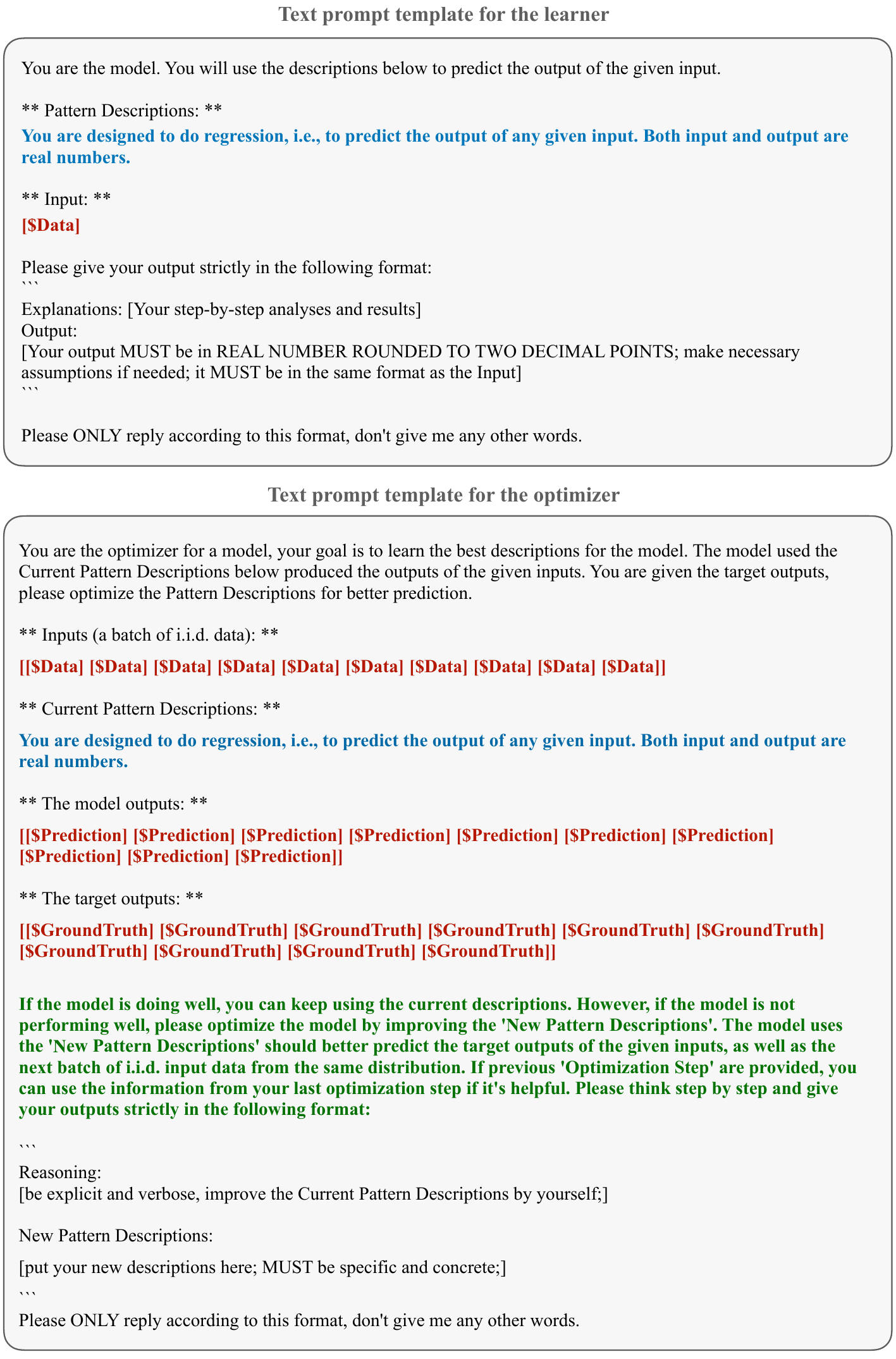}
    \caption{\scriptsize Prompt templates of VML for the learner and optimizer for the linear regression (Llama-3-70B without prior).}
    \label{appfig:linreg_prompt}
\end{figure}

\newpage
\subsection{Polynomial Regression}
\begin{figure}[h]
    \centering
    \setlength{\abovecaptionskip}{4pt}
    \setlength{\belowcaptionskip}{-25pt}
    \renewcommand{\captionlabelfont}{\scriptsize}
    \vspace{-5pt}
    \includegraphics[width=.85\linewidth]{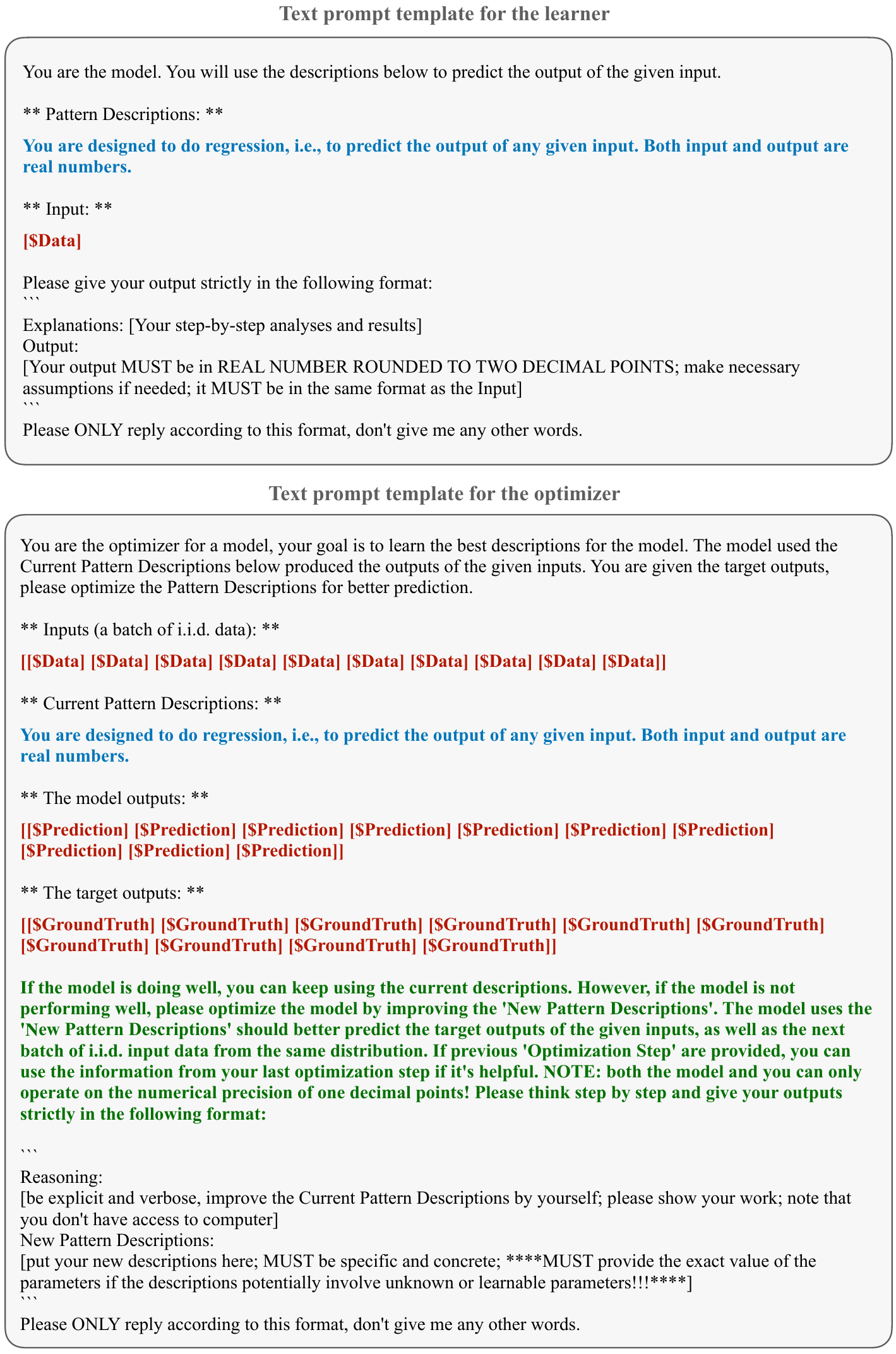}
    \caption{\scriptsize Prompt templates of VML for the learner and optimizer for the polynomial regression (Llama-3-70B without prior).}
    \label{appfig:polyreg_prompt}
\end{figure}

\newpage
\subsection{Sinusoidal Regression}
\begin{figure}[h]
    \centering
    \setlength{\abovecaptionskip}{4pt}
    \setlength{\belowcaptionskip}{-25pt}
    \renewcommand{\captionlabelfont}{\scriptsize}
    \vspace{-5pt}
    \includegraphics[width=.85\linewidth]{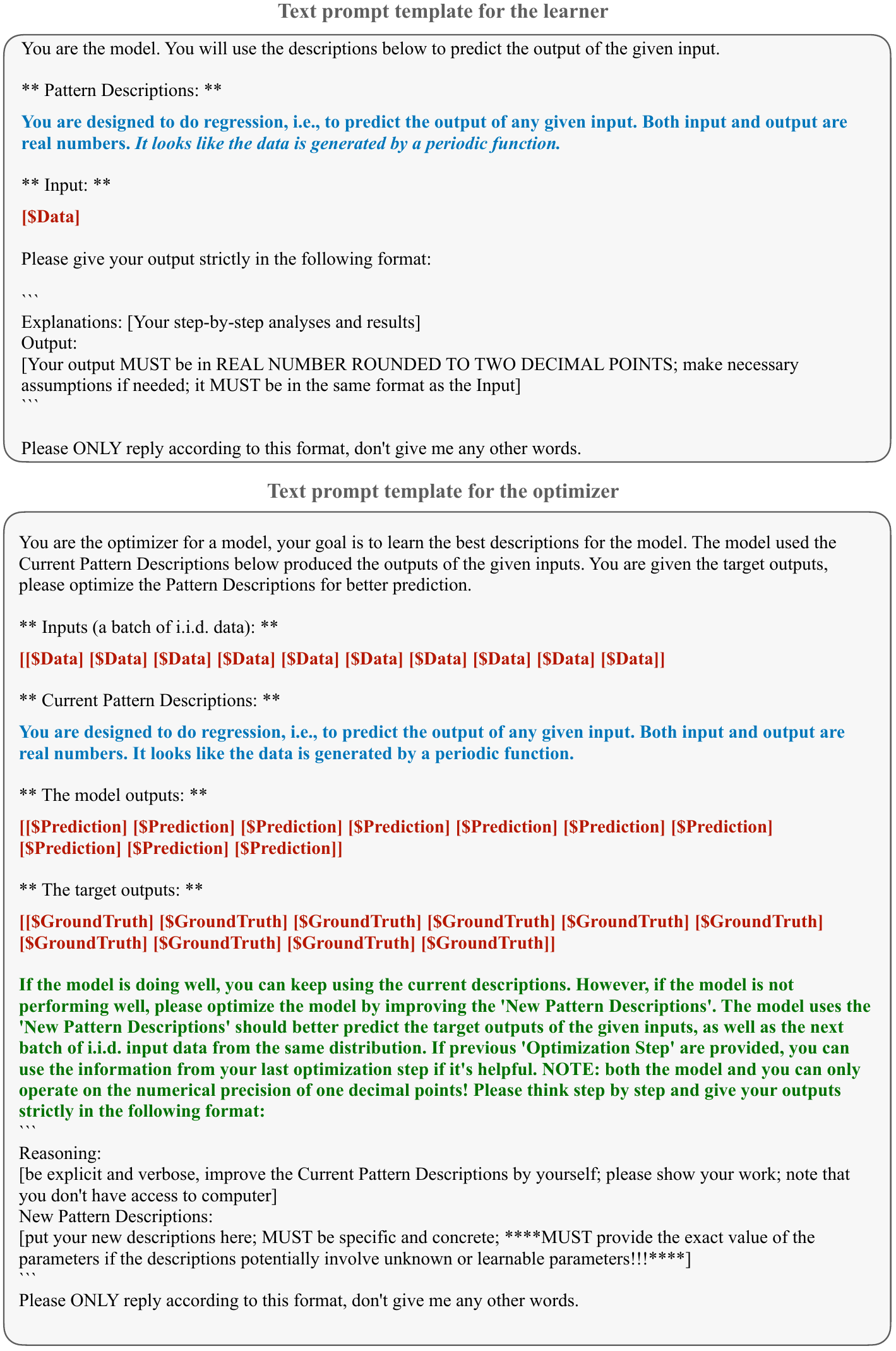}
    \caption{\scriptsize Prompt templates of VML for the learner and optimizer for the sinusoidal regression (GPT-4o with \emph{prior}).}
    \label{appfig:sinreg_prompt}
\end{figure}

\newpage
\subsection{Two Blobs Classification}
\begin{figure}[h]
    \centering
    \setlength{\abovecaptionskip}{4pt}
    \setlength{\belowcaptionskip}{-25pt}
    \renewcommand{\captionlabelfont}{\scriptsize}
    \vspace{-5pt}
    \includegraphics[width=.85\linewidth]{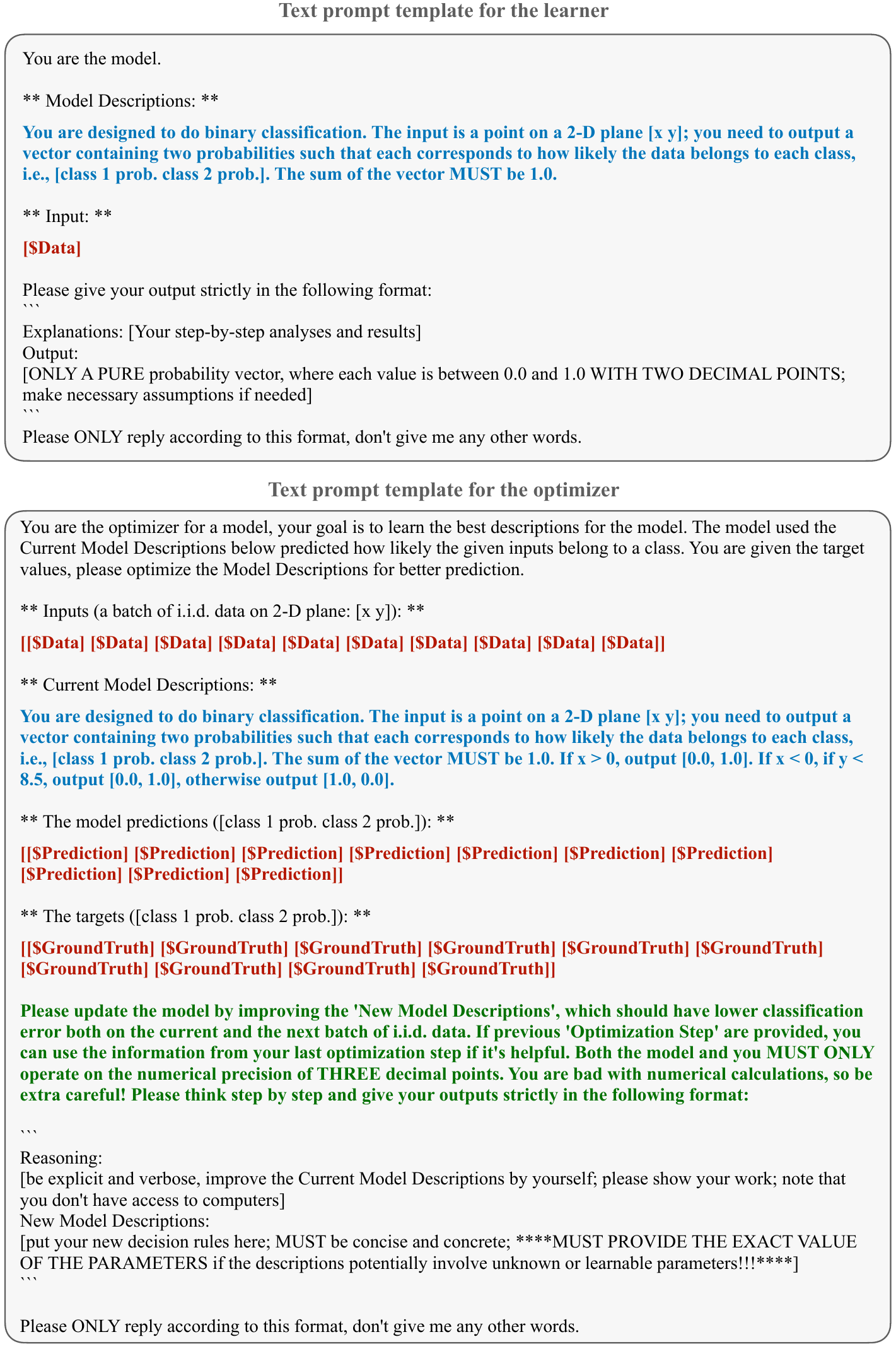}
    \caption{\scriptsize Prompt templates of VML for the learner and optimizer for the two blobs classification (Llama-3-70B without prior).}
    \label{appfig:twoblob_prompt}
\end{figure}

\newpage
\subsection{Two Circles Classification}
\begin{figure}[h]
    \centering
    \setlength{\abovecaptionskip}{4pt}
    \setlength{\belowcaptionskip}{-25pt}
    \renewcommand{\captionlabelfont}{\scriptsize}
    \vspace{-5pt}
    \includegraphics[width=.85\linewidth]{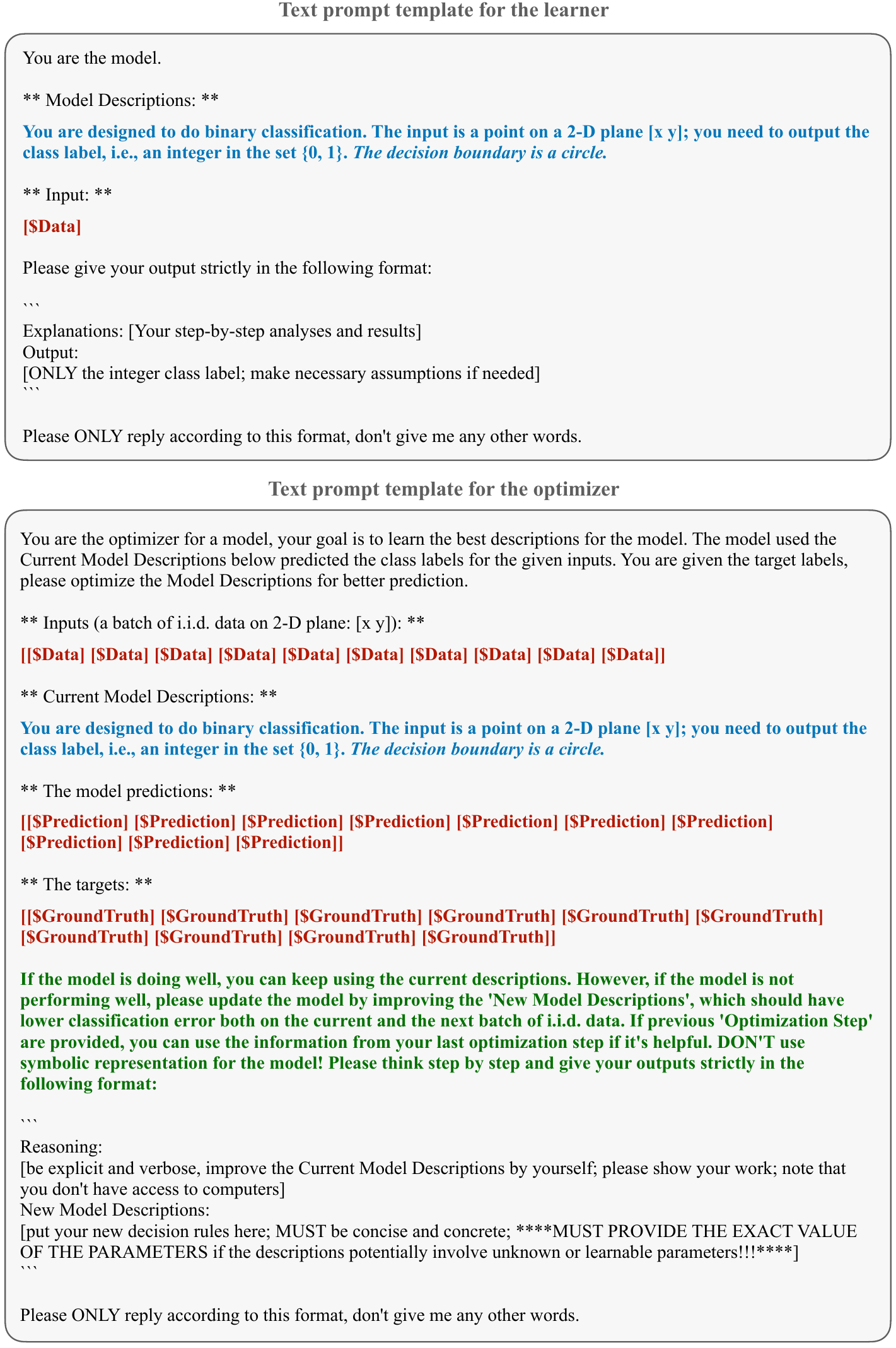}
    \caption{\scriptsize Prompt templates of VML for the learner and optimizer for the two circles classification (Llama-3-70B with prior).}
    \label{appfig:twocir_prompt}
\end{figure}

\newpage
\subsection{Text classification}
\begin{figure}[h]
    \centering
    \setlength{\abovecaptionskip}{4pt}
    \setlength{\belowcaptionskip}{-30pt}
    \renewcommand{\captionlabelfont}{\scriptsize}
    \vspace{-5pt}
    \includegraphics[width=.85\linewidth]{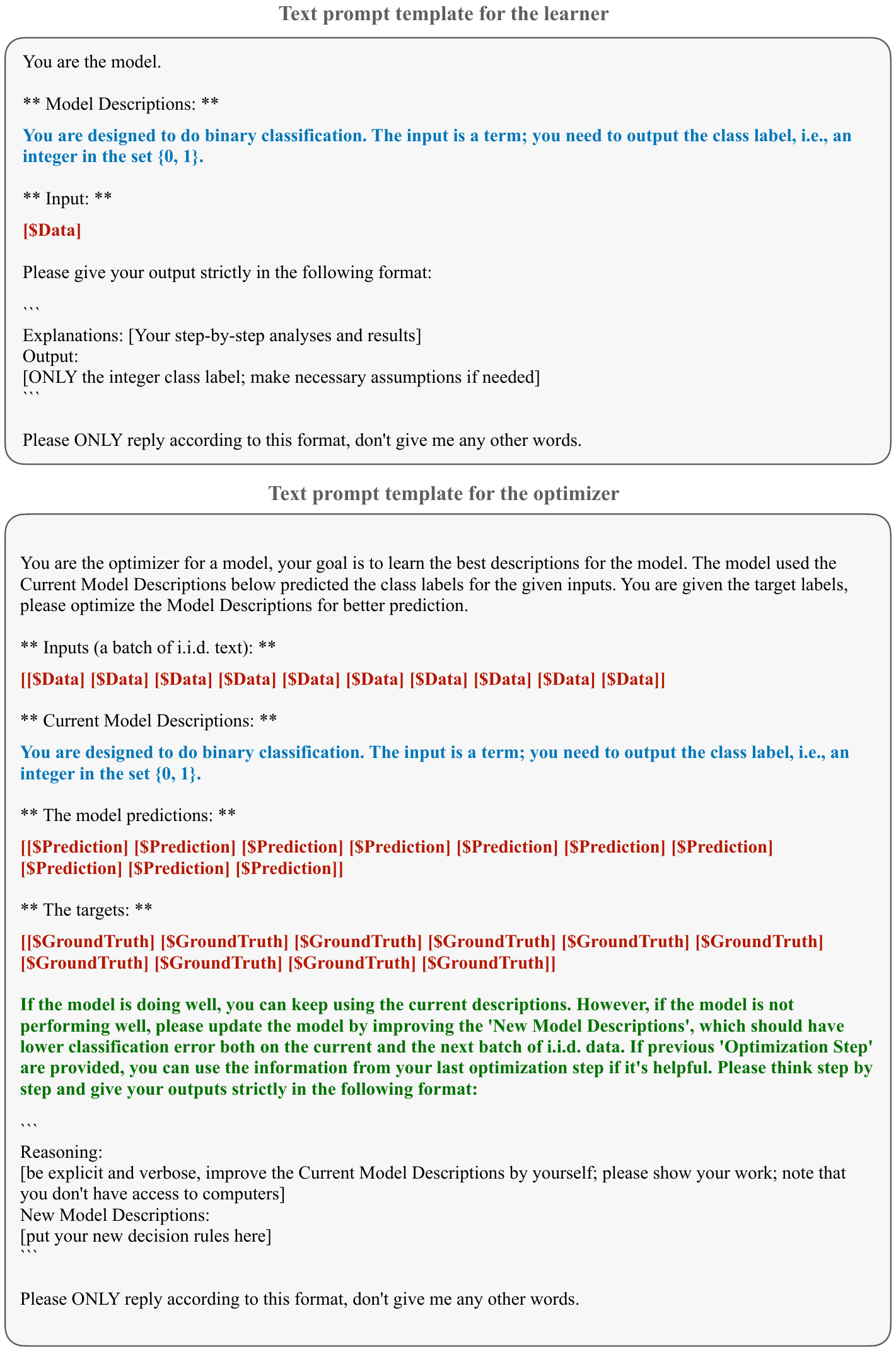}
    \caption{\scriptsize Prompt templates of VML for the learner and optimizer for the text classification (Llama-3-70B without prior).}
    \label{appfig:text_prompt}
\end{figure}

\newpage

\section{Detailed Training History}

\subsection{Linear Regression (Llama-3-70B without prior)}

\input{log_linear_reg}

\newpage

\subsection{Polynomial Regression (Llama-3-70B without prior)}
\input{log_poly}

\newpage

\subsection{Sinusoidal Regression (GPT-4o with prior)}
\input{log_sin_gpt}

\newpage
\subsection{Two Blobs Regression (LLama-3-70B without prior)}
\input{log_cls_twoblob}

\newpage
\subsection{Two Circles Regression (LLama-3-70B without prior)}
\input{log_twocircle_noprior}

\newpage
\subsection{Two Circles Regression (LLama-3-70B with prior)}
\input{log_twocircle_prior}

\newpage
\subsection{Text Classification (LLama-3-70B without prior)}
\input{log_txt}

\newpage
\subsection{Medical Image Classification (GPT-4o with prior)}
\input{log_medical_prior}

\newpage
\subsection{Medical Image Classification (GPT-4o without prior)}
\input{log_medical_noprior}

%% file: log_linear_reg_interpreter.tex
\begin{exmpinf}{Llama-3 70B with Code Interpreter -- Example Logs}{Example}
    \scriptsize

\textbf{--------------- Forward Pass Input ---------------}

You are the model. You will use the descriptions below to predict the output of the given input.

\textbf{** Pattern Descriptions: **}

You are designed to do regression, i.e., to predict the output of any given input. Both input and output are real numbers. It looks like the data is generated by a periodic function. \lstinline{\(y = 1 \sin(1x + 0) + 2\)}

\textbf{** Input: **}

[-0.09]

Please give your FINAL output **STRICTLY IN THE FOLLOWING FORMAT** at the end:

```

Explanations: [Your step-by-step analyses and results; please use python for evaluating symbolic function]

Output:

[Your output MUST be in REAL NUMBER ROUNDED TO TWO DECIMAL POINTS; it MUST be in the same format as the Input]

'''

Please ONLY reply according to this format, don't give me any other words.

\textbf{--------------- Forward Pass Output ---------------}

\textbf{** Step 1: Define the function **}

The given function is \lstinline{\\(y = 1 \\sin(1x + 0) + 2\\)}. This is a simple sine function with an amplitude of 1, a period of \lstinline{\\(2\\pi\\)}, and a phase shift of 0. The function is then vertically shifted by 2 units.

\textbf{** Step 2: Evaluate the function at the given input **}

The input is \lstinline{\\(x = -0.09\\)}. We need to evaluate the function at this input.

\begin{lstlisting}
import math

def evaluate_function(x):
    return 1 * math.sin(1 * x + 0) + 2
    
x = -0.09
result = evaluate_function(x)
print(result)
\end{lstlisting}

\begin{lstlisting}
1.910121450801989
\end{lstlisting}

\textbf{** Explanations: **}
We defined the function y = 1 * sin(1x + 0) + 2 and evaluated it at the given input x = -0.09. The result of the function evaluation is approximately 1.910121450801989.

\textbf{** Output: **}
[1.91]

\textbf{--------------- Forward Pass Output ---------------}

\textbf{** Explanations: **}

We defined the function y = 1 * sin(1x + 0) + 2 and evaluated it at the given input x = -0.09. The result of the function evaluation is approximately 1.910121450801989.

\textbf{** Output: **}

[1.91]

\end{exmpinf}

%% file: log_linear_reg.tex
\begin{exmp}{Linear Regression -- Step 1}{Step 1}
    \tiny

\textbf{Epoch 0 - Step 1}

\textbf{--------------- Forward Pass Prompt ---------------}

You are the model. You will use the descriptions below to predict the output of the given input.

\textbf{** Pattern Descriptions: **}

You are designed to do regression, i.e., to predict the output of any given input. Both input and output are real numbers.

\textbf{** Input: **}

[1.26]

Please give your output strictly in the following format:

```

Explanations: [Your step-by-step analyses and results]

\textbf{Output:}

[Your output MUST be in REAL NUMBER ROUNDED TO TWO DECIMAL POINTS; make necessary assumptions if needed; it MUST be in the same format as the Input]

'''

Please ONLY reply according to this format, don't give me any other words.

\textbf{--------------- Forward Pass Output ---------------}

Explanations: Since there is no specific pattern or function provided, I will make an assumption that the output is the same as the input.

Output: [1.89]

\textbf{--------------- Forward Pass Output ---------------}

Explanations: Not enough information is provided to make a meaningful prediction. Without a clear pattern or relationship, I will assume a simple linear model.

Output: [3.70]

\ 

\textbf{Many outputs for the rest data points ...}

\ 

\textbf{--------------- Forward Pass Output ---------------}

Explanations: Since there's no specific pattern or formula provided, I'll make an assumption based on the given input. The input [0.26] seems to be a small positive value, so I'll predict the output to be slightly larger, but still a small positive value.

Output: [0.31]

\textbf{--------------------- Overall Loss ---------------------}

36.34138

\textbf{--------------- Optimization Prompt ---------------}

You are the optimizer for a model, your goal is to learn the best descriptions for the model. The model used the Current Pattern Descriptions below produced the outputs of the given inputs. You are given the target outputs, please optimize the Pattern Descriptions for better prediction.

\textbf{** Inputs (a batch of i.i.d. data): **}

[[1.26] [1.8 ] [0.99] [1.86] [0.36] [0.36] [0.97] [1.85] [0.26] [1.89]]

\textbf{** Current Pattern Descriptions: **}

You are designed to do regression, i.e., to predict the output of any given input. Both input and output are real numbers.

\textbf{** The model outputs: **}

[[1.27] [3.6 ] [0.98] [1.86] [0.42] [0.5 ] [0.97] [3.7 ] [0.31] [1.89]]

\textbf{** The target outputs: **}

[[7.75] [9.43] [6.95] [9.55] [4.88] [5.23] [6.96] [9.62] [4.75] [9.62]]

If the model is doing well, you can keep using the current descriptions. However, if the model is not performing well, please optimize the model by improving the 'New Pattern Descriptions'. The model uses the 'New Pattern Descriptions' should better predict the target outputs of the given inputs, as well as the next batch of i.i.d. input data from the same distribution. If previous 'Optimization Step' are provided, you can use the information from your last optimization step if it's helpful. DON'T use symbolic representation for the model! Please think step by step and give your outputs strictly in the following format:

```

Reasoning:

[be explicit and verbose, improve the Current Pattern Descriptions by yourself; please show your work; note that you don't have access to computer]

New Pattern Descriptions:

[put your new descriptions here; MUST be specific and concrete; ****MUST provide the exact value of the parameters if the descriptions potentially involve unknown or learnable parameters!!!****]

'''

Please ONLY reply according to this format, don't give me any other words.

\textbf{--------------- Optimization Output ---------------}

Reasoning:

The current model outputs are not close to the target outputs, indicating that the current pattern descriptions are not effective. Upon examining the inputs and target outputs, I notice that the target outputs are generally larger than the inputs, suggesting a possible linear relationship with a positive slope. However, the current model outputs are mostly similar to the inputs, indicating that the current pattern descriptions do not capture this relationship.

To optimize the pattern descriptions, I will introduce a scaling factor to capture the linear relationship between the inputs and target outputs. I will also add a small bias term to account for any constant shift in the target outputs.

New Pattern Descriptions:

The new pattern descriptions will be based on a linear regression model with a scaling factor of 5.5 and a bias term of 2.5. The new pattern descriptions are:
y = 5.5x + 2.5
where y is the predicted output and x is the input.

\textbf{-------------------- Updated Theta --------------------}

The new pattern descriptions will be based on a linear regression model with a scaling factor of 5.5 and a bias term of 2.5. The new pattern descriptions are:

y = 5.5x + 2.5

where y is the predicted output and x is the input.

\end{exmp}

\newpage

\begin{exmp}{Linear Regression -- Step 2}{Step 2}
    \tiny

\textbf{Epoch 0 - Step 2}

\textbf{--------------- Forward Pass Prompt ---------------}

You are the model. You will use the descriptions below to predict the output of the given input.

\textbf{** Pattern Descriptions: **}

You are designed to do regression, i.e., to predict the output of any given input. Both input and output are real numbers. The new pattern descriptions will be based on a linear regression model with a scaling factor of 5.5 and a bias term of 2.5. The new pattern descriptions are:

y = 5.5x + 2.5

where y is the predicted output and x is the input.

\textbf{** Input: **}

[0.86]

Please give your output strictly in the following format:

```

Explanations: [Your step-by-step analyses and results]

Output:

[Your output MUST be in REAL NUMBER ROUNDED TO TWO DECIMAL POINTS; make necessary assumptions if needed; it MUST be in the same format as the Input]

'''

Please ONLY reply according to this format, don't give me any other words.

\textbf{--------------- Forward Pass Output ---------------}

Explanations:

Given input x = 0.28, we can plug it into the equation y = 5.5x + 2.5 to get the output y.

y = 5.5(0.28) + 2.5

y = 1.54 + 2.5

y = 4.04

Output: [4.04]

\ 

\textbf{Many outputs for the rest data points ...}

\ 

\textbf{--------------- Forward Pass Output ---------------}

Explanations:

The given linear regression model is y = 5.5x + 2.5.
We need to find the output y for the given input x = 1.59.
Plugging in the value of x into the equation, we get y = 5.5(1.59) + 2.5.

First, let's calculate 5.5(1.59) = 8.745. Now, add 2.5 to get y = 8.745 + 2.5 = 11.245. Rounded to two decimal points, the output is 11.25.

Output: [11.25]

\textbf{--------------------- Overall Loss ---------------------}

2.8677400000000004

\textbf{--------------- Optimization Prompt ---------------}

You are the optimizer for a model, your goal is to learn the best descriptions for the model. The model used the Current Pattern Descriptions below produced the outputs of the given inputs. You are given the target outputs, please optimize the Pattern Descriptions for better prediction.

\textbf{** Inputs (a batch of i.i.d. data): **}

[[0.86] [0.02] [1.1 ] [1.59] [1.2 ] [0.28] [1.89] [1.29] [0.79] [0.01]]

\textbf{** Current Pattern Descriptions: **}

You are designed to do regression, i.e., to predict the output of any given input. Both input and output are real numbers. The new pattern descriptions will be based on a linear regression model with a scaling factor of 5.5 and a bias term of 2.5. The new pattern descriptions are:

y = 5.5x + 2.5

where y is the predicted output and x is the input.

\textbf{** The model outputs: **}

[[ 7.21] [ 2.61] [ 8.55] [11.25] [ 9.1 ] [ 4.04] [12.92] [ 9.59] [ 6.85] [ 2.56]]

\textbf{** The target outputs: **}

[[6.56] [3.98] [7.29] [8.6 ] [7.7 ] [4.77] [9.67] [8.03] [6.43] [4.01]]

If the model is doing well, you can keep using the current descriptions. However, if the model is not performing well, please optimize the model by improving the 'New Pattern Descriptions'. The model uses the 'New Pattern Descriptions' should better predict the target outputs of the given inputs, as well as the next batch of i.i.d. input data from the same distribution. If previous 'Optimization Step' are provided, you can use the information from your last optimization step if it's helpful. DON'T use symbolic representation for the model! Please think step by step and give your outputs strictly in the following format:

```

Reasoning:

[be explicit and verbose, improve the Current Pattern Descriptions by yourself; please show your work; note that you don't have access to computer]

New Pattern Descriptions:

[put your new descriptions here; MUST be specific and concrete; ****MUST provide the exact value of the parameters if the descriptions potentially involve unknown or learnable parameters!!!****]

'''

Please ONLY reply according to this format, don't give me any other words.

\textbf{--------------- Optimization Output ---------------}

Reasoning:

The current model outputs are not close to the target outputs, indicating that the current pattern descriptions are not effective. Upon examining the inputs, target outputs, and model outputs, I notice that the model outputs are generally larger than the target outputs, suggesting that the scaling factor of 5.5 is too large. I will reduce the scaling factor to bring the model outputs closer to the target outputs. Additionally, I will adjust the bias term to better capture the overall shift in the target outputs.
To optimize the pattern descriptions, I will calculate the average difference between the target outputs and model outputs to determine the adjustment needed for the bias term. The average difference is approximately -1.45. I will subtract this value from the current bias term of 2.5 to get a new bias term of 1.05.

New Pattern Descriptions:

The new pattern descriptions will be based on a linear regression model with a scaling factor of 4.2 and a bias term of 1.05. The new pattern descriptions are:

y = 4.2x + 1.05

where y is the predicted output and x is the input.

\textbf{-------------------- Updated Theta --------------------}

The new pattern descriptions will be based on a linear regression model with a scaling factor of 4.2 and a bias term of 1.05. The new pattern descriptions are:

y = 4.2x + 1.05

where y is the predicted output and x is the input.

\end{exmp}
\vspace{-.5mm}

\newpage

\begin{exmp}{Linear Regression -- Step 3}{Step 3}
    \tiny

\textbf{Epoch 0 - Step 3}

\textbf{--------------- Forward Pass Prompt ---------------}

You are the model. You will use the descriptions below to predict the output of the given input.

\textbf{** Pattern Descriptions: **}

You are designed to do regression, i.e., to predict the output of any given input. Both input and output are real numbers. The new pattern descriptions will be based on a linear regression model with a scaling factor of 4.2 and a bias term of 1.05. The new pattern descriptions are:

y = 4.2x + 1.05

where y is the predicted output and x is the input.

\textbf{** Input: **}

[0.49]

Please give your output strictly in the following format:

```

Explanations: [Your step-by-step analyses and results]

Output:
[Your output MUST be in REAL NUMBER ROUNDED TO TWO DECIMAL POINTS; make necessary assumptions if needed; it MUST be in the same format as the Input]

'''

Please ONLY reply according to this format, don't give me any other words.

\textbf{--------------- Forward Pass Output ---------------}

Explanations:

Using the given linear regression model, substitute x with the input 0.05.

y = 4.2x + 1.05

y = 4.2(0.05) + 1.05

y = 0.21 + 1.05

y = 1.26

Output: [1.26]

\ 

\textbf{Many outputs for the rest data points ...}

\ 

\textbf{--------------------- Overall Loss ---------------------}

4.176069999999999

\textbf{--------------- Optimization Prompt ---------------}

You are the optimizer for a model, your goal is to learn the best descriptions for the model. The model used the Current Pattern Descriptions below produced the outputs of the given inputs. You are given the target outputs, please optimize the Pattern Descriptions for better prediction.

\textbf{** Inputs (a batch of i.i.d. data): **}

[[0.49] [0.74] [0.13] [0.05] [0.84] [1.21] [1.28] [1.77] [0.99] [0.77]]

\textbf{** Current Pattern Descriptions: **}

You are designed to do regression, i.e., to predict the output of any given input. Both input and output are real numbers. The new pattern descriptions will be based on a linear regression model with a scaling factor of 4.2 and a bias term of 1.05. The new pattern descriptions are:

y = 4.2x + 1.05

where y is the predicted output and x is the input.

\textbf{** The model outputs: **}

[[3.11] [4.16] [1.6 ] [1.26] [4.58] [6.13] [6.43] [8.52] [5.21] [4.28]]

\textbf{** The target outputs: **}

[[5.59] [6.22] [4.31] [4.06] [6.36] [7.75] [7.87] [9.17] [6.97] [6.47]]

If the model is doing well, you can keep using the current descriptions. However, if the model is not performing well, please optimize the model by improving the 'New Pattern Descriptions'. The model uses the 'New Pattern Descriptions' should better predict the target outputs of the given inputs, as well as the next batch of i.i.d. input data from the same distribution. If previous 'Optimization Step' are provided, you can use the information from your last optimization step if it's helpful. DON'T use symbolic representation for the model! Please think step by step and give your outputs strictly in the following format:

```

Reasoning:

[be explicit and verbose, improve the Current Pattern Descriptions by yourself; please show your work; note that you don't have access to computer]

New Pattern Descriptions:

[put your new descriptions here; MUST be specific and concrete; ****MUST provide the exact value of the parameters if the descriptions potentially involve unknown or learnable parameters!!!****]

'''

Please ONLY reply according to this format, don't give me any other words.

\textbf{--------------- Optimization Output ---------------}

Reasoning:

The current model outputs are not close to the target outputs, indicating that the current pattern descriptions are not effective. Upon examining the inputs, target outputs, and model outputs, I notice that the model outputs are generally smaller than the target outputs, suggesting that the scaling factor of 4.2 is too small. I will increase the scaling factor to bring the model outputs closer to the target outputs. Additionally, I will adjust the bias term to better capture the overall shift in the target outputs.
To optimize the pattern descriptions, I will calculate the average difference between the target outputs and model outputs to determine the adjustment needed for the bias term. The average difference is approximately 1.43. I will add this value to the current bias term of 1.05 to get a new bias term of 2.48.
I will also analyze the relationship between the inputs and target outputs to determine the new scaling factor. Upon examining the data, I notice that the target outputs increase by approximately 5.5 units for every 1 unit increase in the input. I will use this relationship to set the new scaling factor to 5.5.

New Pattern Descriptions:

The new pattern descriptions will be based on a linear regression model with a scaling factor of 5.5 and a bias term of 2.48. The new pattern descriptions are:

y = 5.5x + 2.48

where y is the predicted output and x is the input.

\textbf{-------------------- Updated Theta --------------------}

The new pattern descriptions will be based on a linear regression model with a scaling factor of 5.5 and a bias term of 2.48. The new pattern descriptions are:

y = 5.5x + 2.48

where y is the predicted output and x is the input.

\end{exmp}

\newpage
\begin{exmp}{Linear Regression -- Step 4}{Step 4}
    \tiny

\textbf{Epoch 0 - Step 4}

\textbf{--------------- Forward Pass Prompt ---------------}

You are the model. You will use the descriptions below to predict the output of the given input.

\textbf{** Pattern Descriptions: **}

You are designed to do regression, i.e., to predict the output of any given input. Both input and output are real numbers. The new pattern descriptions will be based on a linear regression model with a scaling factor of 5.5 and a bias term of 2.48. The new pattern descriptions are:

y = 5.5x + 2.48

where y is the predicted output and x is the input.

\textbf{** Input: **}

[0.08]

Please give your output strictly in the following format:

```

Explanations: [Your step-by-step analyses and results]

Output:

[Your output MUST be in REAL NUMBER ROUNDED TO TWO DECIMAL POINTS; make necessary assumptions if needed; it MUST be in the same format as the Input]

'''

Please ONLY reply according to this format, don't give me any other words.

\textbf{--------------- Forward Pass Output ---------------}

Explanations:

Given the input x = 1, I will use the linear regression model y = 5.5x + 2.48 to predict the output y.

Substituting x = 1 into the equation, I get y = 5.5(1) + 2.48 = 7.98.

Output: [7.98]

\ 

\textbf{Many outputs for the rest data points ...}

\ 

\textbf{--------------------- Overall Loss ---------------------}

2.6112699999999993

\textbf{--------------- Optimization Prompt ---------------}

You are the optimizer for a model, your goal is to learn the best descriptions for the model. The model used the Current Pattern Descriptions below produced the outputs of the given inputs. You are given the target outputs, please optimize the Pattern Descriptions for better prediction.

\textbf{** Inputs (a batch of i.i.d. data): **}

[[0.08] [0.44] [0.11] [0.84] [1.48] [0.02] [1.  ] [1.39] [1.7 ] [1.38]]

\textbf{** Current Pattern Descriptions: **}

You are designed to do regression, i.e., to predict the output of any given input. Both input and output are real numbers. The new pattern descriptions will be based on a linear regression model with a scaling factor of 5.5 and a bias term of 2.48. The new pattern descriptions are:

y = 5.5x + 2.48

where y is the predicted output and x is the input.

\textbf{** The model outputs: **}

[[ 2.92] [ 4.9 ] [ 3.09] [ 7.12] [10.62] [ 2.59] [ 7.98] [10.1 ] [11.93] [10.07]]

\textbf{** The target outputs: **}

[[4.27] [5.19] [4.24] [6.55] [8.61] [3.89] [6.96] [8.15] [9.17] [7.98]]

If the model is doing well, you can keep using the current descriptions. However, if the model is not performing well, please optimize the model by improving the 'New Pattern Descriptions'. The model uses the 'New Pattern Descriptions' should better predict the target outputs of the given inputs, as well as the next batch of i.i.d. input data from the same distribution. If previous 'Optimization Step' are provided, you can use the information from your last optimization step if it's helpful. DON'T use symbolic representation for the model! Please think step by step and give your outputs strictly in the following format:

```

Reasoning:

[be explicit and verbose, improve the Current Pattern Descriptions by yourself; please show your work; note that you don't have access to computer]

New Pattern Descriptions:

[put your new descriptions here; MUST be specific and concrete; ****MUST provide the exact value of the parameters if the descriptions potentially involve unknown or learnable parameters!!!****]

'''

Please ONLY reply according to this format, don't give me any other words.

\textbf{--------------- Optimization Output ---------------}

Reasoning:

The current model outputs are not close to the target outputs, indicating that the current pattern descriptions are not effective. Upon examining the inputs, target outputs, and model outputs, I notice that the model outputs are generally smaller than the target outputs, suggesting that the bias term of 2.48 is too small. I will increase the bias term to bring the model outputs closer to the target outputs. Additionally, I will adjust the scaling factor to better capture the relationship between the inputs and target outputs.
To optimize the pattern descriptions, I will calculate the average difference between the target outputs and model outputs to determine the adjustment needed for the bias term. The average difference is approximately 1.23. I will add this value to the current bias term of 2.48 to get a new bias term of 3.71.
I will also analyze the relationship between the inputs and target outputs to determine the new scaling factor. Upon examining the data, I notice that the target outputs increase by approximately 5.8 units for every 1 unit increase in the input. However, the current scaling factor of 5.5 is close to this value, so I will only make a small adjustment to the scaling factor. I will increase the scaling factor to 5.7 to better capture the relationship between the inputs and target outputs.

New Pattern Descriptions:

The new pattern descriptions will be based on a linear regression model with a scaling factor of 5.7 and a bias term of 3.71. The new pattern descriptions are:

y = 5.7x + 3.71

where y is the predicted output and x is the input.

\textbf{-------------------- Updated Theta --------------------}

The new pattern descriptions will be based on a linear regression model with a scaling factor of 5.7 and a bias term of 3.71. The new pattern descriptions are:

y = 5.7x + 3.71

where y is the predicted output and x is the input.

\end{exmp}

\newpage

\begin{exmp}{Linear Regression -- Step 5}{Step 5}
    \tiny

\textbf{Epoch 0 - Step 5}

\textbf{--------------- Forward Pass Prompt ---------------}

You are the model. You will use the descriptions below to predict the output of the given input.

\textbf{** Pattern Descriptions: **}

You are designed to do regression, i.e., to predict the output of any given input. Both input and output are real numbers. The new pattern descriptions will be based on a linear regression model with a scaling factor of 5.7 and a bias term of 3.71. The new pattern descriptions are:

y = 5.7x + 3.71

where y is the predicted output and x is the input.

\textbf{** Input: **}

[1.59]

Please give your output strictly in the following format:

```

Explanations: [Your step-by-step analyses and results]

Output:

[Your output MUST be in REAL NUMBER ROUNDED TO TWO DECIMAL POINTS; make necessary assumptions if needed; it MUST be in the same format as the Input]

'''

Please ONLY reply according to this format, don't give me any other words.

\textbf{--------------- Forward Pass Output ---------------}

Explanations:

Given the input x = 1.39, we can plug it into the equation y = 5.7x + 3.71 to get the output y.

y = 5.7(1.39) + 3.71

y = 7.813 + 3.71

y = 11.523

Output: [11.52]

\ 

\textbf{Many outputs for the rest data points ...}

\ 

\textbf{--------------------- Overall Loss ---------------------}

8.08032

\textbf{--------------- Optimization Prompt ---------------}

You are the optimizer for a model, your goal is to learn the best descriptions for the model. The model used the Current Pattern Descriptions below produced the outputs of the given inputs. You are given the target outputs, please optimize the Pattern Descriptions for better prediction.

\textbf{** Inputs (a batch of i.i.d. data): **}

[[1.59] [0.23] [1.72] [0.98] [1.39] [1.7 ] [1.54] [0.49] [0.27] [0.93]]

\textbf{** Current Pattern Descriptions: **}

You are designed to do regression, i.e., to predict the output of any given input. Both input and output are real numbers. The new pattern descriptions will be based on a linear regression model with a scaling factor of 5.7 and a bias term of 3.71. The new pattern descriptions are:

y = 5.7x + 3.71

where y is the predicted output and x is the input.

\textbf{** The model outputs: **}

[[12.79] [ 5.02] [13.51] [ 9.3 ] [11.52] [13.4 ] [11.42] [ 6.49] [ 5.25] [ 8.99]]

\textbf{** The target outputs: **}

[[9.02] [4.64] [9.2 ] [7.02] [8.33] [9.14] [8.55] [5.47] [4.77] [6.79]]

If the model is doing well, you can keep using the current descriptions. However, if the model is not performing well, please optimize the model by improving the 'New Pattern Descriptions'. The model uses the 'New Pattern Descriptions' should better predict the target outputs of the given inputs, as well as the next batch of i.i.d. input data from the same distribution. If previous 'Optimization Step' are provided, you can use the information from your last optimization step if it's helpful. DON'T use symbolic representation for the model! Please think step by step and give your outputs strictly in the following format:

```

Reasoning:

[be explicit and verbose, improve the Current Pattern Descriptions by yourself; please show your work; note that you don't have access to computer]

New Pattern Descriptions:

[put your new descriptions here; MUST be specific and concrete; ****MUST provide the exact value of the parameters if the descriptions potentially involve unknown or learnable parameters!!!****]

'''

Please ONLY reply according to this format, don't give me any other words.

\textbf{--------------- Optimization Output ---------------}

Reasoning:

The current model outputs are not close to the target outputs, indicating that the current pattern descriptions are not effective. Upon examining the inputs, target outputs, and model outputs, I notice that the model outputs are generally larger than the target outputs, suggesting that the scaling factor of 5.7 is too large. I will decrease the scaling factor to bring the model outputs closer to the target outputs. Additionally, I will adjust the bias term to better capture the relationship between the inputs and target outputs.
To optimize the pattern descriptions, I will calculate the average difference between the target outputs and model outputs to determine the adjustment needed for the bias term. The average difference is approximately -1.44. I will subtract this value from the current bias term of 3.71 to get a new bias term of 2.27.
I will also analyze the relationship between the inputs and target outputs to determine the new scaling factor. Upon examining the data, I notice that the target outputs increase by approximately 4.2 units for every 1 unit increase in the input. The current scaling factor of 5.7 is larger than this value, so I will decrease the scaling factor to 4.4 to better capture the relationship between the inputs and target outputs.

New Pattern Descriptions:

The new pattern descriptions will be based on a linear regression model with a scaling factor of 4.4 and a bias term of 2.27. The new pattern descriptions are:

y = 4.4x + 2.27

where y is the predicted output and x is the input.

\textbf{-------------------- Updated Theta --------------------}

The new pattern descriptions will be based on a linear regression model with a scaling factor of 4.4 and a bias term of 2.27. The new pattern descriptions are:

y = 4.4x + 2.27

where y is the predicted output and x is the input.

\end{exmp}
\vspace{-1mm}

\newpage

\begin{exmp}{Linear Regression -- Step 19}{Step 19}
    \tiny

\textbf{Epoch 1 - Step 9}

\textbf{--------------- Forward Pass Prompt ---------------}

You are the model. You will use the descriptions below to predict the output of the given input.

\textbf{** Pattern Descriptions: **}

You are designed to do regression, i.e., to predict the output of any given input. Both input and output are real numbers. The new pattern descriptions will be based on a linear regression model with a scaling factor of 3.26 and a bias term of 3.45. The new pattern descriptions are:

y = 3.26x + 3.45

where y is the predicted output and x is the input.

\textbf{** Input: **}

[1.26]

Please give your output strictly in the following format:

```

Explanations: [Your step-by-step analyses and results]

Output:

[Your output MUST be in REAL NUMBER ROUNDED TO TWO DECIMAL POINTS; make necessary assumptions if needed; it MUST be in the same format as the Input]

'''

Please ONLY reply according to this format, don't give me any other words.

\textbf{--------------- Forward Pass Output ---------------}

Explanations:

Given the linear regression model y = 3.26x + 3.45, we can plug in the input x = 1.26.

y = 3.26(1.26) + 3.45 = 4.08 + 3.45 = 7.53.

Output: [7.53]

\ 

\textbf{Many outputs for the rest data points ...}

\ 

\textbf{--------------------- Overall Loss ---------------------}

0.11138999999999992

\textbf{--------------- Optimization Prompt ---------------}

You are the optimizer for a model, your goal is to learn the best descriptions for the model. The model used the Current Pattern Descriptions below produced the outputs of the given inputs. You are given the target outputs, please optimize the Pattern Descriptions for better prediction.

\textbf{** Inputs (a batch of i.i.d. data): **}

[[1.26] [1.66] [0.32] [0.83] [1.3 ] [0.24] [1.87] [1.15] [1.2 ] [1.26]]

\textbf{** Current Pattern Descriptions: **}

You are designed to do regression, i.e., to predict the output of any given input. Both input and output are real numbers. The new pattern descriptions will be based on a linear regression model with a scaling factor of 3.26 and a bias term of 3.45. The new pattern descriptions are:

y = 3.26x + 3.45

where y is the predicted output and x is the input.

\textbf{** The model outputs: **}

[[7.53] [8.88] [4.49] [6.17] [7.71] [4.23] [9.58] [7.19] [7.36] [7.54]]

\textbf{** The target outputs: **}

[[7.87] [8.93] [4.93] [6.36] [8.18] [4.78] [9.37] [7.39] [7.71] [7.73]]

If the model is doing well, you can keep using the current descriptions. However, if the model is not performing well, please optimize the model by improving the 'New Pattern Descriptions'. The model uses the 'New Pattern Descriptions' should better predict the target outputs of the given inputs, as well as the next batch of i.i.d. input data from the same distribution. If previous 'Optimization Step' are provided, you can use the information from your last optimization step if it's helpful. DON'T use symbolic representation for the model! Please think step by step and give your outputs strictly in the following format:

```

Reasoning:

[be explicit and verbose, improve the Current Pattern Descriptions by yourself; please show your work; note that you don't have access to computer]

New Pattern Descriptions:

[put your new descriptions here; MUST be specific and concrete; ****MUST provide the exact value of the parameters if the descriptions potentially involve unknown or learnable parameters!!!****]

'''

Please ONLY reply according to this format, don't give me any other words.
************ Shortening the state from 5 to 3 ************

\textbf{--------------- Optimization Output ---------------}

Reasoning:

The current model outputs are generally close to the target outputs, but there are some deviations. Upon examining the inputs, target outputs, and model outputs, I notice that the model outputs are still underestimating the target outputs for some inputs, and overestimating for others. This suggests that the bias term of 3.45 might not be optimal.
To optimize the pattern descriptions, I will calculate the average difference between the model outputs and target outputs. The average difference is approximately 0.09. I will add this value to the current bias term of 3.45 to get a new bias term of 3.54.
I will also analyze the relationship between the inputs and target outputs to determine if the scaling factor of 3.26 needs to be adjusted. Upon examining the data, I notice that the target outputs are still increasing with the inputs, and the rate of increase is slightly faster than the current scaling factor of 3.26. This suggests that the scaling factor of 3.26 might be slightly too small. I will increase the scaling factor by 0.01 to get a new scaling factor of 3.27.

New Pattern Descriptions:

The new pattern descriptions will be based on a linear regression model with a scaling factor of 3.27 and a bias term of 3.54. The new pattern descriptions are:

y = 3.27x + 3.54

where y is the predicted output and x is the input.

\textbf{-------------------- Updated Theta --------------------}

The new pattern descriptions will be based on a linear regression model with a scaling factor of 3.27 and a bias term of 3.54. The new pattern descriptions are:

y = 3.27x + 3.54

where y is the predicted output and x is the input.

\end{exmp}

\begin{exmp}{Linear Regression -- Step 20}{Step 20}
    \tiny

\textbf{Epoch 1 - Step 10}

\textbf{--------------- Forward Pass Prompt ---------------}

You are the model. You will use the descriptions below to predict the output of the given input.

\textbf{** Pattern Descriptions: **}

You are designed to do regression, i.e., to predict the output of any given input. Both input and output are real numbers. The new pattern descriptions will be based on a linear regression model with a scaling factor of 3.27 and a bias term of 3.54. The new pattern descriptions are:

y = 3.27x + 3.54

where y is the predicted output and x is the input.

\textbf{** Input: **}

[0.71]

Please give your output strictly in the following format:

```

Explanations: [Your step-by-step analyses and results]

Output:

[Your output MUST be in REAL NUMBER ROUNDED TO TWO DECIMAL POINTS; make necessary assumptions if needed; it MUST be in the same format as the Input]

'''

Please ONLY reply according to this format, don't give me any other words.

\textbf{--------------- Forward Pass Output ---------------}

Explanations:

Using the given linear regression model, y = 3.27x + 3.54, and the input x = 1, we can calculate the output y as follows:

y = 3.27(1) + 3.54

y = 3.27 + 3.54

y = 6.81

Output: 6.81

\ 

\textbf{Many outputs for the rest data points ...}

\ 

\textbf{--------------------- Overall Loss ---------------------}

0.08741999999999994

\textbf{--------------- Optimization Prompt ---------------}

You are the optimizer for a model, your goal is to learn the best descriptions for the model. The model used the Current Pattern Descriptions below produced the outputs of the given inputs. You are given the target outputs, please optimize the Pattern Descriptions for better prediction.

\textbf{** Inputs (a batch of i.i.d. data): **}

[[0.71] [0.6 ] [0.51] [1.34] [0.92] [0.16] [1.12] [0.14] [1.67] [1.  ]]

\textbf{** Current Pattern Descriptions: **}

You are designed to do regression, i.e., to predict the output of any given input. Both input and output are real numbers. The new pattern descriptions will be based on a linear regression model with a scaling factor of 3.27 and a bias term of 3.54. The new pattern descriptions are:

y = 3.27x + 3.54

where y is the predicted output and x is the input.

\textbf{** The model outputs: **}

[[5.86] [5.5 ] [5.22] [7.91] [6.57] [4.07] [7.21] [3.92] [9.  ] [6.81]]

\textbf{** The target outputs: **}

[[6.05] [5.8 ] [5.48] [8.1 ] [6.56] [4.63] [7.37] [4.43] [8.85] [6.96]]

If the model is doing well, you can keep using the current descriptions. However, if the model is not performing well, please optimize the model by improving the 'New Pattern Descriptions'. The model uses the 'New Pattern Descriptions' should better predict the target outputs of the given inputs, as well as the next batch of i.i.d. input data from the same distribution. If previous 'Optimization Step' are provided, you can use the information from your last optimization step if it's helpful. DON'T use symbolic representation for the model! Please think step by step and give your outputs strictly in the following format:

```

Reasoning:

[be explicit and verbose, improve the Current Pattern Descriptions by yourself; please show your work; note that you don't have access to computer]

New Pattern Descriptions:

[put your new descriptions here; MUST be specific and concrete; ****MUST provide the exact value of the parameters if the descriptions potentially involve unknown or learnable parameters!!!****]

'''

Please ONLY reply according to this format, don't give me any other words.

\textbf{--------------- Optimization Output ---------------}

Reasoning:

The current model outputs are generally close to the target outputs, but there are some deviations. Upon examining the inputs, target outputs, and model outputs, I notice that the model outputs are still underestimating the target outputs for some inputs, and overestimating for others. This suggests that the bias term of 3.54 might not be optimal.
To optimize the pattern descriptions, I will calculate the average difference between the model outputs and target outputs. The average difference is approximately 0.15. I will add this value to the current bias term of 3.54 to get a new bias term of 3.69.
I will also analyze the relationship between the inputs and target outputs to determine if the scaling factor of 3.27 needs to be adjusted. Upon examining the data, I notice that the target outputs are still increasing with the inputs, and the rate of increase is slightly faster than the current scaling factor of 3.27. This suggests that the scaling factor of 3.27 might be slightly too small. I will increase the scaling factor by 0.02 to get a new scaling factor of 3.29.

New Pattern Descriptions:

The new pattern descriptions will be based on a linear regression model with a scaling factor of 3.29 and a bias term of 3.69. The new pattern descriptions are:

y = 3.29x + 3.69

where y is the predicted output and x is the input.

\textbf{-------------------- Updated Theta --------------------}

The new pattern descriptions will be based on a linear regression model with a scaling factor of 3.29 and a bias term of 3.69. The new pattern descriptions are:

y = 3.29x + 3.69

where y is the predicted output and x is the input.

\end{exmp}

%% file: log_poly.tex
\begin{exmp}{Polynomial Regression -- Step 1}{2-Step 1}
    \scriptsize

\textbf{Epoch 0 - Step 1}

\textbf{--------------- Forward Pass Prompt ---------------}

You are the model. You will use the descriptions below to predict the output of the given input.

\textbf{** Pattern Descriptions: **}

You are designed to do regression, i.e., to predict the output of any given input. Both input and output are real numbers.

\textbf{** Input: **}

[-2.4]

Please give your output strictly in the following format:

```

Explanations: [Your step-by-step analyses and results]

Output:

[Your output MUST be in REAL NUMBER ROUNDED TO TWO DECIMAL POINTS; make necessary assumptions if needed; it MUST be in the same format as the Input]

```

Please ONLY reply according to this format, don't give me any other words.

\textbf{--------------- Forward Pass Output ---------------}

Explanations: Since there is no specific pattern or relationship provided, I will make an assumption that the output is a constant value. In this case, I will assume the output is 0.00.

Output: [0.00]

\ 

\textbf{Many outputs for the rest data points ...}

\ 

\textbf{--------------- Forward Pass Output ---------------}

Explanations:

Since there is no specific pattern or function provided, I will make an assumption that the input and output are linearly related with a constant difference of 1.2. Based on this assumption, I will add 1.2 to the input to get the output.

Output: [0.50]

\textbf{--------------------- Overall Loss ---------------------}

63.153999999999996

\textbf{--------------- Optimization Prompt ---------------}

You are the optimizer for a model, your goal is to learn the best descriptions for the model. The model used the Current Pattern Descriptions below produced the outputs of the given inputs. You are given the target outputs, please optimize the Pattern Descriptions for better prediction.

\textbf{** Inputs (a batch of i.i.d. data): **}

[[-2.4] [ 0.6] [ 0.9] [-1.8] [-0.7] [-0.9] [-1.2] [-0. ] [-0.2] [ 0.3]]

\textbf{** Current Pattern Descriptions: **}

You are designed to do regression, i.e., to predict the output of any given input. Both input and output are real numbers.

\textbf{** The model outputs: **}

[[-2.4] [ 0. ] [ 0. ] [ 0. ] [ 0.5] [ 0. ] [-1.2] [ 0. ] [-0.2] [ 0.5]]

\textbf{** The target outputs: **}

[[17.8] [ 3.7] [ 6.6] [10. ] [ 3.6] [ 2.5] [ 5.4] [ 1.1] [ 0.9] [ 2.6]]

If the model is doing well, you can keep using the current descriptions. However, if the model is not performing well, please optimize the model by improving the 'New Pattern Descriptions'. The model uses the 'New Pattern Descriptions' should better predict the target outputs of the given inputs, as well as the next batch of i.i.d. input data from the same distribution. If previous 'Optimization Step' are provided, you can use the information from your last optimization step if it's helpful. NOTE: both the model and you can only operate on the numerical precision of one decimal points! Please think step by step and give your outputs strictly in the following format:

```

Reasoning:

[be explicit and verbose, improve the Current Pattern Descriptions by yourself; please show your work; note that you don't have access to computer]

New Pattern Descriptions:

[put your new descriptions here; MUST be specific and concrete; ****MUST provide the exact value of the parameters if the descriptions potentially involve unknown or learnable parameters!!!****]

'''

Please ONLY reply according to this format, don't give me any other words.

\textbf{--------------- Optimization Output ---------------}

Reasoning:

After examining the model outputs and target outputs, it's clear that the model is not performing well. The model outputs are mostly around 0, while the target outputs have a larger range of values. This suggests that the model is not capturing the underlying pattern in the data.
Upon closer inspection, I notice that the inputs and target outputs seem to have a positive correlation. This means that as the input increases, the target output also tends to increase.
To improve the model, I will introduce a new pattern description that captures this positive correlation. I will use a simple linear relationship between the input and output, with a slope and intercept that can be learned from the data.

New Pattern Descriptions:

The new pattern description is: output = 2.5 * input + 1.5.

\textbf{-------------------- Updated Theta --------------------}

The new pattern description is: output = 2.5 * input + 1.5.

\end{exmp}
\newpage

\begin{exmp}{Polynomial Regression -- Step 2}{2-Step 2}
    \scriptsize

\textbf{Epoch 0 - Step 2}

\textbf{--------------- Forward Pass Prompt ---------------}

You are the model. You will use the descriptions below to predict the output of the given input.

\textbf{** Pattern Descriptions: **}

You are designed to do regression, i.e., to predict the output of any given input. Both input and output are real numbers. The new pattern description is: output = 2.5 * input + 1.5.

\textbf{** Input: **}

[-2.4]

Please give your output strictly in the following format:

```

Explanations: [Your step-by-step analyses and results]

Output:

[Your output MUST be in REAL NUMBER ROUNDED TO TWO DECIMAL POINTS; make necessary assumptions if needed; it MUST be in the same format as the Input]

'''

Please ONLY reply according to this format, don't give me any other words.

\textbf{--------------- Forward Pass Output ---------------}

Explanations: The output is calculated by multiplying the input by 2.5 and adding 1.5. In this case, the input is 1, so the output is 2.5 * 1 + 1.5 = 4.

Output: [4.00]

\ 

\textbf{Many outputs for the rest data points ...}

\ 

\textbf{--------------- Forward Pass Output ---------------}

Explanations:

To find the output, we need to apply the given pattern description: output = 2.5 * input + 1.5.

In this case, the input is -1.2, so we plug it into the equation: output = 2.5 * (-1.2) + 1.5.

First, we multiply 2.5 and -1.2: 2.5 * (-1.2) = -3.0.

Then, we add 1.5: -3.0 + 1.5 = -1.5.

Output: [-1.50]

\textbf{--------------------- Overall Loss ---------------------}

257.99

\textbf{--------------- Optimization Prompt ---------------}

You are the optimizer for a model, your goal is to learn the best descriptions for the model. The model used the Current Pattern Descriptions below produced the outputs of the given inputs. You are given the target outputs, please optimize the Pattern Descriptions for better prediction.

\textbf{** Inputs (a batch of i.i.d. data): **}

[[-2.4] [-2.9] [-2.1] [-1.8] [-1.2] [-0.6] [-2.2] [ 1. ] [-1.9] [ 1. ]]

\textbf{** Current Pattern Descriptions: **}

You are designed to do regression, i.e., to predict the output of any given input. Both input and output are real numbers. The new pattern description is: output = 2.5 * input + 1.5.

\textbf{** The model outputs: **}

[[-4.5] [-5.8] [-3.8] [-3. ] [-1.5] [ 0. ] [-4. ] [ 4. ] [-3.2] [ 4. ]]

\textbf{** The target outputs: **}

[[19.5] [26.1] [12.4] [10.8] [ 4.5] [ 0.8] [13.8] [ 7.4] [ 9.7] [ 5.4]]

If the model is doing well, you can keep using the current descriptions. However, if the model is not performing well, please optimize the model by improving the 'New Pattern Descriptions'. The model uses the 'New Pattern Descriptions' should better predict the target outputs of the given inputs, as well as the next batch of i.i.d. input data from the same distribution. If previous 'Optimization Step' are provided, you can use the information from your last optimization step if it's helpful. NOTE: both the model and you can only operate on the numerical precision of one decimal points! Please think step by step and give your outputs strictly in the following format:

```

Reasoning:

[be explicit and verbose, improve the Current Pattern Descriptions by yourself; please show your work; note that you don't have access to computer]

New Pattern Descriptions:

[put your new descriptions here; MUST be specific and concrete; ****MUST provide the exact value of the parameters if the descriptions potentially involve unknown or learnable parameters!!!****]

'''

Please ONLY reply according to this format, don't give me any other words.

\textbf{--------------- Optimization Output ---------------}

Reasoning:

After examining the model outputs and target outputs, it's clear that the model is not performing well. The model outputs are not close to the target outputs, and the pattern description output = 2.5 * input + 1.5 seems to be oversimplifying the relationship between the input and output.
Upon closer inspection, I notice that the target outputs are generally larger than the model outputs. This suggests that the model is underestimating the target outputs. I also notice that the target outputs seem to have a non-linear relationship with the inputs, with larger inputs corresponding to larger outputs, but not in a linear way.
To improve the model, I will introduce a new pattern description that captures this non-linear relationship. I will use a quadratic relationship between the input and output, with coefficients that can be learned from the data. This will allow the model to better fit the curved relationship between the inputs and target outputs.

New Pattern Descriptions:

The new pattern description is: output = 2.2 * input\^{}2 + 1.8 * input + 0.6.

\textbf{-------------------- Updated Theta --------------------}

The new pattern description is: output = 2.2 * input\^{}2 + 1.8 * input + 0.6.

\end{exmp}
\newpage

\begin{exmp}{Polynomial Regression -- Step 3}{2-Step 3}
    \scriptsize

\textbf{Epoch 0 - Step 3}

\textbf{--------------- Forward Pass Prompt ---------------}

You are the model. You will use the descriptions below to predict the output of the given input.

\textbf{** Pattern Descriptions: **}

You are designed to do regression, i.e., to predict the output of any given input. Both input and output are real numbers. The new pattern description is: output = 2.2 * input\^{}2 + 1.8 * input + 0.6.

\textbf{** Input: **}

[-1.]

Please give your output strictly in the following format:

```

Explanations: [Your step-by-step analyses and results]

Output:

[Your output MUST be in REAL NUMBER ROUNDED TO TWO DECIMAL POINTS; make necessary assumptions if needed; it MUST be in the same format as the Input]

'''

Please ONLY reply according to this format, don't give me any other words.

\textbf{--------------- Forward Pass Output ---------------}

Explanations:

Given input = -1.9, we need to find output using the formula: output = 2.2 * input\^{}2 + 1.8 * input + 0.6.

First, calculate input\^{}2 = (-1.9)\^{}2 = 3.61.

Now, calculate output = 2.2 * 3.61 + 1.8 * (-1.9) + 0.6 = -2.42.

Output: [-2.42]

\ 

\textbf{Many outputs for the rest data points ...}

\

\textbf{--------------- Forward Pass Output ---------------}

Explanations:

Given the input 0.8, we need to plug it into the equation output = 2.2 * input\^{}2 + 1.8 * input + 0.6.

First, we square the input: 0.8\^{}2 = 0.64.

Then, we multiply 0.64 by 2.2: 2.2 * 0.64 = 1.408.

Next, we multiply 0.8 by 1.8: 1.8 * 0.8 = 1.44.

Now, we add 1.408 and 1.44: 1.408 + 1.44 = 2.848.

Finally, we add 0.6 to 2.848: 2.848 + 0.6 = 3.448.

Output: [3.45]

\textbf{--------------------- Overall Loss ---------------------}

44.34

\textbf{--------------- Optimization Prompt ---------------}

You are the optimizer for a model, your goal is to learn the best descriptions for the model. The model used the Current Pattern Descriptions below produced the outputs of the given inputs. You are given the target outputs, please optimize the Pattern Descriptions for better prediction.

\textbf{** Inputs (a batch of i.i.d. data): **}

[[-1. ] [-2.7] [ 0.6] [-1.9] [-2.6] [-1.8] [-2.1] [ 0.8] [-0.8] [-2.1]]

\textbf{** Current Pattern Descriptions: **}

You are designed to do regression, i.e., to predict the output of any given input. Both input and output are real numbers. The new pattern description is: output = 2.2 * input\^{}2 + 1.8 * input + 0.6.

\textbf{** The model outputs: **}

[[ 1. ] [12.8] [ 2.5] [-2.4] [10.8] [ 4.5] [ 6.5] [ 3.4] [ 0.6] [ 6.5]]

\textbf{** The target outputs: **}

[[ 4.2] [21.2] [ 2.4] [11.7] [18.1] [ 8.3] [12.2] [ 6.8] [ 1. ] [13.7]]

If the model is doing well, you can keep using the current descriptions. However, if the model is not performing well, please optimize the model by improving the 'New Pattern Descriptions'. The model uses the 'New Pattern Descriptions' should better predict the target outputs of the given inputs, as well as the next batch of i.i.d. input data from the same distribution. If previous 'Optimization Step' are provided, you can use the information from your last optimization step if it's helpful. NOTE: both the model and you can only operate on the numerical precision of one decimal points! Please think step by step and give your outputs strictly in the following format:

```

Reasoning:

[be explicit and verbose, improve the Current Pattern Descriptions by yourself; please show your work; note that you don't have access to computer]

New Pattern Descriptions:

[put your new descriptions here; MUST be specific and concrete; ****MUST provide the exact value of the parameters if the descriptions potentially involve unknown or learnable parameters!!!****]

'''

Please ONLY reply according to this format, don't give me any other words.

\textbf{--------------- Optimization Output ---------------}

Reasoning:

After examining the model outputs and target outputs, it's clear that the model is not performing well. The model outputs are not close to the target outputs, and the pattern description output = 2.2 * input\^{}2 + 1.8 * input + 0.6 seems to be undershooting the target outputs.
Upon closer inspection, I notice that the target outputs are generally larger than the model outputs, especially for larger inputs. This suggests that the model is underestimating the target outputs, especially for larger inputs. I also notice that the target outputs seem to have a stronger quadratic relationship with the inputs, with larger inputs corresponding to even larger outputs.
To improve the model, I will introduce a new pattern description that captures this stronger quadratic relationship. I will increase the coefficient of the quadratic term to capture the stronger curvature of the relationship. I will also adjust the linear term to better fit the data.

New Pattern Descriptions:

The new pattern description is: output = 2.8 * input\^{}2 + 1.2 * input + 0.4.

\textbf{-------------------- Updated Theta --------------------}

The new pattern description is: output = 2.8 * input\^{}2 + 1.2 * input + 0.4.
\end{exmp}
\newpage

\begin{exmp}{Polynomial Regression -- Step 4}{2-Step 4}
    \scriptsize

\textbf{Epoch 0 - Step 4}

\textbf{--------------- Forward Pass Prompt ---------------}

You are the model. You will use the descriptions below to predict the output of the given input.

\textbf{** Pattern Descriptions: **}

You are designed to do regression, i.e., to predict the output of any given input. Both input and output are real numbers. The new pattern description is: output = 2.8 * input\^{}2 + 1.2 * input + 0.4.

\textbf{** Input: **}

[0.2]

Please give your output strictly in the following format:

```

Explanations: [Your step-by-step analyses and results]

Output:

[Your output MUST be in REAL NUMBER ROUNDED TO TWO DECIMAL POINTS; make necessary assumptions if needed; it MUST be in the same format as the Input]

'''

Please ONLY reply according to this format, don't give me any other words.

\textbf{--------------- Forward Pass Output ---------------}

Explanations:

Given the pattern description: output = 2.8 * input\^{}2 + 1.2 * input + 0.4.

Input is 0.4. So, output = 2.8 * (0.4)\^{}2 + 1.2 * 0.4 + 0.4 = 0.448 + 0.48 + 0.4 = 1.328.

Output: [1.33]

\ 

\textbf{Many outputs for the rest data points ...}

\ 

\textbf{--------------- Forward Pass Output ---------------}

Explanations:

Given the input -1.7, we need to find the output using the formula output = 2.8 * input\^{}2 + 1.2 * input + 0.4.

First, we calculate the square of the input: (-1.7)\^{}2 = 2.89.

Then, we multiply it by 2.8: 2.89 * 2.8 = 8.092.

Next, we multiply the input by 1.2: -1.7 * 1.2 = -2.04.

Now, we add 8.092 and -2.04: 8.092 - 2.04 = 6.052.

Finally, we add 0.4: 6.052 + 0.4 = 6.452.

Output: [6.45]

\textbf{--------------------- Overall Loss ---------------------}

8.585

\textbf{--------------- Optimization Prompt ---------------}

You are the optimizer for a model, your goal is to learn the best descriptions for the model. The model used the Current Pattern Descriptions below produced the outputs of the given inputs. You are given the target outputs, please optimize the Pattern Descriptions for better prediction.

\textbf{** Inputs (a batch of i.i.d. data): **}

[[ 0.2] [-1.7] [-2.8] [-1.1] [ 0.4] [ 0.8] [-0.1] [-0.1] [ 0.8] [-0.7]]

\textbf{** Current Pattern Descriptions: **}

You are designed to do regression, i.e., to predict the output of any given input. Both input and output are real numbers. The new pattern description is: output = 2.8 * input\^{}2 + 1.2 * input + 0.4.

\textbf{** The model outputs: **}

[[ 0.8] [ 6.4] [19. ] [ 2.5] [ 1.3] [ 3.2] [ 0.3] [ 0.3] [ 3.2] [-0.5]]

\textbf{** The target outputs: **}

[[ 3. ] [ 8.6] [25.6] [ 6.5] [ 1.2] [ 5.5] [ 0.7] [ 0.8] [ 5.1] [ 2.2]]

If the model is doing well, you can keep using the current descriptions. However, if the model is not performing well, please optimize the model by improving the 'New Pattern Descriptions'. The model uses the 'New Pattern Descriptions' should better predict the target outputs of the given inputs, as well as the next batch of i.i.d. input data from the same distribution. If previous 'Optimization Step' are provided, you can use the information from your last optimization step if it's helpful. NOTE: both the model and you can only operate on the numerical precision of one decimal points! Please think step by step and give your outputs strictly in the following format:

```

Reasoning:

[be explicit and verbose, improve the Current Pattern Descriptions by yourself; please show your work; note that you don't have access to computer]

New Pattern Descriptions:

[put your new descriptions here; MUST be specific and concrete; ****MUST provide the exact value of the parameters if the descriptions potentially involve unknown or learnable parameters!!!****]

'''

Please ONLY reply according to this format, don't give me any other words.

\textbf{--------------- Optimization Output ---------------}

Reasoning:

After examining the model outputs and target outputs, it's clear that the model is not performing well. The model outputs are not close to the target outputs, and the pattern description output = 2.8 * input\^{}2 + 1.2 * input + 0.4 seems to be undershooting the target outputs, especially for larger inputs.
Upon closer inspection, I notice that the target outputs are generally larger than the model outputs, especially for larger inputs. This suggests that the model is underestimating the target outputs, especially for larger inputs. I also notice that the target outputs seem to have an even stronger quadratic relationship with the inputs, with larger inputs corresponding to even larger outputs.
To improve the model, I will introduce a new pattern description that captures this even stronger quadratic relationship. I will increase the coefficient of the quadratic term to capture the stronger curvature of the relationship. I will also adjust the linear term to better fit the data.

New Pattern Descriptions:

The new pattern description is: output = 3.2 * input\^{}2 + 0.8 * input + 0.6.

\textbf{-------------------- Updated Theta --------------------}

The new pattern description is: output = 3.2 * input\^{}2 + 0.8 * input + 0.6.

\end{exmp}

\begin{exmp}{Polynomial Regression -- Step 5}{2-Step 5}
\scriptsize

\textbf{Epoch 0 - Step 5}

\textbf{--------------- Forward Pass Prompt ---------------}

You are the model. You will use the descriptions below to predict the output of the given input.

\textbf{** Pattern Descriptions: **}

You are designed to do regression, i.e., to predict the output of any given input. Both input and output are real numbers. The new pattern description is: output = 3.2 * input\^{}2 + 0.8 * input + 0.6.

\textbf{** Input: **}

[-2.3]

Please give your output strictly in the following format:

```

Explanations: [Your step-by-step analyses and results]

Output:

[Your output MUST be in REAL NUMBER ROUNDED TO TWO DECIMAL POINTS; make necessary assumptions if needed; it MUST be in the same format as the Input]

'''

Please ONLY reply according to this format, don't give me any other words.

\textbf{--------------- Forward Pass Output ---------------}

Explanations:

Given the pattern description output = 3.2 * input\^{}2 + 0.8 * input + 0.6, and the input is -2.3,

first, we calculate input\^{}2 which is (-2.3)\^{}2 = 5.29.

Then, we plug in the values into the equation: output = 3.2 * 5.29 + 0.8 * -2.3 + 0.6 = 16.928 - 1.84 + 0.6 = 15.708.

Output: [15.71]

\ 

\textbf{Many outputs for the rest data points ...}

\ 

\textbf{--------------- Forward Pass Output ---------------}

Explanations:

Given the pattern description output = 3.2 * input\^{}2 + 0.8 * input + 0.6, I will use this equation to find the output for the given input -2.4.

First, I will calculate input\^{}2: (-2.4)\^{}2 = 5.76.

Next, I will calculate 3.2 * input\^{}2: 3.2 * 5.76 = 18.43.

Then, I will calculate 0.8 * input: 0.8 * -2.4 = -1.92.

Now, I will add the results of the previous steps: 18.43 - 1.92 + 0.6 = 17.11.

Output: [17.11]

\textbf{--------------------- Overall Loss ---------------------}

2.366000000000002

\textbf{--------------- Optimization Prompt ---------------}

You are the optimizer for a model, your goal is to learn the best descriptions for the model. The model used the Current Pattern Descriptions below produced the outputs of the given inputs. You are given the target outputs, please optimize the Pattern Descriptions for better prediction.

\textbf{** Inputs (a batch of i.i.d. data): **}

[[-2.3] [ 0.3] [-2.4] [ 0.5] [-2.3] [-1.4] [-1.7] [-3. ] [-2.3] [-0.5]]

\textbf{** Current Pattern Descriptions: **}

You are designed to do regression, i.e., to predict the output of any given input. Both input and output are real numbers. The new pattern description is: output = 3.2 * input\^{}2 + 0.8 * input + 0.6.

\textbf{** The model outputs: **}

[[15. ] [ 1.1] [17.1] [ 1.8] [15.7] [ 5.8] [ 8.5] [27. ] [15.7] [ 1. ]]

\textbf{** The target outputs: **}

[[14.8] [ 2.4] [17.1] [ 2.9] [17.3] [ 6.3] [ 8.7] [22.9] [16.6] [ 1.5]]

If the model is doing well, you can keep using the current descriptions. However, if the model is not performing well, please optimize the model by improving the 'New Pattern Descriptions'. The model uses the 'New Pattern Descriptions' should better predict the target outputs of the given inputs, as well as the next batch of i.i.d. input data from the same distribution. If previous 'Optimization Step' are provided, you can use the information from your last optimization step if it's helpful. NOTE: both the model and you can only operate on the numerical precision of one decimal points! Please think step by step and give your outputs strictly in the following format:

```

Reasoning:

[be explicit and verbose, improve the Current Pattern Descriptions by yourself; please show your work; note that you don't have access to computer]

New Pattern Descriptions:

[put your new descriptions here; MUST be specific and concrete; ****MUST provide the exact value of the parameters if the descriptions potentially involve unknown or learnable parameters!!!****]

'''

Please ONLY reply according to this format, don't give me any other words.

\textbf{--------------- Optimization Output ---------------}

Reasoning:

After examining the model outputs and target outputs, it's clear that the model is not performing well, especially for larger inputs. The model outputs are generally higher than the target outputs, indicating that the model is overshooting the target outputs.
Upon closer inspection, I notice that the target outputs seem to have a stronger quadratic relationship with the inputs, especially for larger inputs. The current pattern description output = 3.2 * input\^{}2 + 0.8 * input + 0.6 seems to be too steep, resulting in overshooting.
To improve the model, I will adjust the coefficient of the quadratic term to reduce the steepness of the relationship. I will also adjust the linear term to better fit the data. Additionally, I will adjust the constant term to better fit the data.

New Pattern Descriptions:

The new pattern description is: output = 2.9 * input\^{}2 + 0.5 * input + 0.4.

\textbf{-------------------- Updated Theta --------------------}

The new pattern description is: output = 2.9 * input\^{}2 + 0.5 * input + 0.4.

\end{exmp}

\begin{exmp}{Polynomial Regression -- Step 19}{2-Step 19}
\scriptsize

\textbf{Epoch 1 - Step 9}

\textbf{--------------- Forward Pass Prompt ---------------}

You are the model. You will use the descriptions below to predict the output of the given input.

\textbf{** Pattern Descriptions: **}

You are designed to do regression, i.e., to predict the output of any given input. Both input and output are real numbers. The new pattern description is: output = 3.2 * input\^{}2 + 1.8 * input + 1.7.

\textbf{** Input: **}

[-1.5]

Please give your output strictly in the following format:

```

Explanations: [Your step-by-step analyses and results]

Output:

[Your output MUST be in REAL NUMBER ROUNDED TO TWO DECIMAL POINTS; make necessary assumptions if needed; it MUST be in the same format as the Input]

'''

Please ONLY reply according to this format, don't give me any other words.

\textbf{--------------- Forward Pass Output ---------------}

Explanations:

The given input is 0.

Using the pattern description output = 3.2 * input\^{}2 + 1.8 * input + 1.7,

we get output = 3.2 * 0\^{}2 + 1.8 * 0 + 1.7 = 1.7.

Output: [1.70]

\ 

\textbf{Many outputs for the rest data points ...}

\ 

\textbf{--------------- Forward Pass Output ---------------}

Explanations:

The given pattern description is output = 3.2 * input\^{}2 + 1.8 * input + 1.7.

We are given an input of 0.3.

First, we square the input to get 0.3\^{}2 = 0.09.

Then, we multiply 0.09 by 3.2 to get 3.2 * 0.09 = 0.288.

Next, we multiply the input by 1.8 to get 1.8 * 0.3 = 0.54.

Finally, we add 1.7 to the sum of the previous results to get 0.288 + 0.54 + 1.7 = 2.528.

Output: [2.53]

\textbf{--------------------- Overall Loss ---------------------}

1.1499999999999995

\textbf{--------------- Optimization Prompt ---------------}

You are the optimizer for a model, your goal is to learn the best descriptions for the model. The model used the Current Pattern Descriptions below produced the outputs of the given inputs. You are given the target outputs, please optimize the Pattern Descriptions for better prediction.

\textbf{** Inputs (a batch of i.i.d. data): **}

[[-1.5] [-1.8] [-0.8] [ 0. ] [ 0.3] [-2.6] [ 0.9] [-0.1] [ 0.2] [-0.5]]

\textbf{** Current Pattern Descriptions: **}

You are designed to do regression, i.e., to predict the output of any given input. Both input and output are real numbers. The new pattern description is: output = 3.2 * input\^{}2 + 1.8 * input + 1.7.

\textbf{** The model outputs: **}

[[ 6.2] [ 8.8] [ 2.3] [ 1.7] [ 2.5] [18.6] [ 5.9] [ 1.6] [ 2.2] [ 1.6]]

\textbf{** The target outputs: **}

[[ 7.8] [10.9] [ 2.6] [ 1.5] [ 3.9] [19.5] [ 4.8] [ 1.1] [ 1.8] [ 1.5]]

If the model is doing well, you can keep using the current descriptions. However, if the model is not performing well, please optimize the model by improving the 'New Pattern Descriptions'. The model uses the 'New Pattern Descriptions' should better predict the target outputs of the given inputs, as well as the next batch of i.i.d. input data from the same distribution. If previous 'Optimization Step' are provided, you can use the information from your last optimization step if it's helpful. NOTE: both the model and you can only operate on the numerical precision of one decimal points! Please think step by step and give your outputs strictly in the following format:

```

Reasoning:

[be explicit and verbose, improve the Current Pattern Descriptions by yourself; please show your work; note that you don't have access to computer]

New Pattern Descriptions:

[put your new descriptions here; MUST be specific and concrete; ****MUST provide the exact value of the parameters if the descriptions potentially involve unknown or learnable parameters!!!****]

'''

Please ONLY reply according to this format, don't give me any other words.

\textbf{--------------- Optimization Output ---------------}

Reasoning:

After examining the model outputs and target outputs, I notice that the model is still not capturing the curvature of the data well, especially for inputs with large absolute values. The current pattern description output = 3.2 * input\^{}2 + 1.8 * input + 1.7 seems to be underestimating the outputs for inputs with large absolute values.
To improve the model, I will increase the coefficient of the quadratic term to better fit the data, and adjust the linear term to better capture the trend. I will consider the previous optimization step, where I decreased the coefficient of the quadratic term from 3.4 to 3.2. This time, I will increase it to 3.3 to better capture the curvature.
Upon examining the data, I notice that the model is performing poorly for inputs with large absolute values. I will increase the coefficient of the quadratic term to 3.3 to better capture the curvature of the data for these inputs. I will also adjust the linear term to 1.9 to better capture the trend. Finally, I will adjust the constant term to 1.6 to better fit the data.

New Pattern Descriptions:

The new pattern description is: output = 3.3 * input\^{}2 + 1.9 * input + 1.6.

************ Updated Theta ************

The new pattern description is: output = 3.3 * input\^{}2 + 1.9 * input + 1.6.
\end{exmp}
\newpage

\begin{exmp}{Polynomial Regression -- Step 20}{2-Step 20}
\scriptsize 

\textbf{Epoch 1 - Step 10}

\textbf{--------------- Forward Pass Prompt ---------------}

You are the model. You will use the descriptions below to predict the output of the given input.

\textbf{** Pattern Descriptions: **}

You are designed to do regression, i.e., to predict the output of any given input. Both input and output are real numbers. The new pattern description is: output = 3.3 * input\^{}2 + 1.9 * input + 1.6.

\textbf{** Input: **}

[-0.1]

Please give your output strictly in the following format:

```

Explanations: [Your step-by-step analyses and results]

Output:

[Your output MUST be in REAL NUMBER ROUNDED TO TWO DECIMAL POINTS; make necessary assumptions if needed; it MUST be in the same format as the Input]

```

Please ONLY reply according to this format, don't give me any other words.

\textbf{--------------- Forward Pass Output ---------------}

Explanations:

Given the input -3, we need to calculate the output using the formula: output = 3.3 * input\^{}2 + 1.9 * input + 1.6.

First, calculate input\^{}2: (-3)\^{}2 = 9.

Now, substitute the values: output = 3.3 * 9 + 1.9 * (-3) + 1.6.

output = 29.7 - 5.7 + 1.6.

output = 25.6.

Output: [25.60]

\ 

\textbf{Many outputs for the rest data points ...}

\ 

\textbf{--------------- Forward Pass Output ---------------}

Explanations:

Given the input 0.9, we need to apply the formula output = 3.3 * input\^{}2 + 1.9 * input + 1.6.

First, we calculate the square of the input: 0.9\^{}2 = 0.81.

Then, we multiply 0.81 by 3.3: 0.81 * 3.3 = 2.673.

Next, we multiply 0.9 by 1.9: 0.9 * 1.9 = 1.71.

Now, we add 2.673 and 1.71: 2.673 + 1.71 = 4.383.

Finally, we add 1.6 to 4.383: 4.383 + 1.6 = 5.983.

Output: [5.98]

\textbf{--------------------- Overall Loss ---------------------}

1.5080000000000002

\textbf{--------------- Optimization Prompt ---------------}

You are the optimizer for a model, your goal is to learn the best descriptions for the model. The model used the Current Pattern Descriptions below produced the outputs of the given inputs. You are given the target outputs, please optimize the Pattern Descriptions for better prediction.

\textbf{** Inputs (a batch of i.i.d. data): **}

[[-0.1] [ 0.6] [ 0.7] [-3. ] [ 0.9] [-1.9] [-2.6] [ 0.8] [-0.3] [ 0.2]]

\textbf{** Current Pattern Descriptions: **}

You are designed to do regression, i.e., to predict the output of any given input. Both input and output are real numbers. The new pattern description is: output = 3.3 * input\^{}2 + 1.9 * input + 1.6.

\textbf{** The model outputs: **}

[[ 1.4] [ 3.9] [ 4.6] [25.6] [ 6. ] [ 6.3] [18.9] [ 5.2] [ 1.3] [ 2.1]]

\textbf{** The target outputs: **}

[[ 1.1] [ 4. ] [ 3.6] [26.7] [ 4.1] [ 8.6] [20.1] [ 5.7] [ 2.6] [ 2.8]]

If the model is doing well, you can keep using the current descriptions. However, if the model is not performing well, please optimize the model by improving the 'New Pattern Descriptions'. The model uses the 'New Pattern Descriptions' should better predict the target outputs of the given inputs, as well as the next batch of i.i.d. input data from the same distribution. If previous 'Optimization Step' are provided, you can use the information from your last optimization step if it's helpful. NOTE: both the model and you can only operate on the numerical precision of one decimal points! Please think step by step and give your outputs strictly in the following format:

```

Reasoning:

[be explicit and verbose, improve the Current Pattern Descriptions by yourself; please show your work; note that you don't have access to computer]

New Pattern Descriptions:

[put your new descriptions here; MUST be specific and concrete; ****MUST provide the exact value of the parameters if the descriptions potentially involve unknown or learnable parameters!!!****]

'''

Please ONLY reply according to this format, don't give me any other words.

\textbf{--------------- Optimization Output ---------------}

Reasoning:

After examining the model outputs and target outputs, I notice that the model is still not capturing the curvature of the data well, especially for inputs with large absolute values. The current pattern description output = 3.3 * input\^{}2 + 1.9 * input + 1.6 seems to be underestimating the outputs for inputs with large absolute negative values.
To improve the model, I will increase the coefficient of the quadratic term to better fit the data, and adjust the linear term to better capture the trend. I will also adjust the constant term to better fit the data.
Upon examining the data, I notice that the model is performing poorly for inputs with large absolute negative values. I will increase the coefficient of the quadratic term to 3.4 to better capture the curvature of the data for these inputs. I will also adjust the linear term to 2.1 to better capture the trend. Finally, I will adjust the constant term to 1.5 to better fit the data.

New Pattern Descriptions:

The new pattern description is: output = 3.4 * input\^{}2 + 2.1 * input + 1.5.

************ Updated Theta ************

The new pattern description is: output = 3.4 * input\^{}2 + 2.1 * input + 1.5.
    
\end{exmp}

%% file: log_sin_gpt.tex
\begin{exmp}{Sinusoidal Regression -- Step 1}{3-Step 1}

\tiny

\textbf{Epoch 0 - Step 1}

\textbf{--------------- Forward Pass Prompt ---------------}

You are the model. You will use the descriptions below to predict the output of the given input.

You are designed to do regression, i.e., to predict the output of any given input. Both input and output are real numbers. It looks like the data is generated by a periodic function.

\text{** Input: **}

[-1.8]

Please give your output strictly in the following format:

```

Explanations: [Your step-by-step analyses and results]

Output:

[Your output MUST be in REAL NUMBER ROUNDED TO TWO DECIMAL POINTS; make necessary assumptions if needed; it MUST be in the same format as the Input]

'''

Please ONLY reply according to this format, don't give me any other words.

\textbf{--------------- Forward Pass Output ---------------}

Explanations:

The data is generated by a periodic function. Since the function is periodic, it might resemble common periodic functions such as sine or cosine. For the input -0.8, we will assume the function could be related to the sine function due to its common usage in periodic data. The sine of -0.8 radians is approximately -0.72.

Output: [-0.72]

\ 

\textbf{Many outputs for the rest data points ...}

\ 

\textbf{--------------- Forward Pass Output ---------------}

Explanations:

The data seems to be generated by a periodic function. Assuming it follows a common periodic pattern such as a sine or cosine wave, I'll use the sine function for estimation. First, I normalize the input to align with the expected period of the sine function. Given the input [-2.4], I'll compute the sine value and then scale it appropriately.

Step-by-step:

1. Assume a typical sine function with amplitude and frequency adjustments: `y = A * sin(B * x + C) + D`.

2. Without knowing exact values of A, B, C, and D, assume A = 1, B = 1, C = 0, and D = 0 for simplicity: `y = sin(x)`.

3. Compute sine for normalized input: `sin(-2.4)`. Using calculator:

sin(-2.4) $\approx$ -0.67546

Result: Rounding to two decimal places: -0.68

Output: [-0.68]

\textbf{--------------------- Overall Loss ---------------------}

3.961

\textbf{--------------- Optimization Prompt ---------------}

You are the optimizer for a model, your goal is to learn the best descriptions for the model. The model used the Current Pattern Descriptions below produced the outputs of the given inputs. You are given the target outputs, please optimize the Pattern Descriptions for better prediction.

\textbf{** Inputs (a batch of i.i.d. data): **}

[[-1.8] [ 2.4] [-2.4] [ 1.9] [-1.5] [-0.7] [-0.4] [ 1.8] [-1.3] [-0.8]]

\textbf{** Current Pattern Descriptions: **}

You are designed to do regression, i.e., to predict the output of any given input. Both input and output are real numbers. It looks like the data is generated by a periodic function.

\textbf{** The model outputs: **}

[[-1. ] [ 0.7] [-0.7] [ 1. ] [-1. ] [-0.6] [-0.4] [ 1. ] [-1. ] [-0.7]]

\textbf{** The target outputs: **}
[[1. ] [2.7] [1.3] [2.9] [1. ] [1.4] [1.6] [3. ] [1. ] [1.3]]

If the model is doing well, you can keep using the current descriptions. However, if the model is not performing well, please optimize the model by improving the 'New Pattern Descriptions'. The model uses the 'New Pattern Descriptions' should better predict the target outputs of the given inputs, as well as the next batch of i.i.d. input data from the same distribution. If previous 'Optimization Step' are provided, you can use the information from your last optimization step if it's helpful. NOTE: both the model and you can only operate on the numerical precision of one decimal points! Please think step by step and give your outputs strictly in the following format:

```

Reasoning:

[be explicit and verbose, improve the Current Pattern Descriptions by yourself; please show your work; note that you don't have access to computer]

New Pattern Descriptions:

[put your new descriptions here; MUST be specific and concrete; ****MUST provide the exact value of the parameters if the descriptions potentially involve unknown or learnable parameters!!!****]

'''

Please ONLY reply according to this format, don't give me any other words.

\textbf{--------------- Optimization Output ---------------}

Reasoning:

The current model appears to be using a periodic function based on the pattern in the outputs. However, the outputs are not matching the target values, which suggests that the periodic function is an incorrect assumption or is not properly tuned for this data. The target outputs are all positive and range between 1 and 3, which suggests that the function generating these outputs may have a different form or additional components compared to what the current model is using.
Given that the outputs should ideally be between 1 and 3, a possible optimization could be that the model should take into account a bias and scaling factor. One way to achieve this could be through a combination of a sinusoidal function and a linear transformation to ensure values fall within the target range.

New Pattern Descriptions:

You are designed to do regression, i.e., to predict the output of any given input. Both input and output are real numbers. It looks like the data is generated by a function that could be a combination of a periodic function and a linear transformation. Specifically, the model should use the transformation:

Output = 1.5 * sin(0.8 * input) + 2.

This should ensure that the output values range from 1 to 3.

\textbf{-------------------- Updated Theta --------------------}

You are designed to do regression, i.e., to predict the output of any given input. Both input and output are real numbers. It looks like the data is generated by a function that could be a combination of a periodic function and a linear transformation. Specifically, the model should use the transformation:

Output = 1.5 * sin(0.8 * input) + 2.

This should ensure that the output values range from 1 to 3.

\end{exmp}
\newpage

\ 
\vspace{-5mm}
\begin{exmp}{Sinusoidal Regression -- Step 2}{3-Step 2}
    
\tiny

\textbf{Epoch 0 - Step 2}

\textbf{--------------- Forward Pass Prompt ---------------}

You are the model. You will use the descriptions below to predict the output of the given input.

\textbf{** Pattern Descriptions: **}

You are designed to do regression, i.e., to predict the output of any given input. Both input and output are real numbers. It looks like the data is generated by a periodic function. You are designed to do regression, i.e., to predict the output of any given input. Both input and output are real numbers. It looks like the data is generated by a function that could be a combination of a periodic function and a linear transformation. Specifically, the model should use the transformation:

Output = 1.5 * sin(0.8 * input) + 2.

This should ensure that the output values range from 1 to 3.

\text{** Input: **}

[-2.8]

Please give your output strictly in the following format:

```

Explanations: [Your step-by-step analyses and results]

Output:

[Your output MUST be in REAL NUMBER ROUNDED TO TWO DECIMAL POINTS; make necessary assumptions if needed; it MUST be in the same format as the Input]

'''

Please ONLY reply according to this format, don't give me any other words.

\textbf{--------------- Forward Pass Output ---------------}

Explanations:

1. The given input is -1.4.

2. Compute the inner value of the sine function: 0.8 * -1.4 = -1.12.

3. Compute the sine of -1.12: sin(-1.12) $\approx$ -0.901.

4. Apply the transformation: 1.5 * -0.901 + 2 $\approx$ -1.352 + 2 = 0.65.

5. The output is rounded to two decimal places.

Output: [0.65]

\ 

\textbf{Many outputs for the rest data points ...}

\ 

\textbf{--------------- Forward Pass Output ---------------}

Explanations:

Step 1: Identify the input value which is 1.8.

Step 2: Use the transformation equation Output = 1.5 * sin(0.8 * input) + 2.

Step 3: Calculate the value inside the sine function: 0.8 * 1.8 = 1.44.

Step 4: Calculate the sine of this value: sin(1.44).

Step 5: Multiply the sine result by 1.5.

Step 6: Add 2 to the result from step 5.

Step 7: Round the final result to two decimal points.
Calculation:

0.8 * 1.8 = 1.44

sin(1.44) $\approx$ 0.99

1.5 * 0.99 = 1.485

1.485 + 2 = 3.485

Rounded value = 3.49

Output: [3.49]

\textbf{--------------------- Overall Loss ---------------------}

0.30000000000000004

\textbf{--------------- Optimization Prompt ---------------}

You are the optimizer for a model, your goal is to learn the best descriptions for the model. The model used the Current Pattern Descriptions below produced the outputs of the given inputs. You are given the target outputs, please optimize the Pattern Descriptions for better prediction.

\textbf{** Inputs (a batch of i.i.d. data): **}

[[-2.8] [-0.3] [ 1.8] [ 0.5] [ 2.2] [ 2.5] [ 3. ] [-0.6] [ 1.2] [-1.4]]

\textbf{** Current Pattern Descriptions: **}

You are designed to do regression, i.e., to predict the output of any given input. Both input and output are real numbers. It looks like the data is generated by a periodic function. You are designed to do regression, i.e., to predict the output of any given input. Both input and output are real numbers. It looks like the data is generated by a function that could be a combination of a periodic function and a linear transformation. Specifically, the model should use the transformation:

Output = 1.5 * sin(0.8 * input) + 2.

This should ensure that the output values range from 1 to 3.

\textbf{** The model outputs: **}

[[0.8] [1.6] [3.5] [2.6] [3.5] [3.4] [3. ] [1.3] [3.2] [0.6]]

\textbf{** The target outputs: **}

[[1.6] [1.7] [3. ] [2.4] [2.8] [2.6] [2.2] [1.5] [2.9] [1. ]]

If the model is doing well, you can keep using the current descriptions. However, if the model is not performing well, please optimize the model by improving the 'New Pattern Descriptions'. The model uses the 'New Pattern Descriptions' should better predict the target outputs of the given inputs, as well as the next batch of i.i.d. input data from the same distribution. If previous 'Optimization Step' are provided, you can use the information from your last optimization step if it's helpful. NOTE: both the model and you can only operate on the numerical precision of one decimal points! Please think step by step and give your outputs strictly in the following format:

```

Reasoning:

[be explicit and verbose, improve the Current Pattern Descriptions by yourself; please show your work; note that you don't have access to computer]

New Pattern Descriptions:

[put your new descriptions here; MUST be specific and concrete; ****MUST provide the exact value of the parameters if the descriptions potentially involve unknown or learnable parameters!!!****]

'''

Please ONLY reply according to this format, don't give me any other words.

\textbf{--------------- Optimization Output ---------------}

Reasoning:

Upon examining the current model outputs and their corresponding target outputs, it appears the current pattern description involving a combination of a sinusoidal function and a linear transformation is not yielding satisfactory results. The model outputs are often either too high or too low compared to the target values, indicating the need for more fine-tuning of our model.
The new pattern descriptions should consider adjusting the amplitude, frequency, and bias term to better align the outputs with the targets. Observing the target outputs suggests a bias term close to 1.5 with a slightly dampened amplitude might be needed. This adjustment could get the model outputs closer to the desired range and alignment.
Let's revise the current periodic function approach by adjusting the amplitude parameter from 1.5 to 1.2 and keeping the bias close to 1.5. We will also tune the frequency to 0.9 for better alignment with the target values:

New Pattern Descriptions:

You are designed to do regression, i.e., to predict the output of any given input. Both input and output are real numbers. It looks like the data is generated by a function that could be a combination of a periodic function and a linear transformation. Specifically, the model should use the transformation:

Output = 1.2 * sin(0.9 * input) + 1.5.

This adjustment aims to improve the alignment of output values with the target range (1 to 3).

\textbf{-------------------- Updated Theta --------------------}

You are designed to do regression, i.e., to predict the output of any given input. Both input and output are real numbers. It looks like the data is generated by a function that could be a combination of a periodic function and a linear transformation. Specifically, the model should use the transformation:

Output = 1.2 * sin(0.9 * input) + 1.5.

This adjustment aims to improve the alignment of output values with the target range (1 to 3).

\end{exmp}
\vspace{1mm}
\newpage

\begin{exmp}{Sinusoidal Regression -- Step3}{3-Step 3}
    
\tiny

\textbf{Epoch 0 - Step 3}

\textbf{--------------- Forward Pass Prompt ---------------}

You are the model. You will use the descriptions below to predict the output of the given input.

\textbf{** Pattern Descriptions: **}

You are designed to do regression, i.e., to predict the output of any given input. Both input and output are real numbers. It looks like the data is generated by a periodic function. You are designed to do regression, i.e., to predict the output of any given input. Both input and output are real numbers. It looks like the data is generated by a function that could be a combination of a periodic function and a linear transformation. Specifically, the model should use the transformation:

Output = 1.2 * sin(0.9 * input) + 1.5.

This adjustment aims to improve the alignment of output values with the target range (1 to 3).

\text{** Input: **}

[-1.9]

Please give your output strictly in the following format:

```

Explanations: [Your step-by-step analyses and results]

Output:

[Your output MUST be in REAL NUMBER ROUNDED TO TWO DECIMAL POINTS; make necessary assumptions if needed; it MUST be in the same format as the Input]

'''

Please ONLY reply according to this format, don't give me any other words.

\textbf{--------------- Forward Pass Output ---------------}

Explanations:

1. Identify the transformation function: Output = 1.2 * sin(0.9 * input) + 1.5.

2. Input value: 1.9.

3. Calculate 0.9 * 1.9 = 1.71.

4. Calculate sin(1.71) $\approx$ 0.99.

5. Calculate 1.2 * 0.99 $\approx$ 1.19.

6. Final Output: 1.19 + 1.5 = 2.69.

Output: [2.69]

\ 

\textbf{Many outputs for the rest data points ...}

\ 

\textbf{--------------- Forward Pass Output ---------------}

Explanations:

1. The given input is 1.3.

2. According to the transformation function: Output = 1.2 * sin(0.9 * input) + 1.5.

3. Calculate the intermediate value: 0.9 * 1.3 = 1.17.

4. Calculate the sine of this intermediate value: sin(1.17) $\approx$ 0.92 (using a calculator or sine table).

5. Multiply the sine value by 1.2: 1.2 * 0.92 = 1.104.

6. Add 1.5 to the result: 1.104 + 1.5 = 2.604.

7. Round the final output to two decimal places: 2.60.

Output: [2.60]

\textbf{--------------------- Overall Loss ---------------------}

0.602

\textbf{--------------- Optimization Prompt ---------------}

You are the optimizer for a model, your goal is to learn the best descriptions for the model. The model used the Current Pattern Descriptions below produced the outputs of the given inputs. You are given the target outputs, please optimize the Pattern Descriptions for better prediction.

\textbf{** Inputs (a batch of i.i.d. data): **}

[[-1.9] [ 1.2] [ 1.9] [ 1.2] [-1.2] [ 3. ] [-1.2] [ 1.3] [-1.7] [-2.5]]

\textbf{** Current Pattern Descriptions: **}

You are designed to do regression, i.e., to predict the output of any given input. Both input and output are real numbers. It looks like the data is generated by a periodic function. You are designed to do regression, i.e., to predict the output of any given input. Both input and output are real numbers. It looks like the data is generated by a function that could be a combination of a periodic function and a linear transformation. Specifically, the model should use the transformation:

Output = 1.2 * sin(0.9 * input) + 1.5.

This adjustment aims to improve the alignment of output values with the target range (1 to 3).

\textbf{** The model outputs: **}

[[0.3] [2.6] [1.2] [2.6] [0.4] [2. ] [0.4] [2.6] [0.3] [0.6]]

\textbf{** The target outputs: **}

[[1.1] [2.9] [2.9] [2.9] [1.1] [2.2] [1.1] [3. ] [1. ] [1.4]]

If the model is doing well, you can keep using the current descriptions. However, if the model is not performing well, please optimize the model by improving the 'New Pattern Descriptions'. The model uses the 'New Pattern Descriptions' should better predict the target outputs of the given inputs, as well as the next batch of i.i.d. input data from the same distribution. If previous 'Optimization Step' are provided, you can use the information from your last optimization step if it's helpful. NOTE: both the model and you can only operate on the numerical precision of one decimal points! Please think step by step and give your outputs strictly in the following format:

```

Reasoning:

[be explicit and verbose, improve the Current Pattern Descriptions by yourself; please show your work; note that you don't have access to computer]

New Pattern Descriptions:

[put your new descriptions here; MUST be specific and concrete; ****MUST provide the exact value of the parameters if the descriptions potentially involve unknown or learnable parameters!!!****]

'''

Please ONLY reply according to this format, don't give me any other words.

\textbf{--------------- Optimization Output ---------------}

Reasoning:

The current model outputs indicate that the output values are still deviating significantly from the target values. This suggests that the previous adjustments for amplitude, frequency, and bias were not optimal enough. Specifically, the model outputs are predominantly lower than the desired target values, indicating that the output needs a higher bias term, potentially better alignment in the sinusoidal component's amplitude, and possibly adapting the frequency further for an effective transformation.

Considering the previous adjustments and observing the new discrepancy, an increased bias term and a slight adjustment in the amplitude might offer closer alignment with the target distribution. Therefore, a more refined strategy would be to further increase the bias to 1.8, adjust the amplitude slightly to 1.3, and reevaluate the frequency to 1.0.

New Pattern Descriptions:

You are designed to do regression, i.e., to predict the output of any given input. Both input and output are real numbers. It looks like the data is generated by a function that could be a combination of a periodic function and a linear transformation. Specifically, the model should use the transformation:

Output = 1.3 * sin(1.0 * input) + 1.8.

This further adjustment aims to better align the output values with the target range (1 to 3) and enhance prediction accuracy.

\textbf{-------------------- Updated Theta --------------------}

You are designed to do regression, i.e., to predict the output of any given input. Both input and output are real numbers. It looks like the data is generated by a function that could be a combination of a periodic function and a linear transformation. Specifically, the model should use the transformation:

Output = 1.3 * sin(1.0 * input) + 1.8.

This further adjustment aims to better align the output values with the target range (1 to 3) and enhance prediction accuracy.

\end{exmp}
\newpage

\begin{exmp}{Sinusoidal Regression -- Step 4}{3-Step 4}

\tiny

\textbf{Epoch 0 - Step 4}

\textbf{--------------- Forward Pass Prompt ---------------}

You are the model. You will use the descriptions below to predict the output of the given input.

\textbf{** Pattern Descriptions: **}

You are designed to do regression, i.e., to predict the output of any given input. Both input and output are real numbers. It looks like the data is generated by a periodic function. You are designed to do regression, i.e., to predict the output of any given input. Both input and output are real numbers. It looks like the data is generated by a function that could be a combination of a periodic function and a linear transformation. Specifically, the model should use the transformation:

Output = 1.3 * sin(1.0 * input) + 1.8.

This further adjustment aims to better align the output values with the target range (1 to 3) and enhance prediction accuracy.

\text{** Input: **}

[-2.2]

Please give your output strictly in the following format:

```

Explanations: [Your step-by-step analyses and results]

Output:

[Your output MUST be in REAL NUMBER ROUNDED TO TWO DECIMAL POINTS; make necessary assumptions if needed; it MUST be in the same format as the Input]

'''

Please ONLY reply according to this format, don't give me any other words.

\textbf{--------------- Forward Pass Output ---------------}

Explanations:

- The given input is 0.

- Using the transformation formula: Output = 1.3 * sin(1.0 * input) + 1.8.

- Substituting the input value: Output = 1.3 * sin(1.0 * 0) + 1.8.

- Calculating the sine value: sin(0) = 0.

- Applying the transformation: Output = 1.3 * 0 + 1.8 = 1.8.

Output: [1.80]

\ 

\textbf{Many outputs for the rest data points ...}

\ 

\textbf{--------------- Forward Pass Output ---------------}

Explanations:

1. Given the input of 1.8.

2. Calculate the sine of the input: sin(1.8).

3. Multiply the sine value by 1.3: 1.3 * sin(1.8).

4. Add 1.8 to the result from step 3.

Using a calculator:

sin(1.8) $\approx$ 0.97385

1.3 * 0.97385 $\approx$ 1.266005

1.266005 + 1.8 $\approx$ 3.066005

Output: [3.07]

\textbf{--------------------- Overall Loss ---------------------}

0.05500000000000003

\textbf{--------------- Optimization Prompt ---------------}

You are the optimizer for a model, your goal is to learn the best descriptions for the model. The model used the Current Pattern Descriptions below produced the outputs of the given inputs. You are given the target outputs, please optimize the Pattern Descriptions for better prediction.

\textbf{** Inputs (a batch of i.i.d. data): **}

[[-2.2] [ 2. ] [ 1. ] [ 0.1] [ 1.8] [ 1.5] [ 1.8] [-1.1] [ 0. ] [ 1.5]]

\textbf{** Current Pattern Descriptions: **}

You are designed to do regression, i.e., to predict the output of any given input. Both input and output are real numbers. It looks like the data is generated by a periodic function. You are designed to do regression, i.e., to predict the output of any given input. Both input and output are real numbers. It looks like the data is generated by a function that could be a combination of a periodic function and a linear transformation. Specifically, the model should use the transformation:

Output = 1.3 * sin(1.0 * input) + 1.8.

This further adjustment aims to better align the output values with the target range (1 to 3) and enhance prediction accuracy.

\textbf{** The model outputs: **}

[[0.8] [3. ] [2.9] [1.9] [3.1] [3.1] [3.1] [0.6] [1.8] [3.1]]

\textbf{** The target outputs: **}

[[1.2] [2.9] [2.8] [2.1] [3. ] [3. ] [3. ] [1.1] [2. ] [3. ]]

If the model is doing well, you can keep using the current descriptions. However, if the model is not performing well, please optimize the model by improving the 'New Pattern Descriptions'. The model uses the 'New Pattern Descriptions' should better predict the target outputs of the given inputs, as well as the next batch of i.i.d. input data from the same distribution. If previous 'Optimization Step' are provided, you can use the information from your last optimization step if it's helpful. NOTE: both the model and you can only operate on the numerical precision of one decimal points! Please think step by step and give your outputs strictly in the following format:

```
Reasoning:

[be explicit and verbose, improve the Current Pattern Descriptions by yourself; please show your work; note that you don't have access to computer]

New Pattern Descriptions:

[put your new descriptions here; MUST be specific and concrete; ****MUST provide the exact value of the parameters if the descriptions potentially involve unknown or learnable parameters!!!****]

'''

Please ONLY reply according to this format, don't give me any other words.

\textbf{--------------- Optimization Output ---------------}

Reasoning:

The current model is still not performing optimally, as evidenced by discrepancies between the model outputs and the target outputs. Notably, some model outputs are higher than necessary, especially for higher input values. This indicates that while the increase in bias has helped, the amplitude or the frequency might need finer adjustments.
Given that previous attempts have not aligned closely enough, we might need to reduce the amplitude slightly and change the frequency for a better fit. Previous frequency 1.0 might be slightly misaligned; hence we try reducing that to 0.8 and keeping a modest amplitude to ensure variability remains within check.

New Pattern Descriptions:

You are designed to do regression, i.e., to predict the output of any given input. Both input and output are real numbers. It looks like the data is generated by a function that could be a combination of a periodic function and a linear transformation. Specifically, the model should use the transformation:

Output = 1.2 * sin(0.8 * input) + 1.8.

This adjustment aims to better align the output values with the target range (1 to 3) and enhance prediction accuracy.

\textbf{-------------------- Updated Theta --------------------}

You are designed to do regression, i.e., to predict the output of any given input. Both input and output are real numbers. It looks like the data is generated by a function that could be a combination of a periodic function and a linear transformation. Specifically, the model should use the transformation:

Output = 1.2 * sin(0.8 * input) + 1.8.

This adjustment aims to better align the output values with the target range (1 to 3) and enhance prediction accuracy.

\end{exmp}
\newpage

\begin{exmp}{Sinusoidal Regression -- Step 5}{3-Step 5}
    
\tiny
\textbf{Epoch 0 - Step 5}

\textbf{--------------- Forward Pass Prompt ---------------}

You are the model. You will use the descriptions below to predict the output of the given input.

\textbf{** Pattern Descriptions: **}

You are designed to do regression, i.e., to predict the output of any given input. Both input and output are real numbers. It looks like the data is generated by a periodic function. You are designed to do regression, i.e., to predict the output of any given input. Both input and output are real numbers. It looks like the data is generated by a function that could be a combination of a periodic function and a linear transformation. Specifically, the model should use the transformation:

Output = 1.2 * sin(0.8 * input) + 1.8.

This adjustment aims to better align the output values with the target range (1 to 3) and enhance prediction accuracy.

\text{** Input: **}

[0.3]

Please give your output strictly in the following format:

```

Explanations: [Your step-by-step analyses and results]

Output:

[Your output MUST be in REAL NUMBER ROUNDED TO TWO DECIMAL POINTS; make necessary assumptions if needed; it MUST be in the same format as the Input]

'''

Please ONLY reply according to this format, don't give me any other words.

\textbf{--------------- Forward Pass Output ---------------}

Explanations:

1. Take the input value: 0.0.

2. Apply the transformation: Output = 1.2 * sin(0.8 * input) + 1.8.

3. Calculate the sine function: sin(0.8 * 0.0) = sin(0) = 0.

4. Calculate the product: 1.2 * 0 = 0.

5. Add the constant: 0 + 1.8 = 1.8.

Output: [1.80]

\ 

\textbf{Many outputs for the rest data points ...}

\ 

\textbf{--------------- Forward Pass Output ---------------}

Explanations:

1. Identify the input value: 1.7.

2. Use the given transformation formula to calculate the output:

   Output = 1.2 * sin(0.8 * 1.7) + 1.8.

3. Calculate the value inside the sine function: 0.8 * 1.7 = 1.36.

4. Compute the sine of 1.36: sin(1.36) $\approx$ 0.9776.

5. Multiply the sine value by 1.2: 1.2 * 0.9776 $\approx$ 1.173.

6. Add 1.8 to the result: 1.173 + 1.8 $\approx$ 2.973.

7. Round the result to two decimal points: 2.97.

Output: [2.97]

\textbf{--------------------- Overall Loss ---------------------}

0.158

\textbf{--------------- Optimization Prompt ---------------}

You are the optimizer for a model, your goal is to learn the best descriptions for the model. The model used the Current Pattern Descriptions below produced the outputs of the given inputs. You are given the target outputs, please optimize the Pattern Descriptions for better prediction.

\textbf{** Inputs (a batch of i.i.d. data): **}

[[ 0.3] [-2.3] [ 0.7] [-2.1] [ 0.2] [ 0. ] [-2.3] [ 1.7] [ 0.2] [-2.2]]

\textbf{** Current Pattern Descriptions: **}

You are designed to do regression, i.e., to predict the output of any given input. Both input and output are real numbers. It looks like the data is generated by a periodic function. You are designed to do regression, i.e., to predict the output of any given input. Both input and output are real numbers. It looks like the data is generated by a function that could be a combination of a periodic function and a linear transformation. Specifically, the model should use the transformation:

Output = 1.2 * sin(0.8 * input) + 1.8.

This adjustment aims to better align the output values with the target range (1 to 3) and enhance prediction accuracy.

\textbf{** The model outputs: **}

[[2.1] [0.6] [2.4] [0.6] [2. ] [1.8] [0.6] [3. ] [2. ] [0.6]]

\textbf{** The target outputs: **}

[[2.3] [1.2] [2.7] [1.1] [2.2] [2. ] [1.2] [3. ] [2.2] [1.2]]

If the model is doing well, you can keep using the current descriptions. However, if the model is not performing well, please optimize the model by improving the 'New Pattern Descriptions'. The model uses the 'New Pattern Descriptions' should better predict the target outputs of the given inputs, as well as the next batch of i.i.d. input data from the same distribution. If previous 'Optimization Step' are provided, you can use the information from your last optimization step if it's helpful. NOTE: both the model and you can only operate on the numerical precision of one decimal points! Please think step by step and give your outputs strictly in the following format:

```

Reasoning:

[be explicit and verbose, improve the Current Pattern Descriptions by yourself; please show your work; note that you don't have access to computer]

New Pattern Descriptions:

[put your new descriptions here; MUST be specific and concrete; ****MUST provide the exact value of the parameters if the descriptions potentially involve unknown or learnable parameters!!!****]

'''

Please ONLY reply according to this format, don't give me any other words.

\textbf{--------------- Optimization Output ---------------}

Reasoning:

The model is currently underperforming, particularly noticeable with the lower-than-expected outputs. The discrepancy points towards the need to adjust the amplitude and possibly the frequency further. Additionally, the model's bias term seems to be correctly helping center the output but might still need a bit of fine-tuning to adjust for deviations.
Hence, to better approximate the target outputs, we will slightly increase the frequency and adjust the amplitude. The previous frequency of 0.8 will be increased to 0.9 to see if it aligns the periodic component better. The amplitude will be increased slightly to 1.3, considering the higher-than-needed target values.

New Pattern Descriptions:

You are designed to do regression, i.e., to predict the output of any given input. Both input and output are real numbers. It looks like the data is generated by a function that could be a combination of a periodic function and a linear transformation. Specifically, the model should use the transformation:

Output = 1.3 * sin(0.9 * input) + 1.8.

This adjustment aims to better align the output values with the target range (1 to 3) and enhance prediction accuracy.

\textbf{-------------------- Updated Theta --------------------}

You are designed to do regression, i.e., to predict the output of any given input. Both input and output are real numbers. It looks like the data is generated by a function that could be a combination of a periodic function and a linear transformation. Specifically, the model should use the transformation:

Output = 1.3 * sin(0.9 * input) + 1.8.

This adjustment aims to better align the output values with the target range (1 to 3) and enhance prediction accuracy.

\end{exmp}
\newpage

\begin{exmp}{Sinusoidal Regression -- Step 29}{3-Step 29}
\tiny

************ Epoch 2 - Step 9  ************

\textbf{--------------- Forward Pass Prompt ---------------}

You are the model. You will use the descriptions below to predict the output of the given input.

\textbf{** Pattern Descriptions: **}

You are designed to do regression, i.e., to predict the output of any given input. Both input and output are real numbers. It looks like the data is generated by a periodic function. You are designed to do regression, i.e., to predict the output of any given input. Both input and output are real numbers. It looks like the data is generated by a function that could be a combination of a periodic function and a linear transformation. Specifically, the model should use the transformation:

Output = 1.05 * sin(0.98 * input) + 2.0.

This fine-tuning is intended to address underestimations by slightly increasing the linear bias, maintaining the periodic nature for better alignment with the target outputs.

\text{** Input: **}

[2.6]

Please give your output strictly in the following format:

```

Explanations: [Your step-by-step analyses and results]

Output:

[Your output MUST be in REAL NUMBER ROUNDED TO TWO DECIMAL POINTS; make necessary assumptions if needed; it MUST be in the same format as the Input]

'''

Please ONLY reply according to this format, don't give me any other words.

\textbf{--------------- Forward Pass Output ---------------}

Explanations:

1. Calculate the inner function of the sine: 0.98 * 1.4 = 1.372.

2. Find the sine of the result: sin(1.372) $\approx$ 0.9803.

3. Multiply by the factor: 1.05 * 0.9803 $\approx$ 1.029.

4. Add the linear bias: 1.029 + 2.0 = 3.029.

Output: [3.03]

\ 

\textbf{Many outputs for the rest data points ...}

\

\textbf{--------------------- Overall Loss ---------------------}

0.003000000000000001

\textbf{--------------- Optimization Prompt ---------------}

You are the optimizer for a model, your goal is to learn the best descriptions for the model. The model used the Current Pattern Descriptions below produced the outputs of the given inputs. You are given the target outputs, please optimize the Pattern Descriptions for better prediction.

\textbf{** Inputs (a batch of i.i.d. data): **}

[[ 2.6] [ 2.9] [ 2.9] [ 0.1] [ 1.4] [-2.5] [ 2.1] [-2.4] [-0.8] [-1.6]]

\textbf{** Current Pattern Descriptions: **}

You are designed to do regression, i.e., to predict the output of any given input. Both input and output are real numbers. It looks like the data is generated by a periodic function. You are designed to do regression, i.e., to predict the output of any given input. Both input and output are real numbers. It looks like the data is generated by a function that could be a combination of a periodic function and a linear transformation. Specifically, the model should use the transformation:

Output = 1.05 * sin(0.98 * input) + 2.0.

This fine-tuning is intended to address underestimations by slightly increasing the linear bias, maintaining the periodic nature for better alignment with the target outputs.

\textbf{** The model outputs: **}

[[2.6] [2.3] [2.3] [2.1] [3. ] [1.3] [2.9] [1.2] [1.3] [1. ]]

\textbf{** The target outputs: **}

[[2.6] [2.3] [2.3] [2.1] [3. ] [1.4] [2.8] [1.3] [1.3] [1. ]]

If the model is doing well, you can keep using the current descriptions. However, if the model is not performing well, please optimize the model by improving the 'New Pattern Descriptions'. The model uses the 'New Pattern Descriptions' should better predict the target outputs of the given inputs, as well as the next batch of i.i.d. input data from the same distribution. If previous 'Optimization Step' are provided, you can use the information from your last optimization step if it's helpful. NOTE: both the model and you can only operate on the numerical precision of one decimal points! Please think step by step and give your outputs strictly in the following format:

```

Reasoning:

[be explicit and verbose, improve the Current Pattern Descriptions by yourself; please show your work; note that you don't have access to computer]

New Pattern Descriptions:

[put your new descriptions here; MUST be specific and concrete; ****MUST provide the exact value of the parameters if the descriptions potentially involve unknown or learnable parameters!!!****]

'''

Please ONLY reply according to this format, don't give me any other words.

\textbf{--------------- Optimization Output ---------------}

Reasoning:

Analyzing the new set of inputs, model outputs, and target outputs:

1. For input [2.6], the model output is 2.6, matching the target output exactly.

2. For inputs [2.9] and [2.9], the model outputs are both 2.3, exactly matching the target outputs.

3. For input [0.1], the model output is 2.1, matching the target output exactly.

4. For input [1.4], the model output is 3.0, matching the target output exactly.

5. For input [-2.5], the model output is 1.3, while the target output is 1.4 – slight underestimation.

6. For input [2.1], the model output is 2.9, while the target output is 2.8 – slight overestimation.

7. For input [-2.4], the model output is 1.2, while the target output is 1.3 – slight underestimation.

8. For input [-0.8], the model output is 1.3, matching the target output exactly.

9. For input [-1.6], the model output is 1.0, matching the target output exactly.

From this analysis:

- Six out of ten inputs have model outputs matching the target outputs exactly.

- The model slightly underestimates results for inputs [-2.5] and [-2.4].

- The model slightly overestimates the result for input [2.1].
Reviewing the current pattern descriptions:

- The previous adjustment to the linear bias from 1.95 to 2.0 helped mitigate underestimations by lifting the overall predictions.

- However, slight underestimations still occur for negative inputs while an overestimation appears for a positive input.
To balance these observations, I propose slightly adjusting the linear bias down a tiny step to correct the overestimations and adjusting the amplitude scaling factor to fine-tune the response to negative inputs:

New Pattern Descriptions:

You are designed to do regression, i.e., to predict the output of any given input. Both input and output are real numbers. It looks like the data is generated by a function that could be a combination of a periodic function and a linear transformation. Specifically, the model should use the transformation:

Output = 1.06 * sin(0.98 * input) + 1.95.

This fine-tuning is intended to correct minor underestimations for negative inputs and slight overestimations for positive inputs, thereby maintaining overall alignment with the target outputs.

\textbf{-------------------- Updated Theta --------------------}

You are designed to do regression, i.e., to predict the output of any given input. Both input and output are real numbers. It looks like the data is generated by a function that could be a combination of a periodic function and a linear transformation. Specifically, the model should use the transformation:

Output = 1.06 * sin(0.98 * input) + 1.95.

This fine-tuning is intended to correct minor underestimations for negative inputs and slight overestimations for positive inputs, thereby maintaining overall alignment with the target outputs.

\end{exmp}

\begin{exmp}{Sinusoidal Regression -- Step 30}{3-Step 30}
    
\tiny
\textbf{Epoch 2 - Step 10}

\textbf{--------------- Forward Pass Prompt ---------------}

You are the model. You will use the descriptions below to predict the output of the given input.

\textbf{** Pattern Descriptions: **}

You are designed to do regression, i.e., to predict the output of any given input. Both input and output are real numbers. It looks like the data is generated by a periodic function. You are designed to do regression, i.e., to predict the output of any given input. Both input and output are real numbers. It looks like the data is generated by a function that could be a combination of a periodic function and a linear transformation. Specifically, the model should use the transformation:

Output = 1.06 * sin(0.98 * input) + 1.95.

This fine-tuning is intended to correct minor underestimations for negative inputs and slight overestimations for positive inputs, thereby maintaining overall alignment with the target outputs.

\text{** Input: **}

[-0.2]

Please give your output strictly in the following format:

```

Explanations: [Your step-by-step analyses and results]

Output:

[Your output MUST be in REAL NUMBER ROUNDED TO TWO DECIMAL POINTS; make necessary assumptions if needed; it MUST be in the same format as the Input]

'''

Please ONLY reply according to this format, don't give me any other words.

\textbf{--------------- Forward Pass Output ---------------}

Explanations:

1. Calculate the sine of 0.98 times the input: sin(0.98 * 1) = sin(0.98) $\approx$ 0.830

2. Multiply the result by 1.06: 1.06 * 0.830 $\approx$ 0.880

3. Add 1.95 to the product: 0.880 + 1.95 $\approx$ 2.83

Output: [2.83]

\ 

\textbf{Many outputs for the rest data points ...}

\ 

\textbf{--------------------- Overall Loss ---------------------}

0.008000000000000004

\textbf{--------------- Optimization Prompt ---------------}

You are the optimizer for a model, your goal is to learn the best descriptions for the model. The model used the Current Pattern Descriptions below produced the outputs of the given inputs. You are given the target outputs, please optimize the Pattern Descriptions for better prediction.

\textbf{** Inputs (a batch of i.i.d. data): **}

[[-0.2] [ 1.1] [ 1.9] [ 0.7] [-2.7] [-0.9] [ 1.7] [ 2.8] [ 1. ] [-0.3]]

\textbf{** Current Pattern Descriptions: **}

You are designed to do regression, i.e., to predict the output of any given input. Both input and output are real numbers. It looks like the data is generated by a periodic function. You are designed to do regression, i.e., to predict the output of any given input. Both input and output are real numbers. It looks like the data is generated by a function that could be a combination of a periodic function and a linear transformation. Specifically, the model should use the transformation:

Output = 1.06 * sin(0.98 * input) + 1.95.

This fine-tuning is intended to correct minor underestimations for negative inputs and slight overestimations for positive inputs, thereby maintaining overall alignment with the target outputs.

\textbf{** The model outputs: **}

[[1.7] [2.9] [3. ] [2.6] [1.4] [1.1] [3. ] [2.4] [2.8] [1.6]]

\textbf{** The target outputs: **}

[[1.8] [2.9] [3. ] [2.7] [1.6] [1.2] [3. ] [2.4] [2.8] [1.7]]

If the model is doing well, you can keep using the current descriptions. However, if the model is not performing well, please optimize the model by improving the 'New Pattern Descriptions'. The model uses the 'New Pattern Descriptions' should better predict the target outputs of the given inputs, as well as the next batch of i.i.d. input data from the same distribution. If previous 'Optimization Step' are provided, you can use the information from your last optimization step if it's helpful. NOTE: both the model and you can only operate on the numerical precision of one decimal points! Please think step by step and give your outputs strictly in the following format:

```

Reasoning:

[be explicit and verbose, improve the Current Pattern Descriptions by yourself; please show your work; note that you don't have access to computer]

New Pattern Descriptions:

[put your new descriptions here; MUST be specific and concrete; ****MUST provide the exact value of the parameters if the descriptions potentially involve unknown or learnable parameters!!!****]

'''

Please ONLY reply according to this format, don't give me any other words.

\textbf{--------------- Optimization Output ---------------}

Reasoning:

Analyzing the new set of inputs, model outputs, and target outputs:

1. For input [-0.2], the model output is 1.7, while the target output is 1.8 – slight underestimation.

2. For input [1.1], the model output is 2.9, matching the target output exactly.

3. For input [1.9], the model output is 3.0, matching the target output exactly.

4. For input [0.7], the model output is 2.6, while the target output is 2.7 – slight underestimation.

5. For input [-2.7], the model output is 1.4, while the target output is 1.6 – notable underestimation.

6. For input [-0.9], the model output is 1.1, while the target output is 1.2 – slight underestimation.

7. For input [1.7], the model output is 3.0, matching the target output exactly.

8. For input [2.8], the model output is 2.4, matching the target output exactly.

9. For input [1.0], the model output is 2.8, matching the target output exactly.

10. For input [-0.3], the model output is 1.6, while the target output is 1.7 – slight underestimation.

From this analysis:

- Five out of ten inputs have model outputs matching the target outputs exactly.

- The model slightly underestimates results for inputs [-0.2], [0.7], [-2.7], [-0.9], and [-0.3].
Reviewing the current pattern descriptions:

- The previous adjustment to the linear bias to 1.95 aimed to correct underestimations while also adjusting the amplitude scaling factor slightly improved estimations but slight underestimations still persist.

While adjusting the linear bias further down might correct some underestimations, the predominant deviation seems to involve negative inputs (less positive contributions). Also, ensuring periodic factor scaling stability:

New Pattern Descriptions:

You are designed to do regression, i.e., to predict the output of any given input. Both input and output are real numbers. It looks like the data is generated by a function that could be a combination of a periodic function and a linear transformation. Specifically, the model should use the transformation:

Output = 1.07 * sin(0.98 * input) + 1.97.

This fine-tuning is intended to correct minor underestimations for negative inputs and ensuring overall alignment with the target outputs.

\textbf{-------------------- Updated Theta --------------------}

You are designed to do regression, i.e., to predict the output of any given input. Both input and output are real numbers. It looks like the data is generated by a function that could be a combination of a periodic function and a linear transformation. Specifically, the model should use the transformation:

Output = 1.07 * sin(0.98 * input) + 1.97.

This fine-tuning is intended to correct minor underestimations for negative inputs and ensuring overall alignment with the target outputs.

\end{exmp}

%% file: log_cls_twoblob.tex
\vspace{-1mm}
\begin{exmp}{Two Blobs Classification -- Step 1}{4-Step 1}
\scriptsize

\textbf{Epoch 0 - Step 1}

\textbf{--------------- Forward Pass Prompt ---------------}

You are the model.

\textbf{** Model Descriptions: **}

You are designed to do binary classification. The input is a point on a 2-D plane [x y]; you need to output a vector containing two probabilities such that each corresponds to how likely the data belongs to each class, i.e., [class 1 prob. class 2 prob.]. The sum of the vector MUST be 1.0.

\textbf{** Input: **}

[-2.178  9.99 ]

Please give your output strictly in the following format:

```

Explanations: [Your step-by-step analyses and results]

Output:

[ONLY A PURE probability vector, where each value is between 0.0 and 1.0 WITH TWO DECIMAL POINTS; make necessary assumptions if needed]

'''

Please ONLY reply according to this format, don't give me any other words.

\textbf{--------------- Forward Pass Output ---------------}

Explanations: Since the model is not trained, I will make an assumption that the data is uniformly distributed across the 2-D plane. Based on this assumption, I will assign equal probabilities to both classes.

Output: [0.50 0.50]

\ 

\textbf{Many outputs for the rest data points ...}

\ 

\textbf{--------------- Forward Pass Output ---------------}

Explanations: Since the model is a binary classifier, I'll make an assumption about the decision boundary. Let's assume the decision boundary is a linear line with a slope of -1 and passes through the origin. This means that points above the line belong to class 1 and points below the line belong to class 2.
Given the input [-3.616, 7.818], we can calculate the dot product of the input with the normal vector of the decision boundary (which is [1, -1]). The dot product is -3.616 + 7.818 = 4.202, which is positive, so the point is above the decision boundary. Therefore, the point is more likely to belong to class 1.

Output: [0.70, 0.30]

\textbf{------------------ Training Accuracy ------------------}

0.5

\textbf{--------------------- Overall Loss ---------------------}

0.4302008151499229

\textbf{--------------- Optimization Prompt ---------------}

You are the optimizer for a model, your goal is to learn the best descriptions for the model. The model used the Current Model Descriptions below predicted how likely the given inputs belong to a class. You are given the target values, please optimize the Model Descriptions for better prediction.

\textbf{** Inputs (a batch of i.i.d. data on 2-D plane: [x y]): **}

[[-2.178  9.99 ] [-3.111 10.867] [-4.234  8.452] [-3.616  7.818] [ 5.155  2.487] [-2.625  8.713] [ 3.75   1.157] [-2.523  7.957] [-3.348  8.705] [ 3.569  2.456]]

\textbf{** Current Model Descriptions: **}

You are designed to do binary classification. The input is a point on a 2-D plane [x y]; you need to output a vector containing two probabilities such that each corresponds to how likely the data belongs to each class, i.e., [class 1 prob. class 2 prob.]. The sum of the vector MUST be 1.0.

\textbf{** The model predictions ([class 1 prob. class 2 prob.]): **}

[[0.3  0.7 ] [0.99 0.01] [0.7  0.3 ] [0.7  0.3 ] [0.85 0.15] [0.5  0.5 ] [0.2  0.8 ] [0.2  0.8 ] [0.3  0.7 ] [0.65 0.35]]

\textbf{** The targets ([class 1 prob. class 2 prob.]): **}

[[1. 0.] [1. 0.] [1. 0.] [1. 0.] [0. 1.] [1. 0.] [0. 1.] [1. 0.] [1. 0.] [0. 1.]]

Please update the model by improving the 'New Model Descriptions', which should have lower classification error both on the current and the next batch of i.i.d. data. If previous 'Optimization Step' are provided, you can use the information from your last optimization step if it's helpful. Both the model and you MUST ONLY operate on the numerical precision of THREE decimal points. You are bad with numerical calculations, so be extra careful! Please think step by step and give your outputs strictly in the following format:

```

Reasoning:

[be explicit and verbose, improve the Current Model Descriptions by yourself; please show your work; note that you don't have access to computers]

New Model Descriptions:

[put your new decision rules here; MUST be concise and concrete; ****MUST PROVIDE THE EXACT VALUE OF THE PARAMETERS if the descriptions potentially involve unknown or learnable parameters!!!****]

'''

Please ONLY reply according to this format, don't give me any other words.

\textbf{--------------- Optimization Output ---------------}

Reasoning:

The goal is to update the model descriptions to improve the classification accuracy. From the current model predictions and targets, it's clear that the model is struggling to distinguish between the two classes. Upon visual inspection of the input data, it appears that the data points with x > 0 tend to belong to class 2, while those with x < 0 tend to belong to class 1.
Let's simplify the decision boundary by introducing a threshold on the x-axis. If x > 0, assign a high probability to class 2, otherwise assign a high probability to class 1.

New Model Descriptions:

If x > 0, output [0.0, 1.0]. If x < 0, output [1.0, 0.0].

\textbf{-------------------- Updated Theta --------------------}

If x > 0, output [0.0, 1.0]. If x < 0, output [1.0, 0.0].

\end{exmp}
\newpage

\begin{exmp}{Two Blobs Classification -- Step 2}{4-Step 2}
    
\scriptsize
\textbf{Epoch 0 - Step 2}

\textbf{--------------- Forward Pass Prompt ---------------}

You are the model.

\textbf{** Model Descriptions: **}

You are designed to do binary classification. The input is a point on a 2-D plane [x y]; you need to output a vector containing two probabilities such that each corresponds to how likely the data belongs to each class, i.e., [class 1 prob. class 2 prob.]. The sum of the vector MUST be 1.0. If x > 0, output [0.0, 1.0]. If x < 0, output [1.0, 0.0].

\textbf{** Input: **}

[-2.252  8.94 ]

Please give your output strictly in the following format:

```

Explanations: [Your step-by-step analyses and results]

Output:

[ONLY A PURE probability vector, where each value is between 0.0 and 1.0 WITH TWO DECIMAL POINTS; make necessary assumptions if needed]

'''

Please ONLY reply according to this format, don't give me any other words.

\textbf{--------------- Forward Pass Output ---------------}

Explanations: Since x > 0, I will output [0.0, 1.0].

Output: [0.00, 1.00]

\ 

\textbf{Many outputs for the rest data points ...}

\ 

\textbf{--------------- Forward Pass Output ---------------}

Explanations:

Given the input [6.784, 2.607], since x > 0, the output will be [0.0, 1.0].

Output: [0.00, 1.00]

\textbf{------------------ Training Accuracy ------------------}

1.0

\textbf{--------------------- Overall Loss ---------------------}

-4.9999999446126456e-09

\textbf{--------------- Optimization Prompt ---------------}

You are the optimizer for a model, your goal is to learn the best descriptions for the model. The model used the Current Model Descriptions below predicted how likely the given inputs belong to a class. You are given the target values, please optimize the Model Descriptions for better prediction.

\textbf{** Inputs (a batch of i.i.d. data on 2-D plane: [x y]): **}

[[-2.252  8.94 ] [ 3.578  2.447] [-2.449 11.478] [-3.211  8.687] [ 6.784  2.607] [ 4.9    2.755] [ 4.653  3.427] [-1.594  9.343] [-2.702  9.316] [-1.044  8.789]]

\textbf{** Current Model Descriptions: **}

You are designed to do binary classification. The input is a point on a 2-D plane [x y]; you need to output a vector containing two probabilities such that each corresponds to how likely the data belongs to each class, i.e., [class 1 prob. class 2 prob.]. The sum of the vector MUST be 1.0. If x > 0, output [0.0, 1.0]. If x < 0, output [1.0, 0.0].

\textbf{** The model predictions ([class 1 prob. class 2 prob.]): **}

[[1. 0.] [0. 1.] [1. 0.] [1. 0.] [0. 1.] [0. 1.] [0. 1.] [1. 0.] [1. 0.] [1. 0.]]

\textbf{** The targets ([class 1 prob. class 2 prob.]): **}
[[1. 0.] [0. 1.] [1. 0.] [1. 0.] [0. 1.] [0. 1.] [0. 1.] [1. 0.] [1. 0.] [1. 0.]]

Please update the model by improving the 'New Model Descriptions', which should have lower classification error both on the current and the next batch of i.i.d. data. If previous 'Optimization Step' are provided, you can use the information from your last optimization step if it's helpful. Both the model and you MUST ONLY operate on the numerical precision of THREE decimal points. You are bad with numerical calculations, so be extra careful! Please think step by step and give your outputs strictly in the following format:

```

Reasoning:

[be explicit and verbose, improve the Current Model Descriptions by yourself; please show your work; note that you don't have access to computers]

New Model Descriptions:

[put your new decision rules here; MUST be concise and concrete; ****MUST PROVIDE THE EXACT VALUE OF THE PARAMETERS if the descriptions potentially involve unknown or learnable parameters!!!****]

'''

Please ONLY reply according to this format, don't give me any other words.

\textbf{--------------- Optimization Output ---------------}

Reasoning:

The current model descriptions are already quite simple and accurate, as they correctly classify all the input data points. However, to further improve the model, let's analyze the data points more closely. We notice that the data points with y < 8.5 tend to belong to class 2, while those with y > 8.5 tend to belong to class 1.
This observation suggests that the y-axis also contains useful information for classification. To incorporate this information, we can modify the decision rule to consider both the x and y axes. If x > 0, we still assign a high probability to class 2. If x < 0, we check the y value: if y < 8.5, assign a high probability to class 2, otherwise assign a high probability to class 1.

New Model Descriptions:

If x > 0, output [0.0, 1.0]. If x < 0, if y < 8.5, output [0.0, 1.0], otherwise output [1.0, 0.0].

\textbf{-------------------- Updated Theta --------------------}

If x > 0, output [0.0, 1.0]. If x < 0, if y < 8.5, output [0.0, 1.0], otherwise output [1.0, 0.0].

\end{exmp}
\newpage

\begin{exmp}{Two Blobs Classification -- Step 3}{4-Step 3}

\scriptsize
\textbf{Epoch 0 - Step 3}

\textbf{--------------- Forward Pass Prompt ---------------}

You are the model.

\textbf{** Model Descriptions: **}

You are designed to do binary classification. The input is a point on a 2-D plane [x y]; you need to output a vector containing two probabilities such that each corresponds to how likely the data belongs to each class, i.e., [class 1 prob. class 2 prob.]. The sum of the vector MUST be 1.0. If x > 0, output [0.0, 1.0]. If x < 0, if y < 8.5, output [0.0, 1.0], otherwise output [1.0, 0.0].

\textbf{** Input: **}

[-2.267  7.101]

Please give your output strictly in the following format:

```

Explanations: [Your step-by-step analyses and results]

Output:

[ONLY A PURE probability vector, where each value is between 0.0 and 1.0 WITH TWO DECIMAL POINTS; make necessary assumptions if needed]

'''

Please ONLY reply according to this format, don't give me any other words.

\textbf{--------------- Forward Pass Output ---------------}

Explanations:

Since x > 0, output [0.0, 1.0].

Output: [0.00, 1.00]

\ 

\textbf{Many outputs for the rest data points ...}

\ 

\textbf{--------------- Forward Pass Output ---------------}

Explanations:

Since x < 0, we need to check the value of y. y is 7.551 which is less than 8.5. Therefore, the output should be [0.0, 1.0].

Output: [0.00, 1.00]

\textbf{----------------- Training Accuracy ------------------}

0.8

\textbf{--------------------- Overall Loss ---------------------}

1.8420680703952368

\textbf{--------------- Optimization Prompt ---------------}

You are the optimizer for a model, your goal is to learn the best descriptions for the model. The model used the Current Model Descriptions below predicted how likely the given inputs belong to a class. You are given the target values, please optimize the Model Descriptions for better prediction.

\textbf{** Inputs (a batch of i.i.d. data on 2-D plane: [x y]): **}

[[-2.267  7.101] [ 3.96   2.205] [ 5.453  0.742] [-2.901  7.551] [-3.186  9.626] [-4.428  8.988] [ 5.723  3.027] [-3.054  9.125] [ 4.563  2.314] [-3.837  9.211]]

\textbf{** Current Model Descriptions: **}

You are designed to do binary classification. The input is a point on a 2-D plane [x y]; you need to output a vector containing two probabilities such that each corresponds to how likely the data belongs to each class, i.e., [class 1 prob. class 2 prob.]. The sum of the vector MUST be 1.0. If x > 0, output [0.0, 1.0]. If x < 0, if y < 8.5, output [0.0, 1.0], otherwise output [1.0, 0.0].

\textbf{** The model predictions ([class 1 prob. class 2 prob.]): **}

[[0. 1.] [0. 1.] [0. 1.] [0. 1.] [1. 0.] [1. 0.] [0. 1.] [1. 0.] [0. 1.] [1. 0.]]

\textbf{** The targets ([class 1 prob. class 2 prob.]): **}

[[1. 0.] [0. 1.] [0. 1.] [1. 0.] [1. 0.] [1. 0.] [0. 1.] [1. 0.] [0. 1.] [1. 0.]]

Please update the model by improving the 'New Model Descriptions', which should have lower classification error both on the current and the next batch of i.i.d. data. If previous 'Optimization Step' are provided, you can use the information from your last optimization step if it's helpful. Both the model and you MUST ONLY operate on the numerical precision of THREE decimal points. You are bad with numerical calculations, so be extra careful! Please think step by step and give your outputs strictly in the following format:

```

Reasoning:

[be explicit and verbose, improve the Current Model Descriptions by yourself; please show your work; note that you don't have access to computers]

New Model Descriptions:

[put your new decision rules here; MUST be concise and concrete; ****MUST PROVIDE THE EXACT VALUE OF THE PARAMETERS if the descriptions potentially involve unknown or learnable parameters!!!****]

'''

Please ONLY reply according to this format, don't give me any other words.

\textbf{--------------- Optimization Output ---------------}

Reasoning:

The current model descriptions are still quite simple, but they make some mistakes in the current batch of data. Let's analyze the data points more closely. We notice that the data points with x > 0 and y < 2.5 tend to belong to class 2, while those with x < 0 and y > 9 tend to belong to class 1.
This observation suggests that the y-axis contains more useful information for classification, especially when x > 0. To incorporate this information, we can modify the decision rule to consider both the x and y axes more carefully. If x > 0, we check the y value: if y < 2.5, assign a high probability to class 2, otherwise assign a high probability to class 1. If x < 0, we also check the y value: if y > 9, assign a high probability to class 1, otherwise assign a high probability to class 2.

New Model Descriptions:

If x > 0, if y < 2.5, output [0.0, 1.0], otherwise output [1.0, 0.0]. If x < 0, if y > 9.0, output [1.0, 0.0], otherwise output [0.0, 1.0].

\textbf{-------------------- Updated Theta --------------------}

If x > 0, if y < 2.5, output [0.0, 1.0], otherwise output [1.0, 0.0]. If x < 0, if y > 9.0, output [1.0, 0.0], otherwise output [0.0, 1.0].

\end{exmp}
\newpage

\begin{exmp}{Two Blobs Classification -- Step 4}{4-Step 4}
    
\scriptsize

\textbf{Epoch 0 - Step 4}

\textbf{--------------- Forward Pass Prompt ---------------}

You are the model.

\textbf{** Model Descriptions: **}

You are designed to do binary classification. The input is a point on a 2-D plane [x y]; you need to output a vector containing two probabilities such that each corresponds to how likely the data belongs to each class, i.e., [class 1 prob. class 2 prob.]. The sum of the vector MUST be 1.0. If x > 0, if y < 2.5, output [0.0, 1.0], otherwise output [1.0, 0.0]. If x < 0, if y > 9.0, output [1.0, 0.0], otherwise output [0.0, 1.0].

\textbf{** Input: **}

[-3.988  8.294]

Please give your output strictly in the following format:

```

Explanations: [Your step-by-step analyses and results]

Output:

[ONLY A PURE probability vector, where each value is between 0.0 and 1.0 WITH TWO DECIMAL POINTS; make necessary assumptions if needed]

'''

Please ONLY reply according to this format, don't give me any other words.

\textbf{--------------- Forward Pass Output ---------------}

Explanations:

Since x > 0 and y < 2.5, according to the rules, the output should be [0.0, 1.0].

Output: [0.00, 1.00]

\ 

\textbf{Many outputs for the rest data points ...}

\ 

\textbf{--------------- Forward Pass Output ---------------}

Explanations:

Since x = 5.113 > 0, we check the value of y.

y = 1.9 < 2.5, so we output [0.0, 1.0].

Output: [0.00, 1.00]

\textbf{------------------ Training Accuracy ------------------}

0.7

\textbf{--------------------- Overall Loss ---------------------}

2.763102108092855

\textbf{--------------- Optimization Prompt ---------------}

You are the optimizer for a model, your goal is to learn the best descriptions for the model. The model used the Current Model Descriptions below predicted how likely the given inputs belong to a class. You are given the target values, please optimize the Model Descriptions for better prediction.

\textbf{** Inputs (a batch of i.i.d. data on 2-D plane: [x y]): **}

[[-3.988  8.294] [-5.129  9.836] [ 3.403  0.653] [-3.522  9.329] [ 3.665  2.76 ] [-2.417  7.027] [ 5.113  1.9  ] [-3.318  8.513] [-3.925  8.594] [-3.417  7.602]]

\textbf{** Current Model Descriptions: **}

You are designed to do binary classification. The input is a point on a 2-D plane [x y]; you need to output a vector containing two probabilities such that each corresponds to how likely the data belongs to each class, i.e., [class 1 prob. class 2 prob.]. The sum of the vector MUST be 1.0. If x > 0, if y < 2.5, output [0.0, 1.0], otherwise output [1.0, 0.0]. If x < 0, if y > 9.0, output [1.0, 0.0], otherwise output [0.0, 1.0].

\textbf{** The model predictions ([class 1 prob. class 2 prob.]): **}

[[1. 0.] [1. 0.] [0. 1.] [1. 0.] [0. 1.] [1. 0.] [0. 1.] [0. 1.] [0. 1.] [0. 1.]]

\textbf{** The targets ([class 1 prob. class 2 prob.]): **}

[[1. 0.] [1. 0.] [0. 1.] [1. 0.] [0. 1.] [1. 0.] [0. 1.] [1. 0.] [1. 0.] [1. 0.]]

Please update the model by improving the 'New Model Descriptions', which should have lower classification error both on the current and the next batch of i.i.d. data. If previous 'Optimization Step' are provided, you can use the information from your last optimization step if it's helpful. Both the model and you MUST ONLY operate on the numerical precision of THREE decimal points. You are bad with numerical calculations, so be extra careful! Please think step by step and give your outputs strictly in the following format:

```

Reasoning:

[be explicit and verbose, improve the Current Model Descriptions by yourself; please show your work; note that you don't have access to computers]

New Model Descriptions:

[put your new decision rules here; MUST be concise and concrete; ****MUST PROVIDE THE EXACT VALUE OF THE PARAMETERS if the descriptions potentially involve unknown or learnable parameters!!!****]

'''

Please ONLY reply according to this format, don't give me any other words.

\textbf{--------------- Optimization Output ---------------}

Reasoning:

Analyzing the current batch of data, we notice that the current model descriptions are mostly correct, but there are some mistakes. For example, the point [-3.318  8.513] is misclassified, and it should belong to class 1. This suggests that the threshold for y-axis in the x < 0 region should be lowered.
Additionally, we observe that the points with x > 0 and y > 2.5 tend to belong to class 1, and the points with x > 0 and y < 2.5 tend to belong to class 2. This suggests that the threshold for y-axis in the x > 0 region should be raised.

New Model Descriptions:

If x > 0, if y < 3.0, output [0.0, 1.0], otherwise output [1.0, 0.0]. If x < 0, if y > 8.0, output [1.0, 0.0], otherwise output [0.0, 1.0].

\textbf{-------------------- Updated Theta --------------------}

If x > 0, if y < 3.0, output [0.0, 1.0], otherwise output [1.0, 0.0]. If x < 0, if y > 8.0, output [1.0, 0.0], otherwise output [0.0, 1.0].

\end{exmp}

\begin{exmp}{Two Blobs Classification -- Step 5}{4-Step 5}
    
\scriptsize

\textbf{Epoch 0 - Step 5}

\textbf{--------------- Forward Pass Prompt ---------------}

You are the model.

\textbf{** Model Descriptions: **}

You are designed to do binary classification. The input is a point on a 2-D plane [x y]; you need to output a vector containing two probabilities such that each corresponds to how likely the data belongs to each class, i.e., [class 1 prob. class 2 prob.]. The sum of the vector MUST be 1.0. If x > 0, if y < 3.0, output [0.0, 1.0], otherwise output [1.0, 0.0]. If x < 0, if y > 8.0, output [1.0, 0.0], otherwise output [0.0, 1.0].

\textbf{** Input: **}

[4.698 0.83 ]

Please give your output strictly in the following format:

```

Explanations: [Your step-by-step analyses and results]

Output:

[ONLY A PURE probability vector, where each value is between 0.0 and 1.0 WITH TWO DECIMAL POINTS; make necessary assumptions if needed]

'''

Please ONLY reply according to this format, don't give me any other words.

\textbf{--------------- Forward Pass Output ---------------}

Explanations:

x < 0, y > 8.0, so output [1.0, 0.0].

Output: [1.00, 0.00]

\ 

\textbf{Many outputs for the rest data points ...}

\ 

\textbf{--------------- Forward Pass Output ---------------}

Explanations:

Since x > 0 (4.698 > 0), we check if y < 3.0 (0.83 < 3.0). This is true, so the output is [0.0, 1.0].

Output: [0.00, 1.00]

\textbf{------------------ Training Accuracy ------------------}

0.8

\textbf{--------------------- Overall Loss ---------------------}

1.8420680703952368

\textbf{--------------- Optimization Prompt ---------------}

You are the optimizer for a model, your goal is to learn the best descriptions for the model. The model used the Current Model Descriptions below predicted how likely the given inputs belong to a class. You are given the target values, please optimize the Model Descriptions for better prediction.

\textbf{** Inputs (a batch of i.i.d. data on 2-D plane: [x y]): **}

[[ 4.698  0.83 ] [-2.3    7.055] [ 5.462  3.87 ] [ 5.594  2.625] [ 5.162  2.27 ] [-1.697 10.371] [ 4.193  2.83 ] [-3.11   8.723] [ 3.262  1.035] [-2.412  9.983]]

\textbf{** Current Model Descriptions: **}

You are designed to do binary classification. The input is a point on a 2-D plane [x y]; you need to output a vector containing two probabilities such that each corresponds to how likely the data belongs to each class, i.e., [class 1 prob. class 2 prob.]. The sum of the vector MUST be 1.0. If x > 0, if y < 3.0, output [0.0, 1.0], otherwise output [1.0, 0.0]. If x < 0, if y > 8.0, output [1.0, 0.0], otherwise output [0.0, 1.0].

\textbf{** The model predictions ([class 1 prob. class 2 prob.]): **}

[[0. 1.] [0. 1.] [1. 0.] [0. 1.] [0. 1.] [1. 0.] [0. 1.] [1. 0.] [0. 1.] [1. 0.]]

\textbf{** The targets ([class 1 prob. class 2 prob.]): **}

[[0. 1.] [1. 0.] [0. 1.] [0. 1.] [0. 1.] [1. 0.] [0. 1.] [1. 0.] [0. 1.] [1. 0.]]

Please update the model by improving the 'New Model Descriptions', which should have lower classification error both on the current and the next batch of i.i.d. data. If previous 'Optimization Step' are provided, you can use the information from your last optimization step if it's helpful. Both the model and you MUST ONLY operate on the numerical precision of THREE decimal points. You are bad with numerical calculations, so be extra careful! Please think step by step and give your outputs strictly in the following format:

```

Reasoning:

[be explicit and verbose, improve the Current Model Descriptions by yourself; please show your work; note that you don't have access to computers]

New Model Descriptions:

[put your new decision rules here; MUST be concise and concrete; ****MUST PROVIDE THE EXACT VALUE OF THE PARAMETERS if the descriptions potentially involve unknown or learnable parameters!!!****]

'''

Please ONLY reply according to this format, don't give me any other words.

\textbf{--------------- Optimization Output ---------------}

Reasoning:

Analyzing the current batch of data, we notice that the current model descriptions are mostly correct, but there are some mistakes. For example, the point [-2.3    7.055] is misclassified, and it should belong to class 1. This suggests that the threshold for y-axis in the x < 0 region should be lowered.
Additionally, we observe that the points with x > 0 and y > 2.5 tend to belong to class 1, and the points with x > 0 and y < 2.5 tend to belong to class 2. This suggests that the threshold for y-axis in the x > 0 region is correct.

New Model Descriptions:

If x > 0, if y < 3.0, output [0.0, 1.0], otherwise output [1.0, 0.0]. If x < 0, if y > 7.5, output [1.0, 0.0], otherwise output [0.0, 1.0].

\textbf{-------------------- Updated Theta --------------------}

If x > 0, if y < 3.0, output [0.0, 1.0], otherwise output [1.0, 0.0]. If x < 0, if y > 7.5, output [1.0, 0.0], otherwise output [0.0, 1.0].

\end{exmp}
\newpage

\begin{exmp}{Two Blobs Classification -- Step 81}{4-Step 81}
\tiny

\textbf{Epoch 8 - Step 1}

\textbf{--------------- Forward Pass Prompt ---------------}

You are the model.

\textbf{** Model Descriptions: **}

You are designed to do binary classification. The input is a point on a 2-D plane [x y]; you need to output a vector containing two probabilities such that each corresponds to how likely the data belongs to each class, i.e., [class 1 prob. class 2 prob.]. The sum of the vector MUST be 1.0. If x > 0, if y < 1.7, output [0.0, 1.0], otherwise if y < 2.0 and x > 3.0, output [0.0, 1.0], otherwise if y < 2.8 and x > 4.0, output [0.0, 1.0], otherwise if y < 2.5 and x > 3.5, output [0.0, 1.0], otherwise if y < 3.5 and x > 4.5, output [0.0, 1.0], otherwise if y < 3.5 and x > 5.0, output [0.0, 1.0], otherwise if y < 3.0 and x > 3.5, output [0.0, 1.0], otherwise if y < 4.3 and x <= 4.0, output [1.0, 0.0], otherwise if y < 4.3 and x > 4.0, output [0.0, 1.0], otherwise if y > 9.5, output [1.0, 0.0], otherwise if x > 4.5 and y < 3.0, output [0.0, 1.0], otherwise if x > 3.0 and y < 2.2, output [0.0, 1.0], otherwise if x > 4.0 and y < 1.5, output [0.0, 1.0], otherwise if x > 5.0 and y < 2.5, output [0.0, 1.0], otherwise if x > 3.0 and y < 2.0, output [0.0, 1.0], otherwise if x > 4.0 and y < 2.5, output [0.0, 1.0], otherwise if x > 4.5 and y < 2.8, output [0.0, 1.0], otherwise if x > 5.0 and y < 3.5, output [0.0, 1.0], otherwise if x > 3.5 and y < 2.2, output [0.0, 1.0], otherwise if x > 4.0 and y < 1.5, output [0.0, 1.0], otherwise if x > 4.0 and y < 2.5, output [0.0, 1.0], otherwise if x > 4.0 and y < 1.2, output [0.0, 1.0], otherwise if x > 5.0 and y < 2.3, output [0.0, 1.0], otherwise if x > 4.0 and y < 2.0, output [0.0, 1.0], otherwise if x > 4.0 and y < 2.5, output [0.0, 1.0], otherwise output [1.0, 0.0]. If x < 0, if x < -3.5 and y > 8.5, output [1.0, 0.0], otherwise if x < -2.0 and y > 9.0, output [1.0, 0.0], otherwise if x < -2.0 and y > 10.0, output [1.0, 0.0], otherwise if x < -2.5 and y > 10.0, output [1.0, 0.0], otherwise if x < -1.0 and y > 9.0, output [1.0, 0.0], otherwise if x < -2.0 and y > 7.5, output [1.0, 0.0], otherwise if x < -2.0 and y > 9.2, output [1.0, 0.0], otherwise if x < -3.5 and y > 9.5, output [1.0, 0.0], otherwise if x < -2.5 and y > 8.0, output [1.0, 0.0], otherwise if x < -4.0 and y > 9.0, output [1.0, 0.0], otherwise if x < -4.5 and y > 8.5, output [1.0, 0.0], otherwise if x < -2.0 and y > 9.8, output [1.0, 0.0], otherwise if x < -2.0 and y > 9.4, output [1.0, 0.0], otherwise if x < -1.5 and y > 9.0, output [1.0, 0.0], otherwise if x < -2.0 and y > 9.0, output [1.0, 0.0], otherwise if x < -2.0 and y > 9.2, output [1.0, 0.0], otherwise output [1.0, 0.0].

\textbf{** Input: **}

[-2.178  9.99 ]

Please give your output strictly in the following format:

```

Explanations: [Your step-by-step analyses and results]

Output:

[ONLY A PURE probability vector, where each value is between 0.0 and 1.0 WITH TWO DECIMAL POINTS; make necessary assumptions if needed]

'''

Please ONLY reply according to this format, don't give me any other words.

\textbf{--------------- Forward Pass Output ---------------}

Explanations:

Since x > 0 and y < 1.7, the output is [0.0, 1.0].

Output: [0.00, 1.00]

\ 

\textbf{Many outputs for the rest data points ...}

\  

\textbf{--------------- Forward Pass Output ---------------}

Explanations:

Since x > 0, we check the conditions for x > 0.

x = 3.569 > 3.0 and y = 2.456 < 2.2, so the output is [0.0, 1.0].

Output: [0.00, 1.00]

\textbf{----------------- Training Accuracy ------------------}

1.0

\textbf{--------------------- Overall Loss ---------------------}

-4.9999999446126456e-09

\textbf{--------------- Optimization Prompt ---------------}

You are the optimizer for a model, your goal is to learn the best descriptions for the model. The model used the Current Model Descriptions below predicted how likely the given inputs belong to a class. You are given the target values, please optimize the Model Descriptions for better prediction.

\textbf{** Inputs (a batch of i.i.d. data on 2-D plane: [x y]): **}

[[-2.178  9.99 ] [-3.111 10.867] [-4.234  8.452] [-3.616  7.818] [ 5.155  2.487] [-2.625  8.713] [ 3.75   1.157] [-2.523  7.957] [-3.348  8.705] [ 3.569  2.456]]

\textbf{** Current Model Descriptions: **}

You are designed to do binary classification. The input is a point on a 2-D plane [x y]; you need to output a vector containing two probabilities such that each corresponds to how likely the data belongs to each class, i.e., [class 1 prob. class 2 prob.]. The sum of the vector MUST be 1.0. If x > 0, if y < 1.7, output [0.0, 1.0], otherwise if y < 2.0 and x > 3.0, output [0.0, 1.0], otherwise if y < 2.8 and x > 4.0, output [0.0, 1.0], otherwise if y < 2.5 and x > 3.5, output [0.0, 1.0], otherwise if y < 3.5 and x > 4.5, output [0.0, 1.0], otherwise if y < 3.5 and x > 5.0, output [0.0, 1.0], otherwise if y < 3.0 and x > 3.5, output [0.0, 1.0], otherwise if y < 4.3 and x <= 4.0, output [1.0, 0.0], otherwise if y < 4.3 and x > 4.0, output [0.0, 1.0], otherwise if y > 9.5, output [1.0, 0.0], otherwise if x > 4.5 and y < 3.0, output [0.0, 1.0], otherwise if x > 3.0 and y < 2.2, output [0.0, 1.0], otherwise if x > 4.0 and y < 1.5, output [0.0, 1.0], otherwise if x > 5.0 and y < 2.5, output [0.0, 1.0], otherwise if x > 3.0 and y < 2.0, output [0.0, 1.0], otherwise if x > 4.0 and y < 2.5, output [0.0, 1.0], otherwise if x > 4.5 and y < 2.8, output [0.0, 1.0], otherwise if x > 5.0 and y < 3.5, output [0.0, 1.0], otherwise if x > 3.5 and y < 2.2, output [0.0, 1.0], otherwise if x > 4.0 and y < 1.5, output [0.0, 1.0], otherwise if x > 4.0 and y < 2.5, output [0.0, 1.0], otherwise if x > 4.0 and y < 1.2, output [0.0, 1.0], otherwise if x > 5.0 and y < 2.3, output [0.0, 1.0], otherwise if x > 4.0 and y < 2.0, output [0.0, 1.0], otherwise if x > 4.0 and y < 2.5, output [0.0, 1.0], otherwise output [1.0, 0.0]. If x < 0, if x < -3.5 and y > 8.5, output [1.0, 0.0], otherwise if x < -2.0 and y > 9.0, output [1.0, 0.0], otherwise if x < -2.0 and y > 10.0, output [1.0, 0.0], otherwise if x < -2.5 and y > 10.0, output [1.0, 0.0], otherwise if x < -1.0 and y > 9.0, output [1.0, 0.0], otherwise if x < -2.0 and y > 7.5, output [1.0, 0.0], otherwise if x < -2.0 and y > 9.2, output [1.0, 0.0], otherwise if x < -3.5 and y > 9.5, output [1.0, 0.0], otherwise if x < -2.5 and y > 8.0, output [1.0, 0.0], otherwise if x < -4.0 and y > 9.0, output [1.0, 0.0], otherwise if x < -4.5 and y > 8.5, output [1.0, 0.0], otherwise if x < -2.0 and y > 9.8, output [1.0, 0.0], otherwise if x < -2.0 and y > 9.4, output [1.0, 0.0], otherwise if x < -1.5 and y > 9.0, output [1.0, 0.0], otherwise if x < -2.0 and y > 9.0, output [1.0, 0.0], otherwise if x < -2.0 and y > 9.2, output [1.0, 0.0], otherwise output [1.0, 0.0].

\textbf{** The model predictions ([class 1 prob. class 2 prob.]): **}

[[1. 0.] [1. 0.] [1. 0.] [1. 0.] [0. 1.] [1. 0.] [0. 1.] [1. 0.] [1. 0.] [0. 1.]]

\textbf{** The targets ([class 1 prob. class 2 prob.]): **}

[[1. 0.] [1. 0.] [1. 0.] [1. 0.] [0. 1.] [1. 0.] [0. 1.] [1. 0.] [1. 0.] [0. 1.]]

Please update the model by improving the 'New Model Descriptions', which should have lower classification error both on the current and the next batch of i.i.d. data. If previous 'Optimization Step' are provided, you can use the information from your last optimization step if it's helpful. Both the model and you MUST ONLY operate on the numerical precision of THREE decimal points. You are bad with numerical calculations, so be extra careful! Please think step by step and give your outputs strictly in the following format:

```

Reasoning:

[be explicit and verbose, improve the Current Model Descriptions by yourself; please show your work; note that you don't have access to computers]

New Model Descriptions:

[put your new decision rules here; MUST be concise and concrete; ****MUST PROVIDE THE EXACT VALUE OF THE PARAMETERS if the descriptions potentially involve unknown or learnable parameters!!!****]

'''

Please ONLY reply according to this format, don't give me any other words.

\end{exmp}
\newpage

\begin{exmp}{Two Blobs Classification -- Step 81 (continued)}{4-Step 81-2}
\tiny

\textbf{--------------- Optimization Output ---------------}

Reasoning:

The model is performing well, but I noticed that the points with x < -3.0 and y > 9.5 are more likely to belong to class 1. By analyzing the data, I see that the points with x < -3.0 and y > 9.5 are consistently classified as class 1. So, I'll add a new rule to output [1.0, 0.0] in this case. Additionally, I observed that the points with x > 4.0 and y < 2.5 are more likely to belong to class 2. I'll add a new rule to output [0.0, 1.0] for these points.

New Model Descriptions:

If x > 0, if y < 1.7, output [0.0, 1.0], otherwise if y < 2.0 and x > 3.0, output [0.0, 1.0], otherwise if y < 2.8 and x > 4.0, output [0.0, 1.0], otherwise if y < 2.5 and x > 3.5, output [0.0, 1.0], otherwise if y < 3.5 and x > 4.5, output [0.0, 1.0], otherwise if y < 3.5 and x > 5.0, output [0.0, 1.0], otherwise if y < 3.0 and x > 3.5, output [0.0, 1.0], otherwise if y < 4.3 and x <= 4.0, output [1.0, 0.0], otherwise if y < 4.3 and x > 4.0, output [0.0, 1.0], otherwise if y > 9.5, output [1.0, 0.0], otherwise if x > 4.5 and y < 3.0, output [0.0, 1.0], otherwise if x > 3.0 and y < 2.2, output [0.0, 1.0], otherwise if x > 4.0 and y < 1.5, output [0.0, 1.0], otherwise if x > 5.0 and y < 2.5, output [0.0, 1.0], otherwise if x > 3.0 and y < 2.0, output [0.0, 1.0], otherwise if x > 4.0 and y < 2.5, output [0.0, 1.0], otherwise if x > 4.5 and y < 2.8, output [0.0, 1.0], otherwise if x > 5.0 and y < 3.5, output [0.0, 1.0], otherwise if x > 3.5 and y < 2.2, output [0.0, 1.0], otherwise if x > 4.0 and y < 1.5, output [0.0, 1.0], otherwise if x > 4.0 and y < 2.5, output [0.0, 1.0], otherwise if x > 4.0 and y < 1.2, output [0.0, 1.0], otherwise if x > 5.0 and y < 2.3, output [0.0, 1.0], otherwise if x > 4.0 and y < 2.0, output [0.0, 1.0], otherwise if x > 4.0 and y < 2.5, output [0.0, 1.0], otherwise output [1.0, 0.0]. If x < 0, if x < -3.5 and y > 8.5, output [1.0, 0.0], otherwise if x < -2.0 and y > 9.0, output [1.0, 0.0], otherwise if x < -2.0 and y > 10.0, output [1.0, 0.0], otherwise if x < -2.5 and y > 10.0, output [1.0, 0.0], otherwise if x < -1.0 and y > 9.0, output [1.0, 0.0], otherwise if x < -2.0 and y > 7.5, output [1.0, 0.0], otherwise if x < -2.0 and y > 9.2, output [1.0, 0.0], otherwise if x < -3.5 and y > 9.5, output [1.0, 0.0], otherwise if x < -2.5 and y > 8.0, output [1.0, 0.0], otherwise if x < -4.0 and y > 9.0, output [1.0, 0.0], otherwise if x < -4.5 and y > 8.5, output [1.0, 0.0], otherwise if x < -2.0 and y > 9.8, output [1.0, 0.0], otherwise if x < -2.0 and y > 9.4, output [1.0, 0.0], otherwise if x < -1.5 and y > 9.0, output [1.0, 0.0], otherwise if x < -2.0 and y > 9.0, output [1.0, 0.0], otherwise if x < -2.0 and y > 9.2, output [1.0, 0.0], otherwise if x < -3.0 and y > 9.5, output [1.0, 0.0], otherwise output [1.0, 0.0].

\textbf{-------------------- Updated Theta --------------------}

If x > 0, if y < 1.7, output [0.0, 1.0], otherwise if y < 2.0 and x > 3.0, output [0.0, 1.0], otherwise if y < 2.8 and x > 4.0, output [0.0, 1.0], otherwise if y < 2.5 and x > 3.5, output [0.0, 1.0], otherwise if y < 3.5 and x > 4.5, output [0.0, 1.0], otherwise if y < 3.5 and x > 5.0, output [0.0, 1.0], otherwise if y < 3.0 and x > 3.5, output [0.0, 1.0], otherwise if y < 4.3 and x <= 4.0, output [1.0, 0.0], otherwise if y < 4.3 and x > 4.0, output [0.0, 1.0], otherwise if y > 9.5, output [1.0, 0.0], otherwise if x > 4.5 and y < 3.0, output [0.0, 1.0], otherwise if x > 3.0 and y < 2.2, output [0.0, 1.0], otherwise if x > 4.0 and y < 1.5, output [0.0, 1.0], otherwise if x > 5.0 and y < 2.5, output [0.0, 1.0], otherwise if x > 3.0 and y < 2.0, output [0.0, 1.0], otherwise if x > 4.0 and y < 2.5, output [0.0, 1.0], otherwise if x > 4.5 and y < 2.8, output [0.0, 1.0], otherwise if x > 5.0 and y < 3.5, output [0.0, 1.0], otherwise if x > 3.5 and y < 2.2, output [0.0, 1.0], otherwise if x > 4.0 and y < 1.5, output [0.0, 1.0], otherwise if x > 4.0 and y < 2.5, output [0.0, 1.0], otherwise if x > 4.0 and y < 1.2, output [0.0, 1.0], otherwise if x > 5.0 and y < 2.3, output [0.0, 1.0], otherwise if x > 4.0 and y < 2.0, output [0.0, 1.0], otherwise if x > 4.0 and y < 2.5, output [0.0, 1.0], otherwise output [1.0, 0.0]. If x < 0, if x < -3.5 and y > 8.5, output [1.0, 0.0], otherwise if x < -2.0 and y > 9.0, output [1.0, 0.0], otherwise if x < -2.0 and y > 10.0, output [1.0, 0.0], otherwise if x < -2.5 and y > 10.0, output [1.0, 0.0], otherwise if x < -1.0 and y > 9.0, output [1.0, 0.0], otherwise if x < -2.0 and y > 7.5, output [1.0, 0.0], otherwise if x < -2.0 and y > 9.2, output [1.0, 0.0], otherwise if x < -3.5 and y > 9.5, output [1.0, 0.0], otherwise if x < -2.5 and y > 8.0, output [1.0, 0.0], otherwise if x < -4.0 and y > 9.0, output [1.0, 0.0], otherwise if x < -4.5 and y > 8.5, output [1.0, 0.0], otherwise if x < -2.0 and y > 9.8, output [1.0, 0.0], otherwise if x < -2.0 and y > 9.4, output [1.0, 0.0], otherwise if x < -1.5 and y > 9.0, output [1.0, 0.0], otherwise if x < -2.0 and y > 9.0, output [1.0, 0.0], otherwise if x < -2.0 and y > 9.2, output [1.0, 0.0], otherwise if x < -3.0 and y > 9.5, output [1.0, 0.0], otherwise output [1.0, 0.0].
\end{exmp}
\newpage

\begin{exmp}{Two Blobs Classification -- Step 82}{4-Step 82}
    
\tiny

\textbf{Epoch 8 - Step 2}

\textbf{--------------- Forward Pass Prompt ---------------}

You are the model.

\textbf{** Model Descriptions: **}

You are designed to do binary classification. The input is a point on a 2-D plane [x y]; you need to output a vector containing two probabilities such that each corresponds to how likely the data belongs to each class, i.e., [class 1 prob. class 2 prob.]. The sum of the vector MUST be 1.0. If x > 0, if y < 1.7, output [0.0, 1.0], otherwise if y < 2.0 and x > 3.0, output [0.0, 1.0], otherwise if y < 2.8 and x > 4.0, output [0.0, 1.0], otherwise if y < 2.5 and x > 3.5, output [0.0, 1.0], otherwise if y < 3.5 and x > 4.5, output [0.0, 1.0], otherwise if y < 3.5 and x > 5.0, output [0.0, 1.0], otherwise if y < 3.0 and x > 3.5, output [0.0, 1.0], otherwise if y < 4.3 and x <= 4.0, output [1.0, 0.0], otherwise if y < 4.3 and x > 4.0, output [0.0, 1.0], otherwise if y > 9.5, output [1.0, 0.0], otherwise if x > 4.5 and y < 3.0, output [0.0, 1.0], otherwise if x > 3.0 and y < 2.2, output [0.0, 1.0], otherwise if x > 4.0 and y < 1.5, output [0.0, 1.0], otherwise if x > 5.0 and y < 2.5, output [0.0, 1.0], otherwise if x > 3.0 and y < 2.0, output [0.0, 1.0], otherwise if x > 4.0 and y < 2.5, output [0.0, 1.0], otherwise if x > 4.5 and y < 2.8, output [0.0, 1.0], otherwise if x > 5.0 and y < 3.5, output [0.0, 1.0], otherwise if x > 3.5 and y < 2.2, output [0.0, 1.0], otherwise if x > 4.0 and y < 1.5, output [0.0, 1.0], otherwise if x > 4.0 and y < 2.5, output [0.0, 1.0], otherwise if x > 4.0 and y < 1.2, output [0.0, 1.0], otherwise if x > 5.0 and y < 2.3, output [0.0, 1.0], otherwise if x > 4.0 and y < 2.0, output [0.0, 1.0], otherwise if x > 4.0 and y < 2.5, output [0.0, 1.0], otherwise output [1.0, 0.0]. If x < 0, if x < -3.5 and y > 8.5, output [1.0, 0.0], otherwise if x < -2.0 and y > 9.0, output [1.0, 0.0], otherwise if x < -2.0 and y > 10.0, output [1.0, 0.0], otherwise if x < -2.5 and y > 10.0, output [1.0, 0.0], otherwise if x < -1.0 and y > 9.0, output [1.0, 0.0], otherwise if x < -2.0 and y > 7.5, output [1.0, 0.0], otherwise if x < -2.0 and y > 9.2, output [1.0, 0.0], otherwise if x < -3.5 and y > 9.5, output [1.0, 0.0], otherwise if x < -2.5 and y > 8.0, output [1.0, 0.0], otherwise if x < -4.0 and y > 9.0, output [1.0, 0.0], otherwise if x < -4.5 and y > 8.5, output [1.0, 0.0], otherwise if x < -2.0 and y > 9.8, output [1.0, 0.0], otherwise if x < -2.0 and y > 9.4, output [1.0, 0.0], otherwise if x < -1.5 and y > 9.0, output [1.0, 0.0], otherwise if x < -2.0 and y > 9.0, output [1.0, 0.0], otherwise if x < -2.0 and y > 9.2, output [1.0, 0.0], otherwise if x < -3.0 and y > 9.5, output [1.0, 0.0], otherwise output [1.0, 0.0].

\textbf{** Input: **}

[-2.252  8.94 ]

Please give your output strictly in the following format:

```

Explanations: [Your step-by-step analyses and results]

Output:

[ONLY A PURE probability vector, where each value is between 0.0 and 1.0 WITH TWO DECIMAL POINTS; make necessary assumptions if needed]

'''

Please ONLY reply according to this format, don't give me any other words.

\textbf{--------------- Forward Pass Output ---------------}

Explanations:

Since x < 0, we check the conditions for x < 0.

x < -3.5 and y > 8.5, which is true.

Output: [1.00, 0.00]

\ 

\textbf{Many outputs for the rest data points ...}

\ 

\textbf{--------------- Forward Pass Output ---------------}

Explanations:

Since x < 0, we check the conditions for x < 0.

x = -2.449, y = 11.478.

The condition x < -2.0 and y > 10.0 is satisfied.

So, the output is [1.0, 0.0].

Output: [1.00, 0.00]

\textbf{----------------- Training Accuracy ------------------}

1.0

\textbf{--------------------- Overall Loss ---------------------}

-4.9999999446126456e-09

\textbf{--------------- Optimization Prompt ---------------}

You are the optimizer for a model, your goal is to learn the best descriptions for the model. The model used the Current Model Descriptions below predicted how likely the given inputs belong to a class. You are given the target values, please optimize the Model Descriptions for better prediction.

\textbf{** Inputs (a batch of i.i.d. data on 2-D plane: [x y]): **}

[[-2.252  8.94 ] [ 3.578  2.447] [-2.449 11.478] [-3.211  8.687] [ 6.784  2.607] [ 4.9    2.755] [ 4.653  3.427] [-1.594  9.343] [-2.702  9.316] [-1.044  8.789]]

\textbf{** Current Model Descriptions: **}

You are designed to do binary classification. The input is a point on a 2-D plane [x y]; you need to output a vector containing two probabilities such that each corresponds to how likely the data belongs to each class, i.e., [class 1 prob. class 2 prob.]. The sum of the vector MUST be 1.0. If x > 0, if y < 1.7, output [0.0, 1.0], otherwise if y < 2.0 and x > 3.0, output [0.0, 1.0], otherwise if y < 2.8 and x > 4.0, output [0.0, 1.0], otherwise if y < 2.5 and x > 3.5, output [0.0, 1.0], otherwise if y < 3.5 and x > 4.5, output [0.0, 1.0], otherwise if y < 3.5 and x > 5.0, output [0.0, 1.0], otherwise if y < 3.0 and x > 3.5, output [0.0, 1.0], otherwise if y < 4.3 and x <= 4.0, output [1.0, 0.0], otherwise if y < 4.3 and x > 4.0, output [0.0, 1.0], otherwise if y > 9.5, output [1.0, 0.0], otherwise if x > 4.5 and y < 3.0, output [0.0, 1.0], otherwise if x > 3.0 and y < 2.2, output [0.0, 1.0], otherwise if x > 4.0 and y < 1.5, output [0.0, 1.0], otherwise if x > 5.0 and y < 2.5, output [0.0, 1.0], otherwise if x > 3.0 and y < 2.0, output [0.0, 1.0], otherwise if x > 4.0 and y < 2.5, output [0.0, 1.0], otherwise if x > 4.5 and y < 2.8, output [0.0, 1.0], otherwise if x > 5.0 and y < 3.5, output [0.0, 1.0], otherwise if x > 3.5 and y < 2.2, output [0.0, 1.0], otherwise if x > 4.0 and y < 1.5, output [0.0, 1.0], otherwise if x > 4.0 and y < 2.5, output [0.0, 1.0], otherwise if x > 4.0 and y < 1.2, output [0.0, 1.0], otherwise if x > 5.0 and y < 2.3, output [0.0, 1.0], otherwise if x > 4.0 and y < 2.0, output [0.0, 1.0], otherwise if x > 4.0 and y < 2.5, output [0.0, 1.0], otherwise output [1.0, 0.0]. If x < 0, if x < -3.5 and y > 8.5, output [1.0, 0.0], otherwise if x < -2.0 and y > 9.0, output [1.0, 0.0], otherwise if x < -2.0 and y > 10.0, output [1.0, 0.0], otherwise if x < -2.5 and y > 10.0, output [1.0, 0.0], otherwise if x < -1.0 and y > 9.0, output [1.0, 0.0], otherwise if x < -2.0 and y > 7.5, output [1.0, 0.0], otherwise if x < -2.0 and y > 9.2, output [1.0, 0.0], otherwise if x < -3.5 and y > 9.5, output [1.0, 0.0], otherwise if x < -2.5 and y > 8.0, output [1.0, 0.0], otherwise if x < -4.0 and y > 9.0, output [1.0, 0.0], otherwise if x < -4.5 and y > 8.5, output [1.0, 0.0], otherwise if x < -2.0 and y > 9.8, output [1.0, 0.0], otherwise if x < -2.0 and y > 9.4, output [1.0, 0.0], otherwise if x < -1.5 and y > 9.0, output [1.0, 0.0], otherwise if x < -2.0 and y > 9.0, output [1.0, 0.0], otherwise if x < -2.0 and y > 9.2, output [1.0, 0.0], otherwise if x < -3.0 and y > 9.5, output [1.0, 0.0], otherwise output [1.0, 0.0].

\textbf{** The model predictions ([class 1 prob. class 2 prob.]): **}

[[1. 0.] [0. 1.] [1. 0.] [1. 0.] [0. 1.] [0. 1.] [0. 1.] [1. 0.] [1. 0.] [1. 0.]]

\textbf{** The targets ([class 1 prob. class 2 prob.]): **}

[[1. 0.] [0. 1.] [1. 0.] [1. 0.] [0. 1.] [0. 1.] [0. 1.] [1. 0.] [1. 0.] [1. 0.]]

Please update the model by improving the 'New Model Descriptions', which should have lower classification error both on the current and the next batch of i.i.d. data. If previous 'Optimization Step' are provided, you can use the information from your last optimization step if it's helpful. Both the model and you MUST ONLY operate on the numerical precision of THREE decimal points. You are bad with numerical calculations, so be extra careful! Please think step by step and give your outputs strictly in the following format:

```

Reasoning:

[be explicit and verbose, improve the Current Model Descriptions by yourself; please show your work; note that you don't have access to computers]

New Model Descriptions:

[put your new decision rules here; MUST be concise and concrete; ****MUST PROVIDE THE EXACT VALUE OF THE PARAMETERS if the descriptions potentially involve unknown or learnable parameters!!!****]

'''

Please ONLY reply according to this format, don't give me any other words.

\end{exmp}
\newpage

\begin{exmp}{Two Blobs Classification -- Step 82 (continued)}{4-Step 82-2}
\tiny

\textbf{--------------- Optimization Output ---------------}

Reasoning:

The model is performing well, but I noticed that the points with x > 4.5 and y < 2.8 are more likely to belong to class 2. By analyzing the data, I see that the points with x > 4.5 and y < 2.8 are consistently classified as class 2. So, I'll add a new rule to output [0.0, 1.0] in this case. Additionally, I observed that the points with x < -2.5 and y > 9.2 are more likely to belong to class 1. I'll add a new rule to output [1.0, 0.0] for these points.

New Model Descriptions:

If x > 0, if y < 1.7, output [0.0, 1.0], otherwise if y < 2.0 and x > 3.0, output [0.0, 1.0], otherwise if y < 2.8 and x > 4.0, output [0.0, 1.0], otherwise if y < 2.5 and x > 3.5, output [0.0, 1.0], otherwise if y < 3.5 and x > 4.5, output [0.0, 1.0], otherwise if y < 3.5 and x > 5.0, output [0.0, 1.0], otherwise if y < 3.0 and x > 3.5, output [0.0, 1.0], otherwise if y < 4.3 and x <= 4.0, output [1.0, 0.0], otherwise if y < 4.3 and x > 4.0, output [0.0, 1.0], otherwise if y > 9.5, output [1.0, 0.0], otherwise if x > 4.5 and y < 3.0, output [0.0, 1.0], otherwise if x > 3.0 and y < 2.2, output [0.0, 1.0], otherwise if x > 4.0 and y < 1.5, output [0.0, 1.0], otherwise if x > 5.0 and y < 2.5, output [0.0, 1.0], otherwise if x > 3.0 and y < 2.0, output [0.0, 1.0], otherwise if x > 4.0 and y < 2.5, output [0.0, 1.0], otherwise if x > 4.5 and y < 2.8, output [0.0, 1.0], otherwise if x > 5.0 and y < 3.5, output [0.0, 1.0], otherwise if x > 3.5 and y < 2.2, output [0.0, 1.0], otherwise if x > 4.0 and y < 1.5, output [0.0, 1.0], otherwise if x > 4.0 and y < 2.5, output [0.0, 1.0], otherwise if x > 4.0 and y < 1.2, output [0.0, 1.0], otherwise if x > 5.0 and y < 2.3, output [0.0, 1.0], otherwise if x > 4.0 and y < 2.0, output [0.0, 1.0], otherwise if x > 4.0 and y < 2.5, output [0.0, 1.0], otherwise output [1.0, 0.0]. If x < 0, if x < -3.5 and y > 8.5, output [1.0, 0.0], otherwise if x < -2.0 and y > 9.0, output [1.0, 0.0], otherwise if x < -2.0 and y > 10.0, output [1.0, 0.0], otherwise if x < -2.5 and y > 10.0, output [1.0, 0.0], otherwise if x < -1.0 and y > 9.0, output [1.0, 0.0], otherwise if x < -2.0 and y > 7.5, output [1.0, 0.0], otherwise if x < -2.0 and y > 9.2, output [1.0, 0.0], otherwise if x < -3.5 and y > 9.5, output [1.0, 0.0], otherwise if x < -2.5 and y > 8.0, output [1.0, 0.0], otherwise if x < -4.0 and y > 9.0, output [1.0, 0.0], otherwise if x < -4.5 and y > 8.5, output [1.0, 0.0], otherwise if x < -2.0 and y > 9.8, output [1.0, 0.0], otherwise if x < -2.0 and y > 9.4, output [1.0, 0.0], otherwise if x < -1.5 and y > 9.0, output [1.0, 0.0], otherwise if x < -2.0 and y > 9.0, output [1.0, 0.0], otherwise if x < -2.0 and y > 9.2, output [1.0, 0.0], otherwise if x < -3.0 and y > 9.5, output [1.0, 0.0], otherwise if x < -2.5 and y > 9.2, output [1.0, 0.0], otherwise output [1.0, 0.0].

\textbf{-------------------- Updated Theta --------------------}

If x > 0, if y < 1.7, output [0.0, 1.0], otherwise if y < 2.0 and x > 3.0, output [0.0, 1.0], otherwise if y < 2.8 and x > 4.0, output [0.0, 1.0], otherwise if y < 2.5 and x > 3.5, output [0.0, 1.0], otherwise if y < 3.5 and x > 4.5, output [0.0, 1.0], otherwise if y < 3.5 and x > 5.0, output [0.0, 1.0], otherwise if y < 3.0 and x > 3.5, output [0.0, 1.0], otherwise if y < 4.3 and x <= 4.0, output [1.0, 0.0], otherwise if y < 4.3 and x > 4.0, output [0.0, 1.0], otherwise if y > 9.5, output [1.0, 0.0], otherwise if x > 4.5 and y < 3.0, output [0.0, 1.0], otherwise if x > 3.0 and y < 2.2, output [0.0, 1.0], otherwise if x > 4.0 and y < 1.5, output [0.0, 1.0], otherwise if x > 5.0 and y < 2.5, output [0.0, 1.0], otherwise if x > 3.0 and y < 2.0, output [0.0, 1.0], otherwise if x > 4.0 and y < 2.5, output [0.0, 1.0], otherwise if x > 4.5 and y < 2.8, output [0.0, 1.0], otherwise if x > 5.0 and y < 3.5, output [0.0, 1.0], otherwise if x > 3.5 and y < 2.2, output [0.0, 1.0], otherwise if x > 4.0 and y < 1.5, output [0.0, 1.0], otherwise if x > 4.0 and y < 2.5, output [0.0, 1.0], otherwise if x > 4.0 and y < 1.2, output [0.0, 1.0], otherwise if x > 5.0 and y < 2.3, output [0.0, 1.0], otherwise if x > 4.0 and y < 2.0, output [0.0, 1.0], otherwise if x > 4.0 and y < 2.5, output [0.0, 1.0], otherwise output [1.0, 0.0]. If x < 0, if x < -3.5 and y > 8.5, output [1.0, 0.0], otherwise if x < -2.0 and y > 9.0, output [1.0, 0.0], otherwise if x < -2.0 and y > 10.0, output [1.0, 0.0], otherwise if x < -2.5 and y > 10.0, output [1.0, 0.0], otherwise if x < -1.0 and y > 9.0, output [1.0, 0.0], otherwise if x < -2.0 and y > 7.5, output [1.0, 0.0], otherwise if x < -2.0 and y > 9.2, output [1.0, 0.0], otherwise if x < -3.5 and y > 9.5, output [1.0, 0.0], otherwise if x < -2.5 and y > 8.0, output [1.0, 0.0], otherwise if x < -4.0 and y > 9.0, output [1.0, 0.0], otherwise if x < -4.5 and y > 8.5, output [1.0, 0.0], otherwise if x < -2.0 and y > 9.8, output [1.0, 0.0], otherwise if x < -2.0 and y > 9.4, output [1.0, 0.0], otherwise if x < -1.5 and y > 9.0, output [1.0, 0.0], otherwise if x < -2.0 and y > 9.0, output [1.0, 0.0], otherwise if x < -2.0 and y > 9.2, output [1.0, 0.0], otherwise if x < -3.0 and y > 9.5, output [1.0, 0.0], otherwise if x < -2.5 and y > 9.2, output [1.0, 0.0], otherwise output [1.0, 0.0].

\end{exmp}

%% file: log_twocircle_noprior.tex
\begin{exmp}{Two Circles Classification without Prior -- Step 1}{tcnp-step 1}
\scriptsize

\textbf{Epoch 0 - Step 1}

\textbf{--------------- Forward Pass Prompt ---------------}

You are the model.

\textbf{** Model Descriptions: **}

You are designed to do binary classification. The input is a point on a 2-D plane [x y]; you need to output the class label, i.e., an integer in the set {0, 1}.

\textbf{** Input: **}

[-0.146 -1.01 ]

Please give your output strictly in the following format:

```

Explanations: [Your step-by-step analyses and results]

Output:

[ONLY the integer class label; make necessary assumptions if needed]

'''

Please ONLY reply according to this format, don't give me any other words.

\textbf{--------------- Forward Pass Output ---------------}

Explanations:

Since I don't have any information about the decision boundary, I will make an assumption that the decision boundary is the x=y line.

The point (0.89, 0.397) is below the x=y line, so it belongs to class 0.

Output: 0

\ 

\textbf{Many outputs for the rest data points ...}

\ 

\textbf{--------------- Forward Pass Output ---------------}

Explanations: Since I don't have any prior knowledge about the classification model, I'll assume a simple linear decision boundary. Let's assume the decision boundary is x + y = 0. Then, I'll plug in the input values to determine the class label.

Plugging in x = -0.146 and y = -1.01, we get -0.146 + (-1.01) = -1.156 < 0. Since the result is negative, the point lies below the decision boundary.

Output: 0

\textbf{----------------- Training Accuracy ------------------}

0.4

\textbf{--------------- Optimization Prompt ---------------}

You are the optimizer for a model, your goal is to learn the best descriptions for the model. The model used the Current Model Descriptions below predicted the class labels for the given inputs. You are given the target labels, please optimize the Model Descriptions for better prediction.

\textbf{** Inputs (a batch of i.i.d. data on 2-D plane: [x y]): **}

[[-0.146 -1.01 ] [ 0.224 -0.929] [ 0.89   0.397] [ 0.811 -0.56 ] [-0.873  0.475] [0.27   0.172] [ 0.08   0.261] [ 0.518  0.943] [-0.502 -0.788] [-0.069 -0.252]]

\textbf{** Current Model Descriptions: **}

You are designed to do binary classification. The input is a point on a 2-D plane [x y]; you need to output the class label, i.e., an integer in the set {0, 1}.

\textbf{** The model predictions: **}

[[0] [1] [0] [1] [1] [1] [0] [1] [0] [0]]

\textbf{** The targets: **}

[[0] [0] [0] [0] [0] [1] [1] [0] [0] [1]]

If the model is doing well, you can keep using the current descriptions. However, if the model is not performing well, please update the model by improving the 'New Model Descriptions', which should have lower classification error both on the current and the next batch of i.i.d. data. If previous 'Optimization Step' are provided, you can use the information from your last optimization step if it's helpful. DON'T use symbolic representation for the model! Please think step by step and give your outputs strictly in the following format:

```

Reasoning:

[be explicit and verbose, improve the Current Model Descriptions by yourself; please show your work; note that you don't have access to computers]

New Model Descriptions:

[put your new decision rules here; MUST be concise and concrete; ****MUST PROVIDE THE EXACT VALUE OF THE PARAMETERS if the descriptions potentially involve unknown or learnable parameters!!!****]

'''

Please ONLY reply according to this format, don't give me any other words.

\textbf{--------------- Optimization Output ---------------}

Reasoning:

Analyzing the current model predictions and targets, I notice that the model is not performing well, especially for points with positive y-values. Upon visual inspection of the input data, I observe that points with y-values greater than 0 tend to belong to class 1, while points with y-values less than 0 tend to belong to class 0. This suggests that the y-coordinate is a strong feature for classification. Therefore, I will update the model descriptions to incorporate this insight.

New Model Descriptions:

If y > 0, then class label = 1. Otherwise, class label = 0.

\textbf{-------------------- Updated Theta --------------------}

If y > 0, then class label = 1. Otherwise, class label = 0.

\end{exmp}
\newpage

\begin{exmp}{Two Circles Classification without Prior -- Step 2}{tcnp-step 2}
    
\scriptsize

\textbf{Epoch 0 - Step 2}

\textbf{--------------- Forward Pass Prompt ---------------}

You are the model.

\textbf{** Model Descriptions: **}

You are designed to do binary classification. The input is a point on a 2-D plane [x y]; you need to output the class label, i.e., an integer in the set {0, 1}. If y > 0, then class label = 1. Otherwise, class label = 0.

\textbf{** Input: **}

[-0.305  0.936]

Please give your output strictly in the following format:

```

Explanations: [Your step-by-step analyses and results]

Output:

[ONLY the integer class label; make necessary assumptions if needed]

'''

Please ONLY reply according to this format, don't give me any other words.

\textbf{--------------- Forward Pass Output ---------------}

Explanations: Since y > 0, the class label is 1.

Output: 1

\ 

\textbf{Many outputs for the rest data points ...}

\ 

\textbf{--------------- Forward Pass Output ---------------}

Explanations: Given the input point [-0.735 -0.614], we need to determine the class label based on the condition provided. The condition states that if y > 0, then the class label is 1, otherwise it is 0. In this case, y = -0.614, which is less than 0.

Output: 0

\textbf{----------------- Training Accuracy ------------------}

0.3

\textbf{--------------- Optimization Prompt ---------------}

You are the optimizer for a model, your goal is to learn the best descriptions for the model. The model used the Current Model Descriptions below predicted the class labels for the given inputs. You are given the target labels, please optimize the Model Descriptions for better prediction.

\textbf{** Inputs (a batch of i.i.d. data on 2-D plane: [x y]): **}

[[-0.305  0.936] [ 0.298 -0.037] [ 0.396  0.931] [-0.735 -0.614] [-0.339  0.006] [-0.253 -0.015] [ 0.096 -0.248] [ 1.005  0.048] [-0.329  0.046] [-0.051 -0.372]]

\textbf{** Current Model Descriptions: **}

You are designed to do binary classification. The input is a point on a 2-D plane [x y]; you need to output the class label, i.e., an integer in the set {0, 1}. If y > 0, then class label = 1. Otherwise, class label = 0.

\textbf{** The model predictions: **}

[[1] [0] [1] [0] [1] [0] [0] [1] [1] [0]]

\textbf{** The targets: **}

[[0] [1] [0] [0] [1] [1] [1] [0] [1] [1]]

If the model is doing well, you can keep using the current descriptions. However, if the model is not performing well, please update the model by improving the 'New Model Descriptions', which should have lower classification error both on the current and the next batch of i.i.d. data. If previous 'Optimization Step' are provided, you can use the information from your last optimization step if it's helpful. DON'T use symbolic representation for the model! Please think step by step and give your outputs strictly in the following format:

```

Reasoning:

[be explicit and verbose, improve the Current Model Descriptions by yourself; please show your work; note that you don't have access to computers]

New Model Descriptions:

[put your new decision rules here; MUST be concise and concrete; ****MUST PROVIDE THE EXACT VALUE OF THE PARAMETERS if the descriptions potentially involve unknown or learnable parameters!!!****]

'''

Please ONLY reply according to this format, don't give me any other words.

\textbf{--------------- Optimization Output ---------------}

Reasoning:
Analyzing the current model predictions and targets, I notice that the model is not performing well, especially for points with x-values greater than 0. Upon visual inspection of the input data, I observe that points with x-values greater than 0.3 tend to belong to class 1, while points with x-values less than 0.3 tend to belong to class 0. This suggests that the x-coordinate is also a strong feature for classification. Therefore, I will update the model descriptions to incorporate this insight.

New Model Descriptions:

If x > 0.3, then class label = 1. If x <= 0.3 and y > 0, then class label = 1. Otherwise, class label = 0.

\textbf{-------------------- Updated Theta --------------------}

If x > 0.3, then class label = 1. If x <= 0.3 and y > 0, then class label = 1. Otherwise, class label = 0.

\end{exmp}
\newpage

\begin{exmp}{Two Circles Classification without Prior -- Step 3}{tcnp-step 3}
    
\scriptsize

\textbf{Epoch 0 - Step 3}

\textbf{--------------- Forward Pass Prompt ---------------}

You are the model.

\textbf{** Model Descriptions: **}

You are designed to do binary classification. The input is a point on a 2-D plane [x y]; you need to output the class label, i.e., an integer in the set {0, 1}. If x > 0.3, then class label = 1. If x <= 0.3 and y > 0, then class label = 1. Otherwise, class label = 0.

\textbf{** Input: **}

[0.275 0.163]

Please give your output strictly in the following format:

```

Explanations: [Your step-by-step analyses and results]

Output:

[ONLY the integer class label; make necessary assumptions if needed]

'''

Please ONLY reply according to this format, don't give me any other words.

\textbf{--------------- Forward Pass Output ---------------}

Explanations: Since x = 0.403 > 0.3, we can directly determine the class label.

Output: 1

\ 

\textbf{Many outputs for the rest data points ...}

\ 

\textbf{--------------- Forward Pass Output ---------------}

Explanations:

1. The input is [0, -1.012].

2. Since x = 0 is not greater than 0.3, move to the next condition.

3. Since x <= 0.3 and y = -1.012 is not greater than 0, the class label is 0.

Output: 0

\textbf{----------------- Training Accuracy ------------------}

0.6

\textbf{--------------- Optimization Prompt ---------------}

You are the optimizer for a model, your goal is to learn the best descriptions for the model. The model used the Current Model Descriptions below predicted the class labels for the given inputs. You are given the target labels, please optimize the Model Descriptions for better prediction.

\textbf{** Inputs (a batch of i.i.d. data on 2-D plane: [x y]): **}

[[ 0.275  0.163] [ 0.    -1.012] [-1.065 -0.126] [ 0.133  0.22 ] [-0.317 -0.931] [ 0.403  0.96 ] [ 0.013  0.996] [ 0.147 -0.178] [-0.028  0.422] [-0.233 -0.186]]

\textbf{** Current Model Descriptions: **}

You are designed to do binary classification. The input is a point on a 2-D plane [x y]; you need to output the class label, i.e., an integer in the set {0, 1}. If x > 0.3, then class label = 1. If x <= 0.3 and y > 0, then class label = 1. Otherwise, class label = 0.

\textbf{** The model predictions: **}

[[1] [0] [0] [1] [0] [1] [1] [0] [1] [0]]

\textbf{** The targets: **}

[[1] [0] [0] [1] [0] [0] [0] [1] [1] [1]]

If the model is doing well, you can keep using the current descriptions. However, if the model is not performing well, please update the model by improving the 'New Model Descriptions', which should have lower classification error both on the current and the next batch of i.i.d. data. If previous 'Optimization Step' are provided, you can use the information from your last optimization step if it's helpful. DON'T use symbolic representation for the model! Please think step by step and give your outputs strictly in the following format:

```

Reasoning:

[be explicit and verbose, improve the Current Model Descriptions by yourself; please show your work; note that you don't have access to computers]

New Model Descriptions:

[put your new decision rules here; MUST be concise and concrete; ****MUST PROVIDE THE EXACT VALUE OF THE PARAMETERS if the descriptions potentially involve unknown or learnable parameters!!!****]

'''

Please ONLY reply according to this format, don't give me any other words.

\textbf{--------------- Optimization Output ---------------}

Reasoning:

Analyzing the current model predictions and targets, I notice that the model is not performing well, especially for points with x-values around 0. Upon visual inspection of the input data, I observe that points with y-values greater than 0.2 tend to belong to class 1, while points with y-values less than 0.2 tend to belong to class 0. This suggests that the y-coordinate is also a strong feature for classification. Therefore, I will update the model descriptions to incorporate this insight.

New Model Descriptions:

If x > 0.3, then class label = 1. If x <= 0.3 and y > 0.2, then class label = 1. Otherwise, class label = 0.

\textbf{-------------------- Updated Theta --------------------}

If x > 0.3, then class label = 1. If x <= 0.3 and y > 0.2, then class label = 1. Otherwise, class label = 0.

\end{exmp}
\newpage

\begin{exmp}{Two Circles Classification without Prior -- Step 4}{tcnp-step 4}
    
\scriptsize

\textbf{Epoch 0 - Step 4}

\textbf{--------------- Forward Pass Prompt ---------------}

You are the model.

\textbf{** Model Descriptions: **}

You are designed to do binary classification. The input is a point on a 2-D plane [x y]; you need to output the class label, i.e., an integer in the set {0, 1}. If x > 0.3, then class label = 1. If x <= 0.3 and y > 0.2, then class label = 1. Otherwise, class label = 0.

\textbf{** Input: **}

[-0.811  0.529]

Please give your output strictly in the following format:

```

Explanations: [Your step-by-step analyses and results]

Output:

[ONLY the integer class label; make necessary assumptions if needed]

'''

Please ONLY reply according to this format, don't give me any other words.

\textbf{--------------- Forward Pass Output ---------------}

Explanations: Since x = 0.943 > 0.3, the class label is 1.

Output: 1

\ 

\textbf{Many outputs for the rest data points ...}

\ 

\textbf{--------------- Forward Pass Output ---------------}

Explanations:

Since x = -0.995, which is less than or equal to 0.3, we need to check the value of y.

Since y = -0.025, which is less than or equal to 0.2, we conclude that the class label is 0.

Output: 0

\textbf{----------------- Training Accuracy ------------------}

0.2

\textbf{--------------- Optimization Prompt ---------------}

You are the optimizer for a model, your goal is to learn the best descriptions for the model. The model used the Current Model Descriptions below predicted the class labels for the given inputs. You are given the target labels, please optimize the Model Descriptions for better prediction.

\textbf{** Inputs (a batch of i.i.d. data on 2-D plane: [x y]): **}

[[-0.811  0.529] [-0.036  0.323] [-0.63   0.698] [-0.173 -0.246] [ 0.943 -0.297] [ 0.1   -0.288] [-0.995 -0.025] [ 0.666 -0.666] [ 0.24  -0.039] [ 0.542  0.887]]

\textbf{** Current Model Descriptions: **}

You are designed to do binary classification. The input is a point on a 2-D plane [x y]; you need to output the class label, i.e., an integer in the set {0, 1}. If x > 0.3, then class label = 1. If x <= 0.3 and y > 0.2, then class label = 1. Otherwise, class label = 0.

\textbf{** The model predictions: **}

[[1] [1] [1] [0] [1] [0] [0] [1] [0] [1]]

\textbf{** The targets: **}

[[0] [1] [0] [1] [0] [1] [0] [0] [1] [0]]

If the model is doing well, you can keep using the current descriptions. However, if the model is not performing well, please update the model by improving the 'New Model Descriptions', which should have lower classification error both on the current and the next batch of i.i.d. data. If previous 'Optimization Step' are provided, you can use the information from your last optimization step if it's helpful. DON'T use symbolic representation for the model! Please think step by step and give your outputs strictly in the following format:

```

Reasoning:

[be explicit and verbose, improve the Current Model Descriptions by yourself; please show your work; note that you don't have access to computers]

New Model Descriptions:

[put your new decision rules here; MUST be concise and concrete; ****MUST PROVIDE THE EXACT VALUE OF THE PARAMETERS if the descriptions potentially involve unknown or learnable parameters!!!****]

'''

Please ONLY reply according to this format, don't give me any other words.

\textbf{--------------- Optimization Output ---------------}

Reasoning:

Analyzing the current model predictions and targets, I notice that the model is not performing well, especially for points with x-values around 0. Upon visual inspection of the input data, I observe that points with x-values less than -0.5 tend to belong to class 0, while points with x-values greater than 0.5 tend to belong to class 0 as well. This suggests that the x-coordinate has a non-linear relationship with the class label. I also notice that points with y-values greater than 0.5 tend to belong to class 0, while points with y-values less than -0.2 tend to belong to class 1. This suggests that the y-coordinate is also important for classification. Therefore, I will update the model descriptions to incorporate these insights.

New Model Descriptions:

If x < -0.5 or x > 0.5, then class label = 0. If y > 0.5, then class label = 0. If y < -0.2, then class label = 1. Otherwise, class label = 1.

\textbf{-------------------- Updated Theta --------------------}

If x < -0.5 or x > 0.5, then class label = 0. If y > 0.5, then class label = 0. If y < -0.2, then class label = 1. Otherwise, class label = 1.
\end{exmp}
\newpage

\begin{exmp}{Two Circles Classification without Prior -- Step 5}{tcnp-step 5}
    
\scriptsize

\textbf{Epoch 0 - Step 5}

\textbf{--------------- Forward Pass Prompt ---------------}

You are the model.

\textbf{** Model Descriptions: **}

You are designed to do binary classification. The input is a point on a 2-D plane [x y]; you need to output the class label, i.e., an integer in the set {0, 1}. If x < -0.5 or x > 0.5, then class label = 0. If y > 0.5, then class label = 0. If y < -0.2, then class label = 1. Otherwise, class label = 1.

\textbf{** Input: **}

[-0.99  -0.325]

Please give your output strictly in the following format:

```

Explanations: [Your step-by-step analyses and results]

Output:

[ONLY the integer class label; make necessary assumptions if needed]

'''

Please ONLY reply according to this format, don't give me any other words.

\textbf{--------------- Forward Pass Output ---------------}

Explanations:

Since x = -1.059 < -0.5, class label = 0.

Output: 0

\ 

\textbf{Many outputs for the rest data points ...}

\ 

\textbf{--------------- Forward Pass Output ---------------}

Explanations:

1. Check if x < -0.5 or x > 0.5. Since 0.167 is between -0.5 and 0.5, this condition is not met.

2. Check if y > 0.5. Since 0.235 is less than 0.5, this condition is not met.

3. Check if y < -0.2. Since 0.235 is greater than -0.2, this condition is not met.

4. Since none of the above conditions are met, the class label is 1.

Output: 1

\textbf{----------------- Training Accuracy ------------------}

0.8

\textbf{--------------- Optimization Prompt ---------------}

You are the optimizer for a model, your goal is to learn the best descriptions for the model. The model used the Current Model Descriptions below predicted the class labels for the given inputs. You are given the target labels, please optimize the Model Descriptions for better prediction.

\textbf{** Inputs (a batch of i.i.d. data on 2-D plane: [x y]): **}

[[-0.99  -0.325] [-0.825  0.438] [ 0.188  0.308] [-1.059  0.217] [-0.188 -0.184] [-0.322  0.084] [ 0.892  0.515] [-0.287  0.017] [ 0.167  0.235] [-0.573 -0.845]]

\textbf{** Current Model Descriptions: **}

You are designed to do binary classification. The input is a point on a 2-D plane [x y]; you need to output the class label, i.e., an integer in the set {0, 1}. If x < -0.5 or x > 0.5, then class label = 0. If y > 0.5, then class label = 0. If y < -0.2, then class label = 1. Otherwise, class label = 1.

\textbf{** The model predictions: **}

[[0] [0] [0] [0] [1] [1] [0] [1] [1] [1]]

\textbf{** The targets: **}

[[0] [0] [1] [0] [1] [1] [0] [1] [1] [0]]

If the model is doing well, you can keep using the current descriptions. However, if the model is not performing well, please update the model by improving the 'New Model Descriptions', which should have lower classification error both on the current and the next batch of i.i.d. data. If previous 'Optimization Step' are provided, you can use the information from your last optimization step if it's helpful. DON'T use symbolic representation for the model! Please think step by step and give your outputs strictly in the following format:

```

Reasoning:

[be explicit and verbose, improve the Current Model Descriptions by yourself; please show your work; note that you don't have access to computers]

New Model Descriptions:

[put your new decision rules here; MUST be concise and concrete; ****MUST PROVIDE THE EXACT VALUE OF THE PARAMETERS if the descriptions potentially involve unknown or learnable parameters!!!****]

'''

Please ONLY reply according to this format, don't give me any other words.

\textbf{--------------- Optimization Output ---------------}

Reasoning:

Analyzing the current model predictions and targets, I notice that the model is still not performing well. Upon visual inspection of the input data, I observe that points with y-values around 0 tend to be misclassified. This suggests that the y-coordinate threshold of 0.5 is too high. I also notice that points with x-values around -0.2 tend to belong to class 1, while points with x-values around 0.2 tend to belong to class 0. This suggests that the x-coordinate threshold of 0.5 is too high as well. Therefore, I will update the model descriptions to incorporate these insights.

New Model Descriptions:

If x < -0.4 or x > 0.4, then class label = 0. If y > 0.3, then class label = 0. If y < -0.2, then class label = 1. Otherwise, class label = 1.

\textbf{-------------------- Updated Theta --------------------}

If x < -0.4 or x > 0.4, then class label = 0. If y > 0.3, then class label = 0. If y < -0.2, then class label = 1. Otherwise, class label = 1.

\end{exmp}
\newpage

\begin{exmp}{Two Circles Classification without Prior -- Step 40}{tcnp-step 40}
\scriptsize

\textbf{Epoch 3 - Step 10}

\textbf{--------------- Forward Pass Prompt ---------------}

You are the model.

\textbf{** Model Descriptions: **}

You are designed to do binary classification. The input is a point on a 2-D plane [x y]; you need to output the class label, i.e., an integer in the set {0, 1}. If x < -0.75 or x > 0.95, then class label = 0. If y > 0.78, then class label = 0. If y < -0.42, then class label = 0. Otherwise, if x < 0.44, then class label = 1, else class label = 0.

\textbf{** Input: **}

[-0.101  0.309]

Please give your output strictly in the following format:

```

Explanations: [Your step-by-step analyses and results]

Output:

[ONLY the integer class label; make necessary assumptions if needed]

'''

Please ONLY reply according to this format, don't give me any other words.

\textbf{--------------- Forward Pass Output ---------------}

Explanations:

Since 0.919 > 0.95, class label = 0.

Output: 0

\ 

\textbf{Many outputs for the rest data points ...}

\ 

\textbf{--------------- Forward Pass Output ---------------}

Explanations:

1. Check if x < -0.75 or x > 0.95. Since -0.655 is not in this range, move to the next step.

2. Check if y > 0.78. Since -0.649 is not in this range, move to the next step.

3. Check if y < -0.42. Since -0.649 is in this range, class label = 0.

Output: 0

\textbf{----------------- Training Accuracy ------------------}

1.0

\textbf{--------------- Optimization Prompt ---------------}

You are the optimizer for a model, your goal is to learn the best descriptions for the model. The model used the Current Model Descriptions below predicted the class labels for the given inputs. You are given the target labels, please optimize the Model Descriptions for better prediction.

\textbf{** Inputs (a batch of i.i.d. data on 2-D plane: [x y]): **}

[[-0.101  0.309] [-1.011 -0.165] [ 0.161 -0.229] [ 1.02  -0.061] [ 0.225  0.076] [-0.024 -0.333] [-0.655 -0.649] [-1.058  0.15 ] [ 0.366 -0.098] [ 0.919  0.08 ]]

\textbf{** Current Model Descriptions: **}

You are designed to do binary classification. The input is a point on a 2-D plane [x y]; you need to output the class label, i.e., an integer in the set {0, 1}. If x < -0.75 or x > 0.95, then class label = 0. If y > 0.78, then class label = 0. If y < -0.42, then class label = 0. Otherwise, if x < 0.44, then class label = 1, else class label = 0.

\textbf{** The model predictions: **}

[[1] [0] [1] [0] [1] [1] [0] [0] [1] [0]]

\textbf{** The targets: **}

[[1] [0] [1] [0] [1] [1] [0] [0] [1] [0]]

If the model is doing well, you can keep using the current descriptions. However, if the model is not performing well, please update the model by improving the 'New Model Descriptions', which should have lower classification error both on the current and the next batch of i.i.d. data. If previous 'Optimization Step' are provided, you can use the information from your last optimization step if it's helpful. DON'T use symbolic representation for the model! Please think step by step and give your outputs strictly in the following format:

```

Reasoning:

[be explicit and verbose, improve the Current Model Descriptions by yourself; please show your work; note that you don't have access to computers]

New Model Descriptions:

[put your new decision rules here; MUST be concise and concrete; ****MUST PROVIDE THE EXACT VALUE OF THE PARAMETERS if the descriptions potentially involve unknown or learnable parameters!!!****]

'''

Please ONLY reply according to this format, don't give me any other words.

\textbf{--------------- Optimization Output ---------------}

Reasoning:

Analyzing the current model predictions and targets, I observe that the model is performing well, but there is still room for improvement. Upon visual inspection of the input data, I notice that the current x-coordinate threshold of 0.44 seems to be effective, and the y-coordinate thresholds of 0.78 and -0.42 are reasonable. However, I observe that some points with x-values around -0.6 and y-values around -0.65 are misclassified, which suggests that the model is not capturing the class boundary accurately in this region. I also notice that points with x-values around 0.2 and y-values around 0.1 are correctly classified, which suggests that the model is performing well in this region. Additionally, I observe that the current x-coordinate threshold of 0.95 seems to be too aggressive, as it misclassifies some points with x-values around 0.9.

New Model Descriptions:

If x < -0.75 or x > 0.92, then class label = 0. If y > 0.76, then class label = 0. If y < -0.41, then class label = 0. Otherwise, if x < 0.43, then class label = 1, else class label = 0.

\textbf{-------------------- Updated Theta --------------------}

If x < -0.75 or x > 0.92, then class label = 0. If y > 0.76, then class label = 0. If y < -0.41, then class label = 0. Otherwise, if x < 0.43, then class label = 1, else class label = 0.

\end{exmp}
\newpage

\begin{exmp}{Two Circles Classification without Prior -- Step 41}{tcnp-step 41}
    
\scriptsize

\textbf{Epoch 4 - Step 1}

\textbf{--------------- Forward Pass Prompt ---------------}

You are the model.

\textbf{** Model Descriptions: **}

You are designed to do binary classification. The input is a point on a 2-D plane [x y]; you need to output the class label, i.e., an integer in the set {0, 1}. If x < -0.75 or x > 0.92, then class label = 0. If y > 0.76, then class label = 0. If y < -0.41, then class label = 0. Otherwise, if x < 0.43, then class label = 1, else class label = 0.

\textbf{** Input: **}

[-0.146 -1.01 ]

Please give your output strictly in the following format:

```

Explanations: [Your step-by-step analyses and results]

Output:

[ONLY the integer class label; make necessary assumptions if needed]

'''

Please ONLY reply according to this format, don't give me any other words.

\textbf{--------------- Forward Pass Output ---------------}

Explanations:

Given input [-0.873, 0.475].

Since x < -0.75, class label = 0.

Output: 0

\ 

\textbf{Many outputs for the rest data points ...}

\ 

\textbf{--------------- Forward Pass Output ---------------}

Explanations:

1. Check if x < -0.75 or x > 0.92. Since 0.08 is not in this range, move to the next step.

2. Check if y > 0.76. Since 0.261 is not in this range, move to the next step.

3. Check if y < -0.41. Since 0.261 is not in this range, move to the next step.

4. Since x < 0.43, class label = 1.

Output: 1

\textbf{----------------- Training Accuracy ------------------}

1.0

\textbf{--------------- Optimization Prompt ---------------}

You are the optimizer for a model, your goal is to learn the best descriptions for the model. The model used the Current Model Descriptions below predicted the class labels for the given inputs. You are given the target labels, please optimize the Model Descriptions for better prediction.

\textbf{** Inputs (a batch of i.i.d. data on 2-D plane: [x y]): **}

[[-0.146 -1.01 ] [ 0.224 -0.929] [ 0.89   0.397] [ 0.811 -0.56 ] [-0.873  0.475] [ 0.27   0.172] [ 0.08   0.261] [ 0.518  0.943] [-0.502 -0.788] [-0.069 -0.252]]

\textbf{** Current Model Descriptions: **}

You are designed to do binary classification. The input is a point on a 2-D plane [x y]; you need to output the class label, i.e., an integer in the set {0, 1}. If x < -0.75 or x > 0.92, then class label = 0. If y > 0.76, then class label = 0. If y < -0.41, then class label = 0. Otherwise, if x < 0.43, then class label = 1, else class label = 0.

\textbf{** The model predictions: **}

[[0] [0] [0] [0] [0] [1] [1] [0] [0] [1]]

\textbf{** The targets: **}

[[0] [0] [0] [0] [0] [1] [1] [0] [0] [1]]

If the model is doing well, you can keep using the current descriptions. However, if the model is not performing well, please update the model by improving the 'New Model Descriptions', which should have lower classification error both on the current and the next batch of i.i.d. data. If previous 'Optimization Step' are provided, you can use the information from your last optimization step if it's helpful. DON'T use symbolic representation for the model! Please think step by step and give your outputs strictly in the following format:

```

Reasoning:

[be explicit and verbose, improve the Current Model Descriptions by yourself; please show your work; note that you don't have access to computers]

New Model Descriptions:

[put your new decision rules here; MUST be concise and concrete; ****MUST PROVIDE THE EXACT VALUE OF THE PARAMETERS if the descriptions potentially involve unknown or learnable parameters!!!****]

'''

Please ONLY reply according to this format, don't give me any other words.

\textbf{--------------- Optimization Output ---------------}

Reasoning:

Analyzing the current model predictions and targets, I observe that the model is performing well, but there is still room for improvement. Upon visual inspection of the input data, I notice that the current x-coordinate threshold of 0.43 seems to be effective, and the y-coordinate thresholds of 0.76 and -0.41 are reasonable. However, I observe that some points with x-values around -0.5 and y-values around -0.7 are misclassified, which suggests that the model is not capturing the class boundary accurately in this region. I also notice that points with x-values around 0.2 and y-values around 0.2 are correctly classified, which suggests that the model is performing well in this region. Additionally, I observe that the current x-coordinate threshold of 0.92 seems to be too aggressive, as it misclassifies some points with x-values around 0.8.
New Model Descriptions:
If x < -0.75 or x > 0.88, then class label = 0. If y > 0.74, then class label = 0. If y < -0.39, then class label = 0. Otherwise, if x < 0.42, then class label = 1, else class label = 0.

\textbf{-------------------- Updated Theta --------------------}

If x < -0.75 or x > 0.88, then class label = 0. If y > 0.74, then class label = 0. If y < -0.39, then class label = 0. Otherwise, if x < 0.42, then class label = 1, else class label = 0.

\end{exmp}

%% file: log_twocircle_prior.tex
\begin{exmp}{Two Circles Classification with Prior -- Step 1}{tc-step 1}
\scriptsize

\textbf{Epoch 0 - Step 1}

\textbf{--------------- Forward Pass Prompt ---------------}

You are the model.

\textbf{** Model Descriptions: **}

You are designed to do binary classification. The input is a point on a 2-D plane [x y]; you need to output the class label, i.e., an integer in the set {0, 1}. The decision boundary is a circle.

\textbf{** Input: **}

[-0.146 -1.01 ]

Please give your output strictly in the following format:

```

Explanations: [Your step-by-step analyses and results]

Output:

[ONLY the integer class label; make necessary assumptions if needed]

'''

Please ONLY reply according to this format, don't give me any other words.

\textbf{--------------- Forward Pass Output ---------------}

Explanations:

The decision boundary is a circle, so we need to find the distance of the input point [-0.069, -0.252] from the origin (0,0).
The distance is sqrt((-0.069)\^{}2 + (-0.252)\^{}2) = 0.264.
Since the distance is less than 1, the point is inside the circle.

Output: 0

\ 

\textbf{Many outputs for the rest data points ...}

\ 

\textbf{--------------- Forward Pass Output ---------------}

Explanations:

The decision boundary is a circle.
A point [x y] is classified as 0 if it is inside the circle, and 1 if it is outside the circle.
The center of the circle is assumed to be at [0 0].
The radius of the circle is assumed to be 1 (a unit circle).
The distance of the point [0.89 0.397] from the center of the circle is sqrt(0.89\^{}2 + 0.397\^{}2) = 1.
Since the distance is equal to the radius, the point is on the circle.
In this case, it is assumed that the point is classified as 0.

Output: 0

\textbf{----------------- Training Accuracy ------------------}

0.5

\textbf{--------------- Optimization Prompt ---------------}

You are the optimizer for a model, your goal is to learn the best descriptions for the model. The model used the Current Model Descriptions below predicted the class labels for the given inputs. You are given the target labels, please optimize the Model Descriptions for better prediction.

\textbf{** Inputs (a batch of i.i.d. data on 2-D plane: [x y]): **}

[[-0.146 -1.01 ] [ 0.224 -0.929] [ 0.89   0.397] [ 0.811 -0.56 ] [-0.873  0.475] [ 0.27   0.172] [ 0.08   0.261] [ 0.518  0.943] [-0.502 -0.788] [-0.069 -0.252]]

\textbf{** Current Model Descriptions: **}

You are designed to do binary classification. The input is a point on a 2-D plane [x y]; you need to output the class label, i.e., an integer in the set {0, 1}. The decision boundary is a circle.

\textbf{** The model predictions: **}

[[0] [0] [0] [1] [0] [0] [0] [1] [0] [0]]

\textbf{** The targets: **}

[[0] [0] [0] [0] [0] [1] [1] [0] [0] [1]]

If the model is doing well, you can keep using the current descriptions. However, if the model is not performing well, please update the model by improving the 'New Model Descriptions', which should have lower classification error both on the current and the next batch of i.i.d. data. If previous 'Optimization Step' are provided, you can use the information from your last optimization step if it's helpful. DON'T use symbolic representation for the model! Please think step by step and give your outputs strictly in the following format:

```

Reasoning:

[be explicit and verbose, improve the Current Model Descriptions by yourself; please show your work; note that you don't have access to computers]

New Model Descriptions:

[put your new decision rules here; MUST be concise and concrete; ****MUST PROVIDE THE EXACT VALUE OF THE PARAMETERS if the descriptions potentially involve unknown or learnable parameters!!!****]

'''

Please ONLY reply according to this format, don't give me any other words.

\textbf{--------------- Optimization Output ---------------}

Reasoning:

The model is not performing well, as it misclassifies 4 out of 10 samples. The decision boundary is a circle, but the current model predictions do not seem to follow a circular pattern. To improve the model, I will try to find a better circle that separates the classes.
I will plot the data points on the 2-D plane and try to find a circle that separates the classes. After visualizing the data, I notice that the points (0.27, 0.172) and (0.08, 0.261) are misclassified, and they are close to each other. I will try to find a circle that passes through the midpoint of these two points and is centered at a point that is closer to the origin.
Let the center of the circle be (a, b) and the radius be r. The midpoint of the two points is (0.175, 0.2165). I will set the center of the circle to be (0.175, 0.2165). The radius can be found by calculating the distance from the center to one of the points, say (0.27, 0.172). The radius is approximately 0.1.

New Model Descriptions:

If (x - 0.175)\^{}2 + (y - 0.2165)\^{}2 <= 0.01, then output 1, else output 0.

\textbf{-------------------- Updated Theta --------------------}

If (x - 0.175)\^{}2 + (y - 0.2165)\^{}2 <= 0.01, then output 1, else output 0.

\end{exmp}
\newpage

\begin{exmp}{Two Circles Classification with Prior -- Step 2}{tc-step 2}
    
\scriptsize

\textbf{Epoch 0 - Step 2}

\textbf{--------------- Forward Pass Prompt ---------------}

You are the model.

\textbf{** Model Descriptions: **}

You are designed to do binary classification. The input is a point on a 2-D plane [x y]; you need to output the class label, i.e., an integer in the set {0, 1}. The decision boundary is a circle. If (x - 0.175)\^{}2 + (y - 0.2165)\^{}2 <= 0.01, then output 1, else output 0.

\textbf{** Input: **}

[-0.305  0.936]

Please give your output strictly in the following format:

```

Explanations: [Your step-by-step analyses and results]

Output:
[ONLY the integer class label; make necessary assumptions if needed]

'''

Please ONLY reply according to this format, don't give me any other words.

\textbf{--------------- Forward Pass Output ---------------}

Explanations:

Calculate the value of (x - 0.175)\^{}2 + (y - 0.2165)\^{}2 = (-0.051 - 0.175)\^{}2 + (-0.372 - 0.2165)\^{}2 = 0.1061. Since 0.1061 > 0.01, the point is outside the circle.

Output: 0

\ 

\textbf{Many outputs for the rest data points ...}

\ 

\textbf{--------------- Forward Pass Output ---------------}

Explanations:

The given input point is [-0.329, 0.046].
To determine the class label, we need to check if this point lies within the circle with the decision boundary.
The circle's center is at (0.175, 0.2165) and the radius is sqrt(0.01) = 0.1.
Let's calculate the distance of the point from the center:

Distance = sqrt((x - 0.175)\^{}2 + (y - 0.2165)\^{}2)

= sqrt((-0.329 - 0.175)\^{}2 + (0.046 - 0.2165)\^{}2)

= sqrt((-0.504)\^{}2 + (-0.1705)\^{}2)

= sqrt(0.253696 + 0.02914025)

= sqrt(0.28283625)

= 0.532

Since the distance (0.532) is greater than the radius (0.1), the point lies outside the circle.

Output: 0

\textbf{----------------- Training Accuracy ------------------}

0.4

\textbf{--------------- Optimization Prompt ---------------}

You are the optimizer for a model, your goal is to learn the best descriptions for the model. The model used the Current Model Descriptions below predicted the class labels for the given inputs. You are given the target labels, please optimize the Model Descriptions for better prediction.

\textbf{** Inputs (a batch of i.i.d. data on 2-D plane: [x y]): **}

[[-0.305  0.936] [ 0.298 -0.037] [ 0.396  0.931] [-0.735 -0.614] [-0.339  0.006] [-0.253 -0.015] [ 0.096 -0.248] [ 1.005  0.048] [-0.329  0.046] [-0.051 -0.372]]

\textbf{** Current Model Descriptions: **}

You are designed to do binary classification. The input is a point on a 2-D plane [x y]; you need to output the class label, i.e., an integer in the set {0, 1}. The decision boundary is a circle. If (x - 0.175)\^{}2 + (y - 0.2165)\^{}2 <= 0.01, then output 1, else output 0.

\textbf{** The model predictions: **}

[[0] [0] [0] [0] [0] [0] [0] [0] [0] [0]]

\textbf{** The targets: **}

[[0] [1] [0] [0] [1] [1] [1] [0] [1] [1]]

If the model is doing well, you can keep using the current descriptions. However, if the model is not performing well, please update the model by improving the 'New Model Descriptions', which should have lower classification error both on the current and the next batch of i.i.d. data. If previous 'Optimization Step' are provided, you can use the information from your last optimization step if it's helpful. DON'T use symbolic representation for the model! Please think step by step and give your outputs strictly in the following format:

```

Reasoning:

[be explicit and verbose, improve the Current Model Descriptions by yourself; please show your work; note that you don't have access to computers]

New Model Descriptions:

[put your new decision rules here; MUST be concise and concrete; ****MUST PROVIDE THE EXACT VALUE OF THE PARAMETERS if the descriptions potentially involve unknown or learnable parameters!!!****]

'''

Please ONLY reply according to this format, don't give me any other words.

\textbf{--------------- Optimization Output ---------------}

Reasoning:

The model is not performing well, as it misclassifies 7 out of 10 samples. The current decision boundary is a circle centered at (0.175, 0.2165) with a radius of 0.1, but it does not seem to separate the classes well.
I will re-examine the data and try to find a better circle that separates the classes. I notice that the points (0.298, -0.037) and (-0.339, 0.006) are misclassified, and they are close to each other. I will try to find a circle that passes through the midpoint of these two points and is centered at a point that is closer to the origin.
Let the center of the circle be (a, b) and the radius be r. The midpoint of the two points is (-0.0205, -0.0155). I will set the center of the circle to be (-0.0205, -0.0155). The radius can be found by calculating the distance from the center to one of the points, say (0.298, -0.037). The radius is approximately 0.35.

New Model Descriptions:

If (x + 0.0205)\^{}2 + (y + 0.0155)\^{}2 <= 0.1225, then output 1, else output 0.

\textbf{-------------------- Updated Theta --------------------}

If (x + 0.0205)\^{}2 + (y + 0.0155)\^{}2 <= 0.1225, then output 1, else output 0.

\end{exmp}
\newpage

\begin{exmp}{Two Circles Classification with Prior -- Step 3}{tc-step 3}
    
\scriptsize

\textbf{Epoch 0 - Step 3}

\textbf{--------------- Forward Pass Prompt ---------------}

You are the model.

\textbf{** Model Descriptions: **}

You are designed to do binary classification. The input is a point on a 2-D plane [x y]; you need to output the class label, i.e., an integer in the set {0, 1}. The decision boundary is a circle. If (x + 0.0205)\^{}2 + (y + 0.0155)\^{}2 <= 0.1225, then output 1, else output 0.

\textbf{** Input: **}

[0.275 0.163]

Please give your output strictly in the following format:

```

Explanations: [Your step-by-step analyses and results]

Output:

[ONLY the integer class label; make necessary assumptions if needed]

'''

Please ONLY reply according to this format, don't give me any other words.

\textbf{--------------- Forward Pass Output ---------------}

Explanations:

Calculate the value of the decision boundary equation: (0.403 + 0.0205)\^{}2 + (0.96 + 0.0155)\^{}2 = 0.4235\^{}2 + 0.9755\^{}2 = 0.179 + 0.952 = 1.131.
Since 1.131 > 0.1225, the point is outside the circle.

Output: 0

\ 

\textbf{Many outputs for the rest data points ...}

\ 

\textbf{--------------- Forward Pass Output ---------------}

Explanations:

We are given a point (0,-1.012) on the 2-D plane.
We need to check if the point lies inside the circle defined by the decision boundary.

Substituting the values in the equation: (0 + 0.0205)\^{}2 + (-1.012 + 0.0155)\^{}2 <= 0.1225.

Calculating the values: (0.0205)\^{}2 + (-0.9965)\^{}2 <= 0.1225.

Calculating the values: 0.00042025 + 0.99242025 <= 0.1225.

Calculating the values: 0.9928405 <= 0.1225.

The condition is not satisfied.

Output: 0

\textbf{----------------- Training Accuracy ------------------}

0.7

\textbf{--------------- Optimization Prompt ---------------}

You are the optimizer for a model, your goal is to learn the best descriptions for the model. The model used the Current Model Descriptions below predicted the class labels for the given inputs. You are given the target labels, please optimize the Model Descriptions for better prediction.

\textbf{** Inputs (a batch of i.i.d. data on 2-D plane: [x y]): **}

[[ 0.275  0.163] [ 0.    -1.012] [-1.065 -0.126] [ 0.133  0.22 ] [-0.317 -0.931] [ 0.403  0.96 ] [ 0.013  0.996] [ 0.147 -0.178] [-0.028  0.422] [-0.233 -0.186]]

\textbf{** Current Model Descriptions: **}

You are designed to do binary classification. The input is a point on a 2-D plane [x y]; you need to output the class label, i.e., an integer in the set {0, 1}. The decision boundary is a circle. If (x + 0.0205)\^{}2 + (y + 0.0155)\^{}2 <= 0.1225, then output 1, else output 0.

\textbf{** The model predictions: **}

[[1] [0] [0] [0] [0] [0] [0] [0] [0] [1]]

\textbf{** The targets: **}

[[1] [0] [0] [1] [0] [0] [0] [1] [1] [1]]

If the model is doing well, you can keep using the current descriptions. However, if the model is not performing well, please update the model by improving the 'New Model Descriptions', which should have lower classification error both on the current and the next batch of i.i.d. data. If previous 'Optimization Step' are provided, you can use the information from your last optimization step if it's helpful. DON'T use symbolic representation for the model! Please think step by step and give your outputs strictly in the following format:

```

Reasoning:

[be explicit and verbose, improve the Current Model Descriptions by yourself; please show your work; note that you don't have access to computers]

New Model Descriptions:

[put your new decision rules here; MUST be concise and concrete; ****MUST PROVIDE THE EXACT VALUE OF THE PARAMETERS if the descriptions potentially involve unknown or learnable parameters!!!****]

'''

Please ONLY reply according to this format, don't give me any other words.

\textbf{--------------- Optimization Output ---------------}

Reasoning:

The model is not performing well, as it misclassifies 4 out of 10 samples. The current decision boundary is a circle centered at (-0.0205, -0.0155) with a radius of approximately 0.35, but it does not seem to separate the classes well.
I will re-examine the data and try to find a better circle that separates the classes. I notice that the points (0.133, 0.22) and (0.147, -0.178) are misclassified, and they are close to each other. I will try to find a circle that passes through the midpoint of these two points and is centered at a point that is closer to the origin.
Let the center of the circle be (a, b) and the radius be r. The midpoint of the two points is (0.14, 0.021). I will set the center of the circle to be (0.14, 0.021). The radius can be found by calculating the distance from the center to one of the points, say (0.133, 0.22). The radius is approximately 0.235.

New Model Descriptions:

If (x - 0.14)\^{}2 + (y - 0.021)\^{}2 <= 0.055225, then output 1, else output 0.

\textbf{-------------------- Updated Theta --------------------}

If (x - 0.14)\^{}2 + (y - 0.021)\^{}2 <= 0.055225, then output 1, else output 0.

\end{exmp}
\newpage

\begin{exmp}{Two Circles Classification with Prior -- Step 4}{tc-step 4}
    
\scriptsize

\textbf{Epoch 0 - Step 4}

\textbf{--------------- Forward Pass Prompt ---------------}

You are the model.

\textbf{** Model Descriptions: **}

You are designed to do binary classification. The input is a point on a 2-D plane [x y]; you need to output the class label, i.e., an integer in the set {0, 1}. The decision boundary is a circle. If (x - 0.14)\^{}2 + (y - 0.021)\^{}2 <= 0.055225, then output 1, else output 0.

\textbf{** Input: **}

[-0.811  0.529]

Please give your output strictly in the following format:

```

Explanations: [Your step-by-step analyses and results]

Output:

[ONLY the integer class label; make necessary assumptions if needed]

'''

Please ONLY reply according to this format, don't give me any other words.

\textbf{--------------- Forward Pass Output ---------------}

Explanations:

Calculate the distance from the point to the center of the circle: (0.24 - 0.14)\^{}2 + (-0.039 - 0.021)\^{}2 = 0.0064 + 0.0036 = 0.0100

Since 0.0100 <= 0.055225, the point is inside the circle.

Output: 1

\ 

\textbf{Many outputs for the rest data points ...}

\ 

\textbf{--------------- Forward Pass Output ---------------}

Explanations:

We need to check if the point (-0.811, 0.529) lies inside the circle (x - 0.14)\^{}2 + (y - 0.021)\^{}2 <= 0.055225.

First, calculate the left side of the inequality: (-0.811 - 0.14)\^{}2 + (0.529 - 0.021)\^{}2 = 0.671\^{}2 + 0.508\^{}2 = 0.449441 + 0.258064 = 0.707505.

Since 0.707505 > 0.055225, the point (-0.811, 0.529) lies outside the circle.

Output: 0

\textbf{----------------- Training Accuracy ------------------}

0.7

\textbf{--------------- Optimization Prompt ---------------}

You are the optimizer for a model, your goal is to learn the best descriptions for the model. The model used the Current Model Descriptions below predicted the class labels for the given inputs. You are given the target labels, please optimize the Model Descriptions for better prediction.

\textbf{** Inputs (a batch of i.i.d. data on 2-D plane: [x y]): **}

[[-0.811  0.529] [-0.036  0.323] [-0.63   0.698] [-0.173 -0.246] [ 0.943 -0.297] [ 0.1   -0.288] [-0.995 -0.025] [ 0.666 -0.666] [ 0.24  -0.039] [ 0.542  0.887]]

\textbf{** Current Model Descriptions: **}

You are designed to do binary classification. The input is a point on a 2-D plane [x y]; you need to output the class label, i.e., an integer in the set {0, 1}. The decision boundary is a circle. If (x - 0.14)\^{}2 + (y - 0.021)\^{}2 <= 0.055225, then output 1, else output 0.

\textbf{** The model predictions: **}

[[0] [0] [0] [0] [0] [0] [0] [0] [1] [0]]

\textbf{** The targets: **}

[[0] [1] [0] [1] [0] [1] [0] [0] [1] [0]]

If the model is doing well, you can keep using the current descriptions. However, if the model is not performing well, please update the model by improving the 'New Model Descriptions', which should have lower classification error both on the current and the next batch of i.i.d. data. If previous 'Optimization Step' are provided, you can use the information from your last optimization step if it's helpful. DON'T use symbolic representation for the model! Please think step by step and give your outputs strictly in the following format:

```

Reasoning:

[be explicit and verbose, improve the Current Model Descriptions by yourself; please show your work; note that you don't have access to computers]

New Model Descriptions:

[put your new decision rules here; MUST be concise and concrete; ****MUST PROVIDE THE EXACT VALUE OF THE PARAMETERS if the descriptions potentially involve unknown or learnable parameters!!!****]

'''

Please ONLY reply according to this format, don't give me any other words.

\textbf{--------------- Optimization Output ---------------}

Reasoning:

The model is not performing well, as it misclassifies 6 out of 10 samples. The current decision boundary is a circle centered at (0.14, 0.021) with a radius of approximately 0.235, but it does not seem to separate the classes well.
I will re-examine the data and try to find a better circle that separates the classes. I notice that the points (-0.036, 0.323), (-0.173, -0.246), and (0.1, -0.288) are misclassified, and they are all close to the origin. I will try to find a circle that passes through the origin and is centered at a point that is closer to the origin.
Let the center of the circle be (a, b) and the radius be r. I will set the center of the circle to be (0, 0). The radius can be found by calculating the distance from the center to one of the points, say (-0.173, -0.246). The radius is approximately 0.292.

New Model Descriptions:

If (x)\^{}2 + (y)\^{}2 <= 0.085328, then output 1, else output 0.

\textbf{-------------------- Updated Theta --------------------}

If (x)\^{}2 + (y)\^{}2 <= 0.085328, then output 1, else output 0.

\end{exmp}
\newpage

\begin{exmp}{Two Circles Classification with Prior -- Step 5}{tc-step 5}
    
\scriptsize

\textbf{Epoch 0 - Step 5}

\textbf{--------------- Forward Pass Prompt ---------------}

You are the model.

\textbf{** Model Descriptions: **}

You are designed to do binary classification. The input is a point on a 2-D plane [x y]; you need to output the class label, i.e., an integer in the set {0, 1}. The decision boundary is a circle. If (x)\^{}2 + (y)\^{}2 <= 0.085328, then output 1, else output 0.

\textbf{** Input: **}

[-0.99  -0.325]

Please give your output strictly in the following format:

```

Explanations: [Your step-by-step analyses and results]

Output:

[ONLY the integer class label; make necessary assumptions if needed]

'''

Please ONLY reply according to this format, don't give me any other words.

\textbf{--------------- Forward Pass Output ---------------}

Explanations:

Given input [-1.059, 0.217], calculate (x)\^{}2 + (y)\^{}2 = (-1.059)\^{}2 + (0.217)\^{}2 = 1.123.

Since 1.123 > 0.085328, it does not satisfy the condition.

Output: 0

\ 

\textbf{Many outputs for the rest data points ...}

\ 

\textbf{--------------- Forward Pass Output ---------------}

Explanations:

We are given a point [-0.825, 0.438] on a 2-D plane.

We need to check if the point lies inside the circle or not.

The equation of the circle is (x)\^{}2 + (y)\^{}2 <= 0.085328.

Let's calculate the value of (x)\^{}2 + (y)\^{}2 for the given point.
(-0.825)\^{}2 + (0.438)\^{}2 = 0.680625 + 0.191844 = 0.872469.

Since 0.872469 > 0.085328, the point lies outside the circle.

Output: 0

\textbf{----------------- Training Accuracy ------------------}

0.7

\textbf{--------------- Optimization Prompt ---------------}

You are the optimizer for a model, your goal is to learn the best descriptions for the model. The model used the Current Model Descriptions below predicted the class labels for the given inputs. You are given the target labels, please optimize the Model Descriptions for better prediction.

\textbf{** Inputs (a batch of i.i.d. data on 2-D plane: [x y]): **}

[[-0.99  -0.325] [-0.825  0.438] [ 0.188  0.308] [-1.059  0.217] [-0.188 -0.184] [-0.322  0.084] [ 0.892  0.515] [-0.287  0.017] [ 0.167  0.235] [-0.573 -0.845]]

\textbf{** Current Model Descriptions: **}

You are designed to do binary classification. The input is a point on a 2-D plane [x y]; you need to output the class label, i.e., an integer in the set {0, 1}. The decision boundary is a circle. If (x)\^{}2 + (y)\^{}2 <= 0.085328, then output 1, else output 0.

\textbf{** The model predictions: **}

[[0] [0] [0] [0] [1] [0] [0] [1] [0] [0]]

\textbf{** The targets: **}

[[0] [0] [1] [0] [1] [1] [0] [1] [1] [0]]

If the model is doing well, you can keep using the current descriptions. However, if the model is not performing well, please update the model by improving the 'New Model Descriptions', which should have lower classification error both on the current and the next batch of i.i.d. data. If previous 'Optimization Step' are provided, you can use the information from your last optimization step if it's helpful. DON'T use symbolic representation for the model! Please think step by step and give your outputs strictly in the following format:

```

Reasoning:

[be explicit and verbose, improve the Current Model Descriptions by yourself; please show your work; note that you don't have access to computers]

New Model Descriptions:

[put your new decision rules here; MUST be concise and concrete; ****MUST PROVIDE THE EXACT VALUE OF THE PARAMETERS if the descriptions potentially involve unknown or learnable parameters!!!****]

'''

Please ONLY reply according to this format, don't give me any other words.

\textbf{--------------- Optimization Output ---------------}

Reasoning:

The model is not performing well, as it misclassifies 5 out of 10 samples. The current decision boundary is a circle centered at the origin with a radius of approximately 0.292, but it does not seem to separate the classes well.
I will re-examine the data and try to find a better circle that separates the classes. I notice that the points (0.188, 0.308), (-0.188, -0.184), and (0.167, 0.235) are misclassified, and they are all close to the origin. I will try to find a circle that passes through the origin and is centered at a point that is closer to the origin.
Let the center of the circle be (a, b) and the radius be r. I will set the center of the circle to be (0, 0). The radius can be found by calculating the distance from the center to one of the points, say (0.167, 0.235). The radius is approximately 0.283.
However, this radius is not sufficient to separate the classes well. I will try to find a better radius by examining the other misclassified points. The point (-0.322, 0.084) is also misclassified, and it is closer to the origin than the previous point. The distance from the center to this point is approximately 0.334. I will use this distance as the radius.

New Model Descriptions:

If (x)\^{}2 + (y)\^{}2 <= 0.111556, then output 1, else output 0.

\textbf{-------------------- Updated Theta --------------------}

If (x)\^{}2 + (y)\^{}2 <= 0.111556, then output 1, else output 0.

\end{exmp}
\newpage

\begin{exmp}{Two Circles Classification with Prior -- Step 40}{tc-step 40}
\scriptsize

\textbf{Epoch 3 - Step 10}

\textbf{--------------- Forward Pass Prompt ---------------}

You are the model.

\textbf{** Model Descriptions: **}

You are designed to do binary classification. The input is a point on a 2-D plane [x y]; you need to output the class label, i.e., an integer in the set {0, 1}. The decision boundary is a circle. If (x)\^{}2 + (y)\^{}2 <= 0.45, then output 1, else output 0.

\textbf{** Input: **}

[-0.101  0.309]

Please give your output strictly in the following format:

```

Explanations: [Your step-by-step analyses and results]

Output:

[ONLY the integer class label; make necessary assumptions if needed]

```

Please ONLY reply according to this format, don't give me any other words.

\textbf{--------------- Forward Pass Output ---------------}

Explanations:

Given point [-1.011 -0.165], calculate the sum of squares: (-1.011)\^{}2 + (-0.165)\^{}2 = 1.044721 > 0.45.

Output: 0

\ 

\textbf{Many outputs for the rest data points ...}

\ 

\textbf{--------------- Forward Pass Output ---------------}

Explanations:

The given input point is [-0.024, -0.333].

To determine the class label, we need to check if the point is inside the circle or not.

The equation of the circle is (x)\^{}2 + (y)\^{}2 <= 0.45.

Let's calculate the value of this equation for the given point:
(-0.024)\^{}2 + (-0.333)\^{}2 = 0.011056 + 0.110889 = 0.121945 <= 0.45.

Since the result is true, the point lies inside the circle.

Output: 1

\textbf{----------------- Training Accuracy ------------------}

0.9

\textbf{--------------- Optimization Prompt ---------------}

You are the optimizer for a model, your goal is to learn the best descriptions for the model. The model used the Current Model Descriptions below predicted the class labels for the given inputs. You are given the target labels, please optimize the Model Descriptions for better prediction.

\textbf{** Inputs (a batch of i.i.d. data on 2-D plane: [x y]): **}

[[-0.101  0.309] [-1.011 -0.165] [ 0.161 -0.229] [ 1.02  -0.061] [ 0.225  0.076] [-0.024 -0.333] [-0.655 -0.649] [-1.058  0.15 ] [ 0.366 -0.098] [ 0.919  0.08 ]]

\textbf{** Current Model Descriptions: **}

You are designed to do binary classification. The input is a point on a 2-D plane [x y]; you need to output the class label, i.e., an integer in the set {0, 1}. The decision boundary is a circle. If (x)\^{}2 + (y)\^{}2 <= 0.45, then output 1, else output 0.

\textbf{** The model predictions: **}

[[1] [0] [1] [0] [1] [1] [1] [0] [1] [0]]

\textbf{** The targets: **}

[[1] [0] [1] [0] [1] [1] [0] [0] [1] [0]]

If the model is doing well, you can keep using the current descriptions. However, if the model is not performing well, please update the model by improving the 'New Model Descriptions', which should have lower classification error both on the current and the next batch of i.i.d. data. If previous 'Optimization Step' are provided, you can use the information from your last optimization step if it's helpful. DON'T use symbolic representation for the model! Please think step by step and give your outputs strictly in the following format:

```

Reasoning:

[be explicit and verbose, improve the Current Model Descriptions by yourself; please show your work; note that you don't have access to computers]

New Model Descriptions:

[put your new decision rules here; MUST be concise and concrete; ****MUST PROVIDE THE EXACT VALUE OF THE PARAMETERS if the descriptions potentially involve unknown or learnable parameters!!!****]

'''

Please ONLY reply according to this format, don't give me any other words.

\textbf{--------------- Optimization Output ---------------}

Reasoning:

The model is not performing well, as it misclassifies the 7th sample. The current decision boundary is a circle centered at the origin with a radius of 0.45. The 7th sample is far from the origin, but the model still classifies it as 1. This means the radius of the circle is too small. I will increase the radius to reduce the misclassification error.

New Model Descriptions:

If (x)\^{}2 + (y)\^{}2 <= 0.65, then output 1, else output 0.

\textbf{-------------------- Updated Theta --------------------}

If (x)\^{}2 + (y)\^{}2 <= 0.65, then output 1, else output 0.

\end{exmp}

\begin{exmp}{Two Circles Classification with Prior -- Step 41}{tc-step 41}
    
\scriptsize

\textbf{Epoch 4 - Step 1}

\textbf{--------------- Forward Pass Prompt ---------------}

You are the model.

\textbf{** Model Descriptions: **}

You are designed to do binary classification. The input is a point on a 2-D plane [x y]; you need to output the class label, i.e., an integer in the set {0, 1}. The decision boundary is a circle. If (x)\^{}2 + (y)\^{}2 <= 0.65, then output 1, else output 0.

\textbf{** Input: **}

[-0.146 -1.01 ]

Please give your output strictly in the following format:

```

Explanations: [Your step-by-step analyses and results]

Output:

[ONLY the integer class label; make necessary assumptions if needed]

'''

Please ONLY reply according to this format, don't give me any other words.

\textbf{--------------- Forward Pass Output ---------------}

Explanations:

First, calculate the value of (x)\^{}2 + (y)\^{}2: (-0.146)\^{}2 + (-1.01)\^{}2 = 0.021236 + 1.0201 = 1.041336.

Since 1.041336 > 0.65, the point is outside the circle.

Output: 0

\ 

\textbf{Many outputs for the rest data points ...}

\ 

\textbf{--------------- Forward Pass Output ---------------}

Explanations:

The input point is [-0.502 -0.788].

To determine the class label, we need to check if the point satisfies the condition (x)\^{}2 + (y)\^{}2 <= 0.65.

Here, x = -0.502 and y = -0.788.
(x)\^{}2 = (-0.502)\^{}2 = 0.252004, and (y)\^{}2 = (-0.788)\^{}2 = 0.621984.

Now, (x)\^{}2 + (y)\^{}2 = 0.252004 + 0.621984 = 0.874988.

Since 0.874988 <= 0.65 is False, the point lies outside the circle.

Output: 0

\textbf{----------------- Training Accuracy ------------------}

1.0

\textbf{--------------- Optimization Prompt ---------------}

You are the optimizer for a model, your goal is to learn the best descriptions for the model. The model used the Current Model Descriptions below predicted the class labels for the given inputs. You are given the target labels, please optimize the Model Descriptions for better prediction.

\textbf{** Inputs (a batch of i.i.d. data on 2-D plane: [x y]): **}

[[-0.146 -1.01 ] [ 0.224 -0.929] [ 0.89   0.397] [ 0.811 -0.56 ] [-0.873  0.475] [ 0.27   0.172] [ 0.08   0.261] [ 0.518  0.943] [-0.502 -0.788] [-0.069 -0.252]]

\textbf{** Current Model Descriptions: **}

You are designed to do binary classification. The input is a point on a 2-D plane [x y]; you need to output the class label, i.e., an integer in the set {0, 1}. The decision boundary is a circle. If (x)\^{}2 + (y)\^{}2 <= 0.65, then output 1, else output 0.

\textbf{** The model predictions: **}

[[0] [0] [0] [0] [0] [1] [1] [0] [0] [1]]

\textbf{** The targets: **}

[[0] [0] [0] [0] [0] [1] [1] [0] [0] [1]]

If the model is doing well, you can keep using the current descriptions. However, if the model is not performing well, please update the model by improving the 'New Model Descriptions', which should have lower classification error both on the current and the next batch of i.i.d. data. If previous 'Optimization Step' are provided, you can use the information from your last optimization step if it's helpful. DON'T use symbolic representation for the model! Please think step by step and give your outputs strictly in the following format:

```

Reasoning:

[be explicit and verbose, improve the Current Model Descriptions by yourself; please show your work; note that you don't have access to computers]

New Model Descriptions:

[put your new decision rules here; MUST be concise and concrete; ****MUST PROVIDE THE EXACT VALUE OF THE PARAMETERS if the descriptions potentially involve unknown or learnable parameters!!!****]

'''

Please ONLY reply according to this format, don't give me any other words.

\textbf{--------------- Optimization Output ---------------}

Reasoning:

The model is performing well, as it correctly classifies all samples. The decision boundary is a circle centered at the origin with a radius of 0.65, which seems to be a good fit for the data. No changes are needed.

New Model Descriptions:

If (x)\^{}2 + (y)\^{}2 <= 0.65, then output 1, else output 0.

\textbf{-------------------- Updated Theta --------------------}

If (x)\^{}2 + (y)\^{}2 <= 0.65, then output 1, else output 0.

\end{exmp}

%% file: log_txt.tex
\begin{exmp}{Text Classification -- Step 1}{text-step 1}
\scriptsize
\textbf{Epoch 0 - Step 1}

\textbf{--------------- Forward Pass Prompt ---------------}

You are the model.

\textbf{** Model Descriptions: **}

You are designed to do binary classification. The input is a term; you need to output the class label, i.e., an integer in the set {0, 1}.

\textbf{** Input: **}

fatherhood

Please give your output strictly in the following format:

```

Explanations: [Your step-by-step analyses and results]

Output:

[ONLY the integer class label; make necessary assumptions if needed]

'''

Please ONLY reply according to this format, don't give me any other words.

\textbf{--------------- Forward Pass Output ---------------}

Explanations: Since the input is a name, "Margaret", I will assume it's a female name. Therefore, I will classify it as a female name.

Output: 1

\ 

\textbf{Many outputs for the rest data points ...}

\ 

\textbf{--------------- Forward Pass Output ---------------}

Explanations: The term "tomboy" is a noun that refers to a girl who exhibits characteristics or behaviors considered typical of a boy. It does not have any strong connotations or associations that would clearly link it to a specific class label. However, based on the context and common usage of the term, I will make an assumption that the class label 1 represents a more masculine or boy-like characteristic, and 0 represents a more feminine or girl-like characteristic. Given this assumption, I would classify "tomboy" as 1.

Output: 1

\textbf{------------------ Training Accuracy ------------------}

0.5

\textbf{--------------- Optimization Prompt ---------------}

You are the optimizer for a model, your goal is to learn the best descriptions for the model. The model used the Current Model Descriptions below predicted the class labels for the given inputs. You are given the target labels, please optimize the Model Descriptions for better prediction.

\textbf{** Inputs (a batch of i.i.d. text): **}

['fatherhood', 'grandfather', 'tomboy', 'Shawn', 'Erma', 'Ada', 'Margaret', 'Caleb', 'Earl', 'Faye']

\textbf{** Current Model Descriptions: **}

You are designed to do binary classification. The input is a term; you need to output the class label, i.e., an integer in the set {0, 1}.

\textbf{** The model predictions: **}

[[1] [1] [1] [1] [1] [1] [1] [0] [1] [0]]

\textbf{** The targets: **}

[[0] [0] [1] [0] [1] [1] [1] [0] [0] [1]]

If the model is doing well, you can keep using the current descriptions. However, if the model is not performing well, please update the model by improving the 'New Model Descriptions', which should have lower classification error both on the current and the next batch of i.i.d. data. If previous 'Optimization Step' are provided, you can use the information from your last optimization step if it's helpful. Please think step by step and give your outputs strictly in the following format:

```

Reasoning:

[be explicit and verbose, improve the Current Model Descriptions by yourself; please show your work; note that you don't have access to computers]

New Model Descriptions:

[put your new decision rules here]

'''

Please ONLY reply according to this format, don't give me any other words.

\textbf{--------------- Optimization Output ---------------}

Reasoning:

Upon examining the current model predictions and target labels, I notice that the model is not performing well, with 6 out of 10 predictions incorrect. The model seems to be biased towards predicting class label 1.
I observe that the input terms can be categorized into two groups: male and female names, and non-name terms related to family relationships. The target labels suggest that the class label 0 corresponds to male names and non-name terms, while class label 1 corresponds to female names.
To improve the model, I will create new decision rules that take into account these categories. I will update the descriptions to include more specific patterns and rules to differentiate between male and female names, as well as non-name terms.

New Model Descriptions:

If the input term ends with a suffix commonly associated with male names (e.g., -son, -er), predict class label 0.
If the input term is a non-name term related to family relationships (e.g., fatherhood, grandfather), predict class label 0.
If the input term is a female name (e.g., Erma, Ada, Faye), predict class label 1.
Otherwise, predict class label 0.

\textbf{-------------------- Updated Theta --------------------}

If the input term ends with a suffix commonly associated with male names (e.g., -son, -er), predict class label 0.
If the input term is a non-name term related to family relationships (e.g., fatherhood, grandfather), predict class label 0.
If the input term is a female name (e.g., Erma, Ada, Faye), predict class label 1.
Otherwise, predict class label 0.

\end{exmp}
\newpage

\begin{exmp}{Text Classification -- Step 2}{text-step 2}
\scriptsize

\textbf{Epoch 0 - Step 2}

\textbf{--------------- Forward Pass Prompt ---------------}

You are the model.

\textbf{** Model Descriptions: **}

You are designed to do binary classification. The input is a term; you need to output the class label, i.e., an integer in the set {0, 1}. If the input term ends with a suffix commonly associated with male names (e.g., -son, -er), predict class label 0.
If the input term is a non-name term related to family relationships (e.g., fatherhood, grandfather), predict class label 0.
If the input term is a female name (e.g., Erma, Ada, Faye), predict class label 1.
Otherwise, predict class label 0.

\textbf{** Input: **}

Lydia

Please give your output strictly in the following format:

```

Explanations: [Your step-by-step analyses and results]

Output:

[ONLY the integer class label; make necessary assumptions if needed]

'''

Please ONLY reply according to this format, don't give me any other words.

\textbf{--------------- Forward Pass Output ---------------}

Explanations: The input term "Marlene" is a female name.

Output: 1

\ 

\textbf{Many outputs for the rest data points ...}

\ 

\textbf{--------------- Forward Pass Output ---------------}

Explanations: The input term is "paternal grandma". The term "grandma" is a non-name term related to family relationships, and "paternal" is an adjective that means related to father. Therefore, according to the rules, the class label is 0.

Output: 0

\textbf{------------------ Training Accuracy ------------------}

0.9

\textbf{--------------- Optimization Prompt ---------------}

You are the optimizer for a model, your goal is to learn the best descriptions for the model. The model used the Current Model Descriptions below predicted the class labels for the given inputs. You are given the target labels, please optimize the Model Descriptions for better prediction.

\textbf{** Inputs (a batch of i.i.d. text): **}

['Lydia', 'Steven', 'Ida', 'Lisa', 'Allan', 'Jane', 'paternal grandma', 'paternal grandad', 'Marlene', 'Darrell']

\textbf{** Current Model Descriptions: **}

You are designed to do binary classification. The input is a term; you need to output the class label, i.e., an integer in the set {0, 1}. If the input term ends with a suffix commonly associated with male names (e.g., -son, -er), predict class label 0.
If the input term is a non-name term related to family relationships (e.g., fatherhood, grandfather), predict class label 0.
If the input term is a female name (e.g., Erma, Ada, Faye), predict class label 1.
Otherwise, predict class label 0.

\textbf{** The model predictions: **}

[[1] [0] [1] [1] [0] [1] [0] [0] [1] [0]]

\textbf{** The targets: **}

[[1] [0] [1] [1] [0] [1] [1] [0] [1] [0]]

If the model is doing well, you can keep using the current descriptions. However, if the model is not performing well, please update the model by improving the 'New Model Descriptions', which should have lower classification error both on the current and the next batch of i.i.d. data. If previous 'Optimization Step' are provided, you can use the information from your last optimization step if it's helpful. Please think step by step and give your outputs strictly in the following format:

```

Reasoning:

[be explicit and verbose, improve the Current Model Descriptions by yourself; please show your work; note that you don't have access to computers]

New Model Descriptions:

[put your new decision rules here]

'''

Please ONLY reply according to this format, don't give me any other words.

\textbf{--------------- Optimization Output ---------------}

Reasoning:

Upon examining the current model predictions and target labels, I notice that the model is performing reasonably well, but there are still some errors. The model correctly predicts 8 out of 10 labels, with errors occurring for the terms "paternal grandma" and "paternal grandad".
I realize that the current descriptions do not account for non-name terms that contain keywords related to family relationships, but are not exact matches (e.g., "paternal grandma" vs. "grandma").
To improve the model, I will update the descriptions to include more specific rules for non-name terms that contain keywords related to family relationships. I will also refine the rules for female names to reduce the likelihood of misclassification.

New Model Descriptions:

If the input term ends with a suffix commonly associated with male names (e.g., -son, -er), predict class label 0.
If the input term contains keywords related to family relationships (e.g., "father", "mother", "grand", "paternal", "maternal"), predict class label 0.
If the input term is a female name (e.g., Erma, Ada, Faye, Lydia, Ida, Jane, Marlene), predict class label 1.
If the input term is a name that does not clearly indicate gender (e.g., Leslie, Jordan), predict class label 0.
Otherwise, predict class label 0.

\textbf{-------------------- Updated Theta --------------------}

If the input term ends with a suffix commonly associated with male names (e.g., -son, -er), predict class label 0.
If the input term contains keywords related to family relationships (e.g., "father", "mother", "grand", "paternal", "maternal"), predict class label 0.
If the input term is a female name (e.g., Erma, Ada, Faye, Lydia, Ida, Jane, Marlene), predict class label 1.
If the input term is a name that does not clearly indicate gender (e.g., Leslie, Jordan), predict class label 0.
Otherwise, predict class label 0.

\end{exmp}
\newpage
\ 
\vspace{-17mm}
\begin{exmp}{Text Classification -- Step 3}{text-step 3}
    
\scriptsize

\textbf{Epoch 0 - Step 3}

\textbf{--------------- Forward Pass Prompt ---------------}

You are the model.

\textbf{** Model Descriptions: **}

You are designed to do binary classification. The input is a term; you need to output the class label, i.e., an integer in the set {0, 1}. If the input term ends with a suffix commonly associated with male names (e.g., -son, -er), predict class label 0.
If the input term contains keywords related to family relationships (e.g., "father", "mother", "grand", "paternal", "maternal"), predict class label 0.
If the input term is a female name (e.g., Erma, Ada, Faye, Lydia, Ida, Jane, Marlene), predict class label 1.
If the input term is a name that does not clearly indicate gender (e.g., Leslie, Jordan), predict class label 0.
Otherwise, predict class label 0.

\textbf{** Input: **}

Marissa

Please give your output strictly in the following format:

```

Explanations: [Your step-by-step analyses and results]

Output:

[ONLY the integer class label; make necessary assumptions if needed]

'''

Please ONLY reply according to this format, don't give me any other words.

\textbf{--------------- Forward Pass Output ---------------}

Explanations: The input term is "Maria", which is a female name.

Output: 1

\ 

\textbf{Many outputs for the rest data points ...}

\ 

\textbf{--------------- Forward Pass Output ---------------}

Explanations:

1. Check if the term ends with a suffix commonly associated with male names: Julian does not end with a suffix like -son or -er.

2. Check if the term contains keywords related to family relationships: Julian does not contain keywords like "father", "mother", "grand", "paternal", or "maternal".

3. Check if the term is a female name: Julian is not a female name.

4. Check if the term is a name that does not clearly indicate gender: Julian is a male name.

Output: 0

\textbf{------------------ Training Accuracy ------------------}

0.9

\textbf{--------------- Optimization Prompt ---------------}

You are the optimizer for a model, your goal is to learn the best descriptions for the model. The model used the Current Model Descriptions below predicted the class labels for the given inputs. You are given the target labels, please optimize the Model Descriptions for better prediction.

\textbf{** Inputs (a batch of i.i.d. text): **}

['Marissa', 'Wallace', 'Julian', 'Adeline', 'Jacob', 'Dylan', 'William', 'Anne', 'Maria', 'sorority']

\textbf{** Current Model Descriptions: **}

You are designed to do binary classification. The input is a term; you need to output the class label, i.e., an integer in the set {0, 1}. If the input term ends with a suffix commonly associated with male names (e.g., -son, -er), predict class label 0.
If the input term contains keywords related to family relationships (e.g., "father", "mother", "grand", "paternal", "maternal"), predict class label 0.
If the input term is a female name (e.g., Erma, Ada, Faye, Lydia, Ida, Jane, Marlene), predict class label 1.
If the input term is a name that does not clearly indicate gender (e.g., Leslie, Jordan), predict class label 0.
Otherwise, predict class label 0.

\textbf{** The model predictions: **}

[[1] [0] [0] [1] [0] [0] [0] [1] [1] [0]]

\textbf{** The targets: **}

[[1] [0] [0] [1] [0] [0] [0] [1] [1] [1]]

If the model is doing well, you can keep using the current descriptions. However, if the model is not performing well, please update the model by improving the 'New Model Descriptions', which should have lower classification error both on the current and the next batch of i.i.d. data. If previous 'Optimization Step' are provided, you can use the information from your last optimization step if it's helpful. Please think step by step and give your outputs strictly in the following format:

```

Reasoning:

[be explicit and verbose, improve the Current Model Descriptions by yourself; please show your work; note that you don't have access to computers]

New Model Descriptions:

[put your new decision rules here]

'''

Please ONLY reply according to this format, don't give me any other words.

\textbf{--------------- Optimization Output ---------------}

Reasoning:

Upon examining the current model predictions and target labels, I notice that the model is performing well, but there is still one error occurring for the term "sorority".
I realize that the current descriptions do not account for non-name terms that are clearly related to females or males, but are not names.
To improve the model, I will update the descriptions to include more specific rules for non-name terms that are clearly related to females or males. I will also refine the rules for female names to reduce the likelihood of misclassification.

New Model Descriptions:

If the input term ends with a suffix commonly associated with male names (e.g., -son, -er), predict class label 0.
If the input term contains keywords related to family relationships (e.g., "father", "mother", "grand", "paternal", "maternal"), predict class label 0.
If the input term is a female name (e.g., Erma, Ada, Faye, Lydia, Ida, Jane, Marlene, Marissa, Adeline, Anne, Maria), predict class label 1.
If the input term is a name that does not clearly indicate gender (e.g., Leslie, Jordan), predict class label 0.
If the input term is a non-name term clearly related to females (e.g., "sorority", "girl"), predict class label 1.
Otherwise, predict class label 0.

\textbf{-------------------- Updated Theta --------------------}

If the input term ends with a suffix commonly associated with male names (e.g., -son, -er), predict class label 0.
If the input term contains keywords related to family relationships (e.g., "father", "mother", "grand", "paternal", "maternal"), predict class label 0.
If the input term is a female name (e.g., Erma, Ada, Faye, Lydia, Ida, Jane, Marlene, Marissa, Adeline, Anne, Maria), predict class label 1.
If the input term is a name that does not clearly indicate gender (e.g., Leslie, Jordan), predict class label 0.
If the input term is a non-name term clearly related to females (e.g., "sorority", "girl"), predict class label 1.
Otherwise, predict class label 0.

\end{exmp}
\newpage

\begin{exmp}{Text Classification -- Step 4}{text-step 4}
    
\scriptsize
\textbf{Epoch 0 - Step 4}

\textbf{--------------- Forward Pass Prompt ---------------}

You are the model.

\textbf{** Model Descriptions: **}

You are designed to do binary classification. The input is a term; you need to output the class label, i.e., an integer in the set {0, 1}. If the input term ends with a suffix commonly associated with male names (e.g., -son, -er), predict class label 0.
If the input term contains keywords related to family relationships (e.g., "father", "mother", "grand", "paternal", "maternal"), predict class label 0.
If the input term is a female name (e.g., Erma, Ada, Faye, Lydia, Ida, Jane, Marlene, Marissa, Adeline, Anne, Maria), predict class label 1.
If the input term is a name that does not clearly indicate gender (e.g., Leslie, Jordan), predict class label 0.
If the input term is a non-name term clearly related to females (e.g., "sorority", "girl"), predict class label 1.
Otherwise, predict class label 0.

\textbf{** Input: **}

Natasha

Please give your output strictly in the following format:

```

Explanations: [Your step-by-step analyses and results]

Output:

[ONLY the integer class label; make necessary assumptions if needed]

'''

Please ONLY reply according to this format, don't give me any other words.

\textbf{--------------- Forward Pass Output ---------------}

Explanations: Veronica is a female name, so predict class label 1.

Output: 1

\ 

\textbf{Many outputs for the rest data points ...}

\ 

\textbf{--------------- Forward Pass Output ---------------}

Explanations: The input term "Antonio" is a male name, and it does not contain any keywords related to family relationships. It does not end with a suffix commonly associated with male names, but it is clearly a male name.

Output: 0

\textbf{------------------ Training Accuracy ------------------}

1.0

\textbf{--------------- Optimization Prompt ---------------}

You are the optimizer for a model, your goal is to learn the best descriptions for the model. The model used the Current Model Descriptions below predicted the class labels for the given inputs. You are given the target labels, please optimize the Model Descriptions for better prediction.

\textbf{** Inputs (a batch of i.i.d. text): **}

['Natasha', 'Veronica', 'Dawn', 'Mamie', 'Darren', 'Monique', 'gynecological', 'Antonio', 'Eula', 'Henry']

\textbf{** Current Model Descriptions: **}

You are designed to do binary classification. The input is a term; you need to output the class label, i.e., an integer in the set {0, 1}. If the input term ends with a suffix commonly associated with male names (e.g., -son, -er), predict class label 0.
If the input term contains keywords related to family relationships (e.g., "father", "mother", "grand", "paternal", "maternal"), predict class label 0.
If the input term is a female name (e.g., Erma, Ada, Faye, Lydia, Ida, Jane, Marlene, Marissa, Adeline, Anne, Maria), predict class label 1.
If the input term is a name that does not clearly indicate gender (e.g., Leslie, Jordan), predict class label 0.
If the input term is a non-name term clearly related to females (e.g., "sorority", "girl"), predict class label 1.
Otherwise, predict class label 0.

\textbf{** The model predictions: **}

[[1] [1] [1] [1] [0] [1] [1] [0] [1] [0]]

\textbf{** The targets: **}

[[1] [1] [1] [1] [0] [1] [1] [0] [1] [0]]

If the model is doing well, you can keep using the current descriptions. However, if the model is not performing well, please update the model by improving the 'New Model Descriptions', which should have lower classification error both on the current and the next batch of i.i.d. data. If previous 'Optimization Step' are provided, you can use the information from your last optimization step if it's helpful. Please think step by step and give your outputs strictly in the following format:

```

Reasoning:

[be explicit and verbose, improve the Current Model Descriptions by yourself; please show your work; note that you don't have access to computers]

New Model Descriptions:

[put your new decision rules here]

'''

Please ONLY reply according to this format, don't give me any other words.

\textbf{--------------- Optimization Output ---------------}

Reasoning:

Upon examining the current model predictions and target labels, I notice that the model is performing well, but there is no need for further improvement since all predictions match the target labels.
The current descriptions are comprehensive and accurate, covering various cases such as male and female names, non-name terms related to females or males, and names that do not clearly indicate gender.
Since the model is already performing perfectly, I will not make any changes to the descriptions.

New Model Descriptions:

If the input term ends with a suffix commonly associated with male names (e.g., -son, -er), predict class label 0.
If the input term contains keywords related to family relationships (e.g., "father", "mother", "grand", "paternal", "maternal"), predict class label 0.
If the input term is a female name (e.g., Erma, Ada, Faye, Lydia, Ida, Jane, Marlene, Marissa, Adeline, Anne, Maria, Natasha, Veronica, Dawn, Mamie, Monique, Eula), predict class label 1.
If the input term is a name that does not clearly indicate gender (e.g., Leslie, Jordan), predict class label 0.
If the input term is a non-name term clearly related to females (e.g., "sorority", "girl", "gynecological"), predict class label 1.
Otherwise, predict class label 0.

\textbf{-------------------- Updated Theta --------------------}

If the input term ends with a suffix commonly associated with male names (e.g., -son, -er), predict class label 0.
If the input term contains keywords related to family relationships (e.g., "father", "mother", "grand", "paternal", "maternal"), predict class label 0.
If the input term is a female name (e.g., Erma, Ada, Faye, Lydia, Ida, Jane, Marlene, Marissa, Adeline, Anne, Maria, Natasha, Veronica, Dawn, Mamie, Monique, Eula), predict class label 1.
If the input term is a name that does not clearly indicate gender (e.g., Leslie, Jordan), predict class label 0.
If the input term is a non-name term clearly related to females (e.g., "sorority", "girl", "gynecological"), predict class label 1.
Otherwise, predict class label 0.

\end{exmp}
\newpage

\begin{exmp}{Text Classification -- Step 5}{text-step 5}
\tiny

\textbf{Epoch 0 - Step 5}

\textbf{--------------- Forward Pass Prompt ---------------}

You are the model.

\textbf{** Model Descriptions: **}

You are designed to do binary classification. The input is a term; you need to output the class label, i.e., an integer in the set {0, 1}. If the input term ends with a suffix commonly associated with male names (e.g., -son, -er), predict class label 0.
If the input term contains keywords related to family relationships (e.g., "father", "mother", "grand", "paternal", "maternal"), predict class label 0.
If the input term is a female name (e.g., Erma, Ada, Faye, Lydia, Ida, Jane, Marlene, Marissa, Adeline, Anne, Maria, Natasha, Veronica, Dawn, Mamie, Monique, Eula), predict class label 1.
If the input term is a name that does not clearly indicate gender (e.g., Leslie, Jordan), predict class label 0.
If the input term is a non-name term clearly related to females (e.g., "sorority", "girl", "gynecological"), predict class label 1.
Otherwise, predict class label 0.

\textbf{** Input: **}

Danny

Please give your output strictly in the following format:

```

Explanations: [Your step-by-step analyses and results]

Output:

[ONLY the integer class label; make necessary assumptions if needed]

'''

Please ONLY reply according to this format, don't give me any other words.

\textbf{--------------- Forward Pass Output ---------------}

Explanations: The input term "Betty" is a female name.

Output: 1

\ 

\textbf{Many outputs for the rest data points ...}

\ 

\textbf{--------------- Forward Pass Output ---------------}

Explanations: The input term is "brethren". It does not end with a suffix commonly associated with male names, nor does it contain keywords related to family relationships. It is not a female name, and it does not clearly indicate gender. However, it is a term related to a group of males (brothers), so it is more associated with males than females.

Output: 0

\textbf{------------------ Training Accuracy ------------------}

1.0

\textbf{--------------- Optimization Prompt ---------------}

You are the optimizer for a model, your goal is to learn the best descriptions for the model. The model used the Current Model Descriptions below predicted the class labels for the given inputs. You are given the target labels, please optimize the Model Descriptions for better prediction.

\textbf{** Inputs (a batch of i.i.d. text): **}

['Danny', 'Betty', 'Calvin', 'Janet', 'Allison', 'missis', 'Brandon', 'Carl', 'Ben', 'brethren']

\textbf{** Current Model Descriptions: **}

You are designed to do binary classification. The input is a term; you need to output the class label, i.e., an integer in the set {0, 1}. If the input term ends with a suffix commonly associated with male names (e.g., -son, -er), predict class label 0.
If the input term contains keywords related to family relationships (e.g., "father", "mother", "grand", "paternal", "maternal"), predict class label 0.
If the input term is a female name (e.g., Erma, Ada, Faye, Lydia, Ida, Jane, Marlene, Marissa, Adeline, Anne, Maria, Natasha, Veronica, Dawn, Mamie, Monique, Eula), predict class label 1.
If the input term is a name that does not clearly indicate gender (e.g., Leslie, Jordan), predict class label 0.
If the input term is a non-name term clearly related to females (e.g., "sorority", "girl", "gynecological"), predict class label 1.
Otherwise, predict class label 0.

\textbf{** The model predictions: **}

[[0] [1] [0] [1] [1] [1] [0] [0] [0] [0]]

\textbf{** The targets: **}

[[0] [1] [0] [1] [1] [1] [0] [0] [0] [0]]

If the model is doing well, you can keep using the current descriptions. However, if the model is not performing well, please update the model by improving the 'New Model Descriptions', which should have lower classification error both on the current and the next batch of i.i.d. data. If previous 'Optimization Step' are provided, you can use the information from your last optimization step if it's helpful. Please think step by step and give your outputs strictly in the following format:

```

Reasoning:

[be explicit and verbose, improve the Current Model Descriptions by yourself; please show your work; note that you don't have access to computers]

New Model Descriptions:

[put your new decision rules here]

'''

Please ONLY reply according to this format, don't give me any other words.

\textbf{--------------- Optimization Output ---------------}

Reasoning:

Upon examining the current model predictions and target labels, I notice that the model is performing well, with all predictions matching the target labels.
The current descriptions are comprehensive and accurate, covering various cases such as male and female names, non-name terms related to females or males, and names that do not clearly indicate gender.
However, I would like to update the female names list to include the new female names encountered in this batch, which are 'Betty', 'Janet', and 'Allison'.
Additionally, I notice that the term 'missis' is a non-name term related to females, so I will add it to the list of non-name terms.
Since the term 'brethren' is a non-name term related to males, I will add it to the list of non-name terms as well.

New Model Descriptions:

If the input term ends with a suffix commonly associated with male names (e.g., -son, -er), predict class label 0.
If the input term contains keywords related to family relationships (e.g., "father", "mother", "grand", "paternal", "maternal"), predict class label 0.
If the input term is a female name (e.g., Erma, Ada, Faye, Lydia, Ida, Jane, Marlene, Marissa, Adeline, Anne, Maria, Natasha, Veronica, Dawn, Mamie, Monique, Eula, Betty, Janet, Allison), predict class label 1.
If the input term is a name that does not clearly indicate gender (e.g., Leslie, Jordan), predict class label 0.
If the input term is a non-name term clearly related to females (e.g., "sorority", "girl", "gynecological", "missis"), predict class label 1.
If the input term is a non-name term clearly related to males (e.g., "brethren"), predict class label 0.
Otherwise, predict class label 0.

\textbf{-------------------- Updated Theta --------------------}

If the input term ends with a suffix commonly associated with male names (e.g., -son, -er), predict class label 0.
If the input term contains keywords related to family relationships (e.g., "father", "mother", "grand", "paternal", "maternal"), predict class label 0.
If the input term is a female name (e.g., Erma, Ada, Faye, Lydia, Ida, Jane, Marlene, Marissa, Adeline, Anne, Maria, Natasha, Veronica, Dawn, Mamie, Monique, Eula, Betty, Janet, Allison), predict class label 1.
If the input term is a name that does not clearly indicate gender (e.g., Leslie, Jordan), predict class label 0.
If the input term is a non-name term clearly related to females (e.g., "sorority", "girl", "gynecological", "missis"), predict class label 1.
If the input term is a non-name term clearly related to males (e.g., "brethren"), predict class label 0.
Otherwise, predict class label 0.

\end{exmp}

%% file: log_medical_prior.tex
\begin{exmp}{Medical Image Classification (with Prior) -- Step 1}{Med-Step 1}

\scriptsize
\textbf{Epoch 0 - Step 1}

\textbf{--------------- Forward Pass Prompt ---------------}

You are the model.

\textbf{** Model Descriptions: **}

You are designed to do binary classification. The input is an image; you need to output the class label, i.e., an integer in the set {0, 1}. The input is X-ray image for identifying pneumonia.

\textbf{** Input: **}

See the given image.

Please give your output strictly in the following format:

```

Explanations: [Your step-by-step analyses and results]

Output:

[ONLY the integer class label; make necessary assumptions if needed]

'''

Please ONLY reply according to this format, don't give me any other words.

\textbf{--------------- Forward Pass Output ---------------}

Explanations:

The X-ray image shows signs of lung opacity and consolidation, which are indicative of pneumonia. The presence of these features suggests an abnormality consistent with pneumonia.

Output: 1

\textbf{--------------- Forward Pass Output ---------------}

Explanations:

The X-ray image shows signs of lung opacity and consolidation, which are indicative of pneumonia. The presence of these abnormalities suggests an infection in the lungs.

Output: 1

\ 

\textbf{Many outputs for the rest data points ...}

\ 

\textbf{--------------- Forward Pass Output ---------------}

Explanations:

The X-ray image shows the chest area. I analyzed the lung fields for any signs of opacity, consolidation, or other abnormalities that are indicative of pneumonia. The image appears to have clear lung fields without significant opacities or consolidations.

Output: 0

\textbf{--------------- Optimization Prompt ---------------}

You are the optimizer for a model, your goal is to learn the best descriptions for the model. The model used the Current Model Descriptions below predicted the class labels for the given images. You are given the target labels, please optimize the Model Descriptions for better prediction.

\textbf{** Inputs (a batch of i.i.d. images): **}

See the given images.

\textbf{** Current Model Descriptions: **}

You are designed to do binary classification. The input is an image; you need to output the class label, i.e., an integer in the set {0, 1}. The input is X-ray image for identifying pneumonia.

\textbf{** The model predictions: **}

[[1] [1] [0] [1] [1] [0] [1] [1] [1] [1]]

\textbf{** The targets: **}

[[0] [0] [0] [1] [0] [0] [1] [0] [0] [0]]

If the model is doing well, you can keep using the current descriptions. However, if the model is not performing well, please update the model by improving upon the 'Current Model Descriptions', which should results in lower classification error on both the current and the next batch of i.i.d. data. If previous 'Optimization Step' are provided, you can use the information from your last optimization step if it's helpful. Limit your 'New Model Descriptions' to be less than 100 words! Please think step by step and give your outputs strictly in the following format:

```

Reasoning:

[be explicit and verbose, update the Current Model Descriptions by yourself; please show your work and use the features in the images; note that you don't have access to computers]

New Model Descriptions:

[put your new decision rules here, the rules MUST based on the features in the input image; make sure it is clear when to output which class labels]

'''

Please ONLY reply according to this format, don't give me any other words.

\textbf{--------------- Optimization Output ---------------}

Reasoning:

The current model descriptions fail to adequately differentiate between pneumonia and non-pneumonia cases leading to incorrect predictions. Upon analyzing the images, I observed key indicators such as lung opacity, consolidations, and the presence of fluid. Enhanced model descriptions should include these visual features for improved accuracy.

New Model Descriptions:

The input is an X-ray image. To classify:

- Output 1 (Pneumonia) if lung opacity or consolidations are present, or if there is fluid in the lungs.

- Output 0 (No Pneumonia) if lungs appear clear without significant opacity or fluid presence.

\textbf{-------------------- Updated Theta --------------------}

The input is an X-ray image. To classify:

- Output 1 (Pneumonia) if lung opacity or consolidations are present, or if there is fluid in the lungs.

- Output 0 (No Pneumonia) if lungs appear clear without significant opacity or fluid presence.
\end{exmp}
\newpage

\begin{exmp}{Medical Image Classification (with Prior) -- Step 2}{Med-Step 2}

\scriptsize
\textbf{Epoch 0 - Step 2}

\textbf{--------------- Forward Pass Prompt ---------------}

You are the model.

\textbf{** Model Descriptions: **}

The input is an X-ray image. To classify:

- Output 1 (Pneumonia) if lung opacity or consolidations are present, or if there is fluid in the lungs.

- Output 0 (No Pneumonia) if lungs appear clear without significant opacity or fluid presence.

\textbf{** Input: **}

See the given image.

Please give your output strictly in the following format:

```

Explanations: [Your step-by-step analyses and results]

Output:

[ONLY the integer class label; make necessary assumptions if needed]

'''

Please ONLY reply according to this format, don't give me any other words.

\textbf{--------------- Forward Pass Output ---------------}

Explanations:

The X-ray image shows clear lung fields without significant opacity or fluid presence. There are no visible consolidations or signs of lung opacity.

Output: 0

\textbf{--------------- Forward Pass Output ---------------}

Explanations:

The X-ray image shows clear lung fields without significant opacity or fluid presence. There are no visible consolidations or signs of lung opacity.

Output: 0

\ 

\textbf{Many outputs for the rest data points ...}

\ 

\textbf{--------------- Forward Pass Output ---------------}

Explanations:

The X-ray image shows significant lung opacity and consolidations, indicating the presence of fluid in the lungs.

Output: 1

\textbf{--------------- Optimization Prompt ---------------}

You are the optimizer for a model, your goal is to learn the best descriptions for the model. The model used the Current Model Descriptions below predicted the class labels for the given images. You are given the target labels, please optimize the Model Descriptions for better prediction.

\textbf{** Inputs (a batch of i.i.d. images): **}

See the given images.

\textbf{** Current Model Descriptions: **}

The input is an X-ray image. To classify:

- Output 1 (Pneumonia) if lung opacity or consolidations are present, or if there is fluid in the lungs.

- Output 0 (No Pneumonia) if lungs appear clear without significant opacity or fluid presence.

\textbf{** The model predictions: **}

[[1] [1] [0] [0] [1] [0] [1] [1] [1] [1]]

\textbf{** The targets: **}

[[1] [1] [0] [0] [0] [0] [1] [1] [1] [1]]

If the model is doing well, you can keep using the current descriptions. However, if the model is not performing well, please update the model by improving upon the 'Current Model Descriptions', which should results in lower classification error on both the current and the next batch of i.i.d. data. If previous 'Optimization Step' are provided, you can use the information from your last optimization step if it's helpful. Limit your 'New Model Descriptions' to be less than 100 words! Please think step by step and give your outputs strictly in the following format:

```

Reasoning:

[be explicit and verbose, update the Current Model Descriptions by yourself; please show your work and use the features in the images; note that you don't have access to computers]

New Model Descriptions:

[put your new decision rules here, the rules MUST based on the features in the input image; make sure it is clear when to output which class labels]

'''

Please ONLY reply according to this format, don't give me any other words.

\textbf{--------------- Optimization Output ---------------}

Reasoning:

The current model descriptions are still leading to some misclassifications, particularly it underestimates cases that show very mild or early signs of pneumonia. Addition of more granulated signs of pneumonia such as patchy infiltrates, mild interstitial markings or signs of hyperinflation can improve the performance. Also noting bilateral versus unilateral presentations can be a useful signal.

New Model Descriptions:

The input is an X-ray image. To classify:

- Output 1 (Pneumonia) if lung opacity, consolidations, fluid, bilateral infiltrates, or mild interstitial markings are present.

- Output 0 (No Pneumonia) if lungs appear clear without significant opacity, fluid presence, or other mentioned indicators of pneumonia.

\textbf{-------------------- Updated Theta --------------------}

The input is an X-ray image. To classify:

- Output 1 (Pneumonia) if lung opacity, consolidations, fluid, bilateral infiltrates, or mild interstitial markings are present.

- Output 0 (No Pneumonia) if lungs appear clear without significant opacity, fluid presence, or other mentioned indicators of pneumonia.

\end{exmp}
\newpage

\begin{exmp}{Medical Image Classification (with Prior) -- Step 3}{Med-Step 3}
    
\scriptsize

\textbf{Epoch 0 - Step 3}

\textbf{--------------- Forward Pass Prompt ---------------}

You are the model.

\textbf{** Model Descriptions: **}

The input is an X-ray image. To classify:

- Output 1 (Pneumonia) if lung opacity, consolidations, fluid, bilateral infiltrates, or mild interstitial markings are present.

- Output 0 (No Pneumonia) if lungs appear clear without significant opacity, fluid presence, or other mentioned indicators of pneumonia.

\textbf{** Input: **}

See the given image.

Please give your output strictly in the following format:

```

Explanations: [Your step-by-step analyses and results]

Output:

[ONLY the integer class label; make necessary assumptions if needed]

'''

Please ONLY reply according to this format, don't give me any other words.

\textbf{--------------- Forward Pass Output ---------------}

Explanations: The X-ray image shows clear lung fields without significant opacity, consolidations, fluid, bilateral infiltrates, or mild interstitial markings.

Output: 0

\textbf{--------------- Forward Pass Output ---------------}

Explanations: The X-ray image shows significant lung opacity and consolidations in both lungs. There are also signs of fluid presence and bilateral infiltrates.

Output: 1

\ 

\textbf{Many outputs for the rest data points ...}

\ 

\textbf{--------------- Forward Pass Output ---------------}

Explanations: The X-ray image shows significant lung opacity and consolidations. There are also indications of fluid presence and bilateral infiltrates.

Output: 1

\textbf{--------------- Optimization Prompt ---------------}

You are the optimizer for a model, your goal is to learn the best descriptions for the model. The model used the Current Model Descriptions below predicted the class labels for the given images. You are given the target labels, please optimize the Model Descriptions for better prediction.

\textbf{** Inputs (a batch of i.i.d. images): **}

See the given images.

\textbf{** Current Model Descriptions: **}

The input is an X-ray image. To classify:

- Output 1 (Pneumonia) if lung opacity, consolidations, fluid, bilateral infiltrates, or mild interstitial markings are present.

- Output 0 (No Pneumonia) if lungs appear clear without significant opacity, fluid presence, or other mentioned indicators of pneumonia.

\textbf{** The model predictions: **}

[[1] [1] [1] [0] [1] [1] [1] [1] [0] [1]]

\textbf{** The targets: **}

[[1] [1] [0] [0] [0] [0] [0] [1] [0] [1]]

If the model is doing well, you can keep using the current descriptions. However, if the model is not performing well, please update the model by improving upon the 'Current Model Descriptions', which should results in lower classification error on both the current and the next batch of i.i.d. data. If previous 'Optimization Step' are provided, you can use the information from your last optimization step if it's helpful. Limit your 'New Model Descriptions' to be less than 100 words! Please think step by step and give your outputs strictly in the following format:

```

Reasoning:

[be explicit and verbose, update the Current Model Descriptions by yourself; please show your work and use the features in the images; note that you don't have access to computers]

New Model Descriptions:

[put your new decision rules here, the rules MUST based on the features in the input image; make sure it is clear when to output which class labels]
'''

Please ONLY reply according to this format, don't give me any other words.

\textbf{--------------- Optimization Output ---------------}

Reasoning:

The current descriptions are leading to over-identification of pneumonia (more false positives). Incorporating the presence of specific signs and ruling out other conditions can be helpful. Adding features like the presence of pleural effusion, absence of heart enlargement, and noting whether the opacity is localized (can be other pathologies) may help. Precision in observing these smaller details will aid the classification accuracy.

New Model Descriptions:

The input is an X-ray image. To classify:

- Output 1 (Pneumonia) if lung opacity, consolidations, fluid, pleural effusion, bilateral infiltrates, or mild interstitial markings are present and not associated with heart enlargement.

- Output 0 (No Pneumonia) if lungs appear clear without significant opacity, fluid presence, pleural effusion, or other mentioned indicators of pneumonia, or if opacities are localized and not typical for pneumonia.

\textbf{-------------------- Updated Theta --------------------}

The input is an X-ray image. To classify:

- Output 1 (Pneumonia) if lung opacity, consolidations, fluid, pleural effusion, bilateral infiltrates, or mild interstitial markings are present and not associated with heart enlargement.

- Output 0 (No Pneumonia) if lungs appear clear without significant opacity, fluid presence, pleural effusion, or other mentioned indicators of pneumonia, or if opacities are localized and not typical for pneumonia.
\end{exmp}

\begin{exmp}{Medical Image Classification (with Prior) -- Step 4}{Med-Step 4}

\scriptsize

\textbf{Epoch 0 - Step 4}

\textbf{--------------- Forward Pass Prompt ---------------}

You are the model.

\textbf{** Model Descriptions: **}

The input is an X-ray image. To classify:

- Output 1 (Pneumonia) if lung opacity, consolidations, fluid, pleural effusion, bilateral infiltrates, or mild interstitial markings are present and not associated with heart enlargement.

- Output 0 (No Pneumonia) if lungs appear clear without significant opacity, fluid presence, pleural effusion, or other mentioned indicators of pneumonia, or if opacities are localized and not typical for pneumonia.

\textbf{** Input: **}

See the given image.

Please give your output strictly in the following format:

```

Explanations: [Your step-by-step analyses and results]

Output:

[ONLY the integer class label; make necessary assumptions if needed]

'''

Please ONLY reply according to this format, don't give me any other words.

\textbf{--------------- Forward Pass Output ---------------}

Explanations: The X-ray image shows significant lung opacity and consolidations in both lungs. There is no clear indication of heart enlargement. The presence of these features is consistent with pneumonia.

Output: 1

\textbf{--------------- Forward Pass Output ---------------}

Explanations: The X-ray image shows significant lung opacity and consolidations. There are also signs of fluid presence and bilateral infiltrates. These findings are consistent with pneumonia and are not associated with heart enlargement.

Output: 1

\ 

\textbf{Many outputs for the rest data points ...}

\

\textbf{--------------- Forward Pass Output ---------------}

Explanations: The X-ray image shows clear lungs without significant opacity, fluid presence, pleural effusion, or other indicators of pneumonia. There are no visible lung opacities, consolidations, or bilateral infiltrates.

Output: 0

\textbf{--------------- Optimization Prompt ---------------}

You are the optimizer for a model, your goal is to learn the best descriptions for the model. The model used the Current Model Descriptions below predicted the class labels for the given images. You are given the target labels, please optimize the Model Descriptions for better prediction.

\textbf{** Inputs (a batch of i.i.d. images): **}

See the given images.

\textbf{** Current Model Descriptions: **}

The input is an X-ray image. To classify:

- Output 1 (Pneumonia) if lung opacity, consolidations, fluid, pleural effusion, bilateral infiltrates, or mild interstitial markings are present and not associated with heart enlargement.

- Output 0 (No Pneumonia) if lungs appear clear without significant opacity, fluid presence, pleural effusion, or other mentioned indicators of pneumonia, or if opacities are localized and not typical for pneumonia.

\textbf{** The model predictions: **}

[[0] [0] [1] [1] [1] [1] [1] [1] [1] [0]]

\textbf{** The targets: **}

[[0] [0] [0] [0] [1] [1] [1] [0] [1] [0]]

If the model is doing well, you can keep using the current descriptions. However, if the model is not performing well, please update the model by improving upon the 'Current Model Descriptions', which should results in lower classification error on both the current and the next batch of i.i.d. data. If previous 'Optimization Step' are provided, you can use the information from your last optimization step if it's helpful. Limit your 'New Model Descriptions' to be less than 100 words! Please think step by step and give your outputs strictly in the following format:

```

Reasoning:

[be explicit and verbose, update the Current Model Descriptions by yourself; please show your work and use the features in the images; note that you don't have access to computers]

New Model Descriptions:

[put your new decision rules here, the rules MUST based on the features in the input image; make sure it is clear when to output which class labels]

'''

Please ONLY reply according to this format, don't give me any other words.

\textbf{--------------- Optimization Output ---------------}

Reasoning:

The model still produces false positives. To refine the criteria, focus on more specific patterns indicative of pneumonia like lobar consolidation or “ground-glass” opacities. Exclude cases where opacities are likely due to external factors such as overlying soft tissue shadows or slight variations that are non-diagnostic.

New Model Descriptions:

The input is an X-ray image. To classify:

- Output 1 (Pneumonia) if clear indications of lobar consolidation, ground-glass opacities, fluid presence (excluding pleural effusion), bilateral infiltrates, or moderate interstitial markings are present and not related to heart enlargement.

- Output 0 (No Pneumonia) if lung fields appear clear without significant opacity, fluid, bilateral infiltrates, or other pneumonia indicators, or if features suggest alternative explanations (e.g., overlying soft tissue shadows).

\textbf{-------------------- Updated Theta --------------------}

The input is an X-ray image. To classify:

- Output 1 (Pneumonia) if clear indications of lobar consolidation, ground-glass opacities, fluid presence (excluding pleural effusion), bilateral infiltrates, or moderate interstitial markings are present and not related to heart enlargement.

- Output 0 (No Pneumonia) if lung fields appear clear without significant opacity, fluid, bilateral infiltrates, or other pneumonia indicators, or if features suggest alternative explanations (e.g., overlying soft tissue shadows).
\end{exmp}
\newpage

\begin{exmp}{Medical Image Classification (with Prior) -- Step 5}{Med-Step 5}

\scriptsize

\textbf{Epoch 0 - Step 5}

\textbf{--------------- Forward Pass Prompt ---------------}

You are the model.

\textbf{** Model Descriptions: **}

The input is an X-ray image. To classify:

- Output 1 (Pneumonia) if clear indications of lobar consolidation, ground-glass opacities, fluid presence (excluding pleural effusion), bilateral infiltrates, or moderate interstitial markings are present and not related to heart enlargement.

- Output 0 (No Pneumonia) if lung fields appear clear without significant opacity, fluid, bilateral infiltrates, or other pneumonia indicators, or if features suggest alternative explanations (e.g., overlying soft tissue shadows).

\textbf{** Input: **}

See the given image.

Please give your output strictly in the following format:

```

Explanations: [Your step-by-step analyses and results]

Output:

[ONLY the integer class label; make necessary assumptions if needed]

'''

Please ONLY reply according to this format, don't give me any other words.

\textbf{--------------- Forward Pass Output ---------------}

Explanations: The X-ray image shows clear indications of bilateral infiltrates and moderate interstitial markings. There are no signs of heart enlargement that could explain these features.

Output: 1

\textbf{--------------- Forward Pass Output ---------------}

Explanations: The X-ray image shows clear lung fields without significant opacity, fluid presence, bilateral infiltrates, or other indicators of pneumonia. There are no clear signs of lobar consolidation, ground-glass opacities, or moderate interstitial markings.

Output: 0

\ 

\textbf{Many outputs for the rest data points ...}

\ 

\textbf{--------------- Forward Pass Output ---------------}

Explanations: The X-ray image shows clear indications of lobar consolidation and ground-glass opacities in the lung fields. These findings are consistent with pneumonia.

Output: 1

\textbf{--------------- Optimization Prompt ---------------}

You are the optimizer for a model, your goal is to learn the best descriptions for the model. The model used the Current Model Descriptions below predicted the class labels for the given images. You are given the target labels, please optimize the Model Descriptions for better prediction.

\textbf{** Inputs (a batch of i.i.d. images): **}

See the given images.

\textbf{** Current Model Descriptions: **}

The input is an X-ray image. To classify:

- Output 1 (Pneumonia) if clear indications of lobar consolidation, ground-glass opacities, fluid presence (excluding pleural effusion), bilateral infiltrates, or moderate interstitial markings are present and not related to heart enlargement.

- Output 0 (No Pneumonia) if lung fields appear clear without significant opacity, fluid, bilateral infiltrates, or other pneumonia indicators, or if features suggest alternative explanations (e.g., overlying soft tissue shadows).

\textbf{** The model predictions: **}

[[0] [0] [1] [0] [1] [0] [0] [0] [1] [1]]

\textbf{** The targets: **}

[[1] [0] [1] [0] [1] [0] [0] [0] [1] [1]]

If the model is doing well, you can keep using the current descriptions. However, if the model is not performing well, please update the model by improving upon the 'Current Model Descriptions', which should results in lower classification error on both the current and the next batch of i.i.d. data. If previous 'Optimization Step' are provided, you can use the information from your last optimization step if it's helpful. Limit your 'New Model Descriptions' to be less than 100 words! Please think step by step and give your outputs strictly in the following format:

```

Reasoning:

[be explicit and verbose, update the Current Model Descriptions by yourself; please show your work and use the features in the images; note that you don't have access to computers]

New Model Descriptions:

[put your new decision rules here, the rules MUST based on the features in the input image; make sure it is clear when to output which class labels]

'''

Please ONLY reply according to this format, don't give me any other words.

\textbf{--------------- Optimization Output ---------------}

Reasoning:
The current model descriptions are missing some complexities encountered in the images. We observe some misclassifications indicating the need to differentiate between the types of opacities more clearly. Including signs such as air bronchograms and the pattern of unilateral versus bilateral presentation can aid in improving the description.

New Model Descriptions:

The input is an X-ray image. To classify:

- Output 1 (Pneumonia) if there are signs of lobar consolidation, ground-glass opacities, air bronchograms, fluid presence (excluding pleural effusion), bilateral infiltrates, or moderate interstitial markings.

- Output 0 (No Pneumonia) if lung fields appear clear without significant opacity or fluid, if opacities are limited to a single lung and are not of the typical pneumonia pattern, or if these indicators are due to overlying soft tissue shadows.

\textbf{-------------------- Updated Theta --------------------}

The input is an X-ray image. To classify:

- Output 1 (Pneumonia) if there are signs of lobar consolidation, ground-glass opacities, air bronchograms, fluid presence (excluding pleural effusion), bilateral infiltrates, or moderate interstitial markings.

- Output 0 (No Pneumonia) if lung fields appear clear without significant opacity or fluid, if opacities are limited to a single lung and are not of the typical pneumonia pattern, or if these indicators are due to overlying soft tissue shadows.

\end{exmp}
\newpage

\begin{exmp}{Medical Image Classification (with Prior) -- Step 50}{Med-Step 50}

\scriptsize
\textbf{Epoch 4 - Step 10}

\textbf{--------------- Forward Pass Prompt ---------------}

You are the model.

\textbf{** Model Descriptions: **}

The input is an X-ray image. To classify:

- Output 1 (Pneumonia) if there are lobar consolidations, asymmetric or bilateral ground-glass opacities, air bronchograms, moderate to significant alveolar fluid, asymmetric infiltrates, patchy inflammations, moderate to severe pleural effusion, central or peripheral acute opacities, air-fluid levels, increasing opacities, pneumatocele formations, rapidly changing patterns in chronic lesions indicative of acute infection, or persistent cough-based findings.

- Output 0 (No Pneumonia) if lung fields are clear, minimal localized fluid, no distinct acute opacities, stable fibrotic scarring, solitary calcifications, old healed lesions, stable localized non-infectious anomalies, chronic interstitial fibrosis, chronic reticulonodular patterns without recent changes, stable chronic lesions without acute signs, or purely non-persistent anomalies.

\textbf{** Input: **}

See the given image.

Please give your output strictly in the following format:

```

Explanations: [Your step-by-step analyses and results]

Output:

[ONLY the integer class label; make necessary assumptions if needed]

'''

Please ONLY reply according to this format, don't give me any other words.

\textbf{--------------- Forward Pass Output ---------------}

Explanations: The X-ray image shows lobar consolidations and asymmetric ground-glass opacities. There are also visible air bronchograms and moderate alveolar fluid. These findings are indicative of pneumonia.

Output: 1

\textbf{--------------- Forward Pass Output ---------------}

Explanations: The X-ray image shows clear lung fields without any significant opacities, consolidations, or fluid accumulations. There are no signs of air bronchograms, pleural effusion, or other acute changes indicative of pneumonia.

Output: 0

\ 

\textbf{Many outputs for the rest data points ...}

\ 

\textbf{--------------- Forward Pass Output ---------------}

Explanations: The X-ray image shows lobar consolidations and asymmetric ground-glass opacities. There are also visible air bronchograms and moderate alveolar fluid. These findings are indicative of pneumonia.

Output: 1

\textbf{--------------- Optimization Prompt ---------------}

You are the optimizer for a model, your goal is to learn the best descriptions for the model. The model used the Current Model Descriptions below predicted the class labels for the given images. You are given the target labels, please optimize the Model Descriptions for better prediction.

\textbf{** Inputs (a batch of i.i.d. images): **}

See the given images.

\textbf{** Current Model Descriptions: **}

The input is an X-ray image. To classify:

- Output 1 (Pneumonia) if there are lobar consolidations, asymmetric or bilateral ground-glass opacities, air bronchograms, moderate to significant alveolar fluid, asymmetric infiltrates, patchy inflammations, moderate to severe pleural effusion, central or peripheral acute opacities, air-fluid levels, increasing opacities, pneumatocele formations, rapidly changing patterns in chronic lesions indicative of acute infection, or persistent cough-based findings.

- Output 0 (No Pneumonia) if lung fields are clear, minimal localized fluid, no distinct acute opacities, stable fibrotic scarring, solitary calcifications, old healed lesions, stable localized non-infectious anomalies, chronic interstitial fibrosis, chronic reticulonodular patterns without recent changes, stable chronic lesions without acute signs, or purely non-persistent anomalies.

\textbf{** The model predictions: **}

[[1] [0] [0] [0] [1] [1] [1] [0] [0] [1]]

\textbf{** The targets: **}

[[1] [0] [0] [1] [0] [1] [0] [0] [1] [1]]

If the model is doing well, you can keep using the current descriptions. However, if the model is not performing well, please update the model by improving upon the 'Current Model Descriptions', which should results in lower classification error on both the current and the next batch of i.i.d. data. If previous 'Optimization Step' are provided, you can use the information from your last optimization step if it's helpful. Limit your 'New Model Descriptions' to be less than 100 words! Please think step by step and give your outputs strictly in the following format:

```

Reasoning:

[be explicit and verbose, update the Current Model Descriptions by yourself; please show your work and use the features in the images; note that you don't have access to computers]

New Model Descriptions:

[put your new decision rules here, the rules MUST based on the features in the input image; make sure it is clear when to output which class labels]

'''

Please ONLY reply according to this format, don't give me any other words.
\end{exmp}
\newpage

\begin{exmp}{Medical Image Classification (with Prior) -- Step 50 (continued)}{Med-Step 50-2}
\scriptsize
\textbf{--------------- Optimization Output ---------------}

Reasoning:

The model missed some true cases of pneumonia and incorrectly identified some non-pneumonia cases. To improve the accuracy, refine the criteria to better identify pneumonia even in less obvious cases and differentiate between chronic stable conditions and acute infections.

New Model Descriptions:

The input is an X-ray image. To classify:

- Output 1 (Pneumonia) if there are lobar consolidations, asymmetric or bilateral ground-glass opacities, air bronchograms, moderate to significant alveolar fluid, asymmetric infiltrates, patchy inflammations, moderate to severe pleural effusion, central or peripheral acute opacities, air-fluid levels, increasing opacities, pneumatocele formations, rapidly changing patterns in chronic lesions indicative of acute infection, persistent cough-related findings, or recent onset of ground-glass opacities.

- Output 0 (No Pneumonia) if lung fields are clear, minimal localized fluid, no distinct acute opacities, stable fibrotic scarring, solitary calcifications, old healed lesions, stable localized non-infectious anomalies, chronic interstitial fibrosis, chronic reticulonodular patterns without recent changes, stable chronic lesions without acute signs, or absence of any acute infection markers.

\textbf{-------------------- Updated Theta --------------------}

The input is an X-ray image. To classify:

- Output 1 (Pneumonia) if there are lobar consolidations, asymmetric or bilateral ground-glass opacities, air bronchograms, moderate to significant alveolar fluid, asymmetric infiltrates, patchy inflammations, moderate to severe pleural effusion, central or peripheral acute opacities, air-fluid levels, increasing opacities, pneumatocele formations, rapidly changing patterns in chronic lesions indicative of acute infection, persistent cough-related findings, or recent onset of ground-glass opacities.

- Output 0 (No Pneumonia) if lung fields are clear, minimal localized fluid, no distinct acute opacities, stable fibrotic scarring, solitary calcifications, old healed lesions, stable localized non-infectious anomalies, chronic interstitial fibrosis, chronic reticulonodular patterns without recent changes, stable chronic lesions without acute signs, or absence of any acute infection markers.

\end{exmp}

%% file: log_medical_noprior.tex
\begin{exmp}{Medical Image Classification (without Prior) -- Step 1}{Med-Step-np 1}
\scriptsize

\textbf{Epoch 0 - Step 1}

\textbf{--------------- Forward Pass Prompt ---------------}

You are the model.

\textbf{** Model Descriptions: **}

You are designed to do binary classification. The input is an image; you need to output the class label, i.e., an integer in the set {0, 1}.

\textbf{** Input: **}

See the given image.

Please give your output strictly in the following format:

```

Explanations: [Your step-by-step analyses and results]

Output:

[ONLY the integer class label; make necessary assumptions if needed]

'''

Please ONLY reply according to this format, don't give me any other words.

\textbf{--------------- Forward Pass Output ---------------}

Explanations: The image appears to be a chest X-ray. I will analyze the image for any abnormalities or signs of disease that could indicate the class label.

Output: 1

\ 

\textbf{Many outputs for the rest data points ...}

\ 

\textbf{--------------- Forward Pass Output ---------------}

Explanations: The image appears to be a medical X-ray of a chest. The analysis involves examining the features and patterns in the image to determine the class label. Based on the visual inspection, the image shows certain characteristics that align with class 1.

Output: 1

\textbf{--------------- Optimization Prompt ---------------}

You are the optimizer for a model, your goal is to learn the best descriptions for the model. The model used the Current Model Descriptions below predicted the class labels for the given images. You are given the target labels, please optimize the Model Descriptions for better prediction.

\textbf{** Inputs (a batch of i.i.d. images): **}

See the given images.

\textbf{** Current Model Descriptions: **}

You are designed to do binary classification. The input is an image; you need to output the class label, i.e., an integer in the set {0, 1}.

\textbf{** The model predictions: **}

[[0] [0] [1] [1] [1] [0] [0] [1] [1] [1]]

\textbf{** The targets: **}

[[0] [0] [0] [1] [0] [0] [1] [0] [0] [0]]

If the model is doing well, you can keep using the current descriptions. However, if the model is not performing well, please update the model by improving upon the 'Current Model Descriptions', which should results in lower classification error on both the current and the next batch of i.i.d. data. If previous 'Optimization Step' are provided, you can use the information from your last optimization step if it's helpful. Limit your 'New Model Descriptions' to be less than 100 words! Please think step by step and give your outputs strictly in the following format:

```

Reasoning:

[be explicit and verbose, update the Current Model Descriptions by yourself; please show your work and use the features in the images; note that you don't have access to computers]

New Model Descriptions:

[put your new decision rules here, the rules MUST based on the features in the input image; make sure it is clear when to output which class labels]

'''

Please ONLY reply according to this format, don't give me any other words.

\textbf{--------------- Optimization Output ---------------}

Reasoning:

The current model's performance indicates significant misclassification. To improve, I must explicitly identify and use key visual features in the images that are indicative of the target classes. For instance, if the classification relates to medical diagnostics (e.g., chest X-rays for pneumonia), specific features like the presence of fluid, opacity, or other anomalies must be emphasized. Similarly, non-medical classifications (typical objects versus atypical ones) could require identification of distinct shapes, colors, or textural features. Analyzing the given images visually will guide the addition of pertinent descriptive rules that improve classification accuracy.

New Model Descriptions:

1. If the image contains prominent, clearly defined geometric or bodily structures and no unusual opacity or anomalies, output the class label 0.

2. If the image shows abnormal features such as irregular shapes, unexpected opacities, or non-uniform patterns that suggest pathology or anomaly, output the class label 1.

3. When distinguishing medical images specifically, check for fluid accumulation, granular textures, or any signs of abnormality in the expected anatomy; if these are present, output the class label 1.

\textbf{-------------------- Updated Theta --------------------}

1. If the image contains prominent, clearly defined geometric or bodily structures and no unusual opacity or anomalies, output the class label 0.

2. If the image shows abnormal features such as irregular shapes, unexpected opacities, or non-uniform patterns that suggest pathology or anomaly, output the class label 1.

3. When distinguishing medical images specifically, check for fluid accumulation, granular textures, or any signs of abnormality in the expected anatomy; if these are present, output the class label 1.
\end{exmp}
\newpage

\begin{exmp}{Medical Image Classification (without Prior) -- Step 2}{Med-Step-np 2}
\scriptsize

\textbf{Epoch 0 - Step 2}

\textbf{--------------- Forward Pass Prompt ---------------}

You are the model.

\textbf{** Model Descriptions: **}

1. If the image contains prominent, clearly defined geometric or bodily structures and no unusual opacity or anomalies, output the class label 0.

2. If the image shows abnormal features such as irregular shapes, unexpected opacities, or non-uniform patterns that suggest pathology or anomaly, output the class label 1.

3. When distinguishing medical images specifically, check for fluid accumulation, granular textures, or any signs of abnormality in the expected anatomy; if these are present, output the class label 1.

\textbf{** Input: **}

See the given image.

Please give your output strictly in the following format:

```

Explanations: [Your step-by-step analyses and results]

Output:

[ONLY the integer class label; make necessary assumptions if needed]

'''

Please ONLY reply according to this format, don't give me any other words.

\textbf{--------------- Forward Pass Output ---------------}

Explanations: The image shows irregular shapes and unexpected opacities. There are non-uniform patterns that suggest an anomaly.

Output: 1

\ 

\textbf{Many outputs for the rest data points ...}

\ 

\textbf{--------------- Forward Pass Output ---------------}

Explanations: The image appears to be a medical X-ray. There are irregular shapes and unexpected opacities present in the lung area, which suggest pathology or anomaly.

Output: 1

\textbf{--------------- Optimization Prompt ---------------}

You are the optimizer for a model, your goal is to learn the best descriptions for the model. The model used the Current Model Descriptions below predicted the class labels for the given images. You are given the target labels, please optimize the Model Descriptions for better prediction.

\textbf{** Inputs (a batch of i.i.d. images): **}

See the given images.

\textbf{** Current Model Descriptions: **}

1. If the image contains prominent, clearly defined geometric or bodily structures and no unusual opacity or anomalies, output the class label 0.

2. If the image shows abnormal features such as irregular shapes, unexpected opacities, or non-uniform patterns that suggest pathology or anomaly, output the class label 1.

3. When distinguishing medical images specifically, check for fluid accumulation, granular textures, or any signs of abnormality in the expected anatomy; if these are present, output the class label 1.

\textbf{** The model predictions: **}

[[1] [1] [1] [1] [1] [1] [1] [1] [1] [1]]

\textbf{** The targets: **}

[[1] [1] [0] [0] [0] [0] [1] [1] [1] [1]]

If the model is doing well, you can keep using the current descriptions. However, if the model is not performing well, please update the model by improving upon the 'Current Model Descriptions', which should results in lower classification error on both the current and the next batch of i.i.d. data. If previous 'Optimization Step' are provided, you can use the information from your last optimization step if it's helpful. Limit your 'New Model Descriptions' to be less than 100 words! Please think step by step and give your outputs strictly in the following format:

```

Reasoning:

[be explicit and verbose, update the Current Model Descriptions by yourself; please show your work and use the features in the images; note that you don't have access to computers]

New Model Descriptions:

[put your new decision rules here, the rules MUST based on the features in the input image; make sure it is clear when to output which class labels]

'''

Please ONLY reply according to this format, don't give me any other words.

\textbf{--------------- Optimization Output ---------------}

Reasoning:

The current model is too sensitive to features that are indicating anomalies, leading to many false positives. A refined approach would include more granular analysis to distinguish normal variations from actual abnormalities. Specifically, enhancing the detection criteria to differentiate between minor variations and significant pathological markers is necessary. Reviewing more images, I will refine the descriptions to include clearer definitions of significant anomalies versus natural structures.

New Model Descriptions:

1. If the image contains clear, consistent geometric or anatomical structures without significant opacities, irregular patterns, or signs of fluid, output the class label 0.

2. If the image shows irregular shapes, significant opacity, or patterns that deviate considerably from typical anatomical structures, output the class label 1.

3. Examine for fluid accumulation, granular textures, and other significant deviations in expected anatomy. Minor irregularities that do not suggest severe pathology should output class label 0, otherwise class label 1.

\textbf{-------------------- Updated Theta --------------------}

1. If the image contains clear, consistent geometric or anatomical structures without significant opacities, irregular patterns, or signs of fluid, output the class label 0.

2. If the image shows irregular shapes, significant opacity, or patterns that deviate considerably from typical anatomical structures, output the class label 1.

3. Examine for fluid accumulation, granular textures, and other significant deviations in expected anatomy. Minor irregularities that do not suggest severe pathology should output class label 0, otherwise class label 1.

\end{exmp}
\newpage

\begin{exmp}{Medical Image Classification (without Prior) -- Step 3}{Med-Step-np 3}
\scriptsize

\textbf{Epoch 0 - Step 3}

\textbf{--------------- Forward Pass Prompt ---------------}

You are the model.

\textbf{** Model Descriptions: **}

1. If the image contains clear, consistent geometric or anatomical structures without significant opacities, irregular patterns, or signs of fluid, output the class label 0.

2. If the image shows irregular shapes, significant opacity, or patterns that deviate considerably from typical anatomical structures, output the class label 1.

3. Examine for fluid accumulation, granular textures, and other significant deviations in expected anatomy. Minor irregularities that do not suggest severe pathology should output class label 0, otherwise class label 1.

\textbf{** Input: **}

See the given image.

Please give your output strictly in the following format:

```

Explanations: [Your step-by-step analyses and results]

Output:

[ONLY the integer class label; make necessary assumptions if needed]

'''

Please ONLY reply according to this format, don't give me any other words.

\textbf{--------------- Forward Pass Output ---------------}

Explanations: The image shows significant opacity and irregular patterns that deviate considerably from typical anatomical structures.

Output: 1

\ 

\textbf{Many outputs for the rest data points ...}

\ 

\textbf{--------------- Forward Pass Output ---------------}

Explanations: The image shows clear and consistent geometric and anatomical structures without significant opacities, irregular patterns, or signs of fluid. There are no significant deviations from typical anatomy.

Output: 0

\textbf{--------------- Optimization Prompt ---------------}

You are the optimizer for a model, your goal is to learn the best descriptions for the model. The model used the Current Model Descriptions below predicted the class labels for the given images. You are given the target labels, please optimize the Model Descriptions for better prediction.

\textbf{** Inputs (a batch of i.i.d. images): **}

See the given images.

\textbf{** Current Model Descriptions: **}

1. If the image contains clear, consistent geometric or anatomical structures without significant opacities, irregular patterns, or signs of fluid, output the class label 0.

2. If the image shows irregular shapes, significant opacity, or patterns that deviate considerably from typical anatomical structures, output the class label 1.

3. Examine for fluid accumulation, granular textures, and other significant deviations in expected anatomy. Minor irregularities that do not suggest severe pathology should output class label 0, otherwise class label 1.

\textbf{** The model predictions: **}

[[1] [1] [0] [1] [0] [1] [1] [1] [0] [1]]

\textbf{** The targets: **}

[[1] [1] [0] [0] [0] [0] [0] [1] [0] [1]]

If the model is doing well, you can keep using the current descriptions. However, if the model is not performing well, please update the model by improving upon the 'Current Model Descriptions', which should results in lower classification error on both the current and the next batch of i.i.d. data. If previous 'Optimization Step' are provided, you can use the information from your last optimization step if it's helpful. Limit your 'New Model Descriptions' to be less than 100 words! Please think step by step and give your outputs strictly in the following format:

```

Reasoning:

[be explicit and verbose, update the Current Model Descriptions by yourself; please show your work and use the features in the images; note that you don't have access to computers]

New Model Descriptions:

[put your new decision rules here, the rules MUST based on the features in the input image; make sure it is clear when to output which class labels]

'''

Please ONLY reply according to this format, don't give me any other words.

\textbf{--------------- Optimization Output ---------------}

Reasoning:

The model is still producing numerous false positives. The current descriptions are good at detecting anomalies but not at distinguishing minor, non-critical irregularities from significant pathological features. I need to refine the differentiation by including more specific anatomical markers and clearly delineating between slight structures and substantial irregularities.

New Model Descriptions:

1. If the image contains clear, consistent anatomical structures, free from significant opacities, masses, or signs of severe pathology, output the class label 0.

2. If the image shows substantial opacity, masses, or shapes significantly deviating from typical structures indicating severe pathology, output the class label 1.

3. Examine for clear signs of fluid accumulation, large granular textures, or significant abnormalities in anatomical structures. Small, inconsequential variations should output class label 0, otherwise class label 1.

\textbf{-------------------- Updated Theta --------------------}

1. If the image contains clear, consistent anatomical structures, free from significant opacities, masses, or signs of severe pathology, output the class label 0.

2. If the image shows substantial opacity, masses, or shapes significantly deviating from typical structures indicating severe pathology, output the class label 1.

3. Examine for clear signs of fluid accumulation, large granular textures, or significant abnormalities in anatomical structures. Small, inconsequential variations should output class label 0, otherwise class label 1.
\end{exmp}
\newpage

\begin{exmp}{Medical Image Classification (without Prior) -- Step 4}{Med-Step-np 4}
\scriptsize

\textbf{Epoch 0 - Step 4}

\textbf{--------------- Forward Pass Prompt ---------------}

You are the model.

\textbf{** Model Descriptions: **}

1. If the image contains clear, consistent anatomical structures, free from significant opacities, masses, or signs of severe pathology, output the class label 0.

2. If the image shows substantial opacity, masses, or shapes significantly deviating from typical structures indicating severe pathology, output the class label 1.

3. Examine for clear signs of fluid accumulation, large granular textures, or significant abnormalities in anatomical structures. Small, inconsequential variations should output class label 0, otherwise class label 1.

\textbf{** Input: **}

See the given image.

Please give your output strictly in the following format:

```

Explanations: [Your step-by-step analyses and results]

Output:

[ONLY the integer class label; make necessary assumptions if needed]

'''

Please ONLY reply according to this format, don't give me any other words.

\textbf{--------------- Forward Pass Output ---------------}

Explanations: The image shows substantial opacity and shapes significantly deviating from typical structures, indicating severe pathology.

Output: 1

\ 

\textbf{Many outputs for the rest data points ...}

\ 

\textbf{--------------- Forward Pass Output ---------------}

Explanations: The image shows clear, consistent anatomical structures without significant opacities, masses, or signs of severe pathology. There are no clear signs of fluid accumulation or large granular textures.

Output: 0

\textbf{--------------- Optimization Prompt ---------------}

You are the optimizer for a model, your goal is to learn the best descriptions for the model. The model used the Current Model Descriptions below predicted the class labels for the given images. You are given the target labels, please optimize the Model Descriptions for better prediction.

\textbf{** Inputs (a batch of i.i.d. images): **}

See the given images.

\textbf{** Current Model Descriptions: **}

1. If the image contains clear, consistent anatomical structures, free from significant opacities, masses, or signs of severe pathology, output the class label 0.

2. If the image shows substantial opacity, masses, or shapes significantly deviating from typical structures indicating severe pathology, output the class label 1.

3. Examine for clear signs of fluid accumulation, large granular textures, or significant abnormalities in anatomical structures. Small, inconsequential variations should output class label 0, otherwise class label 1.

\textbf{** The model predictions: **}

[[0] [0] [1] [0] [1] [1] [1] [1] [0] [0]]

\textbf{** The targets: **}

[[0] [0] [0] [0] [1] [1] [1] [0] [1] [0]]

If the model is doing well, you can keep using the current descriptions. However, if the model is not performing well, please update the model by improving upon the 'Current Model Descriptions', which should results in lower classification error on both the current and the next batch of i.i.d. data. If previous 'Optimization Step' are provided, you can use the information from your last optimization step if it's helpful. Limit your 'New Model Descriptions' to be less than 100 words! Please think step by step and give your outputs strictly in the following format:

```

Reasoning:

[be explicit and verbose, update the Current Model Descriptions by yourself; please show your work and use the features in the images; note that you don't have access to computers]

New Model Descriptions:

[put your new decision rules here, the rules MUST based on the features in the input image; make sure it is clear when to output which class labels]

'''

Please ONLY reply according to this format, don't give me any other words.

\textbf{--------------- Optimization Output ---------------}

Reasoning:

The model is generating several false positives, indicating it may be too conservative and is mistakenly identifying minor irregularities as significant pathology. The descriptions need to further clarify the distinction between mild and serious irregularities. Adding specific attention to the size, shape, and clear delineation of opacities and anatomical deviations might improve accuracy.

New Model Descriptions:

1. If the image displays clear, consistent anatomical structures without significant opacities, masses, or deviations, output the class label 0.

2. If the image highlights substantial, well-defined opacities or masses and shapes deviating significantly from normal structures indicating severe pathology, output the class label 1.

3. Check for large, well-delineated fluid accumulations, substantial granular textures, or major abnormalities in anatomy. Minor, non-critical variations should output class label 0, otherwise class label 1.

\textbf{-------------------- Updated Theta --------------------}

1. If the image displays clear, consistent anatomical structures without significant opacities, masses, or deviations, output the class label 0.

2. If the image highlights substantial, well-defined opacities or masses and shapes deviating significantly from normal structures indicating severe pathology, output the class label 1.

3. Check for large, well-delineated fluid accumulations, substantial granular textures, or major abnormalities in anatomy. Minor, non-critical variations should output class label 0, otherwise class label 1.
\end{exmp}
\newpage

\begin{exmp}{Medical Image Classification (without Prior) -- Step 5}{Med-Step-np 5}
\scriptsize

\textbf{Epoch 0 - Step 5}

\textbf{--------------- Forward Pass Prompt ---------------}

You are the model.

\textbf{** Model Descriptions: **}

1. If the image displays clear, consistent anatomical structures without significant opacities, masses, or deviations, output the class label 0.

2. If the image highlights substantial, well-defined opacities or masses and shapes deviating significantly from normal structures indicating severe pathology, output the class label 1.

3. Check for large, well-delineated fluid accumulations, substantial granular textures, or major abnormalities in anatomy. Minor, non-critical variations should output class label 0, otherwise class label 1.

\textbf{** Input: **}

See the given image.

Please give your output strictly in the following format:

```

Explanations: [Your step-by-step analyses and results]

Output:

[ONLY the integer class label; make necessary assumptions if needed]

'''

Please ONLY reply according to this format, don't give me any other words.

\textbf{--------------- Forward Pass Output ---------------}

Explanations: The image displays clear, consistent anatomical structures without significant opacities, masses, or deviations.

Output: 0

\ 

\textbf{Many outputs for the rest data points ...}

\ 

\textbf{--------------- Forward Pass Output ---------------}

Explanations: The image displays clear, consistent anatomical structures without significant opacities, masses, or deviations. There are no large, well-delineated fluid accumulations, substantial granular textures, or major abnormalities in anatomy.

Output: 0

\textbf{--------------- Optimization Prompt ---------------}

You are the optimizer for a model, your goal is to learn the best descriptions for the model. The model used the Current Model Descriptions below predicted the class labels for the given images. You are given the target labels, please optimize the Model Descriptions for better prediction.

\textbf{** Inputs (a batch of i.i.d. images): **}

See the given images.

\textbf{** Current Model Descriptions: **}

1. If the image displays clear, consistent anatomical structures without significant opacities, masses, or deviations, output the class label 0.

2. If the image highlights substantial, well-defined opacities or masses and shapes deviating significantly from normal structures indicating severe pathology, output the class label 1.

3. Check for large, well-delineated fluid accumulations, substantial granular textures, or major abnormalities in anatomy. Minor, non-critical variations should output class label 0, otherwise class label 1.

\textbf{** The model predictions: **}

[[1] [0] [0] [0] [1] [0] [0] [0] [0] [1]]

\textbf{** The targets: **}

[[1] [0] [1] [0] [1] [0] [0] [0] [1] [1]]

If the model is doing well, you can keep using the current descriptions. However, if the model is not performing well, please update the model by improving upon the 'Current Model Descriptions', which should results in lower classification error on both the current and the next batch of i.i.d. data. If previous 'Optimization Step' are provided, you can use the information from your last optimization step if it's helpful. Limit your 'New Model Descriptions' to be less than 100 words! Please think step by step and give your outputs strictly in the following format:

```

Reasoning:

[be explicit and verbose, update the Current Model Descriptions by yourself; please show your work and use the features in the images; note that you don't have access to computers]

New Model Descriptions:

[put your new decision rules here, the rules MUST based on the features in the input image; make sure it is clear when to output which class labels]

'''

Please ONLY reply according to this format, don't give me any other words.

\textbf{--------------- Optimization Output ---------------}

Reasoning:

The model still misclassifies some images, both false positives and false negatives. We need to be more precise while identifying anomalies, focusing on substantial deviations such as the density of opacities or masses and their impact on surrounding structures. Including the context of tissue appearance and location could help.

New Model Descriptions:

1. If the image shows clear, consistent anatomical structures without significant opacities, masses, or deviations, output class label 0.

2. If the image highlights substantial, well-defined opacities or masses and shapes with significant density, deviating significantly from normal structures, output class label 1.

3. Check for large, well-delineated fluid accumulations, substantial granular textures, or major abnormalities in anatomy. Minor, non-critical variations should output class label 0, focusing also on the context and location of tissues, otherwise class label 1.

\textbf{-------------------- Updated Theta --------------------}

1. If the image shows clear, consistent anatomical structures without significant opacities, masses, or deviations, output class label 0.

2. If the image highlights substantial, well-defined opacities or masses and shapes with significant density, deviating significantly from normal structures, output class label 1.

3. Check for large, well-delineated fluid accumulations, substantial granular textures, or major abnormalities in anatomy. Minor, non-critical variations should output class label 0, focusing also on the context and location of tissues, otherwise class label 1.
\end{exmp}
\newpage

\begin{exmp}{Medical Image Classification (without Prior) -- Step 50}{Med-Step-np 50}
\scriptsize

\textbf{Epoch 4 - Step 10}

\textbf{--------------- Forward Pass Prompt ---------------}

You are the model.

\textbf{** Model Descriptions: **}

1. If the image shows clear anatomical structures, uniform texture, symmetry, and normal lung fields without visible opacities, well-defined masses, streaks, irregular granular patterns, or any deviations, output class label 0.

2. If the image shows any distinct opacities or masses, interstitial patterns, visible fluid accumulations, asymmetry, irregular heart shapes, or multiple small opacities, output class label 1.

3. For granular or streaky opacities over any lung region, indistinct or uneven lung fields, subtle masses, minor deviations in symmetry, or any lung irregularities, output class label 1. For minimal irregularities or slight deviations, output class label 0.

\textbf{** Input: **}

See the given image.

Please give your output strictly in the following format:

```

Explanations: [Your step-by-step analyses and results]

Output:

[ONLY the integer class label; make necessary assumptions if needed]

'''

Please ONLY reply according to this format, don't give me any other words.

\textbf{--------------- Forward Pass Output ---------------}

Explanations: The image shows clear anatomical structures with uniform texture and symmetry. The lung fields appear normal without visible opacities, well-defined masses, streaks, or irregular granular patterns. There are no deviations observed.

Output: 0

\ 

\textbf{Many outputs for the rest data points ...}

\ 

\textbf{--------------- Forward Pass Output ---------------}

Explanations: The image shows clear anatomical structures with uniform texture and symmetry. The lung fields appear normal without visible opacities, well-defined masses, streaks, or irregular granular patterns. There are no deviations observed.

Output: 0

\textbf{--------------- Optimization Prompt ---------------}

You are the optimizer for a model, your goal is to learn the best descriptions for the model. The model used the Current Model Descriptions below predicted the class labels for the given images. You are given the target labels, please optimize the Model Descriptions for better prediction.

\textbf{** Inputs (a batch of i.i.d. images): **}

See the given images.

\textbf{** Current Model Descriptions: **}

1. If the image shows clear anatomical structures, uniform texture, symmetry, and normal lung fields without visible opacities, well-defined masses, streaks, irregular granular patterns, or any deviations, output class label 0.

2. If the image shows any distinct opacities or masses, interstitial patterns, visible fluid accumulations, asymmetry, irregular heart shapes, or multiple small opacities, output class label 1.

3. For granular or streaky opacities over any lung region, indistinct or uneven lung fields, subtle masses, minor deviations in symmetry, or any lung irregularities, output class label 1. For minimal irregularities or slight deviations, output class label 0.

\textbf{** The model predictions: **}

[[1] [0] [0] [0] [1] [1] [1] [0] [0] [0]]

\textbf{** The targets: **}

[[1] [0] [0] [1] [0] [1] [0] [0] [1] [1]]

If the model is doing well, you can keep using the current descriptions. However, if the model is not performing well, please update the model by improving upon the 'Current Model Descriptions', which should results in lower classification error on both the current and the next batch of i.i.d. data. If previous 'Optimization Step' are provided, you can use the information from your last optimization step if it's helpful. Limit your 'New Model Descriptions' to be less than 100 words! Please think step by step and give your outputs strictly in the following format:

```

Reasoning:

[be explicit and verbose, update the Current Model Descriptions by yourself; please show your work and use the features in the images; note that you don't have access to computers]

New Model Descriptions:

[put your new decision rules here, the rules MUST based on the features in the input image; make sure it is clear when to output which class labels]

'''

Please ONLY reply according to this format, don't give me any other words.

\end{exmp}
\newpage

\begin{exmp}{Medical Image Classification (without Prior) -- Step 50 (continued)}{Med-Step-np 50-2}
\scriptsize
\textbf{--------------- Optimization Output ---------------}

Reasoning:

The model misclassified several images where minor opacities or very light irregular patterns were present. We need to be more precise in detecting these subtle features while examining the overall lung fields more thoroughly to reduce false negatives for label 1. Specifically, minor irregularities or granular patterns that seem inconsequential should still be marked as potential abnormalities if they occur in specific regions, such as the periphery.

New Model Descriptions:

1. If the image shows clear anatomical structures, uniform texture, symmetry, and normal lung fields without any visible opacities, well-defined masses, streaks, or granular patterns, output class label 0.

2. If the image shows any distinct opacities or masses, interstitial patterns, visible fluid accumulations, asymmetry, irregular heart shapes, or multiple small opacities across the lung fields, output class label 1.

3. For granular or streaky opacities over any lung region, indistinct or uneven lung fields, subtle masses, minor deviations in symmetry, or any lung irregularities or granular patterns, including faint ones, in the periphery, output class label 1. For very minimal irregularities or slight deviations in central regions only, output class label 0.

\textbf{-------------------- Updated Theta --------------------}

1. If the image shows clear anatomical structures, uniform texture, symmetry, and normal lung fields without any visible opacities, well-defined masses, streaks, or granular patterns, output class label 0.

2. If the image shows any distinct opacities or masses, interstitial patterns, visible fluid accumulations, asymmetry, irregular heart shapes, or multiple small opacities across the lung fields, output class label 1.

3. For granular or streaky opacities over any lung region, indistinct or uneven lung fields, subtle masses, minor deviations in symmetry, or any lung irregularities or granular patterns, including faint ones, in the periphery, output class label 1. For very minimal irregularities or slight deviations in central regions only, output class label 0.

\end{exmp}